%% file: main.tex
\documentclass{article}
\pdfoutput=1

\usepackage[preprint,nonatbib]{neurips_2020}

\usepackage[utf8]{inputenc} %
\usepackage[T1]{fontenc}    %
\usepackage{hyperref}       %
\usepackage{url}            %
\usepackage{booktabs}       %
\usepackage{amsfonts}       %
\usepackage{nicefrac}       %
\usepackage{microtype}      %
\usepackage{xcolor}
\usepackage{hyperref}
\usepackage{graphicx}
\usepackage{algorithmicx}
\usepackage{algorithm}%
\usepackage{algpseudocode}%
\usepackage{mathrsfs}
\usepackage{amssymb}
\usepackage{graphicx,wrapfig,lipsum}
\usepackage{caption}
\usepackage{float}
\usepackage{subcaption}
\usepackage{enumitem}
\usepackage{siunitx}
\usepackage{multirow}
\usepackage{tikz}
\usepackage{footnote}
\usepackage{pdfpages}
\usetikzlibrary{arrows,shapes,snakes,automata,backgrounds,petri,positioning,calc}
\usepackage{lipsum}
\usepackage[multiple]{footmisc}
\usepackage{placeins}
\input{macros}

\input{choice_macros}

\usepackage{enumitem}
\setlistdepth{5}
\renewlist{itemize}{itemize}{5}
\setlist[itemize]{label=$\cdot$}
\setlist[itemize]{leftmargin=15pt}
\setlist[itemize,1]{label=\textbullet}
\setlist[itemize,2]{label=--}
\setlist[itemize,3]{label=*}
\setlist[itemize,4]{label=$\diamond$}

\setlist[enumerate]{leftmargin=*}

\usepackage[
backend=biber,
style=numeric-comp,
sorting=ynt
]{biblatex}

\title{What Matters In On-Policy Reinforcement Learning? A Large-Scale Empirical Study}

\author{%
  Marcin Andrychowicz,
  Anton Raichuk,
  Piotr Stańczyk,
  Manu Orsini, \\
  \textbf{Sertan Girgin,
  Raphael Marinier,
  Léonard Hussenot,
  Matthieu Geist,}\\
  \textbf{Olivier Pietquin,
  Marcin Michalski,
  Sylvain Gelly,
  Olivier Bachem} \\ \\
  Google Research, Brain Team
}

\begin{document}
\maketitle

\begin{abstract}
  In recent years, on-policy reinforcement learning (RL) has been successfully applied to many different continuous control tasks. While RL algorithms are often conceptually simple, their state-of-the-art implementations take numerous low- and high-level design decisions that strongly affect the performance of the resulting agents. Those choices are usually not extensively discussed in the literature, leading to discrepancy between published descriptions of algorithms and their implementations. This makes it hard to attribute progress in RL and slows down overall progress \cite{implementation_matters}. As a step towards filling that gap, we implement >50 such ``choices'' in a unified on-policy RL framework, allowing us to investigate their impact in a large-scale empirical study. We train over 250'000 agents in five continuous control environments of different complexity and provide insights and practical recommendations for on-policy training of RL agents.
\end{abstract}

\section{Introduction}
Deep reinforcement learning (RL) has seen increased interest in recent years due to its ability to have neural-network-based agents learn to act in environments through interactions.
For continuous control tasks, on-policy algorithms such as REINFORCE \cite{pg}, TRPO \cite{trpo}, A3C \cite{a3c}, PPO \cite{ppo} and off-policy algorithms such as DDPG \cite{ddpg} and SAC \cite{sac} have enabled successful applications such as quadrupedal
locomotion \cite{sac2}, self-driving \cite{kendall2019learning} or dexterous in-hand manipulation \cite{sac2, openai2018learning, openai2019solving}.

Many of these papers investigate in depth different loss functions and learning paradigms.
Yet, it is less visible that behind successful experiments in deep RL there are complicated code bases that contain a large number of low- and high-level design decisions that are usually not discussed in research papers.
While one may assume that such ``choices'' do not matter, there is some evidence that they are in fact crucial for or even driving good performance  \cite{implementation_matters}.

While there are open-source implementations available that can be used by practitioners, this is still unsatisfactory:
In research publications, often different algorithms implemented in different code bases are compared one-to-one.
This makes it impossible to assess whether improvements are due to the algorithms or due to their implementations.
Furthermore, without an understanding of lower-level choices, it is hard to assess the performance of high-level algorithmic choices as performance may strongly depend on the tuning of hyperparameters and implementation-level details.
Overall, this makes it hard to attribute progress in RL and slows down further research
\cite{henderson2018deep, implementation_matters, islam2017reproducibility}.

\textbf{Our contributions.}
Our key goal in this paper is to investigate such lower level choices in depth and to understand their impact on final agent performance.
Hence, as our key contributions, we (1) implement >50 choices in a unified on-policy algorithm implementation, 
(2) conducted a large-scale (more than 250'000 agents trained) experimental study that covers different aspects of the training process,
and (3) analyze the experimental results to provide practical insights and recommendations for the on-policy training of RL agents.

\textbf{Most surprising finding.}
While many of our experimental findings confirm common RL practices,
some of them are quite surprising, e.g.
the policy initialization scheme significantly influences the
performance while it is rarely even mentioned in RL publications.
In particular, we have found that initializing the network
so that the initial action distribution has zero mean, a rather low standard deviation
 and is independent of the observation significantly improves the training speed
(Sec.~\ref{sec:results-arch}).

The rest of of this paper is structured as follows:
We describe our experimental setup and performance metrics used in Sec.~\ref{sec:performance}.
Then, in Sec.~\ref{sec:results} we present and analyse the experimental results and
finish with related work in Sec.~\ref{sec:related} and conclusions in Sec.~\ref{sec:conclusions}.
The appendices contain the detailed description of all design choices we experiment with (App.~\ref{sec:choices}),
default hyperparameters (App.~\ref{sec:default-settings})
and the raw experimental results (App.~\ref{exp_final_losses}~-~\ref{exp_final_regularizer}).

\section{Study design}\label{sec:performance}

\paragraph{Considered setting.}
In this paper, we consider the setting of \emph{on-policy reinforcement learning for continuous control}.
We define on-policy learning in the following loose sense: We consider policy iteration algorithms that iterate between generating experience using the current policy and using the experience to improve the policy.
This is the standard \emph{modus operandi} of algorithms usually considered on-policy such as PPO \cite{ppo}.
However, we note that algorithms often perform several model updates and thus may operate technically on off-policy data within a single policy improvement iteration.
As benchmark environments, we consider five widely used continuous control environments from OpenAI Gym \cite{gym} of varying complexity: Hopper-v1, Walker2d-v1, HalfCheetah-v1, Ant-v1, and Humanoid-v1 \footnote{
It has been noticed that the version of the Mujoco physics simulator \cite{mujoco}
can slightly influence the behaviour of some of the environments --- \url{https://github.com/openai/gym/issues/1541}. We used Mujoco 2.0
in our experiments.}.

\paragraph{Unified on-policy learning algorithm.}
We took the following approach to create a highly configurable unified on-policy learning algorithm with as many choices as possible:
\begin{enumerate}
    \item We researched prior work and popular code bases to make a list of commonly used choices, i.e., different loss functions (both for value functions and policies), architectural choices such as initialization methods, heuristic tricks such as gradient clipping and all their corresponding hyperparameters.
    \item Based on this, we implemented a single, unified on-policy agent and corresponding training protocol starting from the SEED RL code base \cite{seed}.
    Whenever we were faced with implementation decisions that required us to take decisions that could not be clearly motivated or had alternative solutions, we further added such decisions as additional choices.
    \item We verified that when all choices are selected as in the PPO implementation
    from OpenAI baselines, we obtain similar performance as reported in the PPO paper \cite{ppo}.
    We chose PPO because it is probably the most commonly used on-policy RL algorithm at the moment.
\end{enumerate}

The resulting agent implementation is detailed in Appendix~\ref{sec:choices}.
The key property is that the implementation exposes all choices as configuration options in an unified manner.
For convenience, we mark each of the choice in this paper with a number (e.g., \dchoicep{numenvs}) and a fixed name (e.g. \choicet{numenvs}) that can be easily used to find a description of the choice in Appendix~\ref{sec:choices}.

\paragraph{Difficulty of investigating choices.}
The primary goal of this paper is to understand how the different choices affect the final performance of an agent and to derive recommendations for these choices.
There are two key reasons why this is challenging:

First, we are mainly interested in insights on choices for good  hyperparameter configurations. Yet,
if all choices are sampled randomly, the performance is very bad and
little (if any) training progress is made.
This may be explained by the presence of sub-optimal settings (e.g.,  hyperparameters of the wrong scale) that prohibit learning at all.
If there are many choices, the probability of such failure increases exponentially.

Second, many choices may have strong interactions with other related choices, for example the learning rate and the minibatch size.
This means that such choices need to be tuned together and experiments where only a single choice is varied but interacting choices are kept fixed may be misleading.

\paragraph{Basic experimental design.}
To address these issues, we design a series of experiments as follows:
We create groups of choices around thematic groups where we suspect interactions between different choices, for example we group together all choices related to neural network architecture.
We also include \choicet{adamlr} in all of the groups as we
suspect that it may interact with many other choices.

Then, in each experiment, we train a large number of models where we randomly sample
the choices within the corresponding group \footnote{Exact details for the different experiments are provided in Appendices~\ref{exp_final_losses}~-~\ref{exp_final_regularizer}.}. 
All other settings (for choices not in the group) are set to settings of a competitive base configuration (detailed in Appendix~\ref{sec:default-settings}) that is close to the default PPOv2 configuration\footnote{\url{https://github.com/openai/baselines/blob/master/baselines/ppo2/defaults.py}}
scaled up to $256$ parallel environments.
This has two effects: 
First, it ensures that our set of trained models contains good models (as verified by performance statistics in the corresponding results).
Second, it guarantees that we have models that have different combinations of potentially interacting choices.

We then consider two different analyses for each choice (e.g, for \choicet{advantageestimator}):

\emph{Conditional 95th percentile}: For each potential value of that choice (e.g., \choicet{advantageestimator} = \texttt{N-Step}), we look at the performance distribution of sampled configurations with that value.
We report the 95th percentile of the performance as well as a confidence interval based on a binomial approximation \footnote{We compute confidence intervals with a significance level of $\alpha=5\%$ as follows: We find $i_l = icdf\left(\frac\alpha2\right)$ and $i_h = icdf\left(1-\frac\alpha2\right)$ where $icdf$ is the inverse cumulative density function of a binomial distribution with $p=0.95$ (as we consider the 95th percentile) and the number of draws equals the number of samples. We then report the $i_l$th  and $i_h$th highest scores as the confidence interval.}.
Intuitively, this corresponds to a robust estimate of the performance one can expect if all other choices in the group were tuned with random search and a limited budget of roughly 20  hyperparameter configurations.

\emph{Distribution of choice within top 5\% configurations.}
We further consider for each choice the distribution of values among the top 5\%
configurations trained in that experiment.
The reasoning is as follows: 
By design of the experiment, values for each choice are distributed uniformly at random.
Thus, if certain values are over-represented in the top models, this indicates that the specific choice is important in guaranteeing good performance.

\paragraph{Performance measures.}
We employ the following way to compute performance:
For each  hyperparameter configuration, we train $3$ models with independent random seeds where each model is trained for one million (Hopper, HalfCheetah, Walker2d) or two million environment steps (Ant, Humanoid).
We evaluate trained policies every hundred thousand steps by freezing the policy and computing the average undiscounted episode return of 100 episodes (with the stochastic policy).
We then average these score to obtain a single performance score of the seed
which is proportional to the area under the learning curve.
This ensures we assign higher scores to agents that learn quickly.
The performance score of a hyperparameter configuration is finally set to the median performance score across the 3 seeds.
This reduces the impact of training noise, i.e., that certain seeds of the same configuration may train much better than others.

\input{results.tex}

\section{Related Work}\label{sec:related}

Islam~et~al.~\cite{islam2017reproducibility} and
Henderson~et~al.~\cite{henderson2018deep}
point out the reproducibility issues in RL
including the performance differences between different code bases,
the importance of hyperparameter tuning and the high level
of stochasticity due to random seeds. Tucker~et~al.~\cite{tucker2018mirage} showed that the gains, which had been
attributed to one of the recently 
proposed policy gradients improvements,
were, in fact, caused by the implementation details.
The most closely related work to ours is probably Engstrom~et~al.~\cite{implementation_matters}
where the authors investigate code-level improvements in the
PPO \cite{ppo} code base and conclude that they are responsible
for the most of the performance difference between PPO and TRPO
\todo{uncomment for the camera-ready version}
\cite{trpo}.
Our work is also similar to other large-scale studies
done in other fields of Deep Learning, e.g.
model-based RL \cite{langlois2019benchmarking},
GANs \cite{lucic2018gans}, NLP \cite{kaplan2020scaling}, disentangled representations~\cite{locatello2018challenging}
and convolution network architectures \cite{radosavovic2020designing}.

\section{Conclusions}\label{sec:conclusions}
In this paper, we investigated the importance of a broad set of high- and low-level choices that need to be made when designing and implementing on-policy learning algorithms.
Based on more than 250'000 experiments in five continuous control environments, we evaluate the impact of different choices and provide practical recommendations.
One of the surprising insights is that the initial action distribution plays an important role in agent performance.
We expect this to be a fruitful avenue for future research.

\todo{uncomment for the camera-ready version}
\small
{
\small
\printbibliography
}

\newpage
\appendix
\input{rl.tex}
\input{choices.tex}

\input{default.tex}

\input{final_losses/main.tex}
\input{final_arch2/main.tex}
\input{final_stability/main.tex}
\input{final_advantages/main.tex}
\input{final_setup/main.tex}
\input{final_time/main.tex}
\input{final_optimize/main.tex}
\input{final_regularizer/main.tex}

\end{document}

%% file: macros.tex
\renewcommand{\S}{\mathcal{S}}

\newcommand{\A}{\mathcal{A}}
\newcommand{\N}{\mathcal{N}}
\renewcommand{\L}{\mathcal{L}}
\newcommand{\R}{\mathbb{R}}
\newcommand{\E}{\mathbb{E}}
\newcommand{\todo}[1]{}

\newcommand{\grad}{\nabla_\theta}
\newcommand{\sg}{\textbf{sg}}
\newcommand{\kl}{\texttt{KL}}
\newcommand{\ent}{\texttt{H}}

%% file: choice_macros.tex
\newcounter{choice}[section]
\DeclareRobustCommand{\setchoice}[1]{%
   \refstepcounter{choice}%
 \label{choice:#1}}
\newcommand{\choicep}[1]{\texttt{C\ref{choice:#1}}}
\newcommand{\choicet}[1]{\texttt{\csname#1\endcsname{ }(C\ref{choice:#1})}}
\newcommand{\choicetable}[1]{\texttt{C\ref{choice:#1}} & \texttt{\csname#1\endcsname}}

\newcommand{\dchoicep}[1]{\choicep{#1}\setchoice{#1}}
\newcommand{\dchoicet}[1]{\choicet{#1}\setchoice{#1}}
\newcommand{\dchoicetable}[1]{\choicetable{#1}\setchoice{#1}}

%% file: results.tex
\section{Experiments}\label{sec:results}

We run experiments for eight thematic groups:
\emph{Policy Losses} (Sec.~\ref{sec:results-losses}), 
\emph{Networks architecture} (Sec.~\ref{sec:results-arch}),
\emph{Normalization and clipping} (Sec.~\ref{sec:results-stability}),
\emph{Advantage Estimation} (Sec.~\ref{sec:results-advantage}),
\emph{Training setup} (Sec.~\ref{sec:results-setup}),
\emph{Timesteps handling} (Sec.~\ref{sec:results-time}),
\emph{Optimizers} (Sec.~\ref{sec:results-opt}),
and \emph{Regularization} (Sec.~\ref{sec:results-reg}).
For each group, we provide a full experimental design and full experimental plots in Appendices~\ref{exp_final_losses}~-~\ref{exp_final_regularizer} so that the reader can draw their own conclusions from the experimental results.
In the following sections, we provide short descriptions of the experiments, our interpretation of the results, as well as practical recommendations for on-policy training for continuous control.

\subsection{Policy losses (based on the results in Appendix~\ref{exp_final_losses})}\label{sec:results-losses}

\textbf{Study description.} 
We investigate different policy losses (\choicep{policyloss}): vanilla policy gradient (PG), V-trace \cite{impala}, PPO \cite{ppo}, AWR \cite{awr}, V-MPO\footnote{
We used the V-MPO policy loss without the decoupled KL constraint as we investigate
the effects of different policy regularizers separately in Sec.~\ref{sec:results-reg}.} \cite{vmpo} and the limiting case of AWR ($\beta \rightarrow 0$) and V-MPO ($\epsilon_n \rightarrow 0$)
which we call Repeat Positive Advantages (RPA) as it is equivalent to
the negative log-probability of actions with positive advantages.
See App.~\ref{sec:choices-losses} for a detailed description of the different losses.
We further sweep the hyperparameters of each of the losses (\choicep{vtracelossrho}, \choicep{ppoepsilon}, \choicep{awrbeta}, \choicep{awrw}, \choicep{vmpoeps}),
the learning rate (\choicep{adamlr}) and the number of passes over the data (\choicep{numepochsperstep}).

The goal of this study is to better understand the importance of the policy loss function in the on-policy setting considered in this paper.
The goal is \textbf{not} to provide a general statement that one of the losses is better than the others as some of them were specifically designed for other settings
(e.g., the V-trace loss is targeted at near-on-policy data in a distributed setting).

\begin{figure}[h]
  \centering
  \includegraphics[width=1\textwidth]{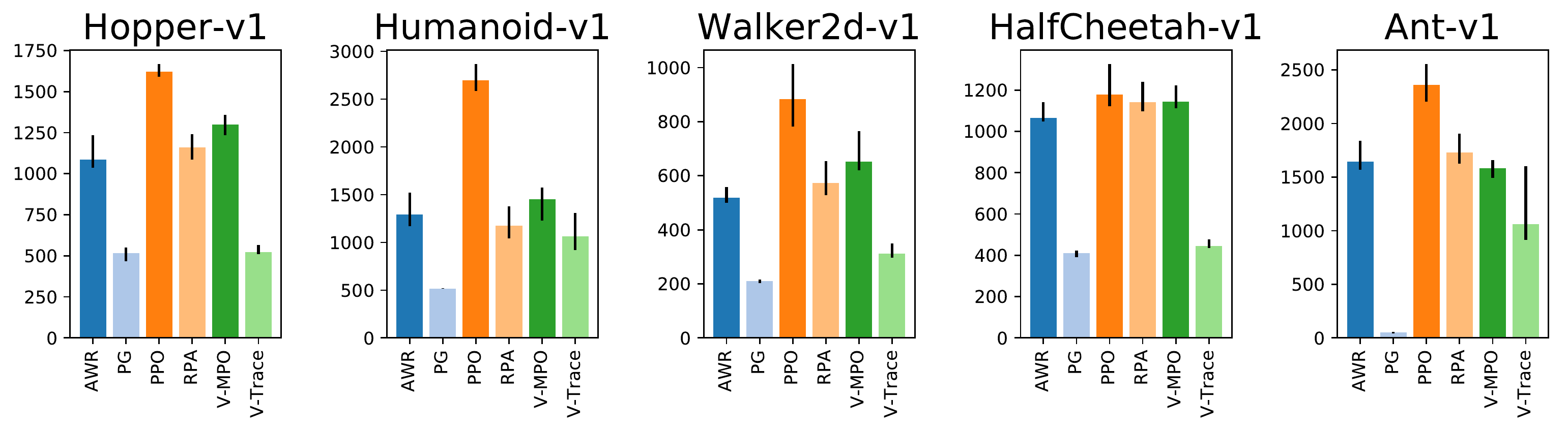}
  \caption{Comparison of different policy losses (\choicep{policyloss}).}
  \label{fig:losses}
\end{figure}

\textbf{Interpretation.} 
Fig.~\ref{fig:losses} shows the 95-th percentile of the average policy score during training for different policy losses  (\choicep{policyloss}).
We observe that PPO performs better than the other losses on 4 out of 5 environments and is
one of the top performing losses on HalfCheetah.
As we randomly sample the loss specific hyperparameters in this analysis, one might argue that our approach favours choices that are not too sensitive to hyperparameters.
At the same time, there might be losses that are sensitive to their hyperparameters but for which good settings may be easily found.
Fig.~\ref{fig:final_losses_custom_loss} shows that even if we condition on choosing the optimal
loss hyperparameters for each loss\footnote{AWR loss has two hyperparameters --- the temperature $\beta$ (\choicep{awrbeta}) and the weight clipping coefficient $\omega_\texttt{max}$ (\choicep{awrw}). We only condition on $\beta$ which is more important.},
PPO still outperforms the other losses on the two hardest tasks --- Humanoid and Ant\footnote{
These two tasks were not included in the original PPO paper \cite{ppo} so the hyperparameters we use were not tuned for them.} and is one of the top
performing losses on the other $3$ tasks.
Moreover, we show the empirical cumulative density functions of agent performance conditioned on the
policy loss used in Fig.~\ref{fig:final_losses__ecdf_standard_policy_losses}.
Perhaps unsurprisingly, PG and V-trace perform worse on all tasks.
This is likely caused by their inability to handle data that become off-policy in one iteration, either due to multiple passes (\choicep{numepochsperstep}) over experience (which can be seen in Fig.~\ref{fig:final_losses__correlation_epochs_per_step_vs_losses}) or a large experience buffer (\choicep{stepsize}) in relation to the batch size (\choicep{batchsize}).
Overall, these results show that trust-region optimization (preventing the current policy from diverging too much from the behavioral one) which is present in all the other policy losses is crucial for good sample complexity.
For PPO and its clipping threshold $\epsilon$ (\choicep{ppoepsilon}), we further observe that $\epsilon=0.2$ and $\epsilon=0.3$ perform reasonably well in all environments but that lower ($\epsilon=0.1$) or higher ($\epsilon=0.5$) values give better performance on some of the environments (See Fig.~\ref{fig:final_losses__gin_study_design_choice_value_sub_standard_policy_losses_ppo_ppo_epsilon} and Fig.~\ref{fig:final_stability__gin_study_design_choice_value_ppo_epsilon}).

\textbf{Recommendation.} Use the PPO policy loss. Start with the clipping threshold
set to $0.25$ but also try lower and higher values if possible.

\subsection{Networks architecture (based on the results in Appendix~\ref{exp_final_arch2})}\label{sec:results-arch}

\textbf{Study description.} 
We investigate the impact of differences in the policy and value function neural network architectures.
We consider choices related to the network structure and size (\choicep{mlpshared}, \choicep{sharedwidth}, \choicep{policywidth}, \choicep{valuewidth}, \choicep{shareddepth}, \choicep{policydepth}, \choicep{policydepth}), activation functions (\choicep{activation}), and initialization of network weights (\choicep{init}, \choicep{policyinit}, \choicep{valueinit}).
We further include choices related to the standard deviation of actions (\choicep{stdind}, \choicep{stdtransform}, \choicep{initialstd}, \choicep{minstd}) and transformations of sampled actions (\choicep{actionpost}).

\textbf{Interpretation.}
Separate value and policy networks (\choicep{mlpshared}) appear to lead to better performance on four out of five environments (Fig.~\ref{fig:final_arch__mlpshared}).
To avoid analyzing the other choices based on bad models, we thus focus for the rest of this experiment only on agents with separate value and policy networks.
Regarding network sizes, the optimal width of the policy MLP depends on the complexity of the environment (Fig.~\ref{fig:final_arch2__gin_study_design_choice_value_policy_mlp_width}) and too low or too high values can cause significant drop in performance while for the value function there seems to be no downside in using wider networks (Fig.~\ref{fig:final_arch2__gin_study_design_choice_value_value_mlp_width}).
Moreover, on some environments it is beneficial to make the value network wider than the policy one, e.g. on HalfCheetah the best results are achieved with $16-32$ units per layer in the policy network and $256$ in the value network.
Two hidden layers appear to work well for policy (Fig.~\ref{fig:final_arch2__gin_study_design_choice_value_policy_mlp_depth}) and value networks (Fig.~\ref{fig:final_arch2__gin_study_design_choice_value_value_mlp_depth}) in all tested environments.
As for activation functions, we observe that \texttt{tanh} activations
perform best and \texttt{relu} worst. (Fig.~\ref{fig:final_arch2__gin_study_design_choice_value_activation}).

\begin{figure}[h]
  \centering
  \includegraphics[width=\textwidth]{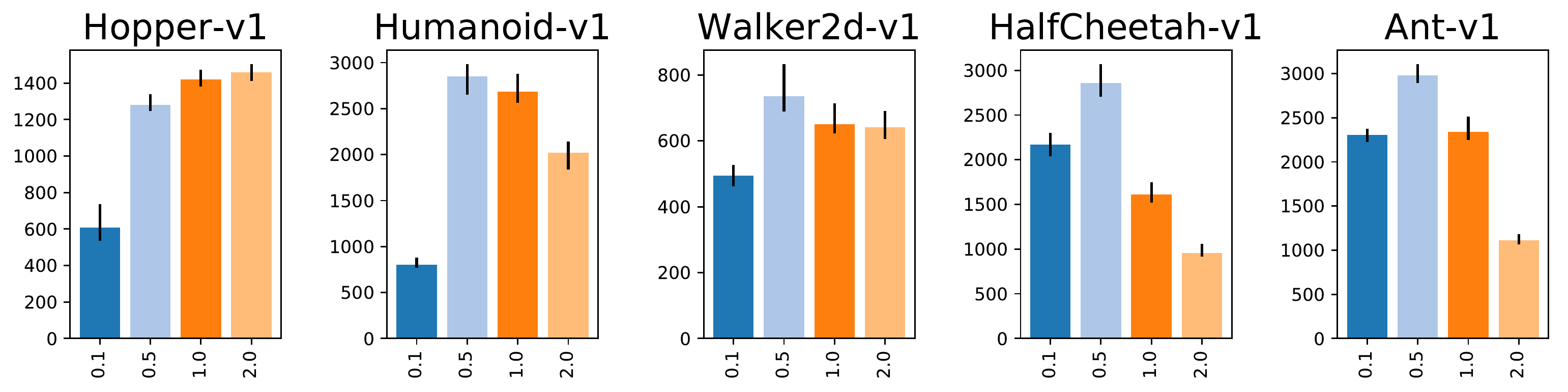}
  \caption{Comparison of different initial standard deviations of actions (\choicep{initialstd}).}
  \label{fig:initialstd}
\end{figure}

Interestingly, the initial policy appears to have a surprisingly high impact on the training performance.
The key recipe appears is to initialize the policy at the beginning of training so that the action distribution is centered around $0$\footnote{All environments expect normalized actions in $[-1, 1]$.} regardless of the observation and has a rather small standard deviation.
This can be achieved by initializing the policy MLP with smaller weights in the last layer (\choicep{policyinit}, Fig.~\ref{fig:final_arch2__gin_study_design_choice_value_last_kernel_init_policy_scaling}, this alone boosts the performance on Humanoid by 66\%)
so that the initial action distribution is almost independent of the observation and by introducing an offset in the standard deviation of actions (\choicep{initialstd}).
Fig.~\ref{fig:initialstd} shows that the performance is very sensitive to the initial action standard deviation
with 0.5 performing best on all environments except Hopper where higher values perform better.

Fig.~\ref{fig:final_arch2__gin_study_design_choice_value_action_postprocessing} compares two approaches to transform unbounded sampled actions into the bounded $ [-1,\,1]$ domain expected by the environment (\choicep{actionpost}): clipping and applying a \texttt{tanh} function.
\texttt{tanh} performs slightly better overall (in particular it improves the performance on HalfCheetah by $30\%$ ). Comparing Fig.~\ref{fig:final_arch2__gin_study_design_choice_value_action_postprocessing}
and Fig.~\ref{fig:initialstd} suggests that the difference might be mostly caused by the decreased magnitude
of initial actions\footnote{\texttt{tanh} can also potentially perform better with entropy regularization
(not used in this experiment) as it bounds the maximum possible policy entropy.}.

Other choices appear to be less important:
The scale of the last layer initialization matters much less for the value MLP (\choicep{valueinit}) than for the policy MLP
(Fig.~\ref{fig:final_arch2__gin_study_design_choice_value_last_kernel_init_value_scaling}).
Apart from the last layer scaling, the network initialization scheme (\choicep{init}) does not matter too much (Fig.~\ref{fig:final_arch2__gin_study_design_choice_value_initializer}). 
Only \texttt{he\_normal} and \texttt{he\_uniform} \cite{he2015delving} appear to be suboptimal choices with the other options performing very similarly.
There also appears to be no clear benefits if the standard deviation of the policy is learned for each state (i.e. outputted by the policy network) or once globally for all states (\choicep{stdind}, Fig.~\ref{fig:final_arch2__gin_study_design_choice_value_std_independent_of_input}).
For the transformation of policy output into action standard deviation (\choicep{stdtransform}), \texttt{softplus} and exponentiation perform very similarly\footnote{We noticed that some
of the training runs with exponentiation resulted in NaNs but clipping
the exponent solves this issue (See Sec.~\ref{sec:choices-action} for the details).}
(Fig.~\ref{fig:final_arch2__gin_study_design_choice_value_scale_function}).
Finally, the minimum action standard deviation (\choicep{minstd}) seems to matter little, if it is not set too large (Fig.~\ref{fig:final_arch2__gin_study_design_choice_value_activation}).

\textbf{Recommendation.} 
Initialize the last policy layer with $100 \times$ smaller weights.
Use $\texttt{softplus}$ to transform network output into action standard deviation and add a (negative) offset to its input to decrease the initial standard deviation of actions.
Tune this offset if possible.
Use \texttt{tanh} both as the activation function (if the networks are not too deep) and to transform the samples from the normal distribution to the bounded action space.
Use a wide value MLP (no layers shared with the policy) but tune the policy width (it might need to be narrower than the value MLP).

\subsection{Normalization and clipping (based on the results in Appendix~\ref{exp_final_stability})}\label{sec:results-stability}

\textbf{Study description.} 
We investigate the impact of different normalization techniques:
observation normalization (\choicep{norminput}),
value function normalization (\choicep{normreward}),
per-minibatch advantage normalization (\choicep{normadv}),
as well as gradient (\choicep{clipgrad}) and
observation (\choicep{clipinput}) clipping.

\textbf{Interpretation.}
Input normalization (\choicep{norminput}) is crucial for good performance on
all environments apart from Hopper (Fig.~\ref{fig:final_stability__gin_study_design_choice_value_input_normalization}).
Quite surprisingly, value function normalization (\choicep{normreward}) also influences the performance
very strongly --- it is crucial for good performance on HalfCheetah and Humanoid,
helps slightly on Hopper and Ant and significantly \emph{hurts} the performance on Walker2d (Fig.~\ref{fig:final_stability__gin_study_design_choice_value_reward_normalization}).
We are not sure why the value function scale matters that much but suspect that
it affects the performance by changing the speed of the value function fitting.\footnote{
Another explanation could be the interaction between the value function normalization
and PPO-style value clipping (\choicep{ppovalueclip}).
We have, however, disable the value clipping in this experiment to avoid
this interaction. The disabling of the value clipping could
also explain why our conclusions are different from \cite{implementation_matters}
where a form of value normalization improved the performance on Walker.}
In contrast to observation and value function normalization, per-minibatch
advantage normalization (\choicep{normadv}) seems not to affect the performance too much
(Fig.~\ref{fig:final_stability__gin_study_design_choice_value_normalize_advantages}).
Similarly, we have found little evidence that clipping normalized\footnote{We only applied clipping if input normalization was enabled.} observations (\choicep{clipinput})
helps (Fig.~\ref{fig:final_stability__gin_study_design_choice_value_sub_input_normalization_avg_compfalse_input_clipping}) but it might be worth
using if there is a risk of extremely high observations due to simulator divergence.
Finally, gradient clipping (\choicep{clipgrad}) provides a small performance boost
with the exact clipping threshold making little difference
(Fig.~\ref{fig:final_stability__gin_study_design_choice_value_gradient_clipping}).

\textbf{Recommendation.} Always use observation normalization and
check if value function normalization improves performance.
Gradient clipping might slightly help
but is of secondary importance.

\subsection{Advantage Estimation (based on the results in Appendix~\ref{exp_final_advantages})}\label{sec:results-advantage}

\textbf{Study description.}
We compare the most commonly used advantage
estimators (\choicep{advantageestimator}):
N-step \cite{sutton1998introduction}, GAE \cite{gae} and V-trace \cite{impala}
and their hyperparameters (\choicep{nstep}, \choicep{gaelambda},
\choicep{vtraceaelambda}, \choicep{vtraceaecrho}).
We also experiment with applying PPO-style pessimistic clipping (\choicep{ppovalueclip})
to the value loss (present in the original PPO implementation but not mentioned
in the PPO paper \cite{ppo}) and using Huber loss \cite{huber} instead of MSE for value learning (\choicep{valueloss}, \choicep{huberdelta}).
Moreover, we varied the number of parallel environments used (\choicep{numenvs}) as it changes the length of the experience fragments collected in each step.

\textbf{Interpretation.} 
GAE and V-trace appear to perform better than N-step returns
(Fig.~\ref{fig:final_advantages__gin_study_design_choice_value_advantage_estimator}~and~\ref{fig:final_advantages_custom_advantage_estimator})
which indicates that it is beneficial to combine the value estimators from multiple
timesteps.
We have not found a significant performance difference between GAE and V-trace
in our experiments.
$\lambda=0.9$ (\choicep{gaelambda}, \choicep{vtraceaelambda})
performed well regardless of whether GAE
(Fig.~\ref{fig:final_advantages__gin_study_design_choice_value_sub_advantage_estimator_gae_gae_lambda})
or V-trace
(Fig.~\ref{fig:final_advantages__gin_study_design_choice_value_sub_advantage_estimator_v_trace_lambda})
was used on all tasks but tuning this value per environment may
lead to modest performance gains.
We have found that PPO-style value loss clipping (\choicep{ppovalueclip})
hurts the performance regardless of the clipping threshold\footnote{
This is consistent with prior work \cite{implementation_matters}.}
(Fig.~\ref{fig:final_advantages__gin_study_design_choice_value_ppo_style_value_clipping_epsilon}).
Similarly, the Huber loss (\choicep{valueloss}) performed worse than MSE in all environments
(Fig.~\ref{fig:final_advantages__gin_study_design_choice_value_value_loss})
regardless of the value of the threshold (\choicep{huberdelta}) used
(Fig.~\ref{fig:final_advantages__gin_study_design_choice_value_sub_value_loss_huber_delta}).

\textbf{Recommendation.} Use GAE with $\lambda=0.9$ but neither Huber loss nor PPO-style value loss clipping.

\subsection{Training setup (based on the results in Appendix~\ref{exp_final_setup})}\label{sec:results-setup}

\textbf{Study description.}
We investigate choices related to the data collection and minibatch handling: the number of parallel
environments used (\choicep{numenvs}),
the number of transitions gathered in each iteration
(\choicep{stepsize}), the number of passes over the data (\choicep{numepochsperstep}),
minibatch size (\choicep{batchsize}) and how the data is split into minibatches (\choicep{batchhandling}).

For the last choice, in addition to standard choices, we also consider a new small modification of the original PPO approach:
The original PPO implementation splits the data in each policy iteration step into individual transitions and then randomly assigns them to minibatches (\choicep{batchhandling}).
This makes it impossible to compute advantages as the temporal structure is broken. Therefore, the advantages are computed once at the beginning of each policy iteration step and then used in minibatch policy and value function optimization. This results in higher diversity of data in each minibatch at the cost of using slightly stale advantage estimations.
As a remedy to this problem, we propose to recompute the advantages at the beginning of each pass over the data instead of just once per iteration.

\textbf{Results.} 
Unsurprisingly, going over the experience multiple times appears to be crucial for good sample complexity
(Fig.~\ref{fig:final_setup__gin_study_design_choice_value_epochs_per_step}).
Often, this is computationally cheap due to the simple models considered, in particular on machines with accelerators such as GPUs and TPUs.
As we increase the number of parallel environments (\choicep{numenvs}),
performance decreases sharply on some of the environments (Fig.~\ref{fig:final_setup__gin_study_design_choice_value_num_actors_in_learner}). This is likely caused by shortened experience chunks (See Sec.~\ref{sec:choices-setup}
for the detailed description of the data collection process) and
earlier value bootstrapping.
Despite that, training with more environments usually leads to faster
training in wall-clock time if enough CPU cores are available.
Increasing the batch size (\choicep{batchsize}) does not appear to hurt the sample complexity
in the range we tested (Fig.~\ref{fig:final_setup__gin_study_design_choice_value_batch_size_transitions}) which suggests that it should be increased for faster iteration speed.
On the other hand, the number of transitions gathered in each iteration (\choicep{stepsize})
influences the performance quite significantly
(Fig.~\ref{fig:final_setup__gin_study_design_choice_value_step_size_transitions}).
Finally, we compare different ways to handle minibatches (See Sec.~\ref{sec:choices-setup}
for the detailed description of different variants) in Fig.~\ref{fig:final_setup__gin_study_design_choice_value_batch_mode}~and~
\ref{fig:final_setup2__gin_study_design_choice_value_batch_mode}.
The plots suggest that stale advantages can in fact hurt performance
and that recomputing them at the beginning of each pass at least partially
mitigates the problem and performs best among all variants.

\textbf{Recommendation.} Go over experience multiple times.
Shuffle individual transitions before assigning them to minibatches
and recompute advantages once per data pass (See App.~\ref{sec:choices-setup} for the details).
For faster wall-clock time training use many parallel environments
and increase the batch size (both might hurt the sample complexity).
Tune the number of transitions in each iteration (\choicep{stepsize})
if possible.

\subsection{Timesteps handling (based on the results in Appendix~\ref{exp_final_time})}\label{sec:results-time}

\textbf{Study description.} We investigate choices related to the handling of timesteps: discount factor\footnote{While the discount factor is sometimes treated as a part of the environment, we assume that the real goal
is to maximize \emph{undiscounted} returns and the discount factor is a part of the algorithm which makes learning easier.}~(\choicep{discount}), frame skip (\choicep{frameskip}), and how episode termination due to timestep limits are handled (\choicep{handleabandon}).
The latter relates to a technical difficulty explained in App.~\ref{sec:choices-time} where one assumes for the algorithm an infinite time horizon
but then trains using a finite time horizon \cite{pardo2017time}.

\textbf{Interpretation.}
Fig.~\ref{fig:final_time__gin_study_design_choice_value_discount_factor} shows that the performance depends heavily on the discount factor $\gamma$ (\choicep{discount}) with $\gamma=0.99$ performing reasonably well in all environments.
Skipping every other frame (\choicep{frameskip}) improves the performance on $2$ out of $5$ environments (Fig.~\ref{fig:final_time__gin_study_design_choice_value_frame_skip}).
Proper handling of episodes abandoned due to the timestep limit seems not to affect the performance
(\choicep{handleabandon}, Fig.~\ref{fig:final_time__gin_study_design_choice_value_handle_abandoned_episodes_properly})
which is probably caused by the fact that the timestep limit
is quite high ($1000$ transitions) in all the environments we considered.

\textbf{Recommendation.} 
Discount factor $\gamma$ is one of the most important hyperparameters and should be tuned per environment (start with $\gamma=0.99$).
Try frame skip if possible. 
There is no need to handle environments step limits in a special way for large step limits.

\subsection{Optimizers (based on the results in Appendix~\ref{exp_final_optimize})}\label{sec:results-opt}
\textbf{Study description.} 
We investigate two gradient-based optimizers commonly used in RL: (\choicep{optimizer}) -- Adam \cite{adam} and RMSprop -- as well as their hyperparameters (\choicep{adamlr}, \choicep{rmslr}, \choicep{adammom}, \choicep{rmsmom}, \choicep{adameps}, \choicep{rmseps}, \choicep{rmscent}) and a linear learning rate decay schedule (\choicep{lrdecay}).

\textbf{Interpretation.}
The differences in performance between the optimizers (\choicep{optimizer}) appear to be rather small with no optimizer consistently outperforming the other across environments (Fig.~\ref{fig:final_optimize__gin_study_design_choice_value_optimizer}).
Unsurprisingly, the learning rate influences the performance very strongly (Fig.~\ref{fig:final_optimize__gin_study_design_choice_value_sub_optimizer_adam_learning_rate}) with the default value of $0.0003$ for Adam (\choicep{adamlr}) performing well on all tasks.
Fig.~\ref{fig:final_optimize__gin_study_design_choice_value_sub_optimizer_adam_momentum} shows that Adam works better with momentum (\choicep{adammom}).
For RMSprop, momentum (\choicep{rmsmom}) makes less difference (Fig.~\ref{fig:final_optimize__gin_study_design_choice_value_sub_optimizer_rmsprop_momentum}) but our results suggest that it might slightly improve performance\footnote{Importantly, switching from no momentum to momentum 0.9 increases the RMSprop step size by approximately 10$\times$ and requires an appropriate adjustment to the learning rate (Fig.~\ref{fig:final_optimize__correlation_rmsprop_momentum_vs_lr}).}.
Whether the centered or uncentered version of RMSprop is used (\choicep{rmscent}) makes no difference (Fig.~\ref{fig:final_optimize__gin_study_design_choice_value_sub_optimizer_rmsprop_centered}) and similarly we did not find any difference between different values of the $\epsilon$ coefficients (\choicep{adameps}, \choicep{rmseps}, Fig.~\ref{fig:final_optimize__gin_study_design_choice_value_sub_optimizer_adam_epsilon}~and~\ref{fig:final_optimize__gin_study_design_choice_value_sub_optimizer_rmsprop_epsilon}).
Linearly decaying the learning rate to $0$ increases the performance on $4$ out of $5$ tasks but the gains are very small apart from Ant, where it leads to $15\%$ higher scores (Fig.~\ref{fig:final_optimize__gin_study_design_choice_value_learning_rate_decay}).

\textbf{Recommendation.} Use Adam \cite{adam} optimizer with momentum $\beta_1=0.9$
and a tuned learning rate ($0.0003$ is a safe default).
Linearly decaying the learning rate may slightly improve performance but is
of secondary importance.

\subsection{Regularization (based on the results in Appendix~\ref{exp_final_regularizer})}\label{sec:results-reg}

\textbf{Study description.} 
We investigate different policy regularizers (\choicep{regularizationtype}),
which can have either the form of a penalty (\choicep{regularizerpenalty}, e.g. bonus for higher entropy)
or a soft constraint (\choicep{regularizerconstraint}, e.g. entropy should not be lower than some threshold) which is enforced with a Lagrange multiplier.
In particular, we consider the following regularization terms: entropy (\choicep{regularizerconstraintentropy}, \choicep{regularizerpenaltyentropy}), the Kullback–Leibler divergence (KL) between a reference $\N(0,1)$ action distribution and the current policy (\choicep{regularizerconstraintklrefpi}, \choicep{regularizerpenaltyklrefpi}) and the KL divergence and reverse KL divergence between the current policy and the behavioral one 
(\choicep{regularizerconstraintklmupi}, \choicep{regularizerpenaltyklmupi}, \choicep{regularizerconstraintklpimu}, \choicep{regularizerpenaltyklpimu}), as well as the ``decoupled'' KL divergence from \cite{mpo,vmpo} (\choicep{regularizerconstraintklmupimean}, \choicep{regularizerconstraintklmupistd}, \choicep{regularizerpenaltyklmupimean}, \choicep{regularizerpenaltyklmupistd}).

\textbf{Interpretation.}
We do not find evidence that any of the investigated regularizers helps significantly
on our environments with the exception of HalfCheetah on which all constraints (especially the entropy constraint) help (Fig.~\ref{fig:final_regularizer_custom_regularization}~and~\ref{fig:final_regularizer__gin_study_design_choice_value_policy_regularization}).
However, the performance boost is largely independent on the constraint threshold
(Fig.~\ref{fig:final_regularizer__gin_study_design_choice_value_sub_regularization_constraint_entropy_entropy_threshold},
\ref{fig:final_regularizer__gin_study_design_choice_value_sub_regularization_constraint_klmupi_kl_mu_pi_threshold},
\ref{fig:final_regularizer__gin_study_design_choice_value_sub_regularization_constraint_klpimu_kl_pi_mu_threshold},
\ref{fig:final_regularizer__gin_study_design_choice_value_sub_regularization_constraint_decoupled_klmupi_kl_mu_pi_mean_threshold},
\ref{fig:final_regularizer__gin_study_design_choice_value_sub_regularization_constraint_decoupled_klmupi_kl_mu_pi_std_threshold} and
\ref{fig:final_regularizer__gin_study_design_choice_value_sub_regularization_constraint_klrefpi_kl_ref_pi_threshold}) which suggests that the effect is caused by the initial high strength of the penalty (before it gets adjusted) and not by the desired constraint.
While it is a bit surprising that regularization does not help at all (apart from HalfCheetah),
we conjecture that regularization might be less important in our experiments
because: (1) the PPO policy loss already enforces the trust region which makes KL penalties or constraints redundant; and
(2) the careful policy initialization (See Sec.~\ref{sec:results-arch})
is enough to guarantee good exploration and makes the entropy bonus or constraint redundant.

%% file: rl.tex
\section{Reinforcement Learning Background}\label{sec:rl}

We consider the standard reinforcement learning formalism
consisting of an agent interacting with an environment.
To simplify the exposition we assume in this section that the environment is fully observable.
An environment
is described by
a set of states $\S$,
a set of actions $\A$,
a distribution of initial states $p(s_0)$,
a reward function $r : \S \times \A \rightarrow \R$,
transition probabilities $p(s_{t+1}|s_t,a_t)$ ($t$ is a timestep index explained later),
termination probabilities $T(s_t,a_t)$
and a discount factor $\gamma \in [0,1]$.

A policy $\pi$ is a mapping from state to a distribution over actions.
Every episode starts by sampling an initial state $s_0$.
At every timestep $t$ the agent produces an action based on the current state:
$a_t \sim \pi(\cdot|s_t)$.
In turn, the agent receives a reward $r_t=r(s_t,a_t)$ and the environment's state is updated.
With probability $T(s_t,a_t)$ the episode is terminated, and otherwise the
new environments state $s_{t+1}$ is sampled from $p(\cdot|s_t,a_t)$.
The discounted sum of future rewards, also referred to as the \emph{return}, is defined as
$R_t=\sum_{i=t}^\infty \gamma^{i-t} r_i$.
The agent's goal is to find the policy $\pi$ which maximizes the expected return $\E_\pi [R_0|s_0]$, where
the expectation is taken over the initial state distribution, the policy, and environment transitions accordingly to the dynamics
specified above.
The \emph{Q-function} or \emph{action-value} function of a given policy $\pi$ is defined as $Q^\pi(s_t,a_t)=\E_\pi[R_t|s_t,a_t]$, while the
\emph{V-function} or \emph{state-value} function is defined as $V^\pi(s_t)=\E_\pi[R_t|s_t]$.
The value $A^\pi(s_t,a_t)=Q^\pi(s_t,a_t)-V^\pi(s_t)$ is called
the \emph{advantage} and tells whether the action $a_t$ is better or worse than an average
action the policy $\pi$ takes in the state~$s_t$.

In practice, the policies and value functions are going to be represented as neural networks.
In particular, RL algorithms we consider maintain two neural networks: one representing the current policy $\pi$
and a value network which approximates the value function of the current policy $V \approx V^\pi$.

%% file: choices.tex
\section{List of Investigated Choices}\label{sec:choices}

In this section we list all algorithmic choices which we consider in our experiments.
See Sec.~\ref{sec:rl} for a very brief introduction to RL and the notation we use
in this section.

\subsection{Data collection and optimization loop}\label{sec:choices-setup}
RL algorithms interleave running the current policy in the environment with policy and value function networks optimization.
In particular, we create \dchoicet{numenvs} environments  \cite{a3c}.
In each iteration, we run all environments synchronously sampling actions from the current policy
until we have gathered \dchoicet{stepsize} transitions total
(this means that we have \choicet{numenvs} experience fragments, each consisting of  $\choicet{stepsize}~/~\choicet{numenvs}$ transitions).
Then, we perform \dchoicet{numepochsperstep} epochs of minibatch updates where in each epoch
we split the data into minibatches of size \dchoicet{batchsize},
and performing gradient-based optimization \cite{ppo}.
Going over collected experience multiple times means that it is not strictly an on-policy RL algorithm but
it may increase the sample complexity of the algorithm at the cost of more computationally expensive optimization step.

We consider four different variants of the above scheme (choice \dchoicep{batchhandling}):
\begin{itemize}
    \item \texttt{Fixed trajectories}: Each minibatch consists of full experience fragments
    and in each epoch we go over exactly the same minibatches in the same order.
    \item \texttt{Shuffle trajectories}: Like \texttt{Fixed trajectories} but we randomly assign full experience fragments to minibatches in each epoch.
    \item \texttt{Shuffle transitions}: We break experience fragments into individual transitions and assign them randomly to minibatches in each epoch.
    This makes the estimation of advantages impossible in each minibatch (most of the advantage estimators use future states, See App.~\ref{sec:choices-adv}) so we precompute all
    advantages at the beginning of \emph{each iteration} using full experience fragments.
    This approach leads to higher diversity of data in each minibatch at the price of somewhat stale advantage estimations.
    The original PPO implementation from OpenAI Baselines\footnote{\url{https://github.com/openai/baselines/tree/master/baselines/ppo2}} works this way
    but this is not mentioned in the PPO paper \cite{ppo}.
    \item \texttt{Shuffle transitions (recompute advantages)}: Like \texttt{Shuffle transitions} but we recompute advantages at the beginning of \emph{each epoch}.
\end{itemize}

\subsection{Advantage estimation}\label{sec:choices-adv}
Let $V$ be an approximator of the value function of some policy, i.e. $V \approx V^\pi$.
We experimented with the three most commonly used advantage estimators in on-policy RL (choice \dchoicep{advantageestimator}):
\begin{itemize}
\item \textbf{N-step} return \cite{sutton1998introduction} is defined as $$\hat{V}_t^{(N)}=\sum_{i=t}^{t+N-1} \gamma^{i-t} r_i + \gamma^{N} V(s_{t+N}) \approx V^\pi(s_t).$$
The parameter $N$ (choice \dchoicep{nstep}) controls the bias--variance tradeoff of the estimator
with bigger values resulting in an estimator closer to empirical returns and having less bias and more variance.
Given N-step returns we can estimate advantages as follows: 
$$\hat{A}_t^\text{(N)} = \hat{V}_t^\text{(N)} - V(s_t) \approx A^\pi(s_t,a_t).$$

\item \textbf{Generalized Advantage Estimator, GAE($\lambda$)} \cite{gae}
is a method that combines multi-step returns in the following way:
$$\hat{V}_t^\text{GAE} = (1-\lambda) \sum_{N>0}\lambda^{N-1} \hat{V}_t^{(N)} \approx V^\pi(s_t),$$
where $0<\lambda<1$ is a hyperparameter (choice \dchoicep{gaelambda}) controlling the bias--variance trade-off.
Using this, we can estimate advantages with:
$$\hat{A}_t^\text{GAE} = \hat{V}_t^\text{GAE} - V(s_t) \approx A^\pi(s_t,a_t).$$
It is possible
to compute the values of this estimator for all states
encountered in an episode in linear time~\cite{gae}.

\item \textbf{V-trace($\lambda,\bar{c},\bar{\rho}$)} \cite{impala} is an extension of GAE which introduces
truncated importance sampling weights to account for the fact that the current
policy might be slightly different from the policy which generated the experience.
It is parameterized by $\lambda$ (choice \dchoicep{vtraceaelambda}) which serves the same role as in GAE and two parameters 
$\bar{c}$ and $\bar{\rho}$ which are truncation thresholds for two different types
of importance weights. 
See \cite{impala} for the detailed description of the V-trace estimator.
All experiments in the original paper \cite{impala} use $\bar{c}=\bar{\rho}=1$.
Similarly, we only consider the case $\bar{c}=\bar{\rho}$, i.e., we consider a single choice \dchoicet{vtraceaecrho}.
\end{itemize}

The value network is trained by fitting one of the returns
described above with an MSE (quadratic) or a Huber \cite{huber}
loss (choice \dchoicep{valueloss}).
Huber loss is a quadratic around zero up to some threshold (choice \dchoicep{huberdelta})
at which point it becomes a linear function.

The original PPO implementation \cite{ppo} uses an additional pessimistic clipping
in the value loss function. See \cite{implementation_matters}
for the description of this technique.
It is parameterized by a clipping threshold (choice \dchoicep{ppovalueclip}).

\subsection{Policy losses}\label{sec:choices-losses}

\newcommand{\pimu}{\frac{\pi(a_t|s_t)}{\mu(a_t|s_t)}}

Let $\pi$ denote the policy being optimized, and $\mu$ the behavioral policy, i.e. the policy which generated the experience.
Moreover, let $\hat{A}_t^\pi$ and $\hat{A_t^\mu}$ be some estimators of the advantage at timestep $t$
for the policies $\pi$ and $\mu$.

We consider optimizing the policy with the following policy losses (choice \dchoicep{policyloss}):
\begin{itemize}
    \item \textbf{Policy Gradients (PG)} \cite{pg} with advantages: $\L_\texttt{PG} = -\log \pi(a_t|s_t) \hat{A}_t^\pi$.
    It can be shown that if $\hat{A}_t^\pi$ estimators are unbiased, then $\grad \L_\texttt{PG}$ is an unbiased estimator of the gradient of the policy performance assuming that experience was generated by the current policy $\pi$.
    
    \item \textbf{V-trace \cite{impala}}: $\L_\texttt{V-trace}^{\bar{\rho}} = \sg(\rho_t) \L_\texttt{PG},$ where
    $\rho_t=\min(\frac{\pi(a_t|s_t)}{\mu(a_t|s_t)}, \bar{\rho})$ is a truncated importance weight, $\sg$
    is the \texttt{stop\_gradient} operator\footnote{Identity function with gradient zero.} and $\bar{\rho}$ is a hyperparameter (choice \dchoicep{vtracelossrho}).
    $\grad L_\texttt{V-trace}^{\bar{\rho}}$ is an unbiased estimator of the gradient of the policy performance if $\bar{\rho} = \infty$
    regardless of the behavioural policy$\mu$\footnote{
    Assuming that advantage estimators are unbiased and $\mu(a_t|s_t) > 0$ for all pairs $(s_t,a_t)$ for which $\pi(a_t|s_t) > 0$.}.
    
    \item \textbf{Proximal Policy Optimization (PPO)} \cite{ppo}: $$\L^\epsilon_{\text{PPO}}=- \min \left[ \frac{\pi(a_t|s_t)}{\mu(a_t|s_t)} \hat{A}_t^\pi,\, \mbox{clip}\left(\frac{\pi(a_t|s_t)}{\mu(a_t|s_t)},\,\frac{1}{1+\epsilon},\,1+\epsilon \right)\hat{A}_t^\pi\right],$$ where
$\epsilon$ is a hyperparameter\footnote{
The original PPO paper used $1-\epsilon$ instead $1/(1+\epsilon)$ as the lower bound for the clipping.
Both variants are used in practice and we have decided to use $1/(1+\epsilon)$ as it is more symmetric.} \dchoicep{ppoepsilon}.
This loss encourages the policy to take actions which are better than average (have positive advantage)
while clipping discourages bigger changes to the policy by limiting how much can be gained
by changing the policy on a particular data point. 

\item \textbf{Advantage-Weighted Regression (AWR)} \cite{awr}: 
$$\L_\texttt{AWR}^{\beta,\,\omega_\texttt{max}} = -\log \pi(a_t|s_t) \min\left(\exp(A_t^\mu/\beta),\, \omega_\texttt{max}\right).$$
It can be shown that for $\omega_\texttt{max}=\infty$ (choice \dchoicep{awrw}) it corresponds to an approximate optimization of the policy $\pi$
under a constraint of the form $\kl(\pi||\mu) < \epsilon$ where the KL bound $\epsilon$ depends on the exponentiation temperature $\beta$ (choice \dchoicep{awrbeta}).
Notice that in contrast to previous policy losses, AWR uses estimates of the advantages for the behavioral policy ($A_t^\mu$) and not the current one ($A_t^\pi$).
AWR was proposed mostly as an off-policy RL algorithm.

\item \textbf{On-Policy Maximum a Posteriori Policy Optimization (V-MPO)} \cite{vmpo}: 
This policy loss is the same as AWR with the following differences:
(1) exponentiation is replaced with the \texttt{softmax} operator and there is no clipping with $\omega_\texttt{max}$;
(2) only samples with the top half advantages in each batch are used;
(3) the temperature $\beta$ is treated as a Lagrange multiplier and adjusted automatically to keep a constraint
on how much the weights (i.e. softmax outputs) diverge from a uniform distribution
with the constraint threshold $\epsilon$ being a hyperparameter (choice \dchoicep{vmpoeps}).
(4) A soft constraint on $\kl(\mu||\pi)$ is added. In our experiments, we did not treat this constraint
as a part of the V-MPO policy loss as policy regularization is considered separately (See Sec.~\ref{sec:choices-reg}). 

\item \textbf{Repeat Positive Advantages (RPA)}: $\L_\texttt{RPA} = -\log \pi(a_t|s_t) [A_t > 0]$\footnote{
$[P]$ denotes the Iverson bracket, i.e. $[P]=1$ if $P$ is true and $[P]=0$ otherwise.}.
This is a new loss we introduce in this paper.
We choose this loss because it is the limiting case of AWR and V-MPO.
In particular,
$\L_\texttt{AWR}^{\beta,\,\omega_\texttt{max}}$ converges to $\omega_\texttt{max} \L_\texttt{RPA}$
for $\beta \rightarrow 0$ and 
for $\epsilon \rightarrow 0$ V-MPO converges to RPA with $[A_t > 0]$ replaced by only taking
the top half advantages in each batch\footnote{
For $\epsilon \rightarrow 0$, we have $\beta \rightarrow \infty$ which results in $\texttt{softmax}(A_t^\mu/\beta) \rightarrow 1$.
} (the two conditions become even more similar if advantage normalization is used, See Sec.~\ref{sec:choices-norm}).
\end{itemize}

\subsection{Handling of timesteps}
\label{sec:choices-time}
The most important hyperparameter controlling how timesteps are handled
is the discount factor $\gamma$ (choice \dchoicep{discount}).
Moreover, we consider the so-called \emph{frame skip}\footnote{
While not too common in continuous control, this technique is standard in
RL for Atari \cite{dqn}.}  (choice \dchoicep{frameskip}).
Frame skip equal to $n$ means that we modify the environment
by repeating each action outputted by the policy $n$ times (unless the episode has terminated in the meantime) and sum the received rewards.
When using frame skip, we also adjust the discount factor appropriately, i.e.
we discount with $\gamma^n$ instead of $\gamma$.

Many reinforcement learning environments (including the ones we
use in our experiments) have \emph{step limits} which means that
an episode is terminated after some fixed number of steps
(assuming it was not terminated earlier for some other reason).
Moreover, the number of remaining environment steps
is not included in policy observations which
makes the environments non-Markovian and can potentially
make learning harder \cite{pardo2017time}.
We consider two ways to handle such \emph{abandoned} episodes.
We either treat the final transition as any other terminal transition, e.g.
the value target for the last state is equal to the final reward, or
we take the fact that we do not know what would happen if the episode was not terminated
into account. In the latter case, we set the advantage for the final state to zero
and its value target to the current value function.
This also influences
the value targets for prior states
as the value targets are computed recursively \cite{gae, impala}.
We denote this choice by \dchoicet{handleabandon}.

\subsection{Optimizers}\label{sec:choices-opt}
We experiment with two most commonly
used gradient-based optimizers in RL (choice \dchoicep{optimizer}): Adam \cite{adam}
and RMSProp.\footnote{RMSProp was proposed by Geoffrey Hinton in one of his Coursera lectures: \url{http://www.cs.toronto.edu/~tijmen/csc321/slides/lecture_slides_lec6.pdf}}
You can find the description of the optimizers and their hyperparameters in the original publications.
For both optimizers, we sweep the learning rate (choices \dchoicep{adamlr} and \dchoicep{rmslr}), momentum (choices \dchoicep{adammom} and \dchoicep{rmsmom}) and
the $\epsilon$ parameters added for numerical stability (choice \dchoicep{adameps} and \dchoicep{rmseps}).
Moreover, for RMSProp we consider both centered and uncentered versions (choice \dchoicep{rmscent}).
For their remaining hyperparameters, we use the default values
from TensorFlow \cite{tf}, i.e. $\beta_2=0.999$ for Adam and $\rho=0.1$
for RMSProp.
Finally, we allow a linear learning rate schedule via the hyperparameter \dchoicet{lrdecay} which defines the terminal learning rate as a fraction of the initial learning rate (i.e., $0.0$ correspond to a decay to zero whereas $1.0$ corresponds to no decay).

\subsection{Policy regularization}\label{sec:choices-reg}
We consider three different modes for regularization (choice \dchoicep{regularizationtype}):
\begin{itemize}
    \item \texttt{No regularization}: We apply no regularization.
    \item \texttt{Penalty}: we apply a regularizer $R$ with fixed strength, i.e., we add to the loss the term $\alpha R$ for some fixed coefficient $\alpha$ which is a hyperparameter.
    \item \texttt{Constraint}: We impose a soft constraint of the form $R < \epsilon$ on the value of the regularizer where $\epsilon$ is a hyperparameter.
This can be achieved by treating $\alpha$ as a Lagrange multiplier.
In this case, we optimize the value of $\alpha > 0$ together with the networks parameter by adding to the loss the term $\alpha \cdot \sg(\epsilon-R)$, where $\sg$ in the \texttt{stop\_gradient} operator.
In practice, we use the following parametrization $\alpha=\exp(c \cdot p)$ where $p$ is a trainable parameter (initialized with $0$) and $c$ is a coefficient controlling how fast $\alpha$ is adjusted (we use $c=10$ in all experiments).
After each gradient step, we clip the value of $p$ to the range $[\log(10^{-6})/10,\,\log(10^6)/10]$ to avoid extremely small and large values.
\end{itemize}

We consider a number of different policy regularizers both for penalty (choice \dchoicep{regularizerpenalty}) and constraint regularization (choice \dchoicep{regularizerconstraint}):
\begin{itemize}
    \item Entropy $\ent(\pi(\cdot|s))$  ---  it encourages the policy to try diverse actions \cite{maxent}.
    \item $\kl(\mu(\cdot|s)||\pi(\cdot|s))$ --- the Kullback–Leibler divergence between the behavioral
    and the current policy \cite{trpo} prevents the probability of taking a given action from \emph{decreasing} too rapidly.
    \item $\kl(\pi(\cdot|s)||\mu(\cdot|s))$ ---  similar to the previous one but prevents too rapid \emph{increase} of probabilities.
    \item $\kl(\texttt{ref}(\cdot|s)||\pi(\cdot|s))$ where $\texttt{ref}$ is some reference distribution. We use $\texttt{ref}=\N(0,1)$ in all experiments.
    This kind of regularization encourages the policy to try all possible actions.
    \item Decoupled $\kl(\mu(\cdot|s)||\pi(\cdot|s))$. For Gaussian distributions we can split $\kl(\mu(\cdot|s)||\pi(\cdot|s))$ into a term which depends on the
    change in the mean of the distribution and another one which depends on the change in the standard deviation:
    $\kl(\mu(\cdot|s)||\pi(\cdot|s)) = \kl(\mu(\cdot|s)||\zeta(\cdot|s)) + \kl(\zeta(\cdot|s)||\pi(\cdot|s))$ where $\zeta(\cdot|s)$
    is a Gaussian distribution with same mean as $\mu(\cdot|s)$ and the same standard deviation as $\pi(\cdot|s)$.
    Therefore, instead of using $\kl(\mu(\cdot|s)||\pi(\cdot|s))$ directly, we can use two separate regularizers, $\kl(\mu(\cdot|s)||\zeta(\cdot|s))$ and $\kl(\zeta(\cdot|s)||\pi(\cdot|s))$,
    with different strengths. The soft constraint version of this regularizer in used in V-MPO\footnote{
    The current arXiv version of the V-MPO paper \cite{vmpo} incorrectly uses the standard deviation of the old policy instead of the new one in the definition of $\kl(\mu(\cdot|s)||\pi(\cdot|s))$
    which leads to a slightly different decomposition. We do not expect this to make any difference in practice.} \cite{vmpo} with the threshold on $\kl(\mu(\cdot|s)||\zeta(\cdot|s))$
    being orders of magnitude lower than the one on $\kl(\zeta(\cdot|s)||\pi(\cdot|s))$.
\end{itemize}
While one could add any linear combination of the above terms to the loss, we have decided to only use a single regularizer in each experiment.
Overall, all these combinations of regularization modes and different hyperparameters lead to the choices detailed in Table~\ref{table:regularization-hps}.

\begin{table}[h]
  \caption{Choices pertaining to regularization.}
  \label{table:regularization-hps}
  \centering
  \begin{tabular}{cl}
    \toprule
    Choice & Name  \\
    \midrule
    \choicetable{regularizationtype} \\
    \choicetable{regularizerpenalty} \\
    \choicetable{regularizerconstraint} \\
\dchoicetable{regularizerconstraintklmupi}\\
\dchoicetable{regularizerconstraintklpimu}\\
\dchoicetable{regularizerconstraintklrefpi}\\
\dchoicetable{regularizerconstraintklmupimean}\\
\dchoicetable{regularizerconstraintklmupistd}\\
\dchoicetable{regularizerconstraintentropy}\\
\dchoicetable{regularizerpenaltyklmupi}\\
\dchoicetable{regularizerpenaltyklpimu}\\
\dchoicetable{regularizerpenaltyklrefpi}\\
\dchoicetable{regularizerpenaltyklmupimean}\\
\dchoicetable{regularizerpenaltyklmupistd}\\
\dchoicetable{regularizerpenaltyentropy}\\
    \bottomrule
  \end{tabular}
\end{table}

\subsection{Neural network architecture}

We use multilayer perceptrons (MLPs) to represent policies and value functions.
We either use separate networks for the policy and value function, or use a single network with two linear heads, one for the policy and one for the value function (choice \dchoicep{mlpshared}).
We consider different widths for the shared MLP (choice \dchoicep{sharedwidth}), the policy MLP (choice \dchoicep{policywidth}) and the value MLP (choice \dchoicep{valuewidth}) as well as different depths for the shared MLP (choice \dchoicep{shareddepth}), the policy MLP (choice \dchoicep{policydepth}) and the value MLP (choice \dchoicep{valuedepth}).
If we use the shared MLP, we further add a hyperparameter \dchoicet{baselinecost} that rescales the contribution of the value loss to the full objective function.
This is important in this case as the shared layers of the MLP affect the loss terms related to both the policy and the value function.
We further consider different activation functions (choice \dchoicep{activation}) and different neural network initializers (choice \dchoicep{init}).
For the initialization of both the last layer in the policy MLP / the policy head (choice \dchoicep{policyinit}) and the last layer in the value MLP / the value head (choice \dchoicep{valueinit}), we further consider a hyperparameter that rescales the network weights of these layers after initialization.

\subsection{Action distribution parameterization}\label{sec:choices-action}

A policy is a mapping from states to distributions of actions.
In practice, a parametric distribution is chosen and
the policy output is treated as the distribution parameters.
The vast majority of RL applications in continuous control use a Gaussian distribution
to represent the action distribution and this is also the approach we take.

This, however, still leaves a few decisions which need to be make in the implementation:
\begin{itemize}
    \item Should the standard deviation of actions be a part of the network output (used e.g. in \cite{sac}) or should it be independent of
    inputs like in \cite{ppo} (choice \dchoicep{stdind})? In the latter case, the standard deviation is still learnable but it is the same for each state.
    \item Gaussian distributions are parameterized with a mean and a standard deviation which has to be non-negative.
    What function should be used to transform network outputs which can be negative into the standard deviation (choice \dchoicep{stdtransform})?
    We consider exponentiation\footnote{
    For numerical stability, we clip the exponent to the range $[-15, 15]$. Notice that due to clipping this function has zero derivative outside of the range $[-15, 15]$ which is undesirable.
    Therefore, we use a ``custom gradient'' for the clipping function, namely we assume that it has derivative equal $1$
    everywhere.} (used e.g. in \cite{ppo}) and the \texttt{softplus}\footnote{$\texttt{softplus}(x) = \log(e^x+1)$} function (used e.g. in~\cite{mbpo}).
    \item What should be the initial standard deviation of the action distribution (choice \dchoicep{initialstd})?
    We can control it by adding some fixed value to the input to the function computing the standard deviation (e.g. $\texttt{softplus}$).
    \item Should we add a small value to the standard deviation to avoid very low values (choice \dchoicep{minstd})?
    \item Most continuous control environments expect actions from a bounded range (often $[-1,\, 1]$) but the commonly used Gaussian distribution can produce values of an arbitrary magnitude. We consider two approaches to handle this (choice \dchoicep{actionpost}):
    The easiest solution is to just clip the action to the allowed range
    when sending it to the environment (used e.g. in \cite{ppo}).
    Another approach is to apply the \texttt{tanh} function to the distribution to bound the range of actions (used e.g. in \cite{openai2018learning}).
    This additional transformation changes the density of actions --- if action $u$ is parameterized as
    $u = \texttt{tanh}(x)$, where $x$ is a sample from a Gaussian distribution with probability density function $p_\theta$,
    than the density of $u$ is $\log p_u(u)=\log p_\theta(x) - \log \tanh'(x)$, where $x = \tanh^{-1}(u)$.
    This additional $\log \tanh'(x)$ term does not affect policy losses because they only use
    $\nabla_\theta \log p_u(u) = \nabla_\theta \log p_\theta(x)$.
    Similarly, this term does not affect the KL divergences which may be used for regularization (See Sec.~\ref{sec:choices-reg})
    because the KL divergence has a form of the difference of two log-probabilities on the same sample and the two $\log \tanh'(x)$
    terms cancel out.\footnote{$\kl(U_1,U_2) = \E_{u \leftarrow U_1} \log U_1(u) - \log U_2(u) = \E_{x \leftarrow X_1} (\log X_1(x) - \log \tanh'(x)) - (\log X_2(x) - \log \tanh'(x)) =  \E_{x \leftarrow X_1} \log X_1(x) - \log X_2(x) = \kl(X_1||X_2)$.}
    The only place where the $\log \tanh'(x)$ term affects the policy gradient computation and should be included
    is the entropy regularization as
    $H(U) = -\E_{u} \log p_u(u) = \E_{x} [-\log p_\theta(x) + \log \tanh'(x)]$.
    This additional $\log \tanh'(x)$ term penalizies the policy for taking extreme actions which prevents
    $\tanh$ saturation and the loss of the gradient. Moreover, it prevents the action entropy from becoming unbounded.
\end{itemize}

To sum up, we parameterize the actions distribution as
$$T_u(\N(x_\mu,\, T_\rho(x_\rho + c_\rho) + \epsilon_\rho)),$$
where
\begin{itemize}
    \item $x_\mu$ is a part of the policy network output,
    \item $x_\rho$ is either a part of the policy network output or a
    separate learnable parameter (one per action dimension),
    \item $\epsilon_\rho$ (\choicep{minstd}) is a hyperparameter controlling minimal standard deviation,
    \item $T_\rho$ (\choicep{stdtransform}) is a standard deviation transformation ($\R \rightarrow \R_{\ge 0}$),
    \item $T_u$ (\choicep{actionpost}) is an action transformation ($\R \rightarrow [-1,\,1]$),
    \item $c_\rho$ is a constant controlling the initial standard deviation
    and computed as $c_\rho = T_\rho^{-1}(i_\rho - \epsilon_\rho)$ where
    $i_\rho$ is the desired initial standard deviation (\choicep{initialstd}).
\end{itemize}

\subsection{Data normalization and clipping}\label{sec:choices-norm}

While it is not always mentioned in RL publications, many RL
implementations perform different types of data normalization.
In particular, we consider the following:
\begin{itemize}
    \item Observation normalization (choice \dchoicep{norminput}). If enabled, we keep the empirical mean $o_\mu$ and standard deviation $o_\rho$
    of each observation coordinate (based on all observations seen so far)
    and normalize observations by subtracting the empirical mean and dividing by
    $\max(o_\rho,\,10^{-6})$. This results in all neural networks inputs having
    approximately zero mean and standard deviation equal to one.
    Moreover, we optionally clip the normalized observations to the range
    $[-o_\texttt{max},\, o_\texttt{max}]$ where $o_\texttt{max}$ is a hyperparameter
    (choice \dchoicep{clipinput}).

    \item Value function normalization (choice \dchoicep{normreward}).
    Similarly to observations, we also maintain the empirical mean $v_\mu$ and standard deviation $v_\rho$ of value function targets (See Sec.~\ref{sec:choices-adv}).
    The value function network predicts normalized targets
    $(\hat{V} - v_\mu) / \max(v_\rho,\, 10^{-6})$
    and its outputs are denormalized accordingly to obtain predicted values:
    $\hat{V} = v_\mu + V_\texttt{out} \max(v_\rho,\, 10^{-6})$
    where $V_\texttt{out}$ is the value network output.
    
    \item Per minibatch advantage normalization  (choice \dchoicep{normadv}).
    We normalize advantages in each minibatch by subtracting their mean and dividing by their standard deviation for the policy loss. 
    
    \item Gradient clipping (choice \dchoicep{clipgrad}). We rescale the
    gradient before feeding it to the optimizer so that its L2 norm does not
    exceed the desired threshold.
\end{itemize}

%% file: default.tex
\clearpage\section{Default settings for experiments}
\label{sec:default-settings}
\label{sec:default_settings}
Table~\ref{table:default} shows the default configuration used for all the experiments in this paper.
We only list sub-choices that are active (e.g.
we use the PPO loss so 
we do not list hyperparameters associated with different policy losses).
\begin{table}[h]
  \caption{Default settings used in experiments.}
  \label{table:default}
  \centering
  \begin{tabular}{clr}
    \toprule
    Choice & Name & Default value \\
    \midrule
    \choicetable{numenvs} & 256\\
\choicetable{stepsize} & 2048\\
\choicetable{numepochsperstep} & 10\\
\choicetable{batchsize} & 64\\
\choicetable{batchhandling} & \texttt{Shuffle transitions}\\
\choicetable{advantageestimator} & \texttt{GAE}\\
\choicetable{gaelambda} & 0.95\\
\choicetable{valueloss} & \texttt{MSE}\\
\choicetable{ppovalueclip} & 0.2\\
\choicetable{policyloss} & \texttt{PPO}\\
\choicetable{ppoepsilon} & 0.2\\
\choicetable{discount} & 0.99\\
\choicetable{frameskip} & 1\\
\choicetable{handleabandon} & \texttt{False}\\
\choicetable{optimizer} & \texttt{Adam}\\
\choicetable{adamlr} & 3e-4\\
\choicetable{adammom} & 0.9\\
\choicetable{adameps} & 1e-7\\
\choicetable{lrdecay} & \texttt{0.0}\\
\choicetable{regularizationtype} & \texttt{None}\\
\choicetable{mlpshared} & \texttt{Shared}\\
\choicetable{policywidth} & 64\\
\choicetable{valuewidth} & 64\\
\choicetable{policydepth} & 2\\
\choicetable{valuedepth} & 2\\
\choicetable{activation} & \texttt{tanh}\\
\choicetable{init} & \texttt{Orthogonal with gain 1.41}\\
\choicetable{policyinit} & 0.01\\
\choicetable{valueinit} & 1.0\\
\choicetable{stdind} & \texttt{True}\\
\choicetable{stdtransform} & \texttt{safe\_exp}\\
\choicetable{initialstd} & 1.0\\
\choicetable{actionpost} & \texttt{clip}\\
\choicetable{minstd} & 1e-3\\
\choicetable{norminput} & \texttt{Average}\\
\choicetable{clipinput} & 10.0\\
\choicetable{normreward} & \texttt{Average} \\
\choicetable{normadv} & \texttt{False}\\
\choicetable{clipgrad} & 0.5\\
    \bottomrule
  \end{tabular}
\end{table}

%% file: final_losses/main.tex
\clearpage
\section{Experiment \texttt{Policy Losses}}
\label{exp_final_losses}
\subsection{Design}
\label{exp_design_final_losses}
For each of the 5 environments, we sampled 2000 choice configurations where we sampled the following choices independently and uniformly from the following ranges:
\begin{itemize}
    \item \choicet{numepochsperstep}: \{1, 3, 10\}
    \item \choicet{policyloss}: \{AWR, PG, PPO, RPA, V-MPO, V-Trace\}
    \begin{itemize}
        \item For the case ``\choicet{policyloss} = AWR'', we further sampled the sub-choices:
        \begin{itemize}
            \item \choicet{awrbeta}: \{0.0001, 0.0003, 0.001, 0.003, 0.01, 0.03, 0.1, 0.3\}
            \item \choicet{awrw}: \{1.1, 1.2, 1.3, 1.5\}
        \end{itemize}
        \item For the case ``\choicet{policyloss} = PPO'', we further sampled the sub-choices:
        \begin{itemize}
            \item \choicet{ppoepsilon}: \{0.1, 0.2, 0.3, 0.5\}
        \end{itemize}
        \item For the case ``\choicet{policyloss} = V-MPO'', we further sampled the sub-choices:
        \begin{itemize}
            \item \choicet{vmpoeps}: \{0.0001, 0.0003, 0.001, 0.003, 0.01, 0.03, 0.1, 0.3, 1.0\}
        \end{itemize}
        \item For the case ``\choicet{policyloss} = V-Trace'', we further sampled the sub-choices:
        \begin{itemize}
            \item \choicet{vtracelossrho}: \{1.0, 1.2, 1.5, 2.0\}
        \end{itemize}
    \end{itemize}
    \item \choicet{adamlr}: \{3e-05, 0.0001, 0.0003, 0.001, 0.003\}
\end{itemize}
All the other choices were set to the default values as described in Appendix~\ref{sec:default_settings}.

For each of the sampled choice configurations, we train 3 agents with different random seeds and compute the performance metric as described in Section~\ref{sec:performance}.
\subsection{Results}
\label{exp_results_final_losses}
\begin{figure}[ht]
\begin{center}
\centerline{\includegraphics[width=1\textwidth]{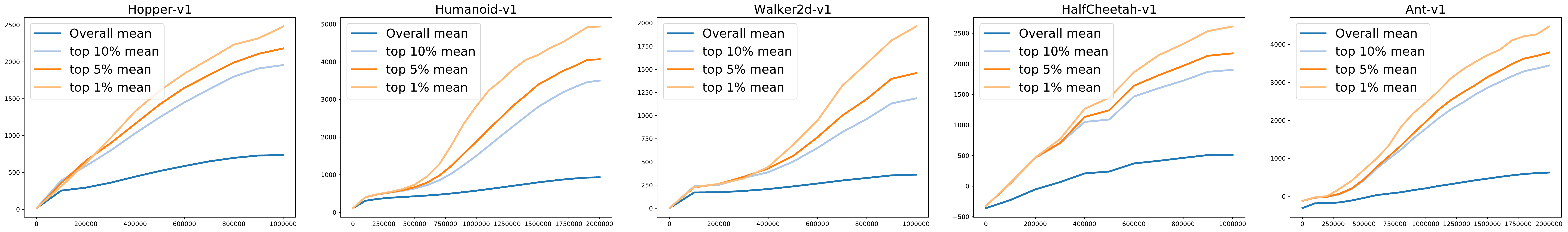}}
\caption{Training curves.}
\label{fig:final_losses_training_curves}
\end{center}
\end{figure}

\begin{figure}[ht]
\begin{center}
\centerline{\includegraphics[width=1\textwidth]{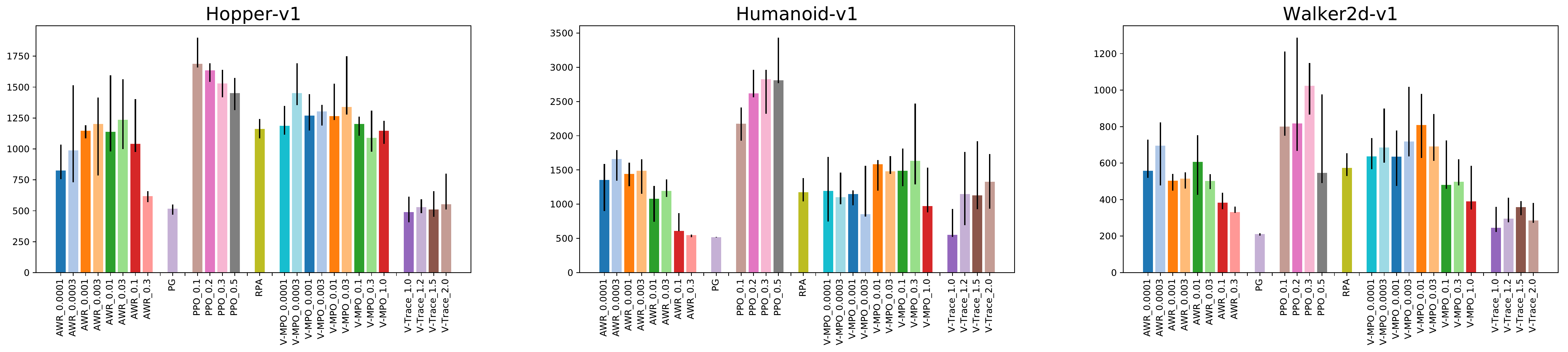}}
\centerline{\includegraphics[width=0.65\textwidth]{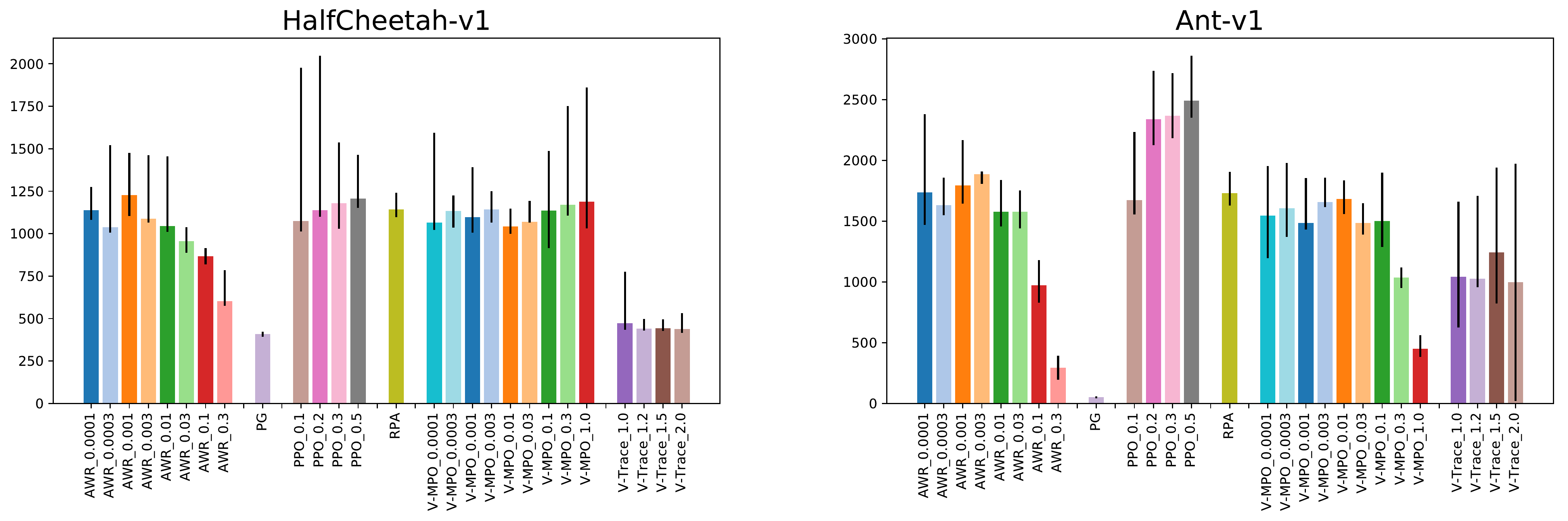}}
\caption{Comparison of 95th percentile of the performance of different policy losses conditioned on their hyperparameters.}
\label{fig:final_losses_custom_loss}
\end{center}
\end{figure}

\begin{figure}[ht]
\begin{center}
\centerline{\includegraphics[width=0.45\textwidth]{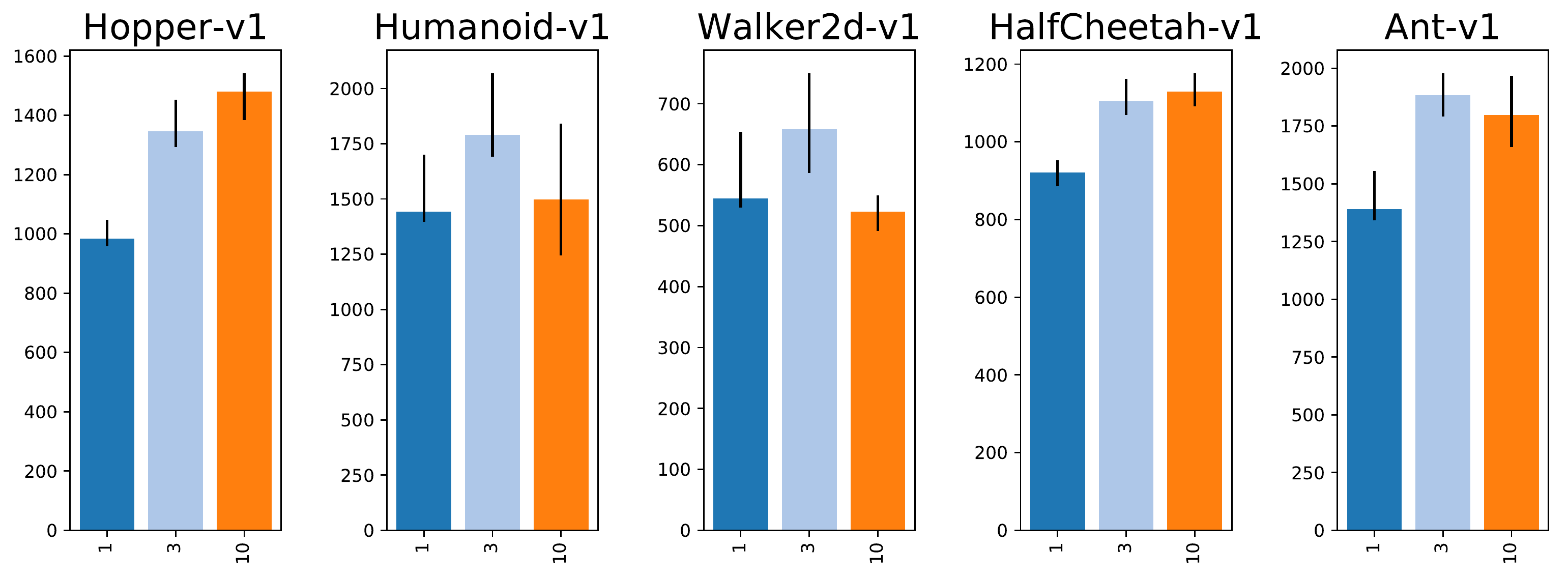}\hspace{1cm}\includegraphics[width=0.45\textwidth]{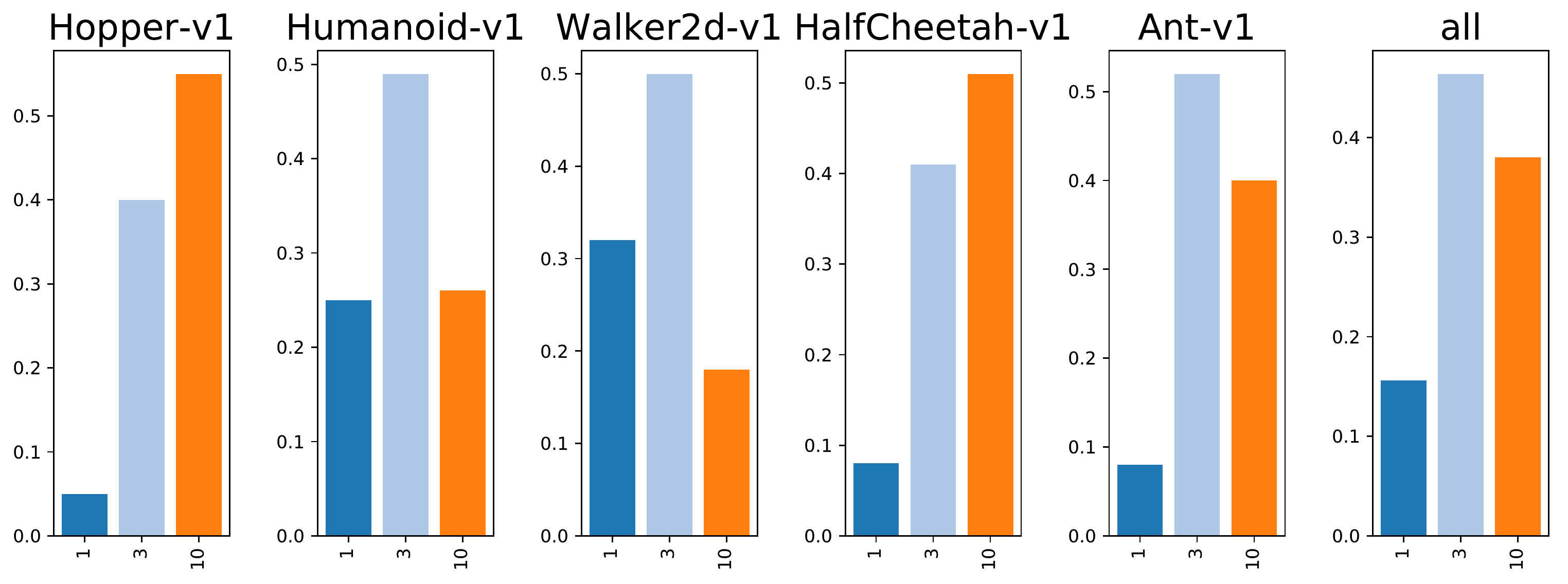}}
\caption{Analysis of choice \choicet{numepochsperstep}: 95th percentile of performance scores conditioned on choice (left) and distribution of choices in top 5\% of configurations (right).}
\label{fig:final_losses__gin_study_design_choice_value_epochs_per_step}
\end{center}
\end{figure}

\begin{figure}[ht]
\begin{center}
\centerline{\includegraphics[width=0.45\textwidth]{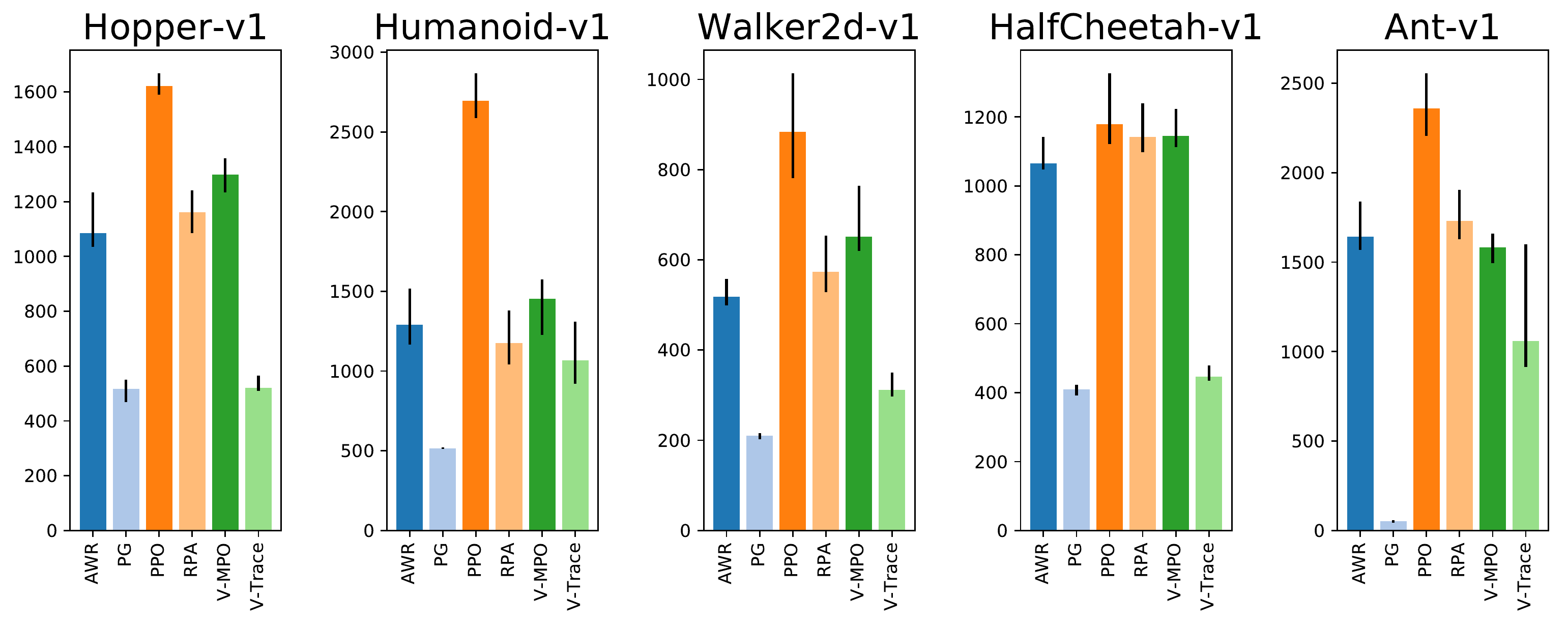}\hspace{1cm}\includegraphics[width=0.45\textwidth]{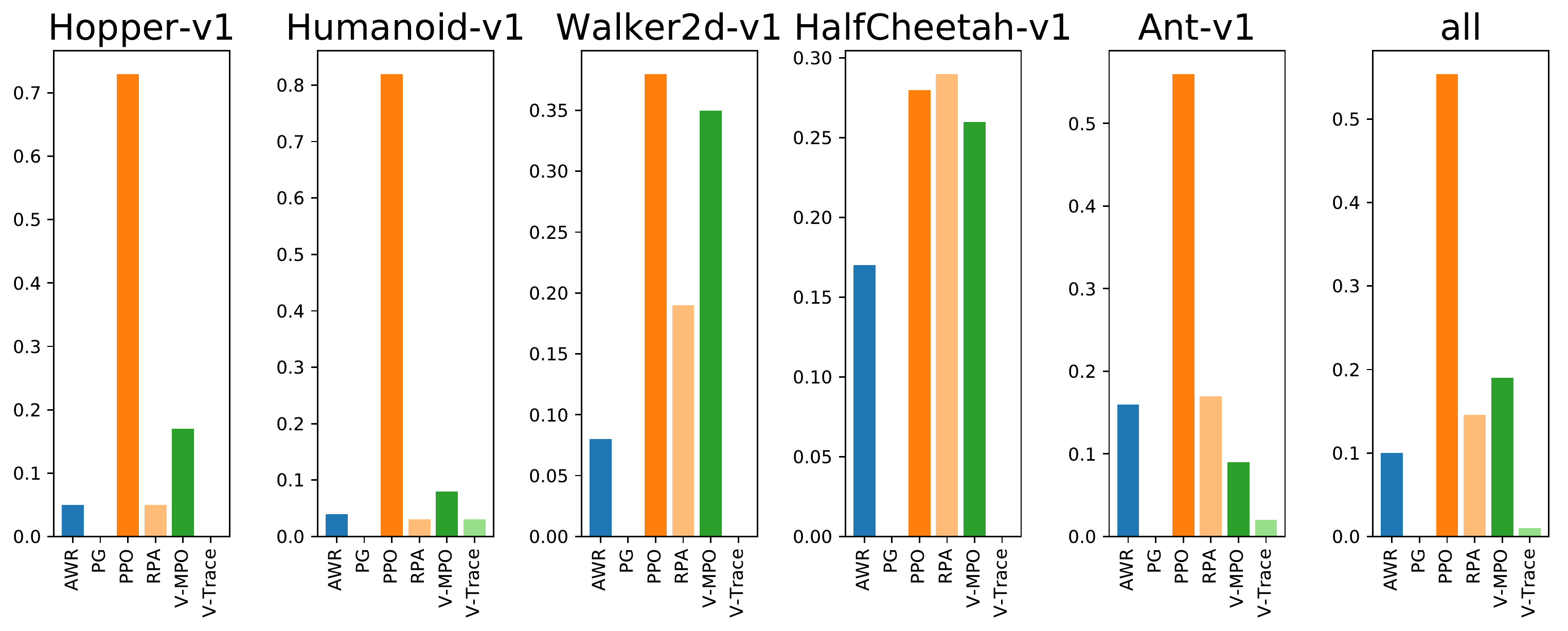}}
\caption{Analysis of choice \choicet{policyloss}: 95th percentile of performance scores conditioned on choice (left) and distribution of choices in top 5\% of configurations (right).}
\label{fig:final_losses__gin_study_design_choice_value_standard_policy_losses}
\end{center}
\end{figure}

\begin{figure}[ht]
\begin{center}
\centerline{\includegraphics[width=0.45\textwidth]{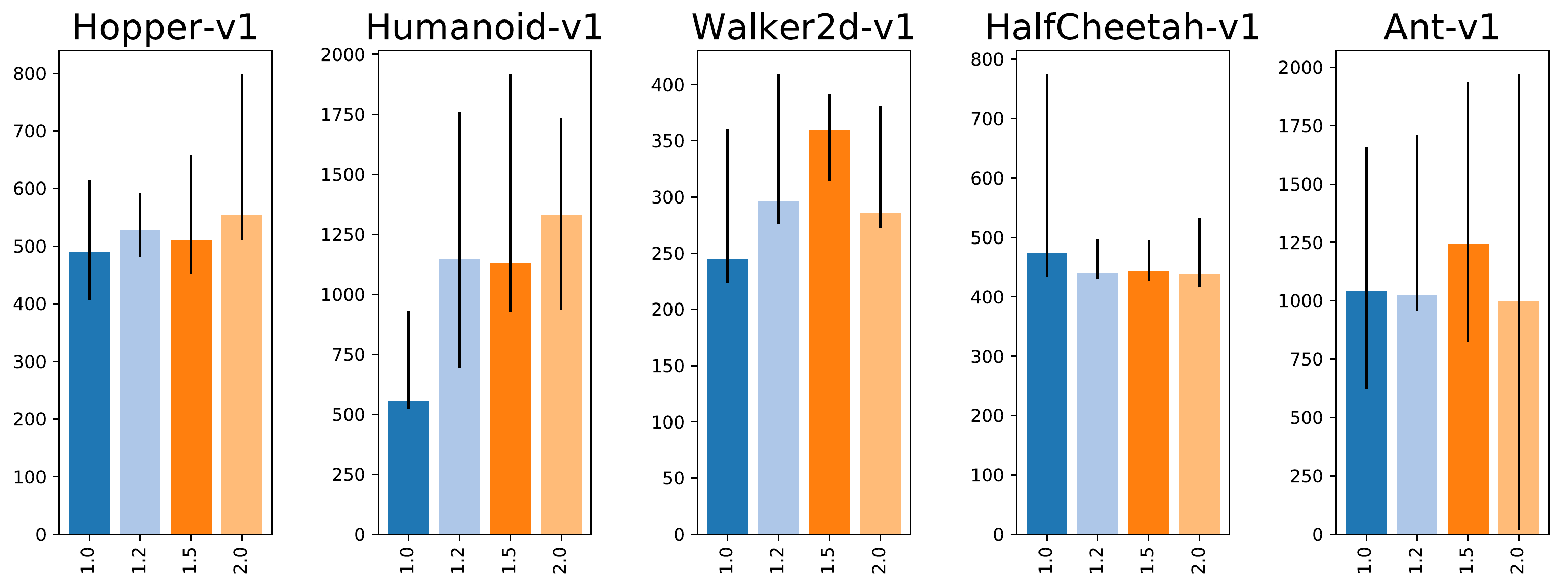}\hspace{1cm}\includegraphics[width=0.45\textwidth]{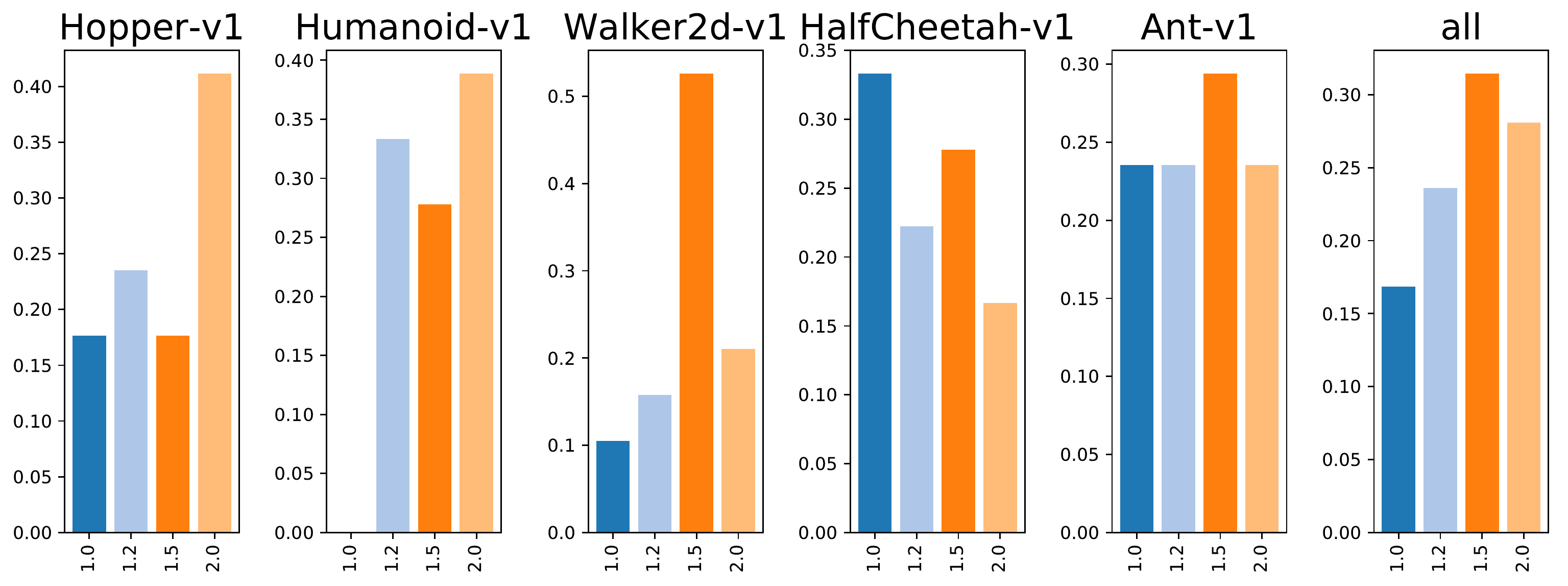}}
\caption{Analysis of choice \choicet{vtracelossrho}: 95th percentile of performance scores conditioned on sub-choice (left) and distribution of sub-choices in top 5\% of configurations (right).}
\label{fig:final_losses__gin_study_design_choice_value_sub_standard_policy_losses_v_trace_vtrace_max_importance_weight}
\end{center}
\end{figure}

\begin{figure}[ht]
\begin{center}
\centerline{\includegraphics[width=0.45\textwidth]{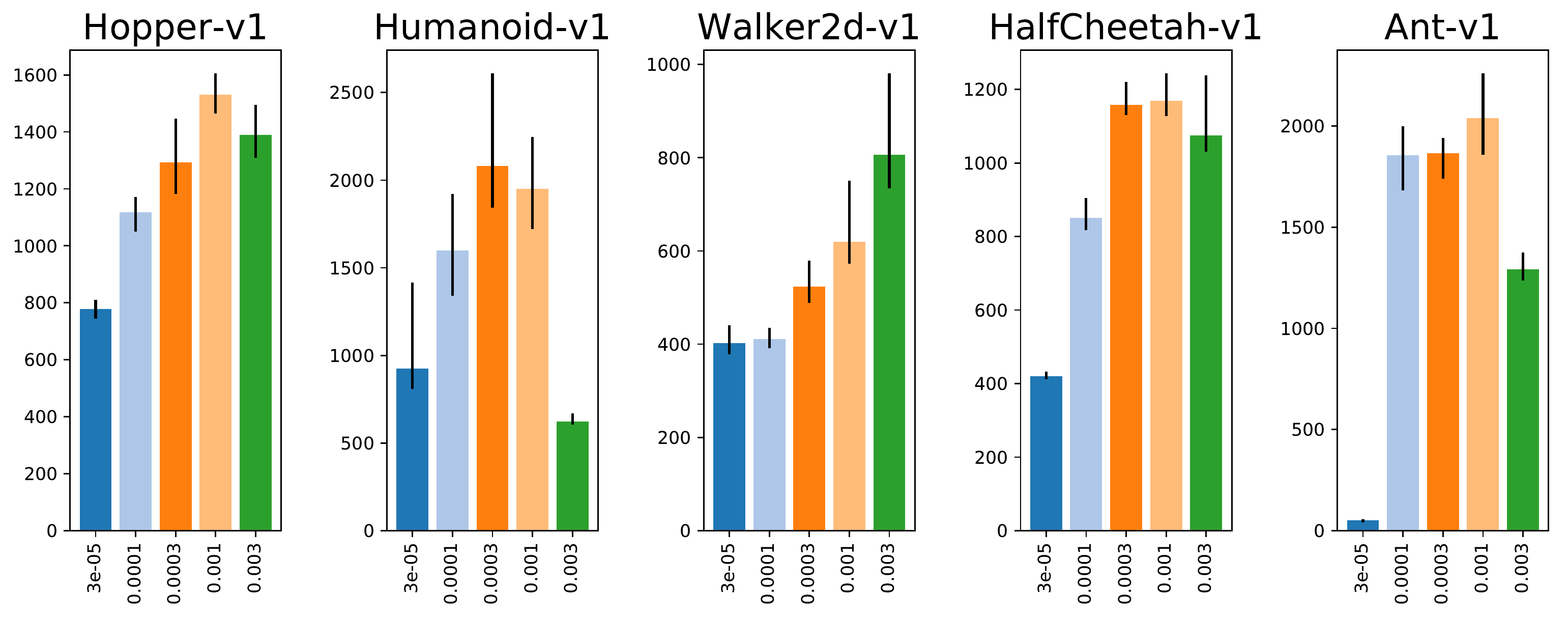}\hspace{1cm}\includegraphics[width=0.45\textwidth]{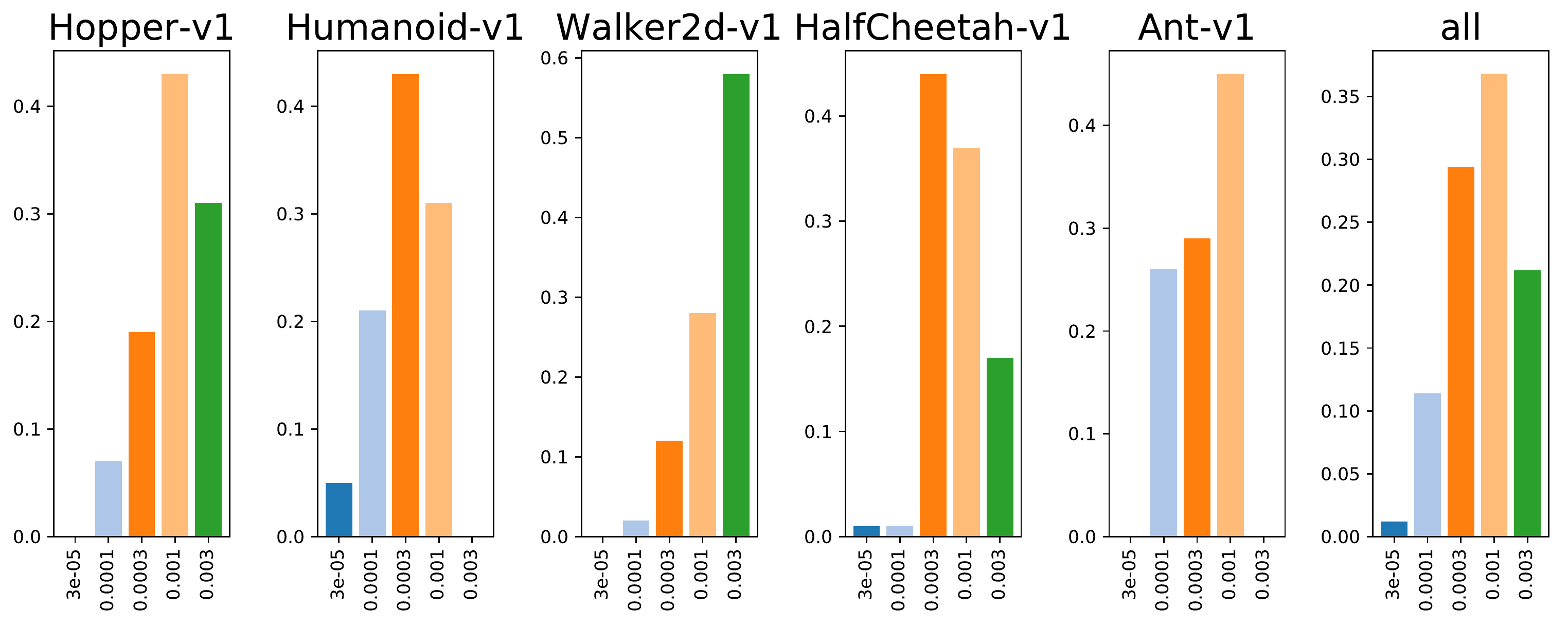}}
\caption{Analysis of choice \choicet{adamlr}: 95th percentile of performance scores conditioned on choice (left) and distribution of choices in top 5\% of configurations (right).}
\label{fig:final_losses__gin_study_design_choice_value_learning_rate}
\end{center}
\end{figure}

\begin{figure}[ht]
\begin{center}
\centerline{\includegraphics[width=0.45\textwidth]{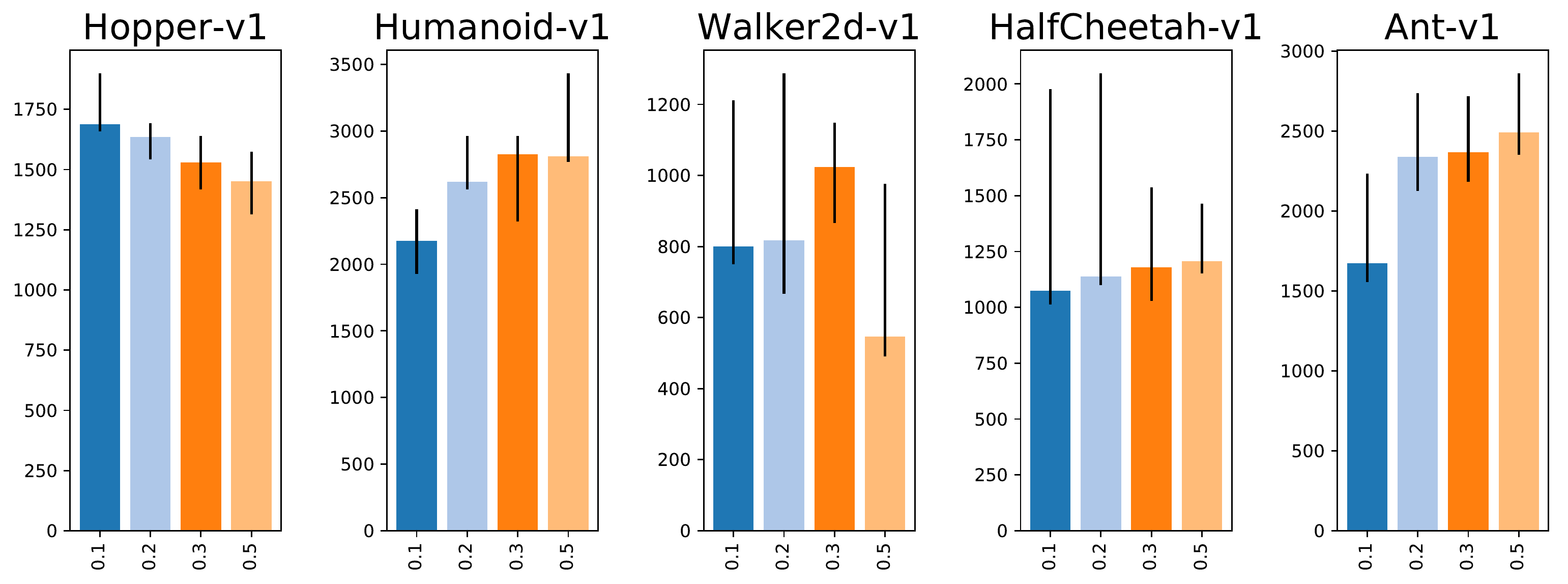}\hspace{1cm}\includegraphics[width=0.45\textwidth]{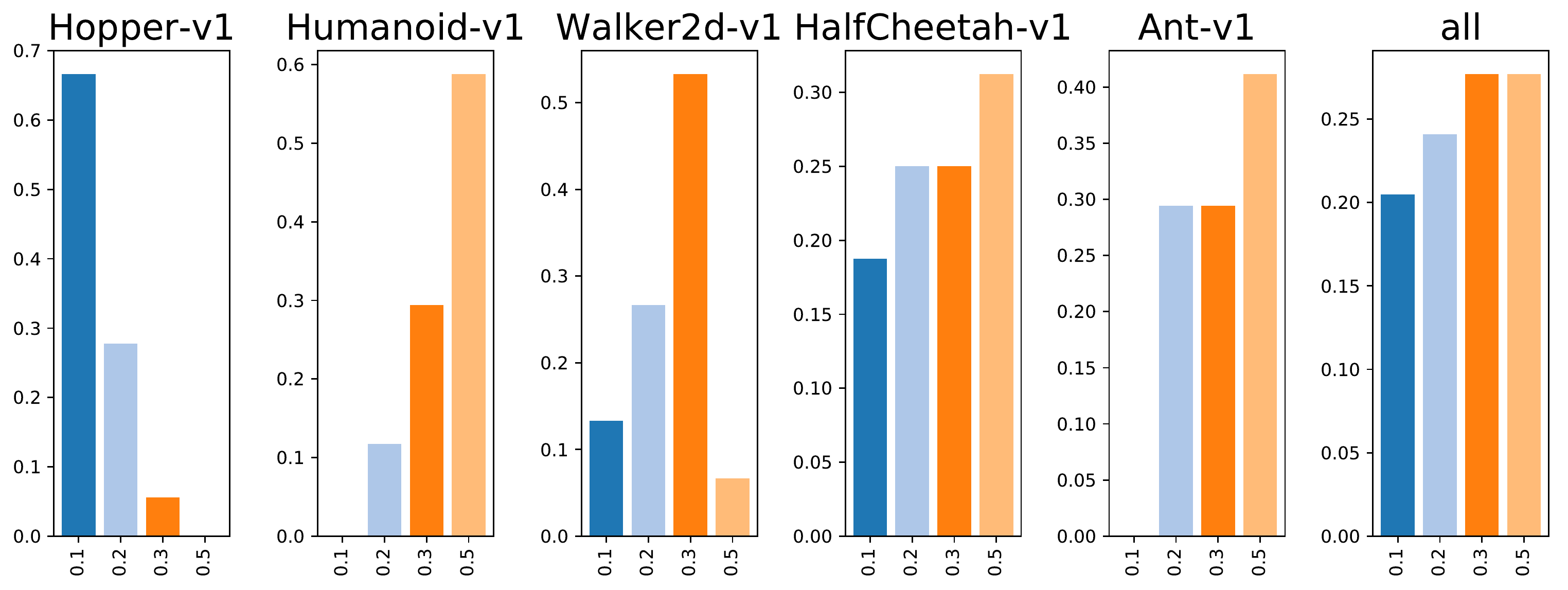}}
\caption{Analysis of choice \choicet{ppoepsilon}: 95th percentile of performance scores conditioned on sub-choice (left) and distribution of sub-choices in top 5\% of configurations (right).}
\label{fig:final_losses__gin_study_design_choice_value_sub_standard_policy_losses_ppo_ppo_epsilon}
\end{center}
\end{figure}

\begin{figure}[ht]
\begin{center}
\centerline{\includegraphics[width=0.45\textwidth]{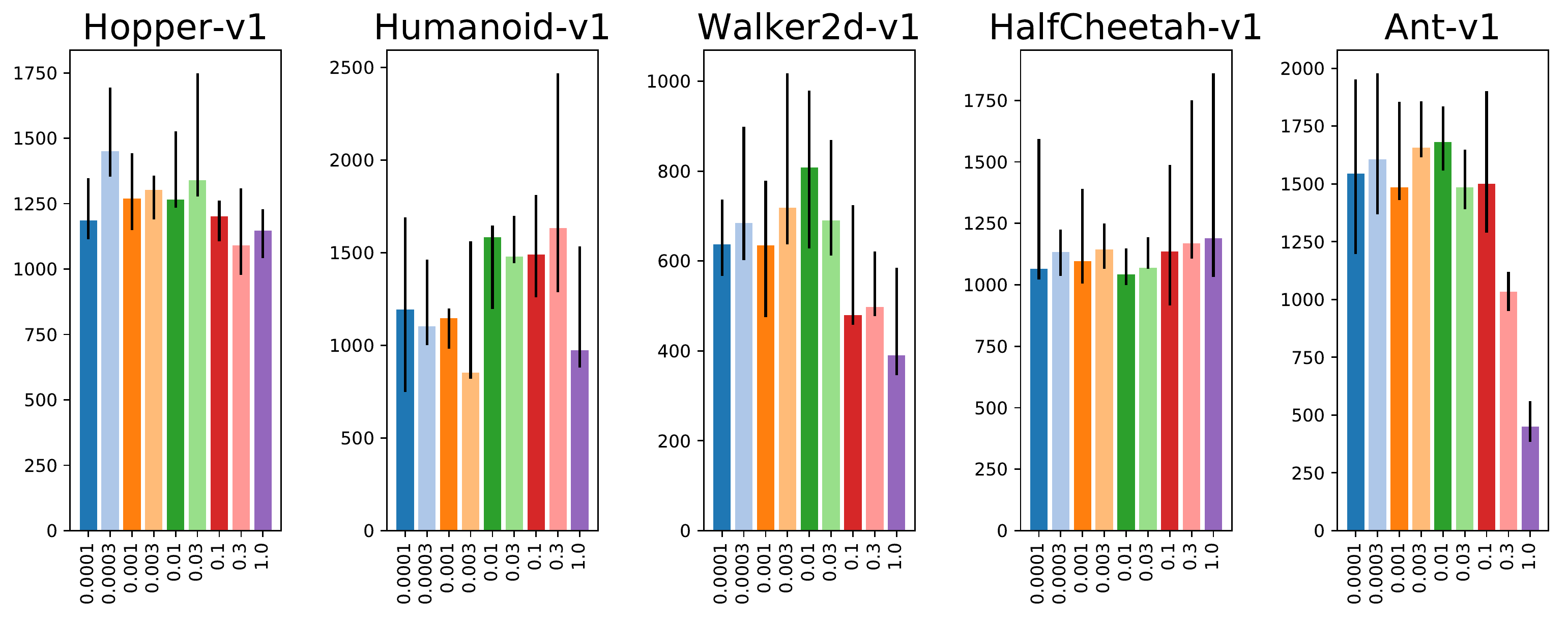}\hspace{1cm}\includegraphics[width=0.45\textwidth]{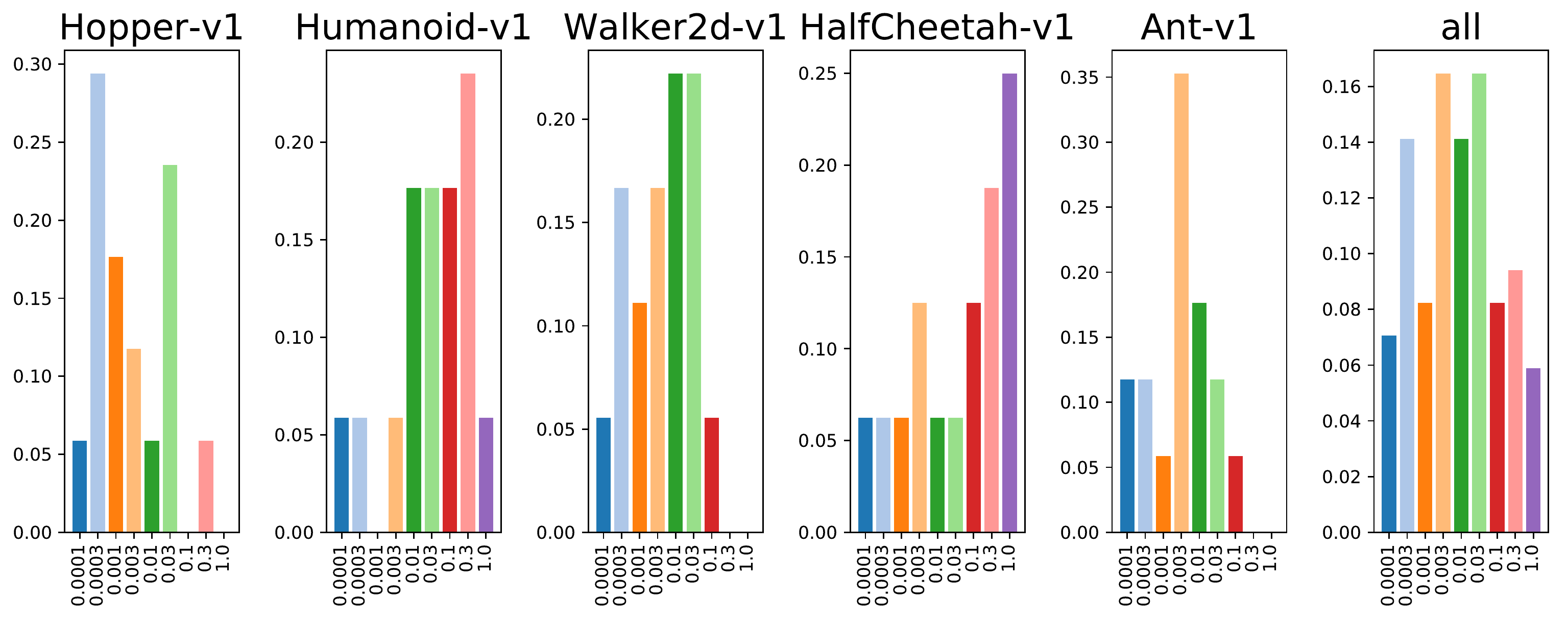}}
\caption{Analysis of choice \choicet{vmpoeps}: 95th percentile of performance scores conditioned on sub-choice (left) and distribution of sub-choices in top 5\% of configurations (right).}
\label{fig:final_losses__gin_study_design_choice_value_sub_standard_policy_losses_v_mpo_vmpo_e_n}
\end{center}
\end{figure}

\begin{figure}[ht]
\begin{center}
\centerline{\includegraphics[width=0.45\textwidth]{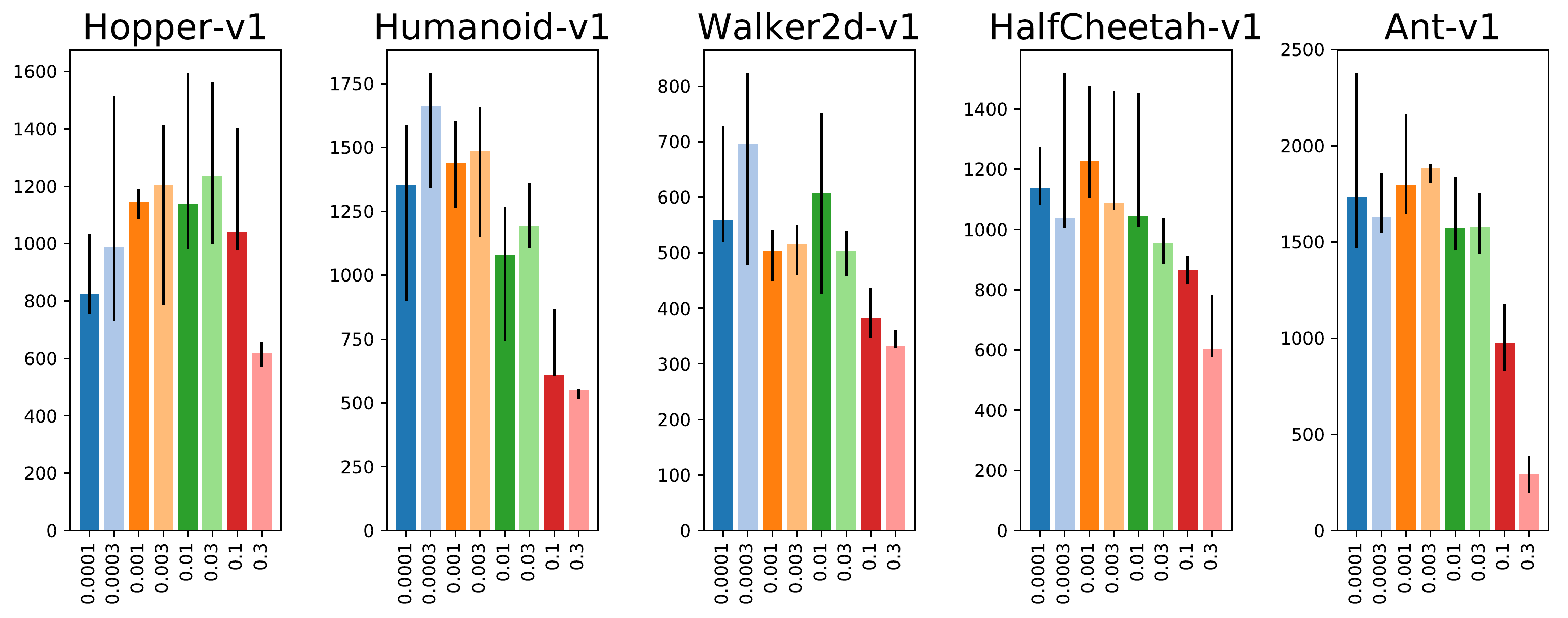}\hspace{1cm}\includegraphics[width=0.45\textwidth]{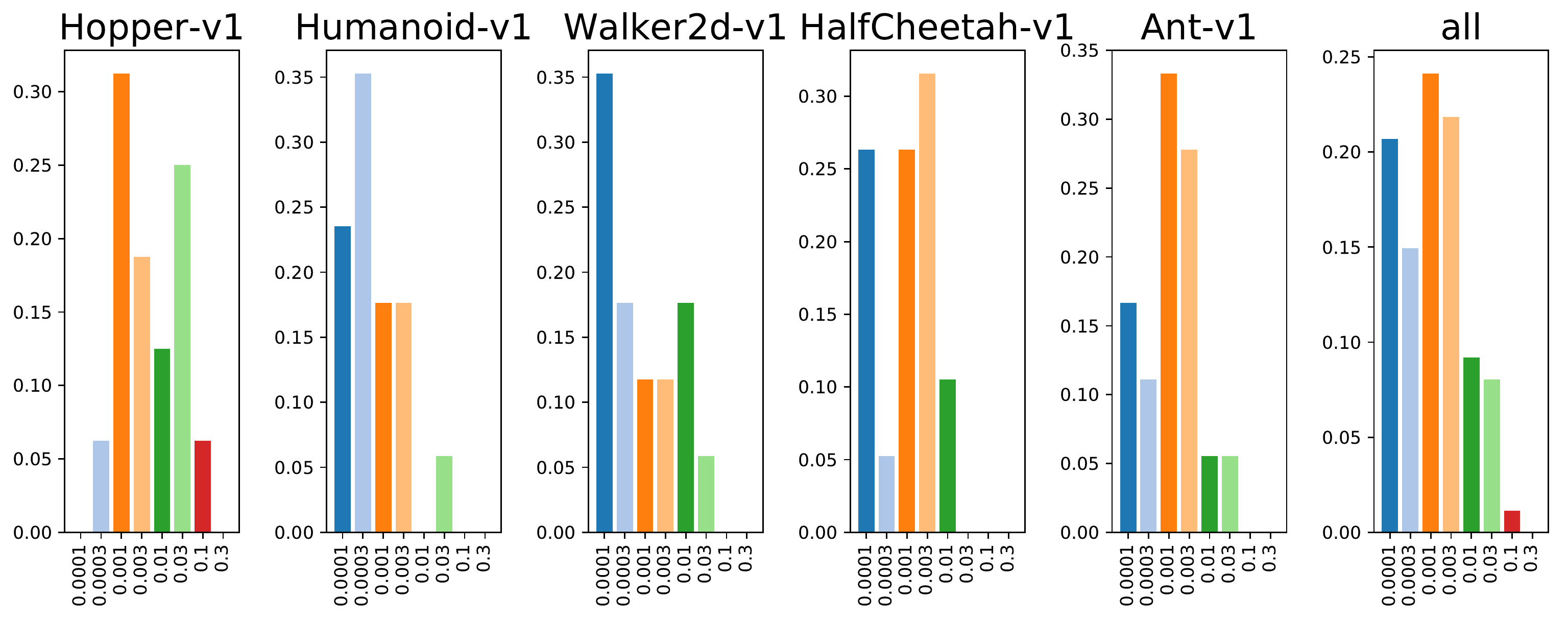}}
\caption{Analysis of choice \choicet{awrbeta}: 95th percentile of performance scores conditioned on sub-choice (left) and distribution of sub-choices in top 5\% of configurations (right).}
\label{fig:final_losses__gin_study_design_choice_value_sub_standard_policy_losses_awr_awr_beta}
\end{center}
\end{figure}

\begin{figure}[ht]
\begin{center}
\centerline{\includegraphics[width=0.45\textwidth]{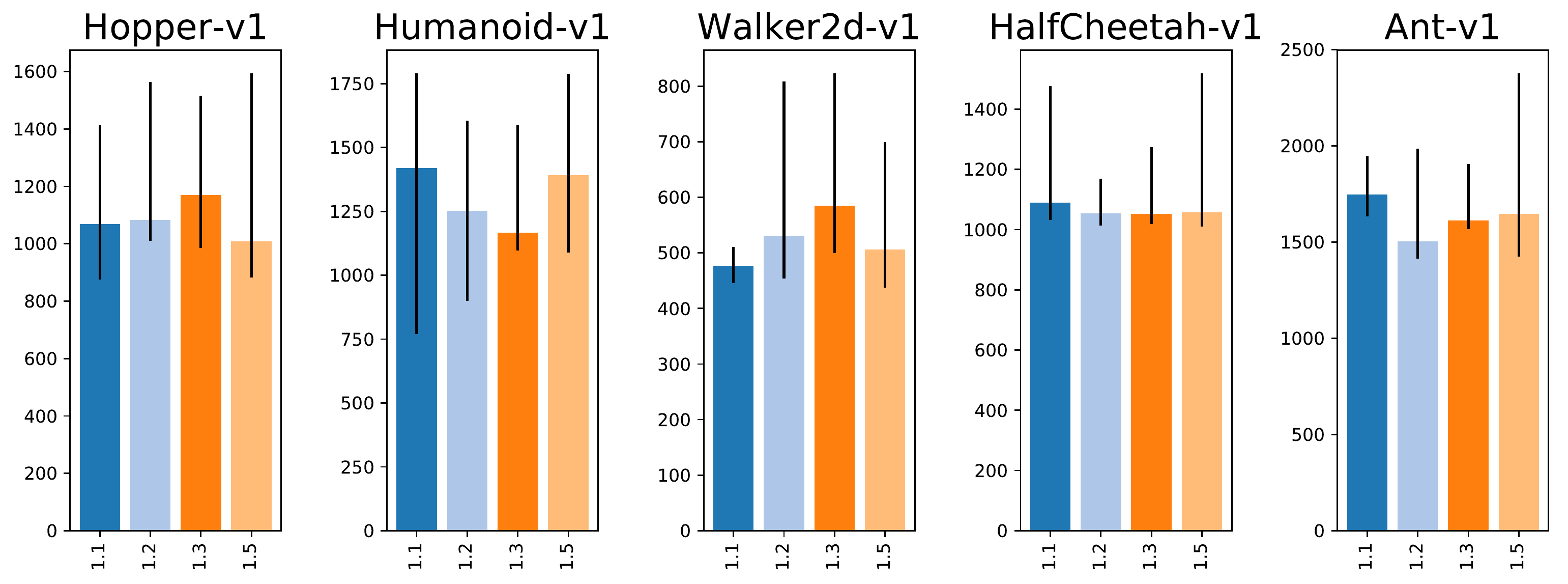}\hspace{1cm}\includegraphics[width=0.45\textwidth]{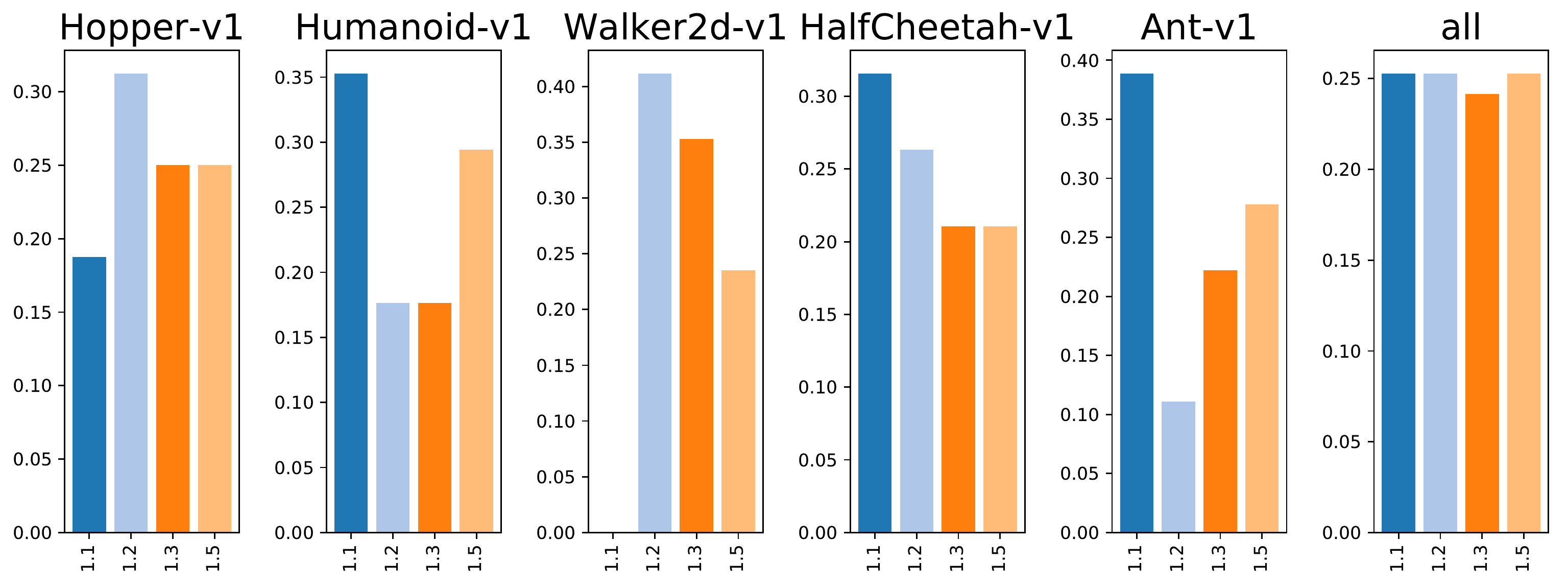}}
\caption{Analysis of choice \choicet{awrw}: 95th percentile of performance scores conditioned on sub-choice (left) and distribution of sub-choices in top 5\% of configurations (right).}
\label{fig:final_losses__gin_study_design_choice_value_sub_standard_policy_losses_awr_awr_w_max}
\end{center}
\end{figure}

We report aggregate statistics of the experiment in Table~\ref{tab:final_losses_overview} as well as training curves in Figure~\ref{fig:final_losses_training_curves}.
For each of the investigated choices in this experiment, we further provide a per-choice analysis in Figures~\ref{fig:final_losses_custom_loss}-\ref{fig:final_losses__gin_study_design_choice_value_sub_standard_policy_losses_awr_awr_w_max}.
\begin{table}[ht]
\begin{center}
\caption{Performance quantiles across choice configurations.}
\label{tab:final_losses_overview}
\begin{tabular}{lrrrrr}
\toprule
{} & Ant-v1 & HalfCheetah-v1 & Hopper-v1 & Humanoid-v1 & Walker2d-v1 \\
\midrule
90th percentile &   1490 &            994 &      1103 &        1224 &         459 \\
95th percentile &   1727 &           1080 &      1297 &        1630 &         565 \\
99th percentile &   2290 &           1363 &      1621 &        2611 &         869 \\
Max             &   2862 &           2048 &      1901 &        3435 &        1351 \\
\bottomrule
\end{tabular}

\end{center}
\end{table}

\begin{figure}[ht]
\begin{center}
\centerline{\includegraphics[width=1\textwidth]{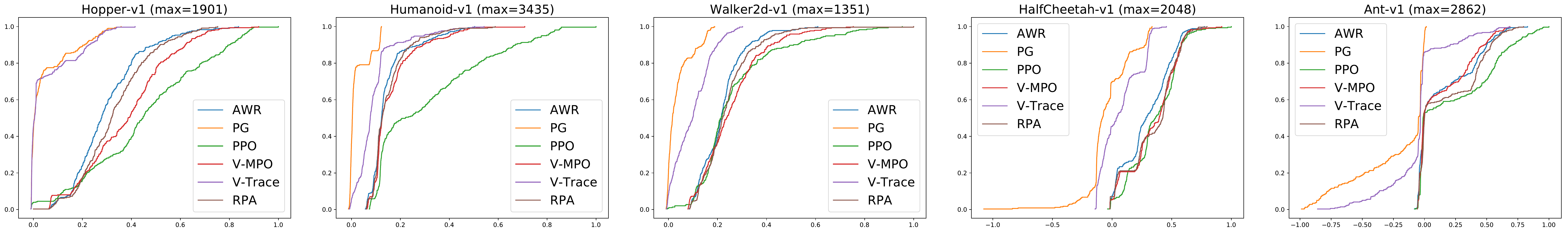}}
\caption{Empirical cumulative density functions of agent performance conditioned on different values of \choicet{policyloss}. The x axis denotes performance rescaled so that 0 corresponds to a random policy and 1 to the best found configuration, and the y axis denotes the quantile.}
\label{fig:final_losses__ecdf_standard_policy_losses}
\end{center}
\end{figure}

\begin{figure}[ht]
\begin{center}
\centerline{\includegraphics[width=1\textwidth]{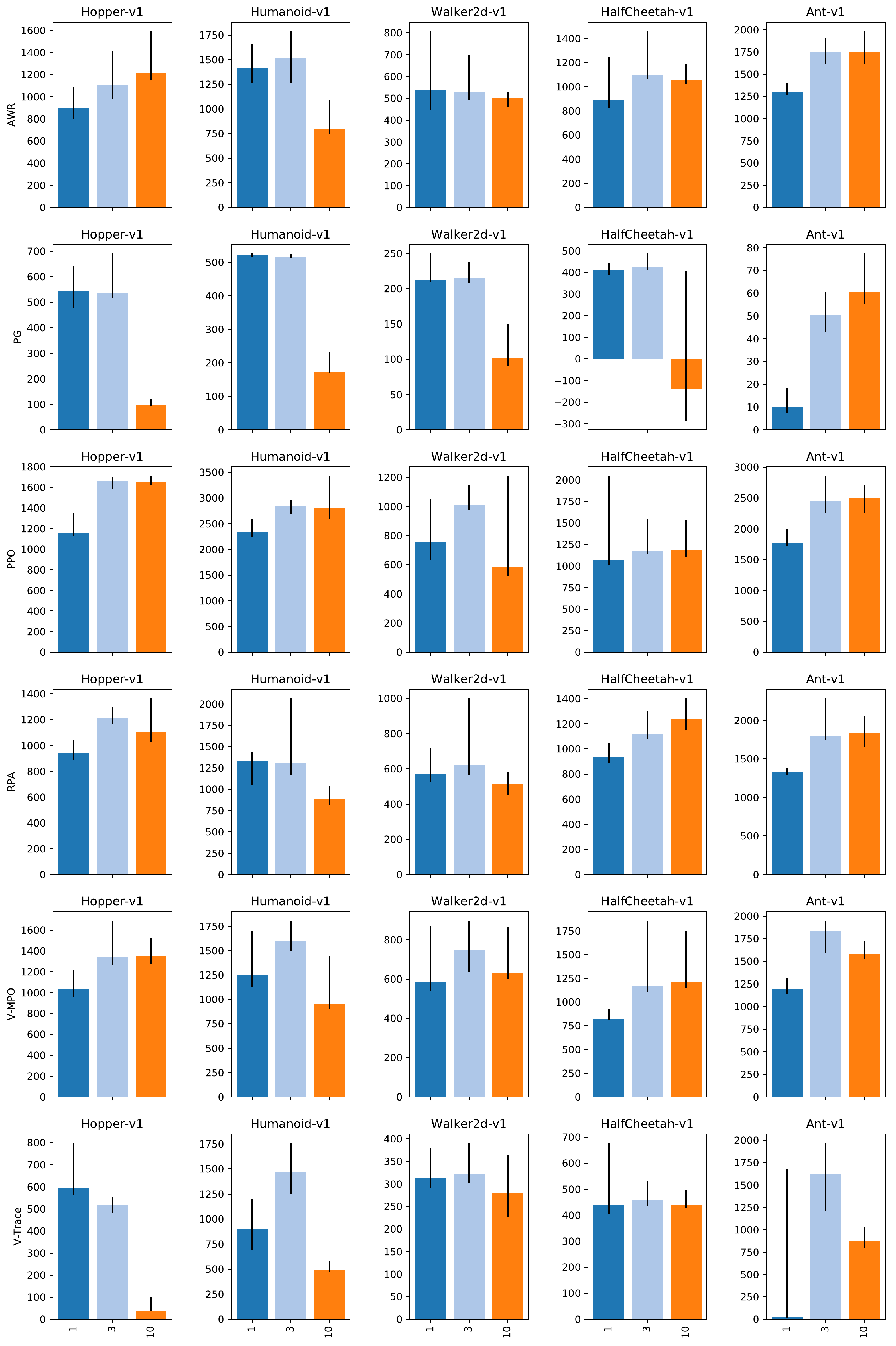}}
\caption{95th percentile of performance scores conditioned on \choicet{policyloss}(rows) and \choicet{numepochsperstep}(bars).}
\label{fig:final_losses__correlation_epochs_per_step_vs_losses}
\end{center}
\end{figure}
\clearpage

%% file: final_arch2/main.tex
\clearpage
\section{Experiment \texttt{Networks architecture}}
\label{exp_final_arch2}
\subsection{Design}
\label{exp_design_final_arch2}
For each of the 5 environments, we sampled 4000 choice configurations where we sampled the following choices independently and uniformly from the following ranges:
\begin{itemize}
    \item \choicet{actionpost}: \{clip, tanh\}
    \item \choicet{valueinit}: \{0.001, 0.01, 0.1, 1.0\}
    \item \choicet{stdind}: \{False, True\}
    \item \choicet{policyinit}: \{0.001, 0.01, 0.1, 1.0\}
    \item \choicet{stdtransform}: \{exp, softplus\}
    \item \choicet{initialstd}: \{0.1, 0.5, 1.0, 2.0\}
    \item \choicet{init}: \{Glorot normal, Glorot uniform, He normal, He uniform, LeCun normal, LeCun uniform, Orthogonal, Orthogonal(gain=1.41)\}
    \item \choicet{mlpshared}: \{separate, shared\}
    \begin{itemize}
        \item For the case ``\choicet{mlpshared} = separate'', we further sampled the sub-choices:
        \begin{itemize}
            \item \choicet{policywidth}: \{16, 32, 64, 128, 256, 512\}
            \item \choicet{policydepth}: \{1, 2, 4, 8\}
            \item \choicet{valuewidth}: \{16, 32, 64, 128, 256, 512\}
            \item \choicet{valuedepth}: \{1, 2, 4, 8\}
        \end{itemize}
        \item For the case ``\choicet{mlpshared} = shared'', we further sampled the sub-choices:
        \begin{itemize}
            \item \choicet{sharedwidth}: \{16, 32, 64, 128, 256, 512\}
            \item \choicet{shareddepth}: \{1, 2, 4, 8\}
            \item \choicet{baselinecost}: \{0.001, 0.1, 1.0, 10.0, 100.0\}
        \end{itemize}
    \end{itemize}
    \item \choicet{minstd}: \{0.0, 0.01, 0.1\}
    \item \choicet{adamlr}: \{3e-05, 0.0001, 0.0003, 0.001\}
    \item \choicet{activation}: \{ELU, Leaky ReLU, ReLU, Sigmoid, Swish, Tanh\}
\end{itemize}
All the other choices were set to the default values as described in Appendix~\ref{sec:default_settings}.

For each of the sampled choice configurations, we train 3 agents with different random seeds and compute the performance metric as described in Section~\ref{sec:performance}.

\begin{figure}[ht]
\begin{center}
\centerline{\includegraphics[width=0.45\textwidth]{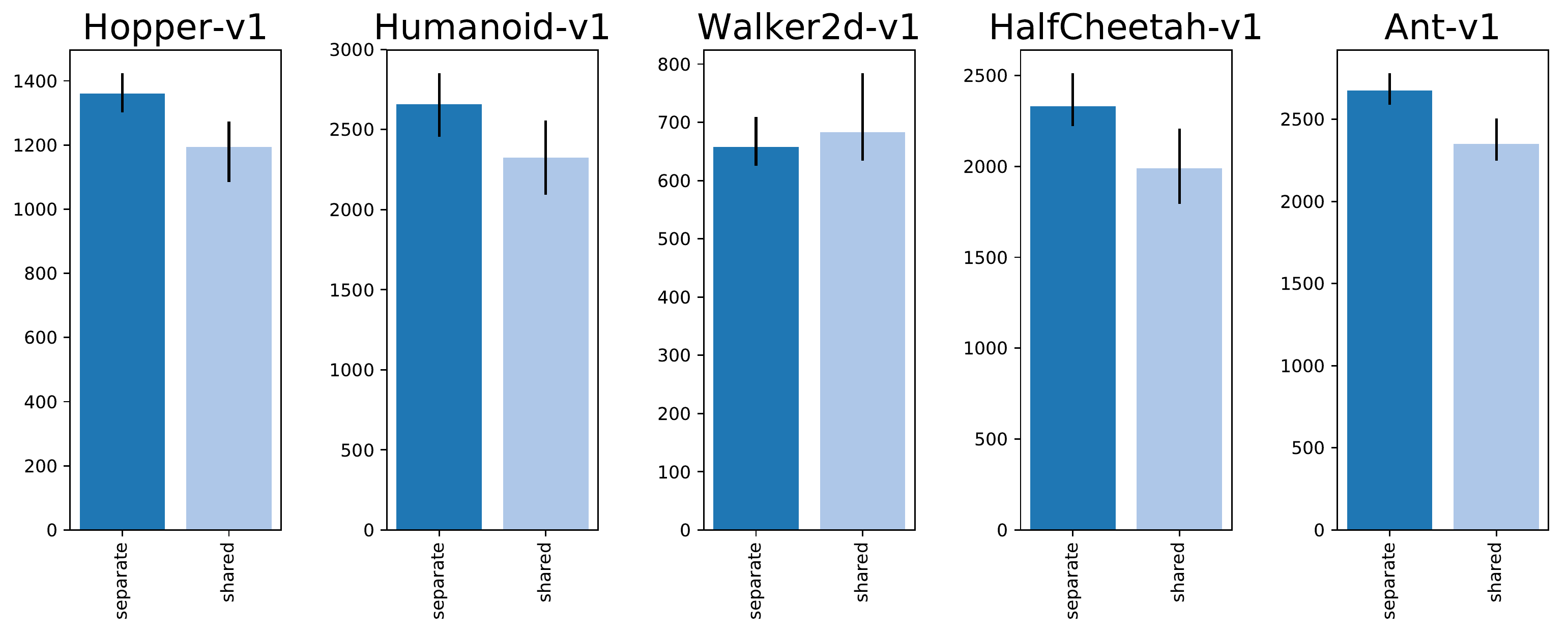}\hspace{1cm}\includegraphics[width=0.45\textwidth]{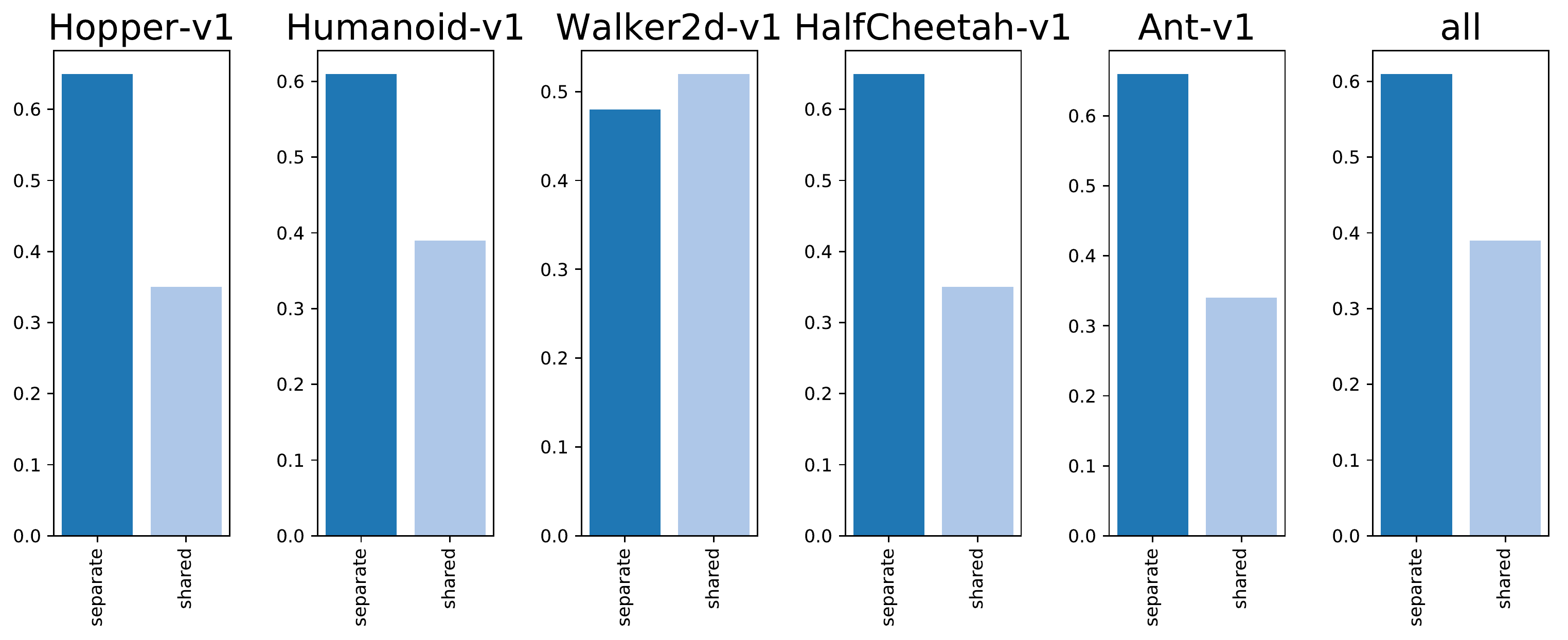}}
\caption{Analysis of choice \choicet{mlpshared}: 95th percentile of performance scores conditioned on choice (left) and distribution of choices in top 5\% of configurations (right).}
\label{fig:final_arch__mlpshared}
\end{center}
\end{figure}

After running the experiment described above we noticed (Fig.~\ref{fig:final_arch__mlpshared}) that separate policy and value function networks (\choicep{mlpshared}) perform better and we have rerun the experiment with only this variant present.

\subsection{Results}
\label{exp_results_final_arch2}
\begin{figure}[ht]
\begin{center}
\centerline{\includegraphics[width=1\textwidth]{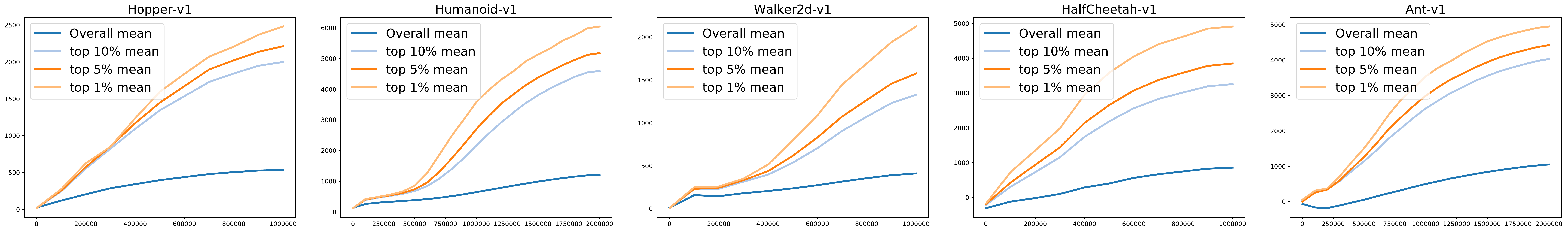}}
\caption{Training curves.}
\label{fig:final_arch2_training_curves}
\end{center}
\end{figure}

\begin{figure}[ht]
\begin{center}
\centerline{\includegraphics[width=0.45\textwidth]{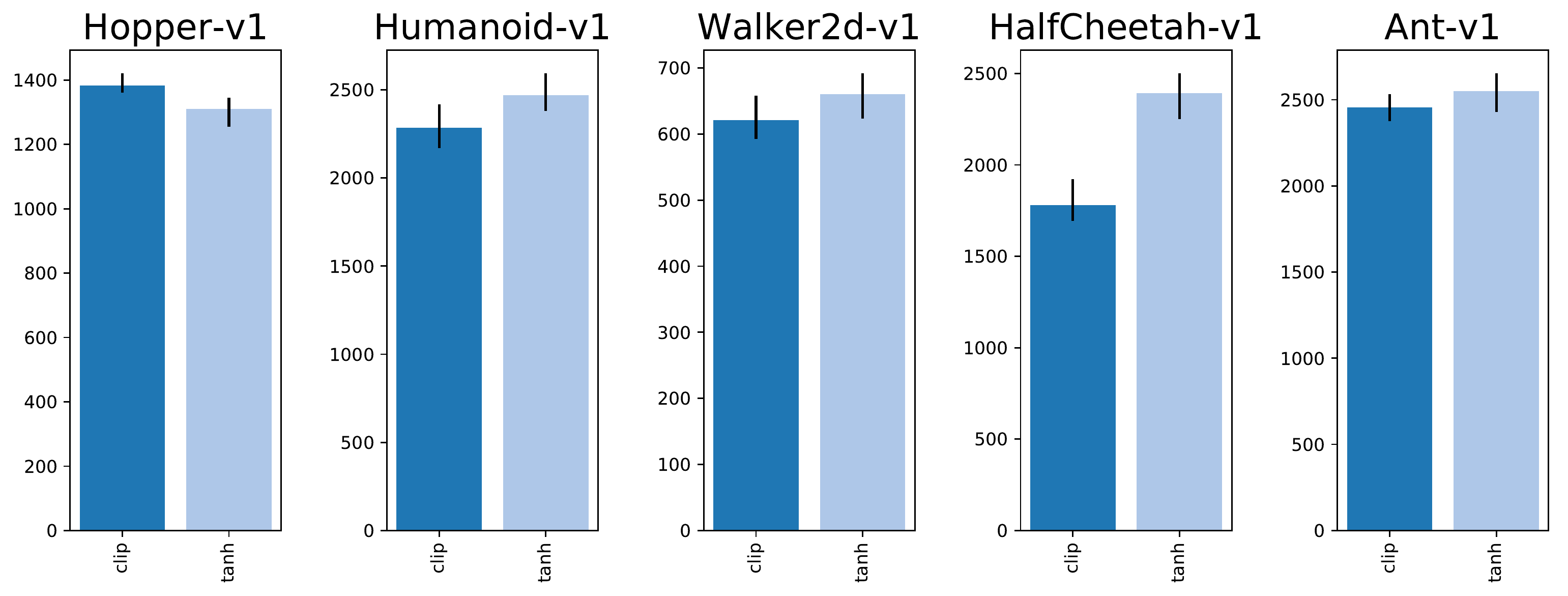}\hspace{1cm}\includegraphics[width=0.45\textwidth]{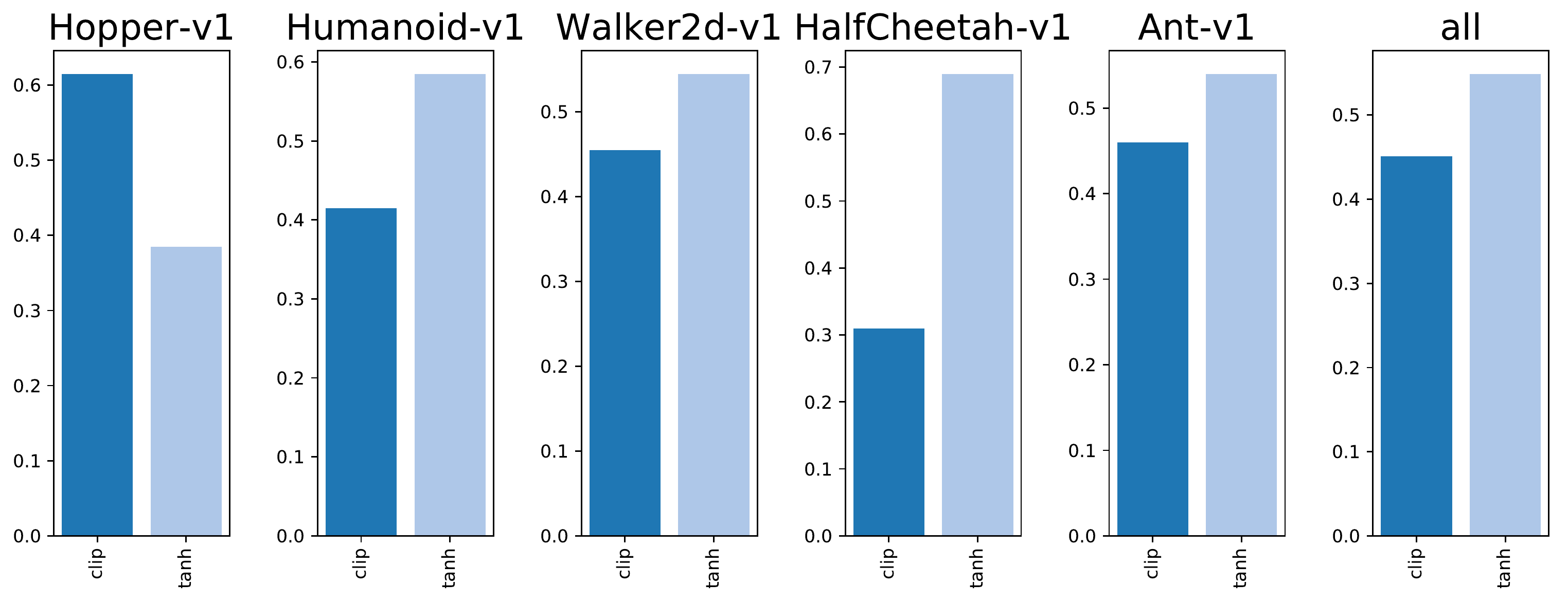}}
\caption{Analysis of choice \choicet{actionpost}: 95th percentile of performance scores conditioned on choice (left) and distribution of choices in top 5\% of configurations (right).}
\label{fig:final_arch2__gin_study_design_choice_value_action_postprocessing}
\end{center}
\end{figure}

\begin{figure}[ht]
\begin{center}
\centerline{\includegraphics[width=0.45\textwidth]{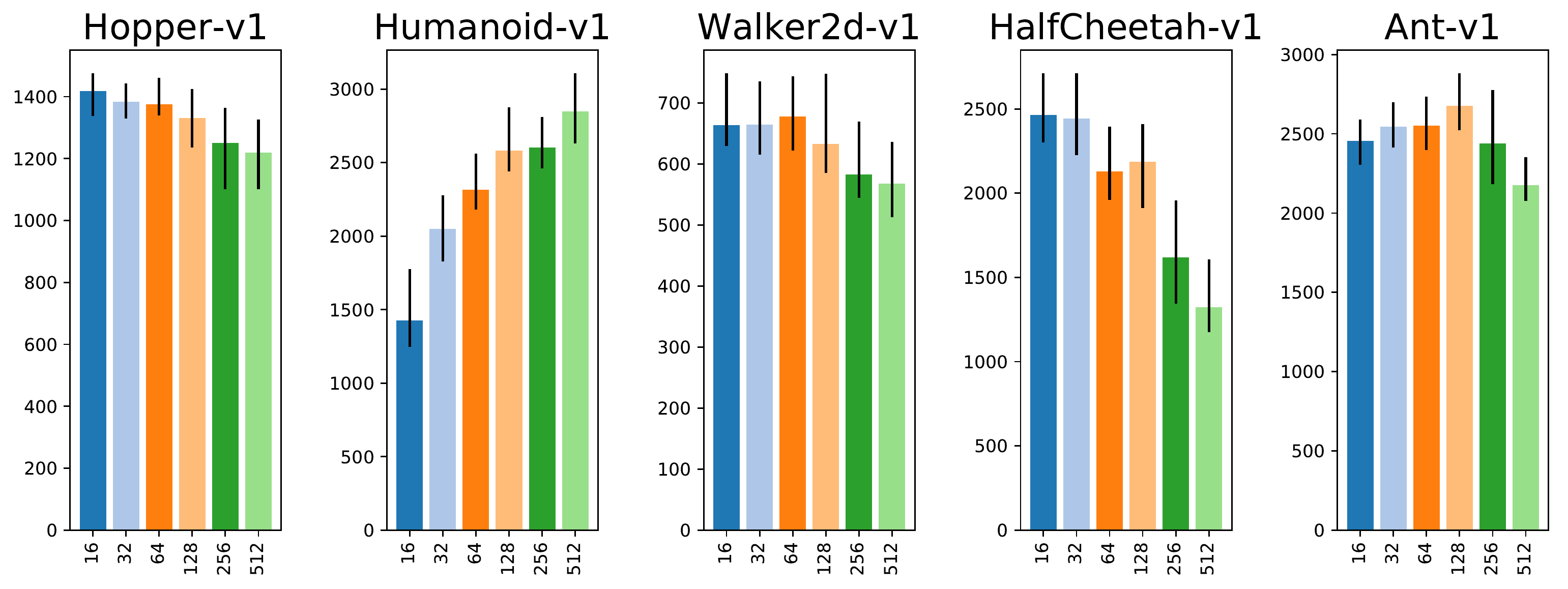}\hspace{1cm}\includegraphics[width=0.45\textwidth]{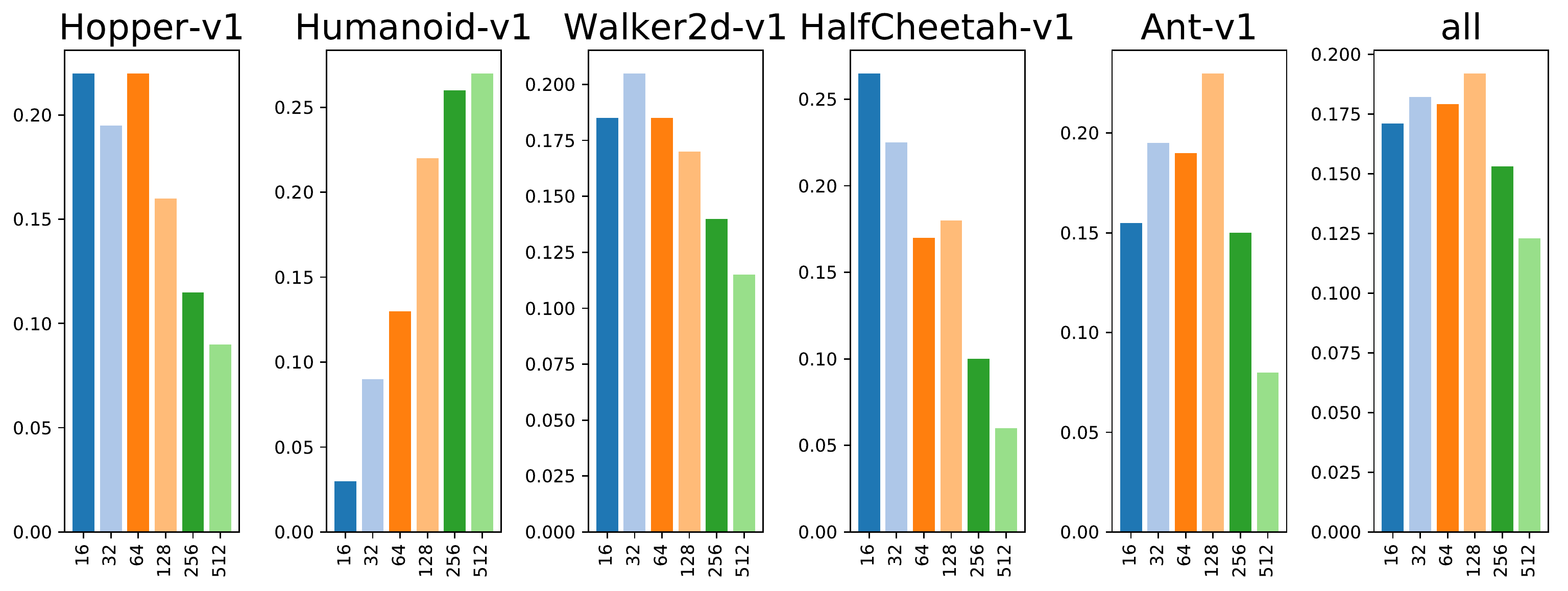}}
\caption{Analysis of choice \choicet{policywidth}: 95th percentile of performance scores conditioned on choice (left) and distribution of choices in top 5\% of configurations (right).}
\label{fig:final_arch2__gin_study_design_choice_value_policy_mlp_width}
\end{center}
\end{figure}

\begin{figure}[ht]
\begin{center}
\centerline{\includegraphics[width=0.45\textwidth]{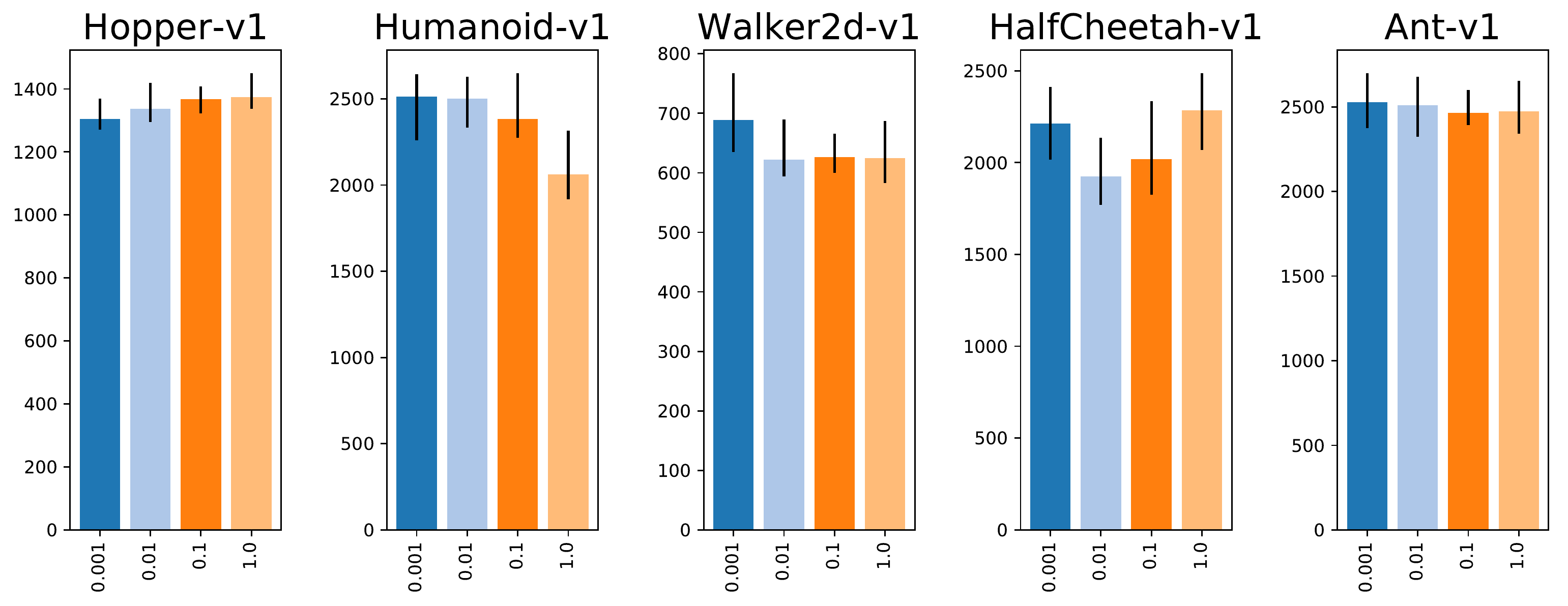}\hspace{1cm}\includegraphics[width=0.45\textwidth]{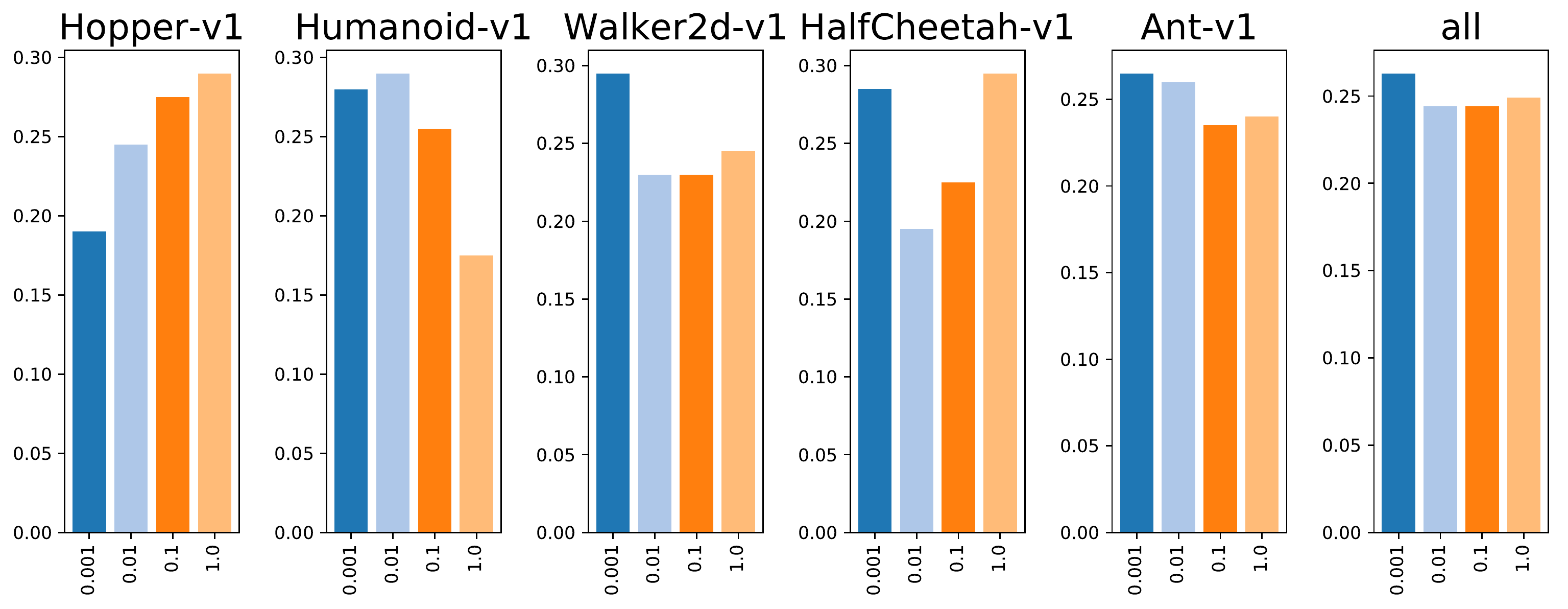}}
\caption{Analysis of choice \choicet{valueinit}: 95th percentile of performance scores conditioned on choice (left) and distribution of choices in top 5\% of configurations (right).}
\label{fig:final_arch2__gin_study_design_choice_value_last_kernel_init_value_scaling}
\end{center}
\end{figure}

\begin{figure}[ht]
\begin{center}
\centerline{\includegraphics[width=0.45\textwidth]{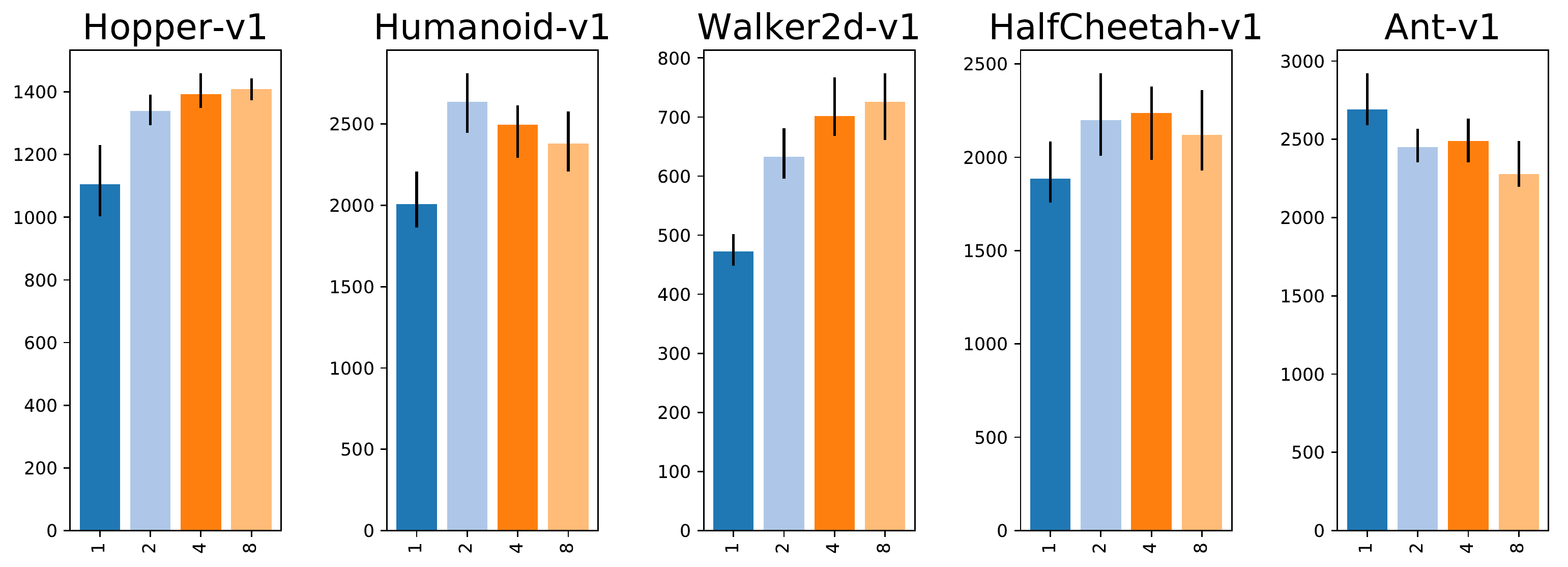}\hspace{1cm}\includegraphics[width=0.45\textwidth]{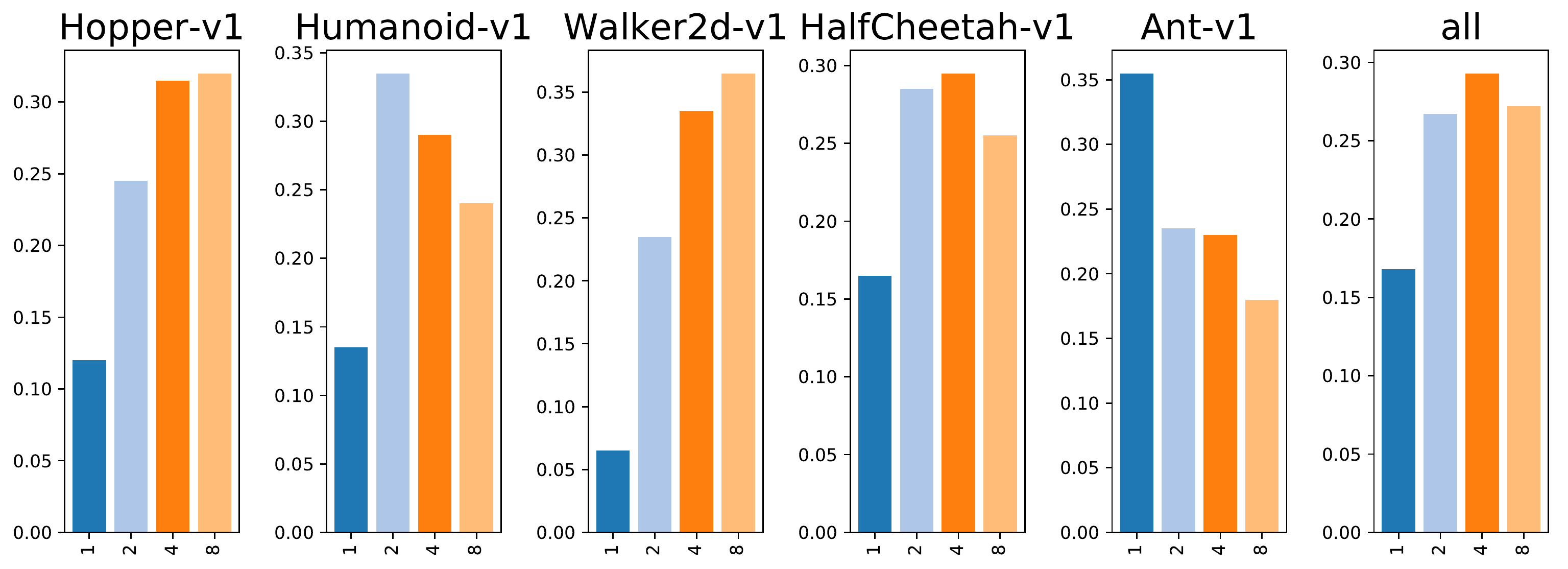}}
\caption{Analysis of choice \choicet{valuedepth}: 95th percentile of performance scores conditioned on choice (left) and distribution of choices in top 5\% of configurations (right).}
\label{fig:final_arch2__gin_study_design_choice_value_value_mlp_depth}
\end{center}
\end{figure}

\begin{figure}[ht]
\begin{center}
\centerline{\includegraphics[width=0.45\textwidth]{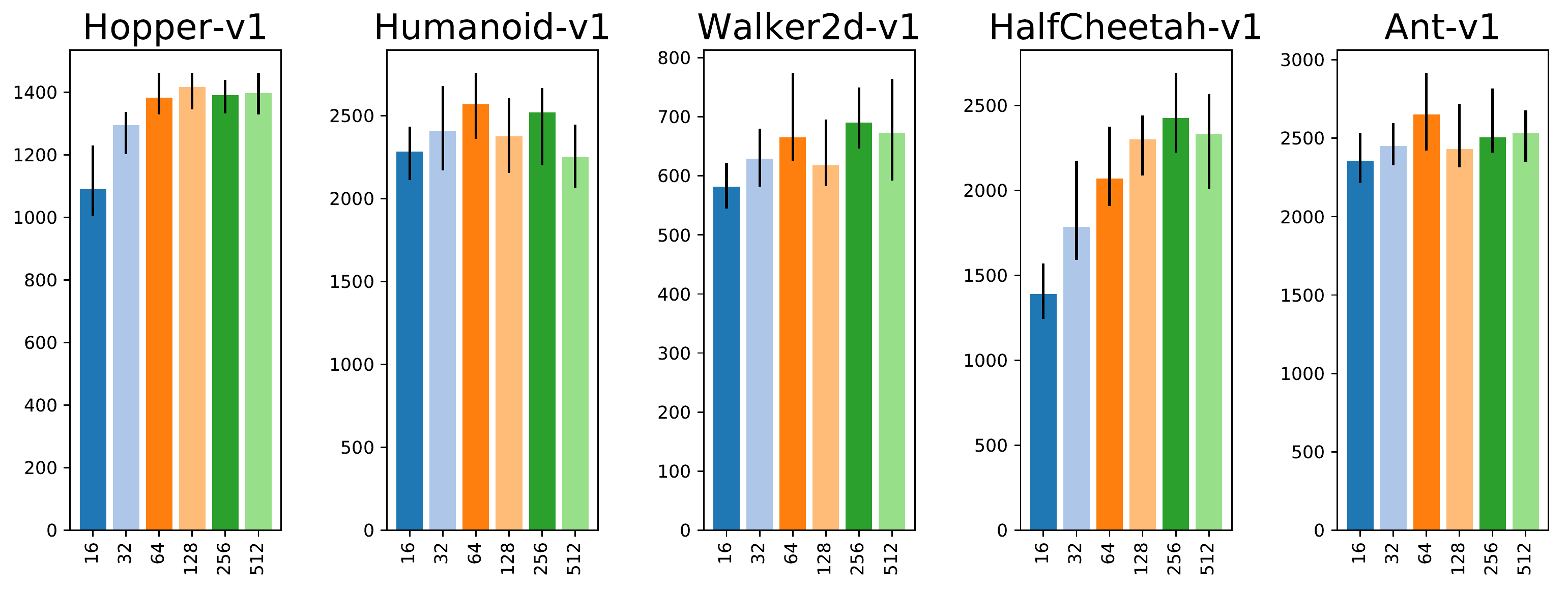}\hspace{1cm}\includegraphics[width=0.45\textwidth]{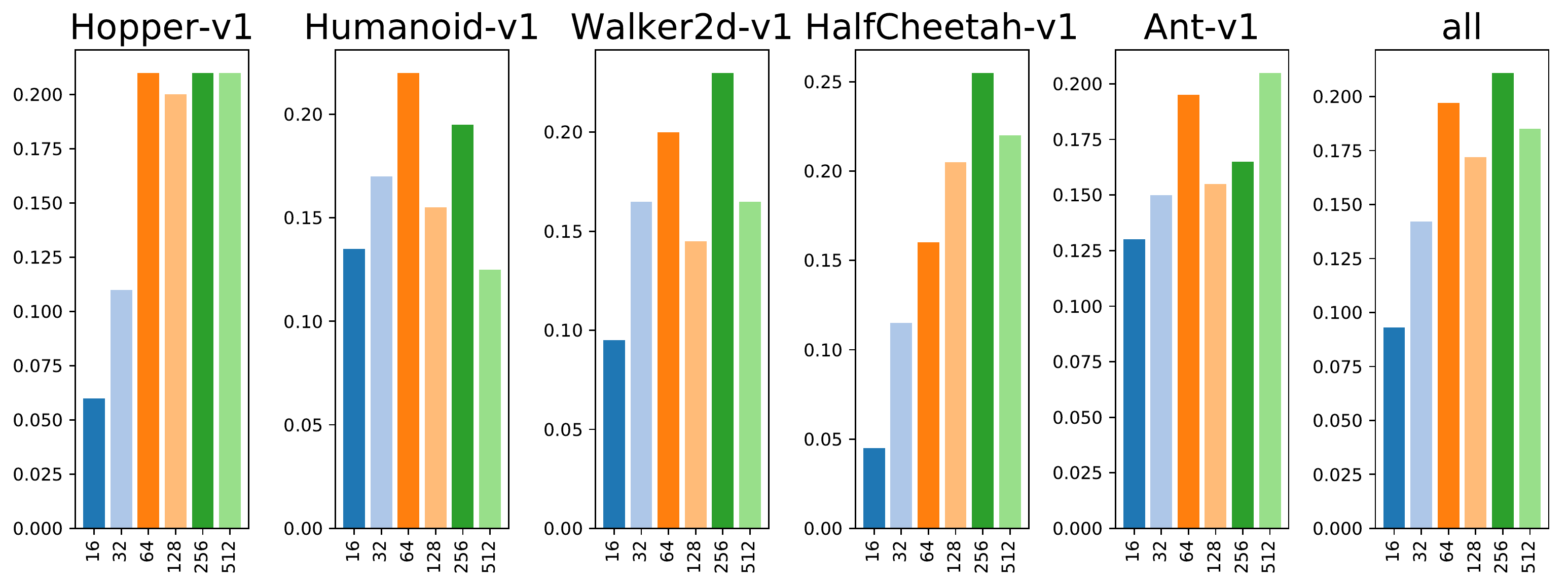}}
\caption{Analysis of choice \choicet{valuewidth}: 95th percentile of performance scores conditioned on choice (left) and distribution of choices in top 5\% of configurations (right).}
\label{fig:final_arch2__gin_study_design_choice_value_value_mlp_width}
\end{center}
\end{figure}

\begin{figure}[ht]
\begin{center}
\centerline{\includegraphics[width=0.45\textwidth]{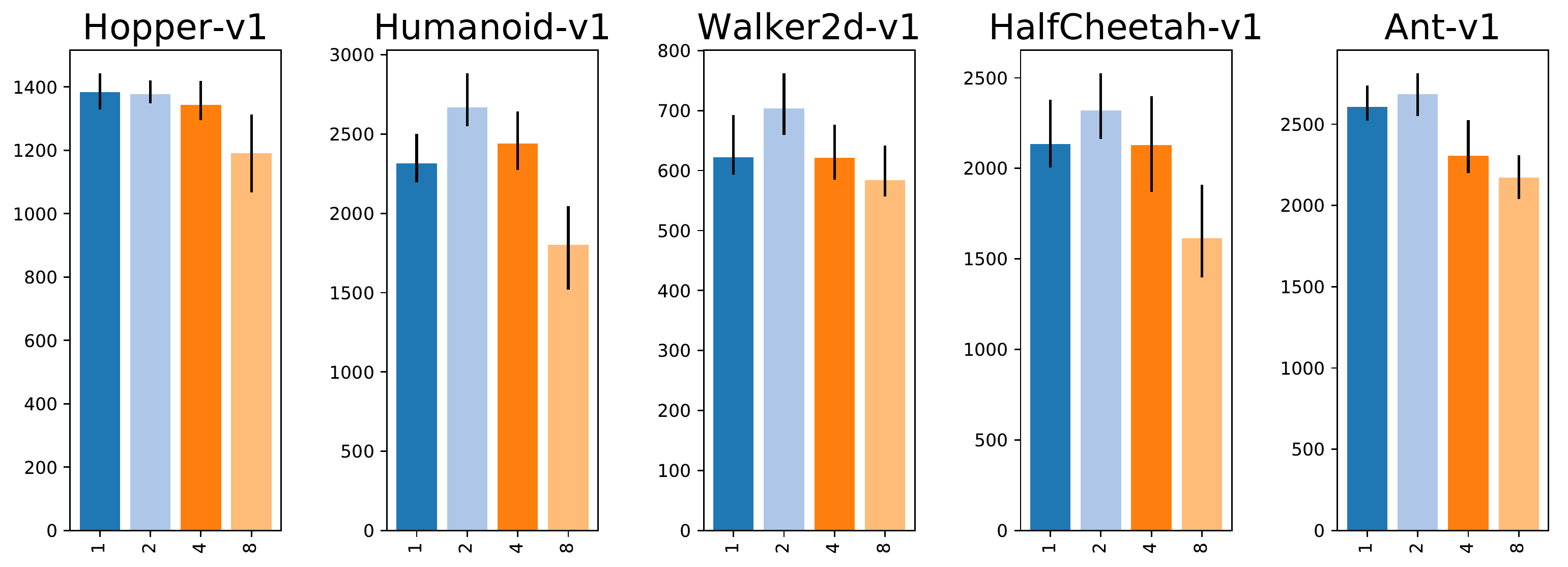}\hspace{1cm}\includegraphics[width=0.45\textwidth]{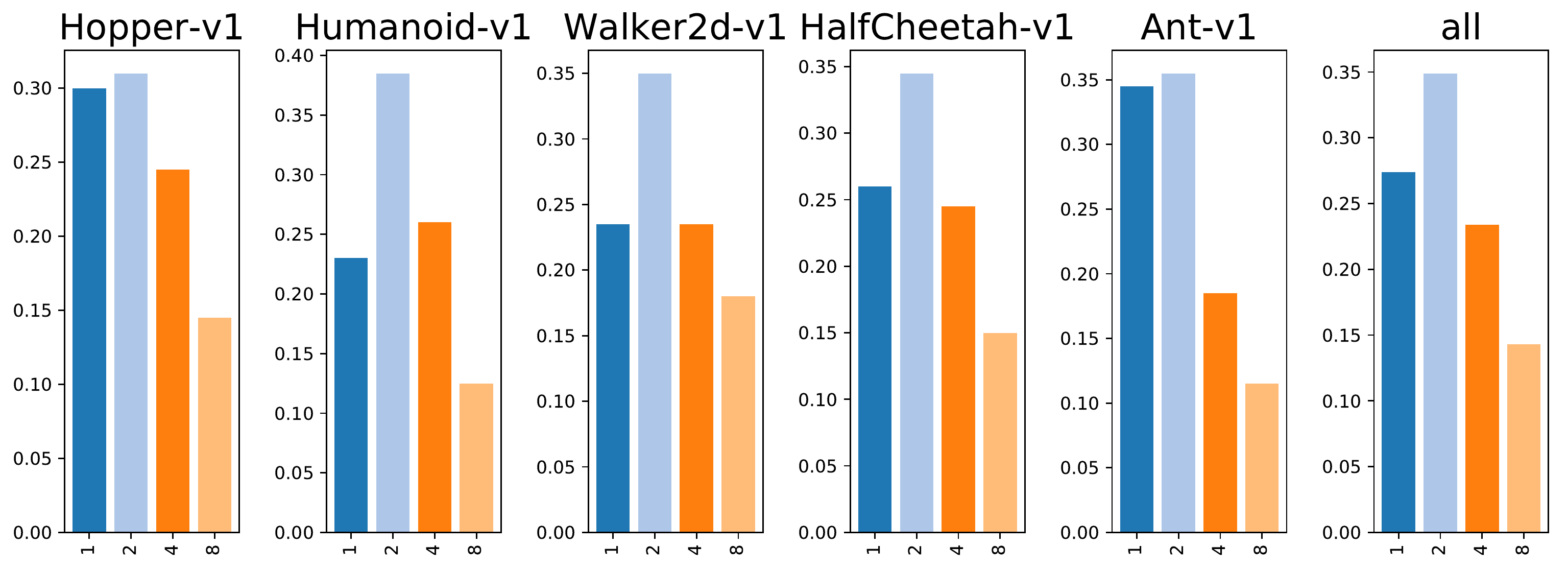}}
\caption{Analysis of choice \choicet{policydepth}: 95th percentile of performance scores conditioned on choice (left) and distribution of choices in top 5\% of configurations (right).}
\label{fig:final_arch2__gin_study_design_choice_value_policy_mlp_depth}
\end{center}
\end{figure}

\begin{figure}[ht]
\begin{center}
\centerline{\includegraphics[width=0.45\textwidth]{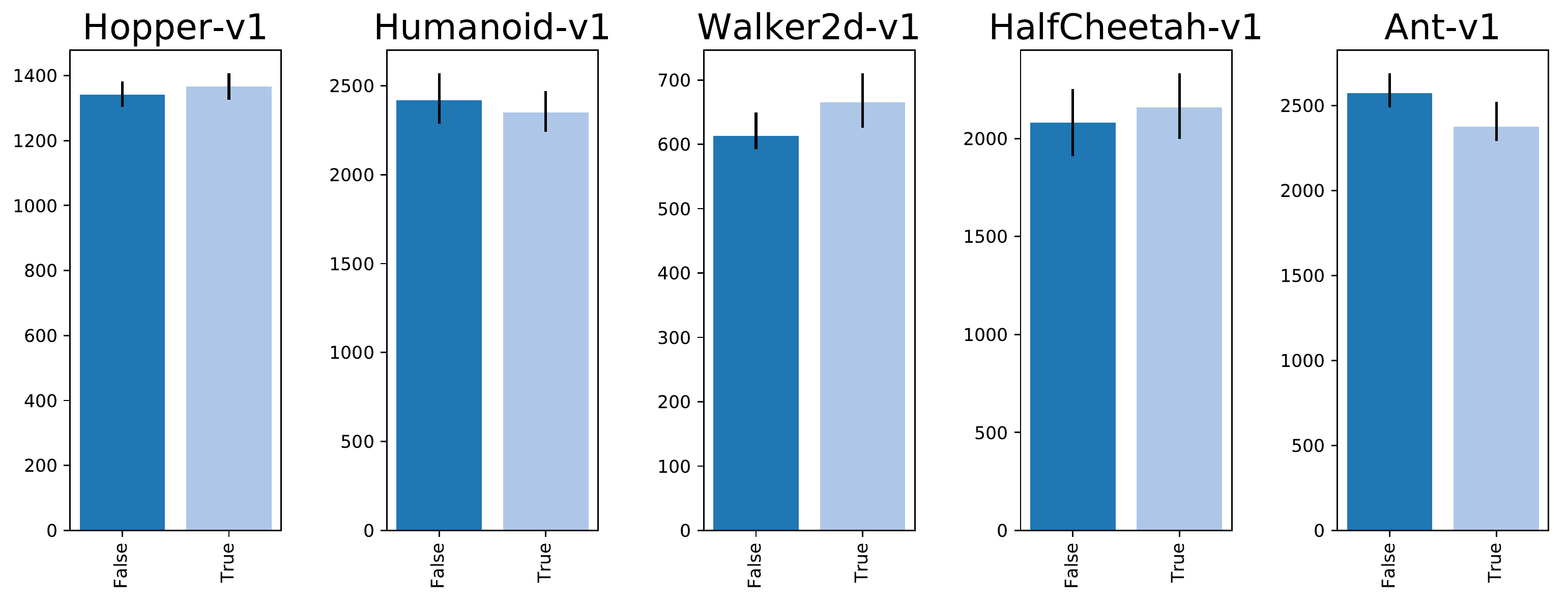}\hspace{1cm}\includegraphics[width=0.45\textwidth]{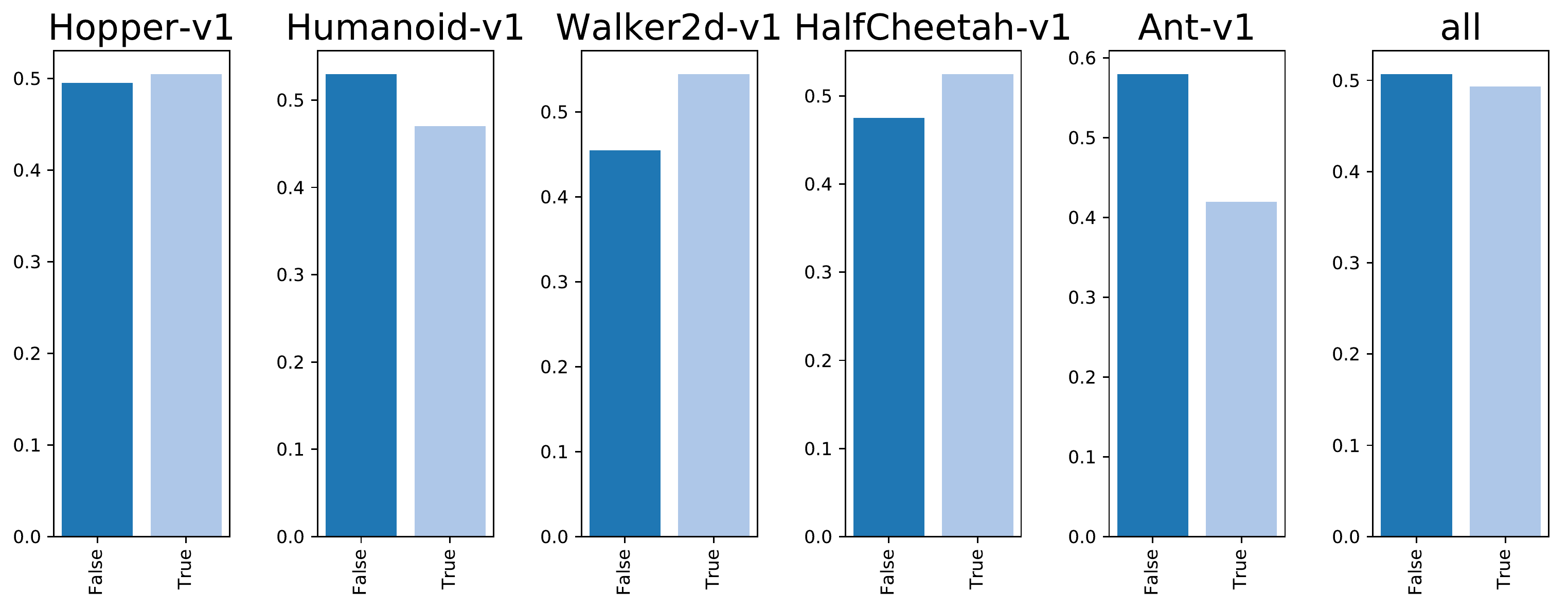}}
\caption{Analysis of choice \choicet{stdind}: 95th percentile of performance scores conditioned on choice (left) and distribution of choices in top 5\% of configurations (right).}
\label{fig:final_arch2__gin_study_design_choice_value_std_independent_of_input}
\end{center}
\end{figure}

\begin{figure}[ht]
\begin{center}
\centerline{\includegraphics[width=0.45\textwidth]{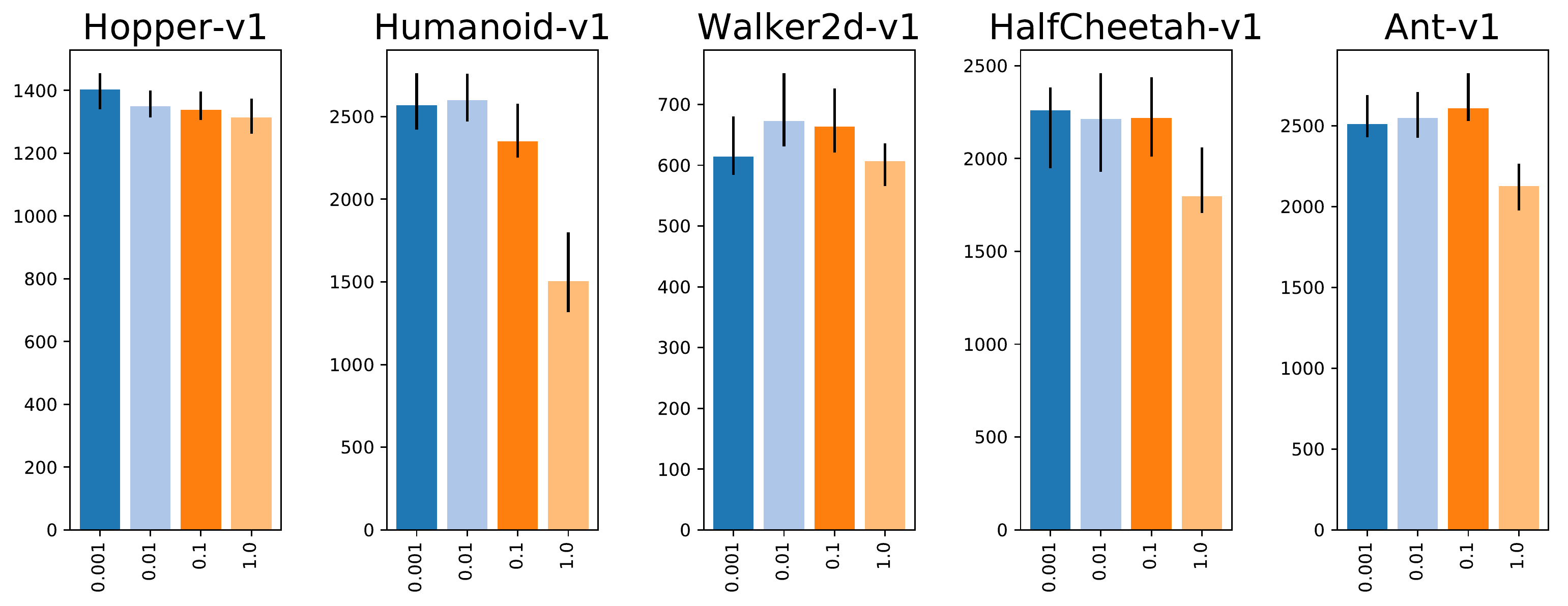}\hspace{1cm}\includegraphics[width=0.45\textwidth]{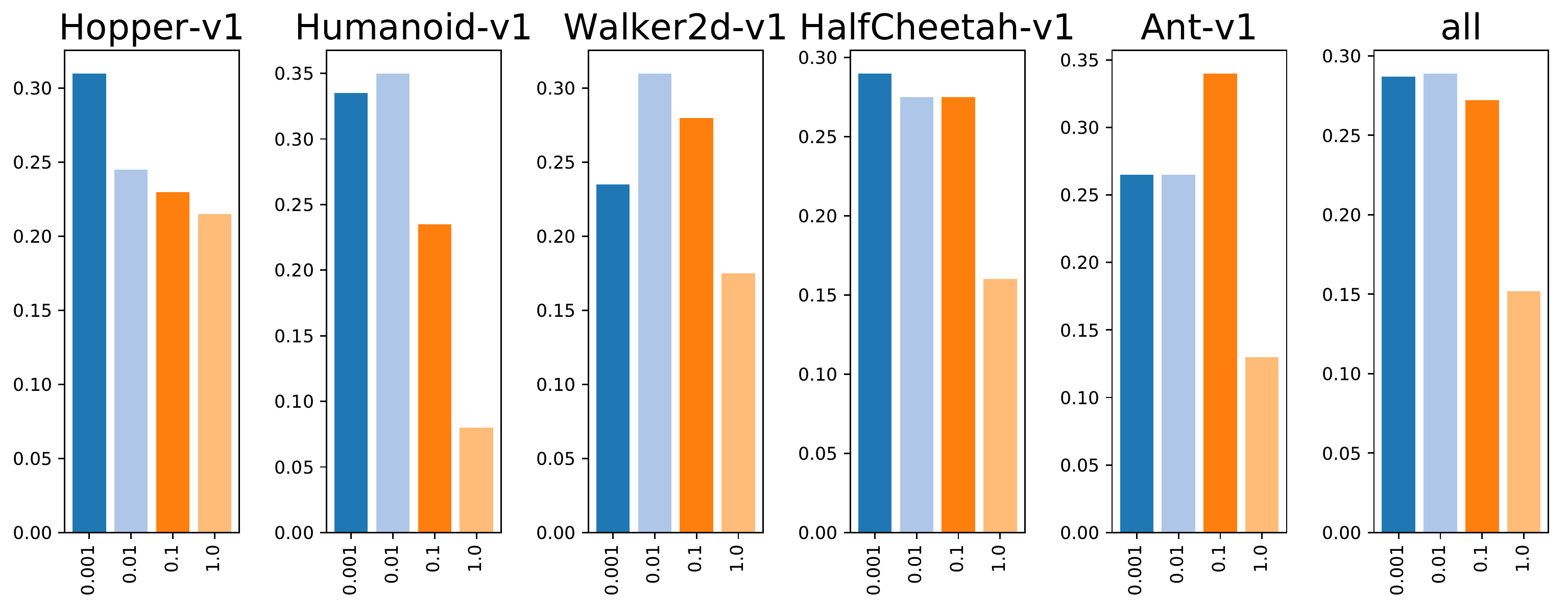}}
\caption{Analysis of choice \choicet{policyinit}: 95th percentile of performance scores conditioned on choice (left) and distribution of choices in top 5\% of configurations (right).}
\label{fig:final_arch2__gin_study_design_choice_value_last_kernel_init_policy_scaling}
\end{center}
\end{figure}

\begin{figure}[ht]
\begin{center}
\centerline{\includegraphics[width=0.45\textwidth]{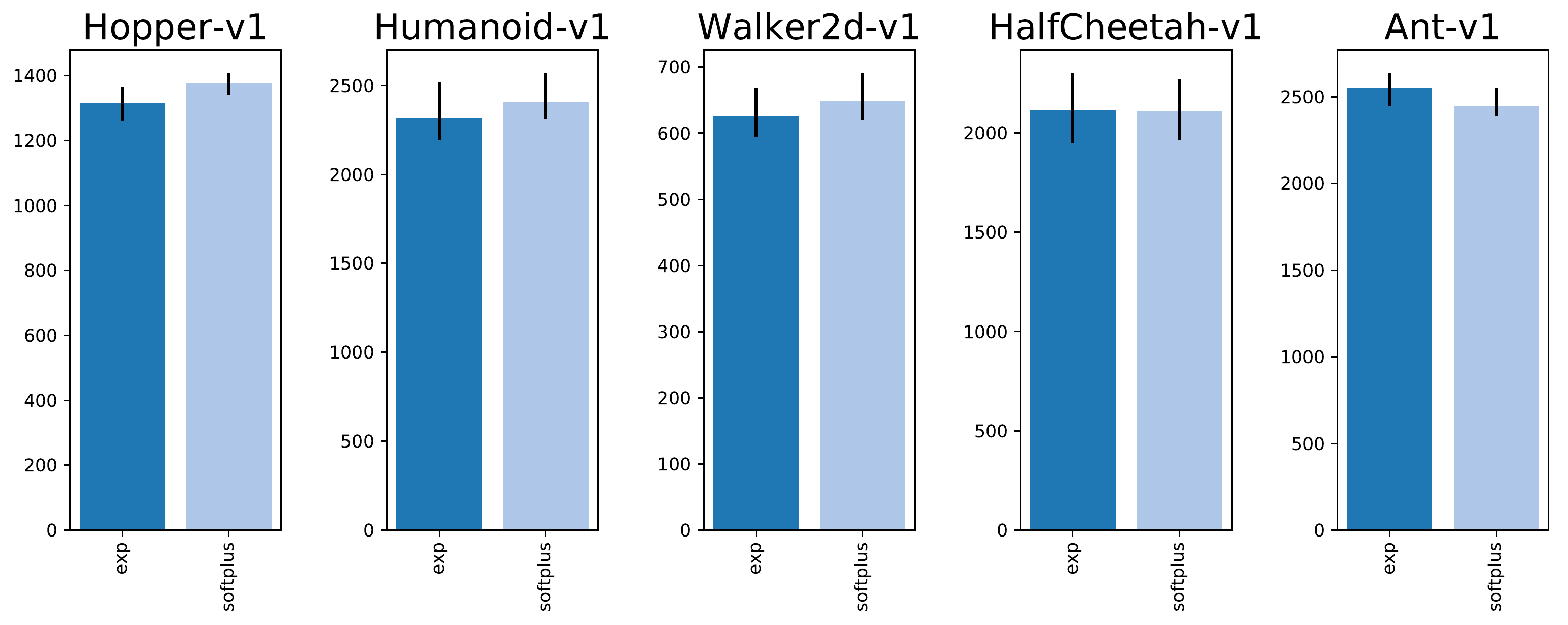}\hspace{1cm}\includegraphics[width=0.45\textwidth]{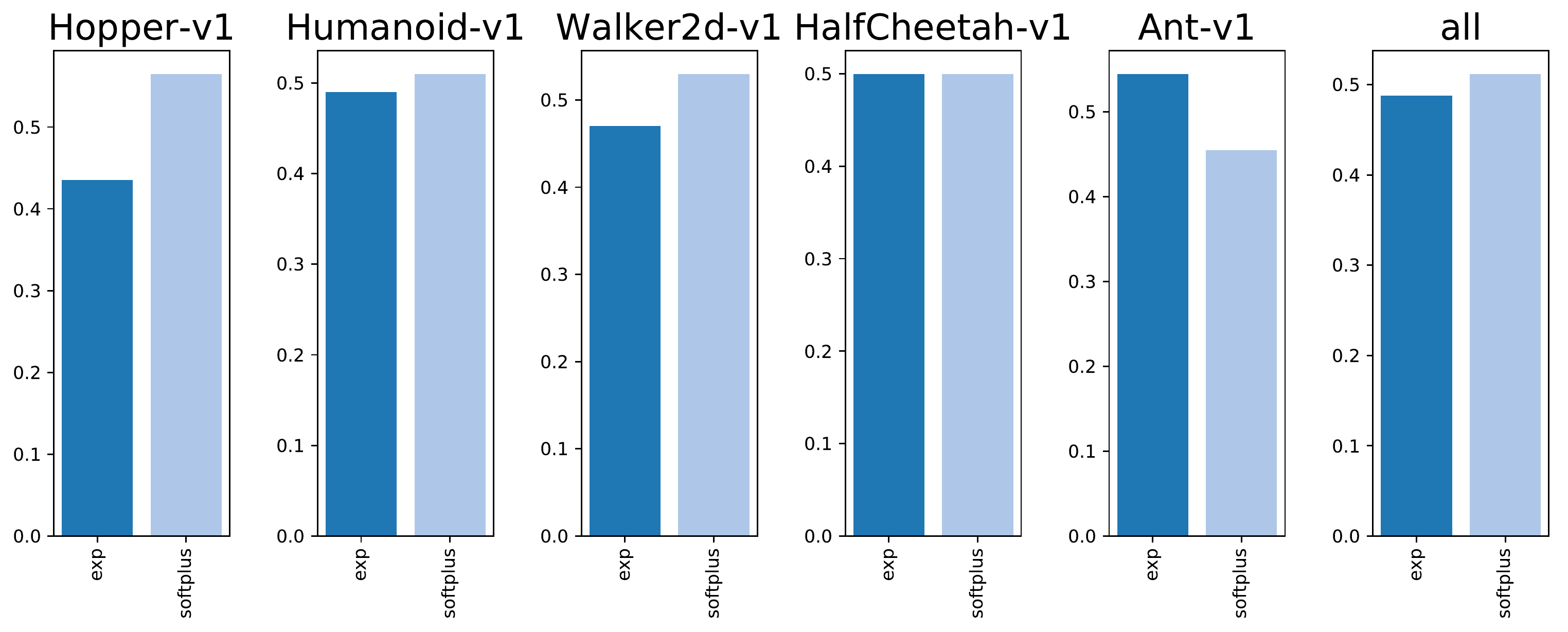}}
\caption{Analysis of choice \choicet{stdtransform}: 95th percentile of performance scores conditioned on choice (left) and distribution of choices in top 5\% of configurations (right).}
\label{fig:final_arch2__gin_study_design_choice_value_scale_function}
\end{center}
\end{figure}

\begin{figure}[ht]
\begin{center}
\centerline{\includegraphics[width=0.45\textwidth]{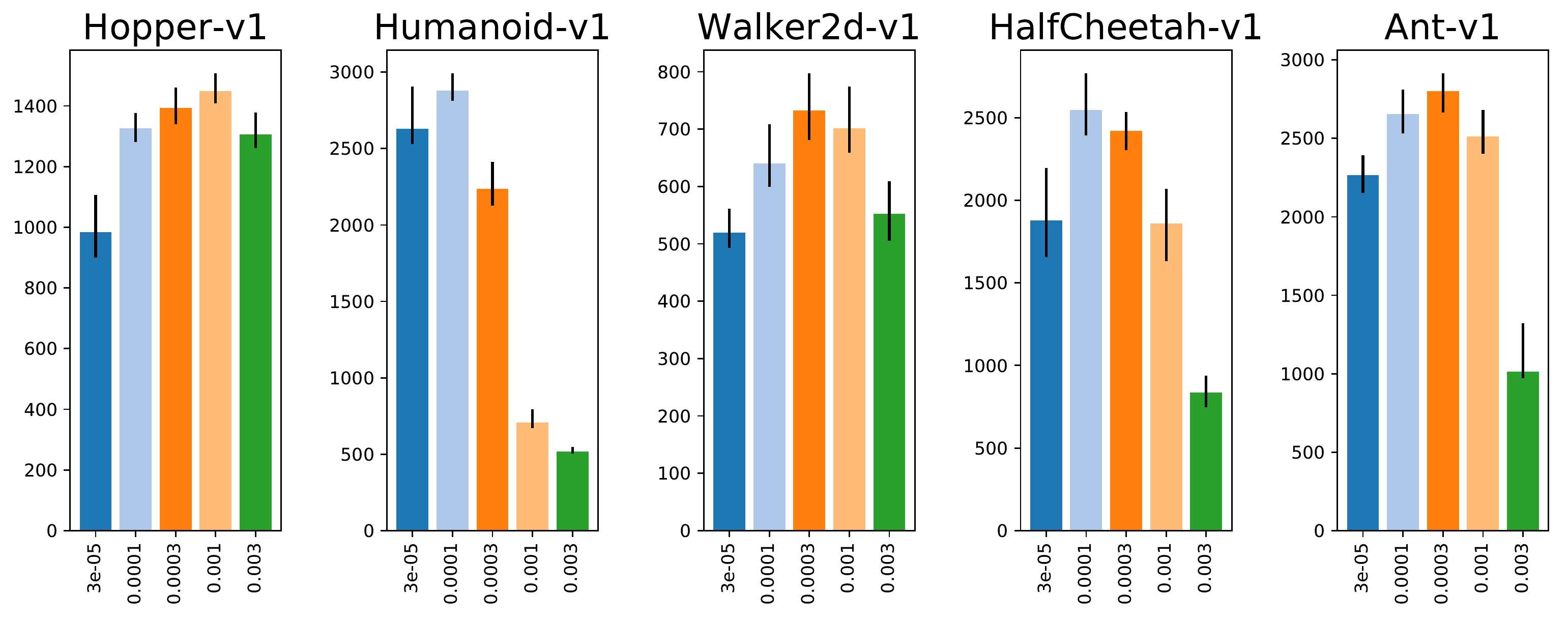}\hspace{1cm}\includegraphics[width=0.45\textwidth]{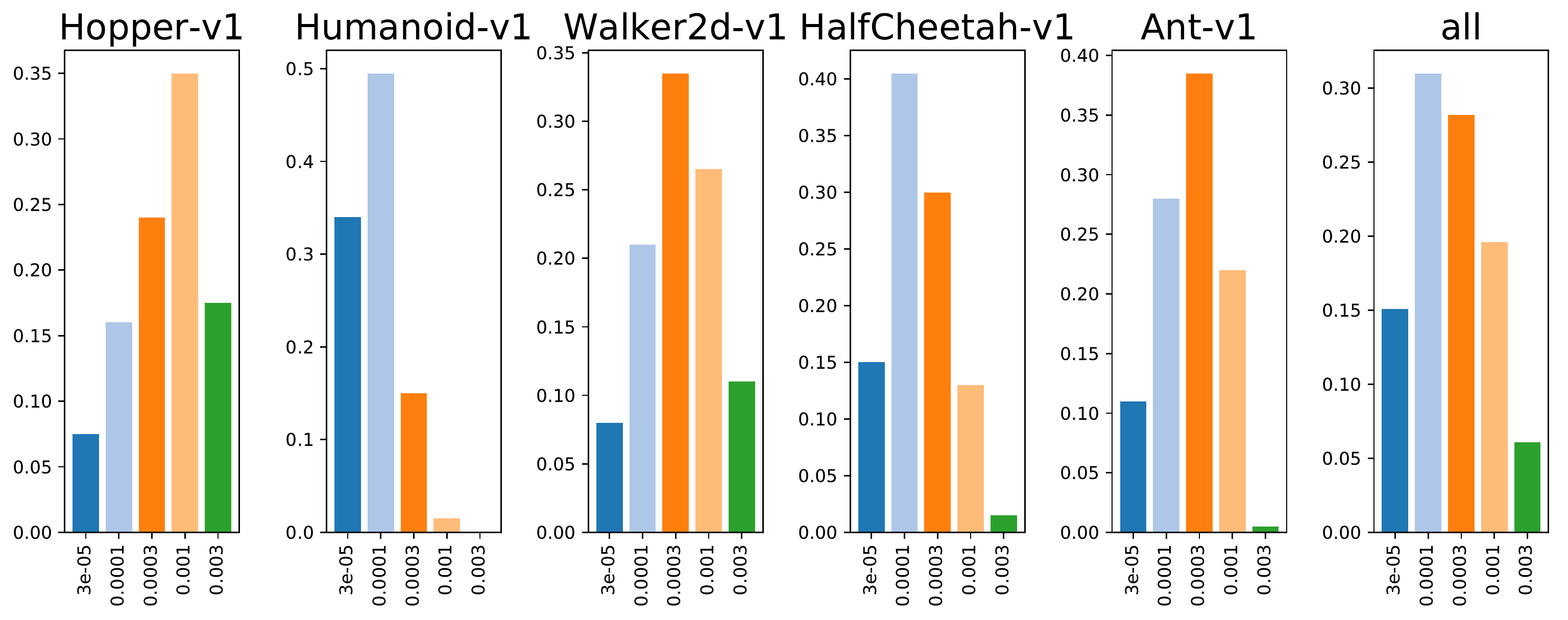}}
\caption{Analysis of choice \choicet{adamlr}: 95th percentile of performance scores conditioned on choice (left) and distribution of choices in top 5\% of configurations (right).}
\label{fig:final_arch2__gin_study_design_choice_value_learning_rate}
\end{center}
\end{figure}

\begin{figure}[ht]
\begin{center}
\centerline{\includegraphics[width=0.45\textwidth]{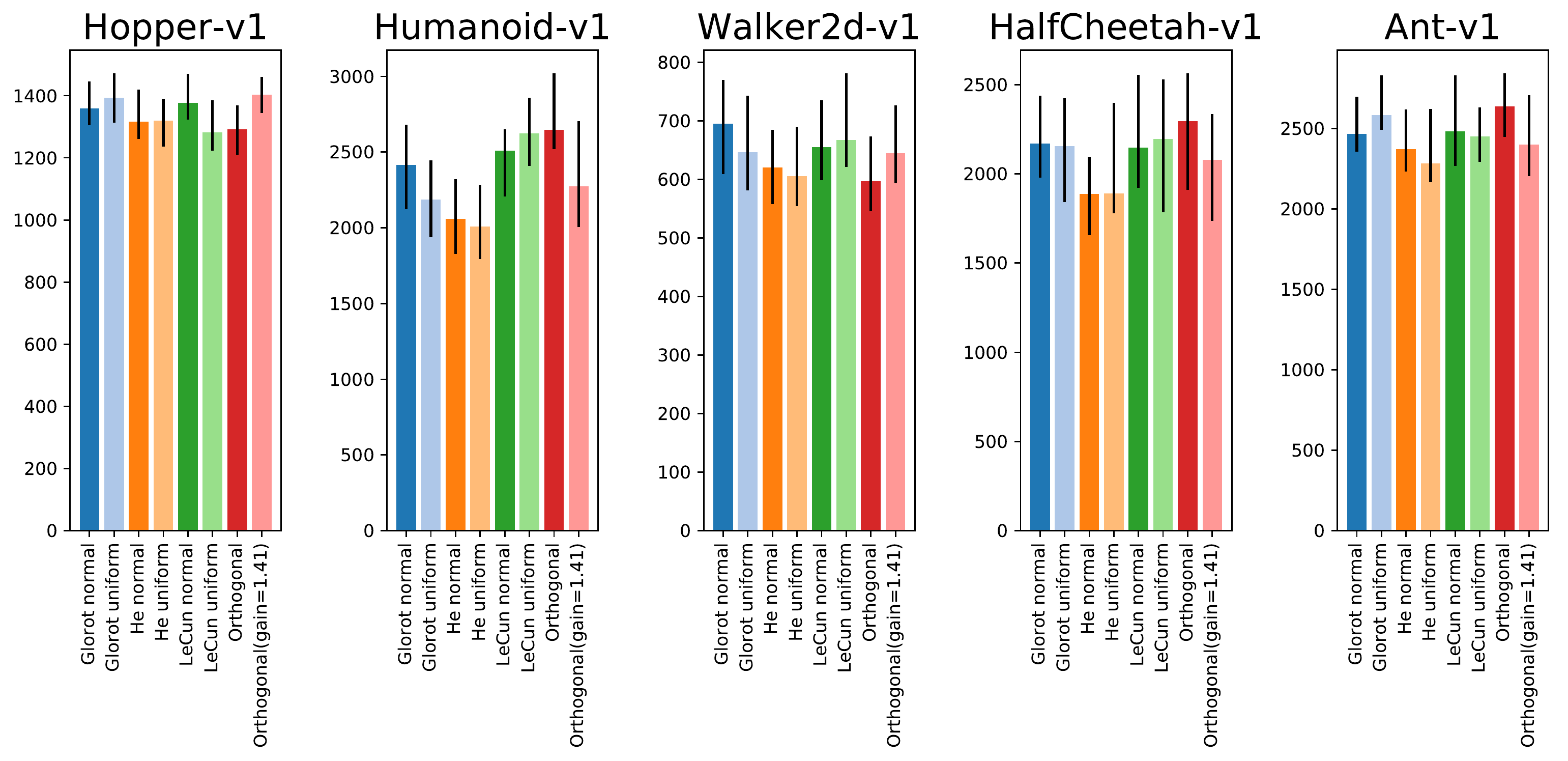}\hspace{1cm}\includegraphics[width=0.45\textwidth]{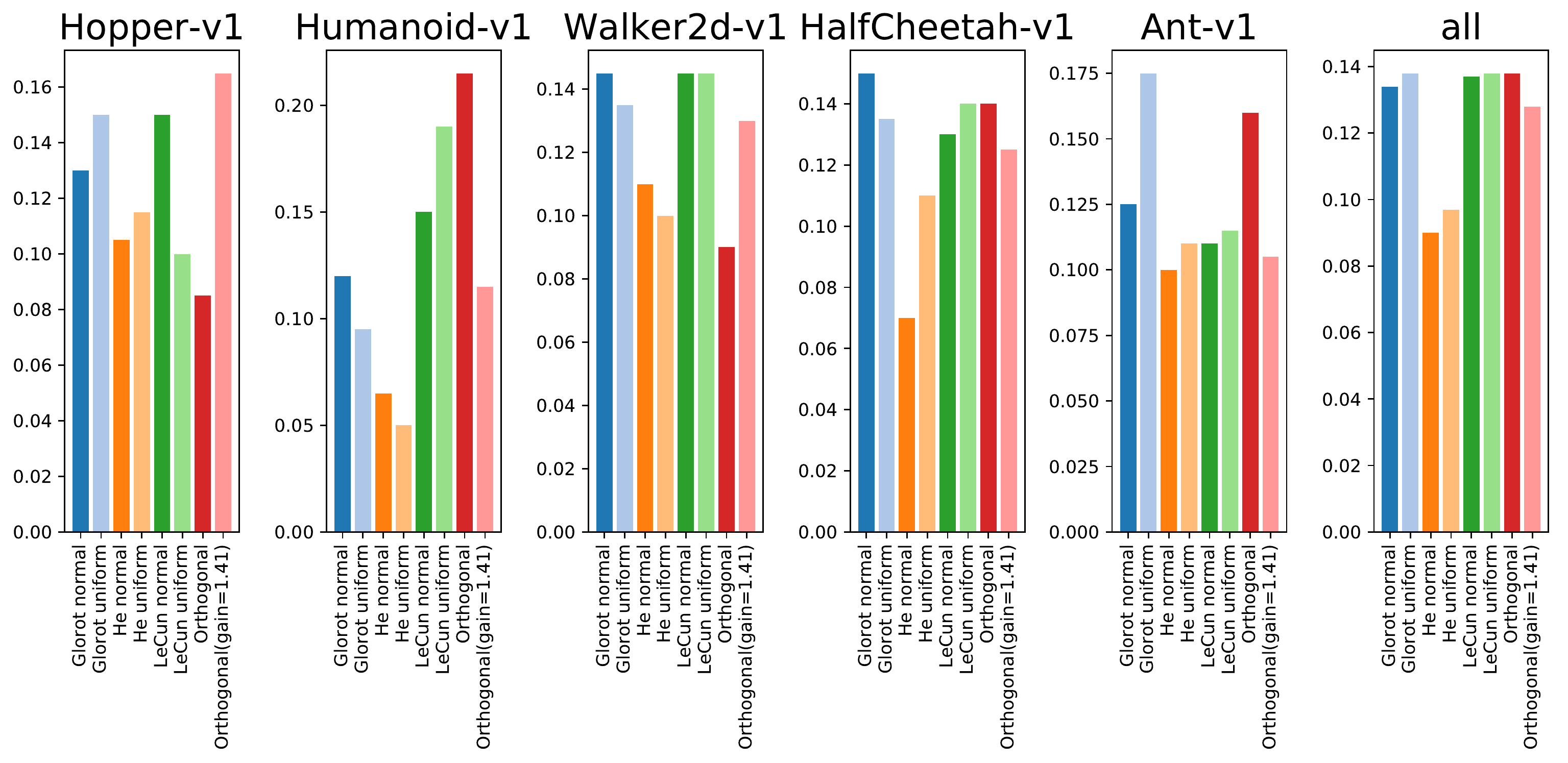}}
\caption{Analysis of choice \choicet{init}: 95th percentile of performance scores conditioned on choice (left) and distribution of choices in top 5\% of configurations (right).}
\label{fig:final_arch2__gin_study_design_choice_value_initializer}
\end{center}
\end{figure}

\begin{figure}[ht]
\begin{center}
\centerline{\includegraphics[width=0.45\textwidth]{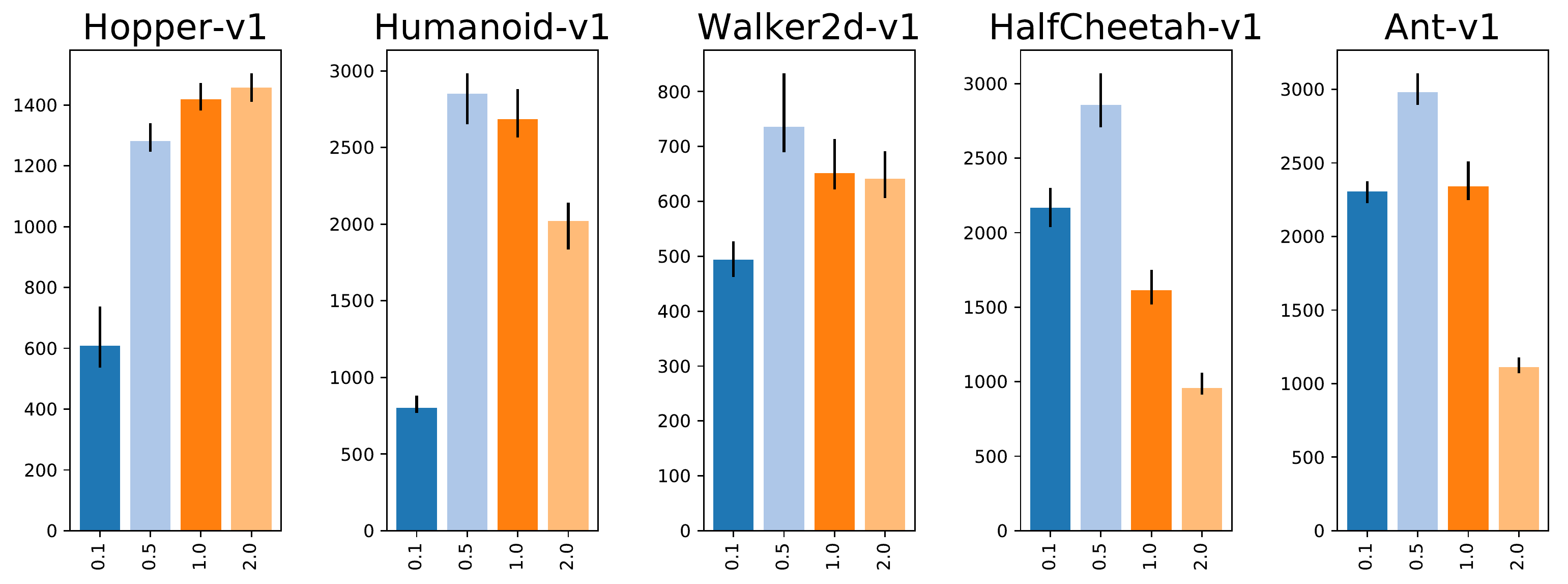}\hspace{1cm}\includegraphics[width=0.45\textwidth]{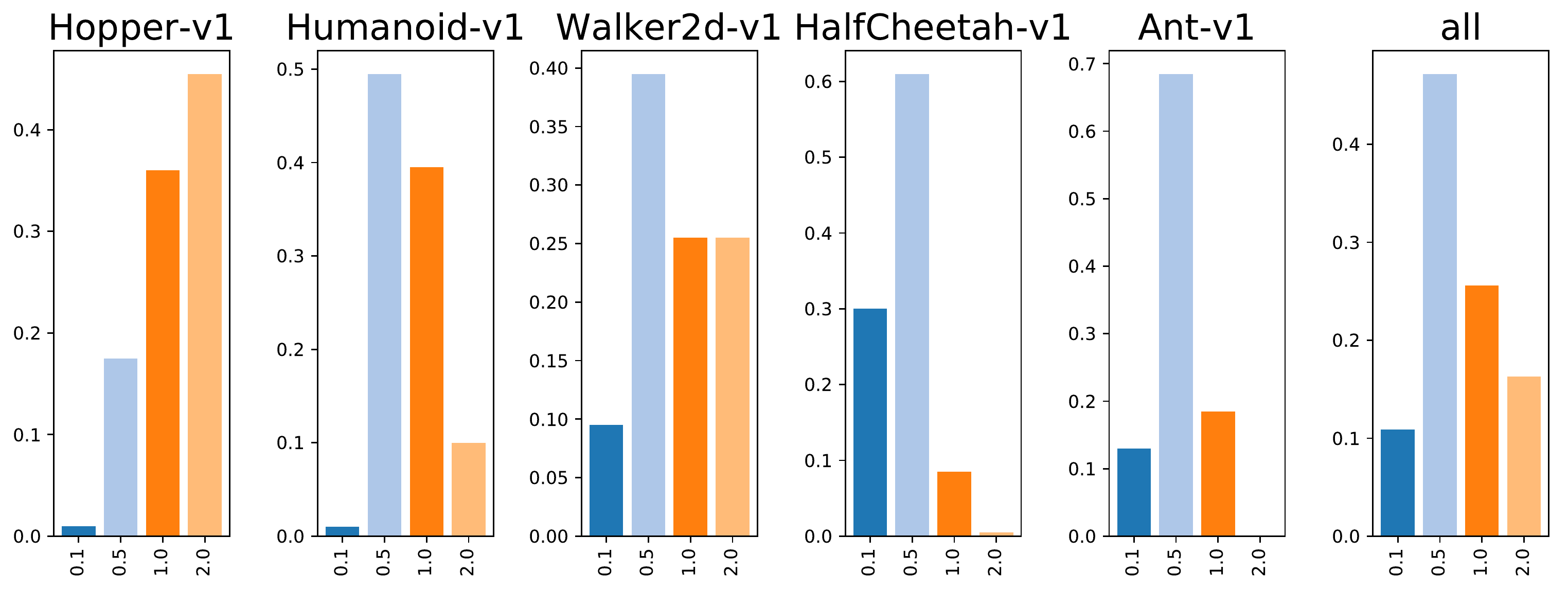}}
\caption{Analysis of choice \choicet{initialstd}: 95th percentile of performance scores conditioned on choice (left) and distribution of choices in top 5\% of configurations (right).}
\label{fig:final_arch2__gin_study_design_choice_value_initial_action_std}
\end{center}
\end{figure}

\begin{figure}[ht]
\begin{center}
\centerline{\includegraphics[width=0.45\textwidth]{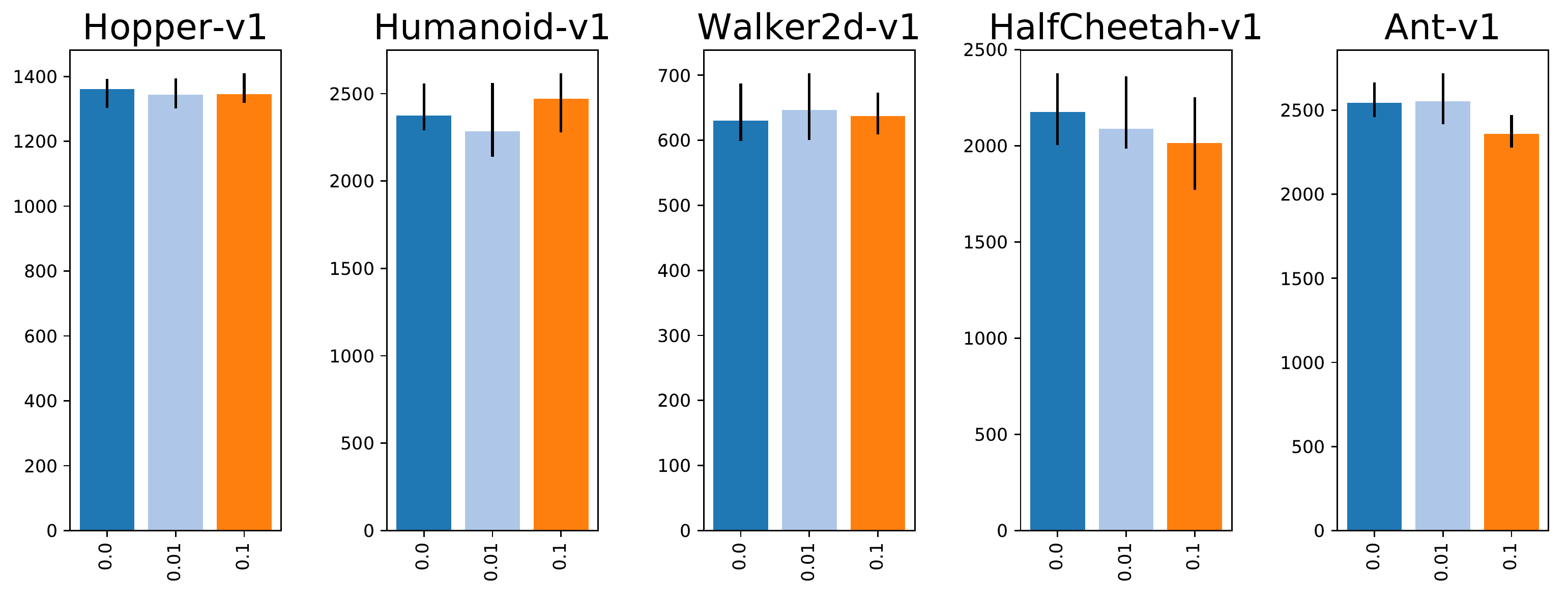}\hspace{1cm}\includegraphics[width=0.45\textwidth]{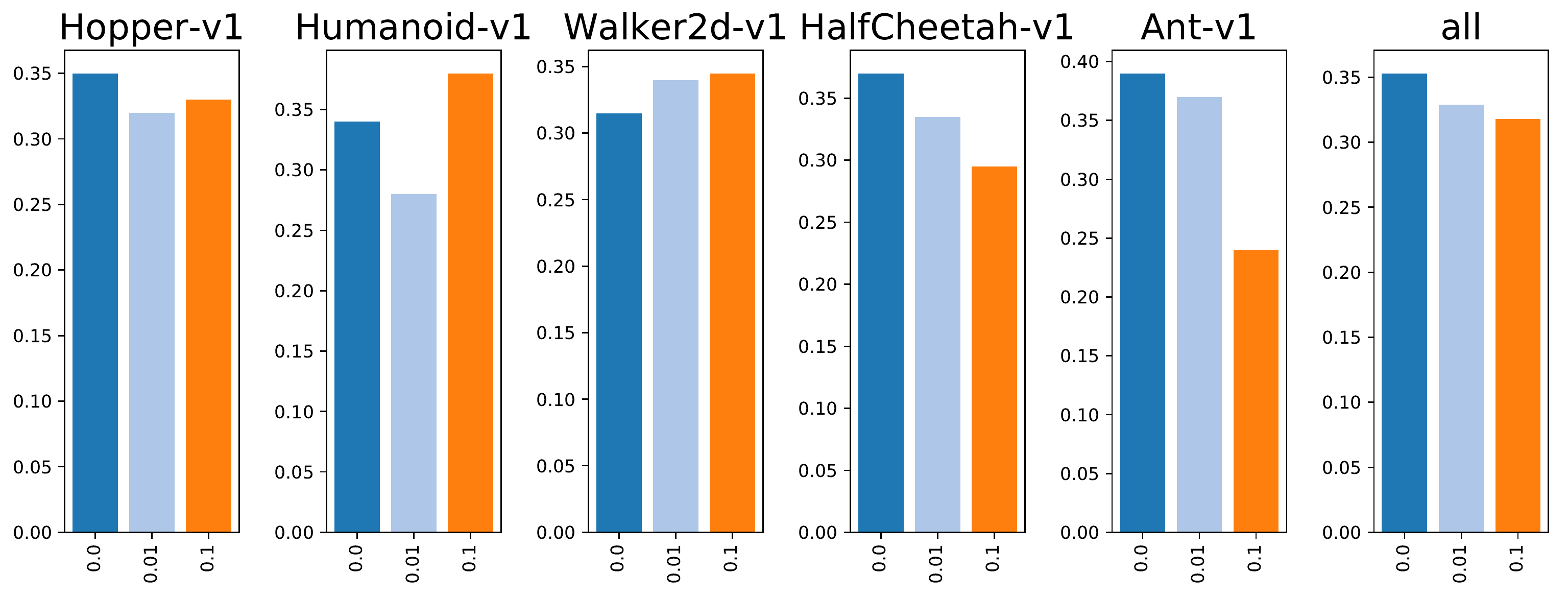}}
\caption{Analysis of choice \choicet{minstd}: 95th percentile of performance scores conditioned on choice (left) and distribution of choices in top 5\% of configurations (right).}
\label{fig:final_arch2__gin_study_design_choice_value_minimum_action_std}
\end{center}
\end{figure}

\begin{figure}[ht]
\begin{center}
\centerline{\includegraphics[width=0.45\textwidth]{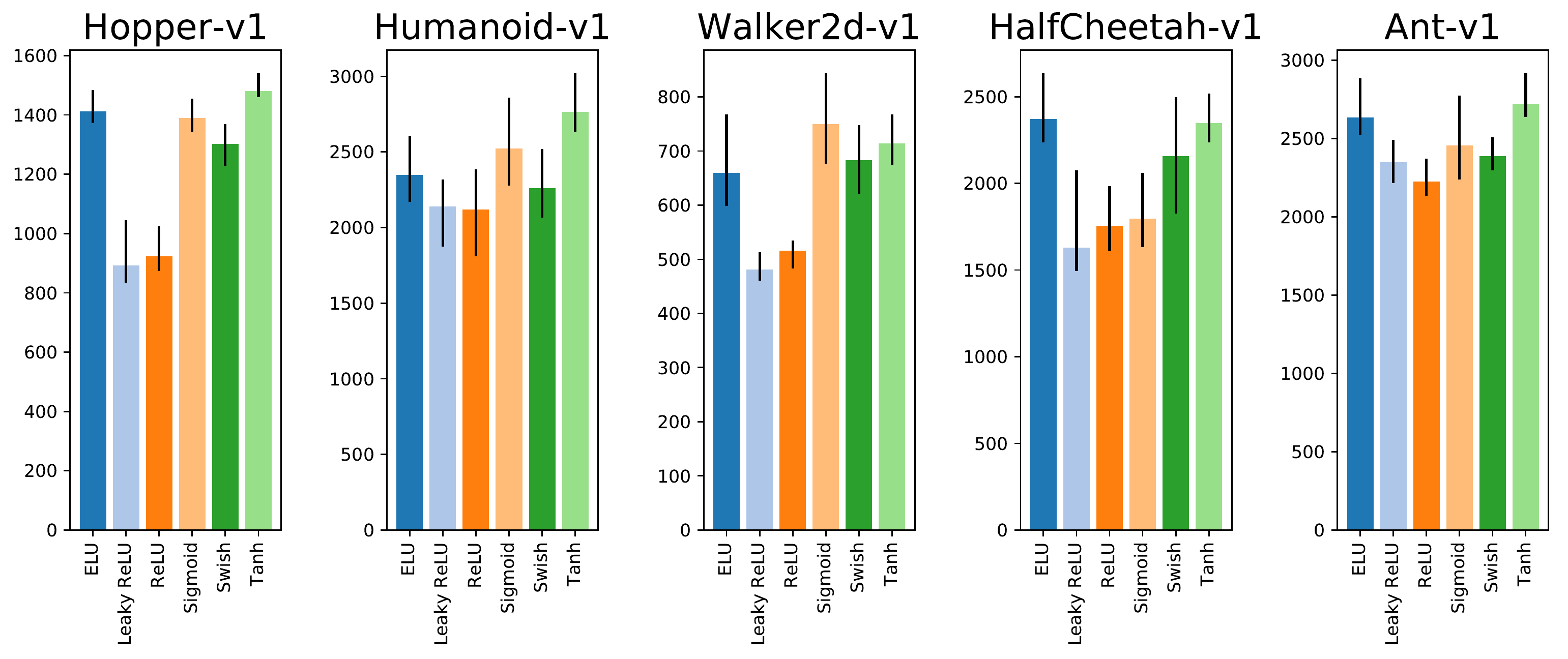}\hspace{1cm}\includegraphics[width=0.45\textwidth]{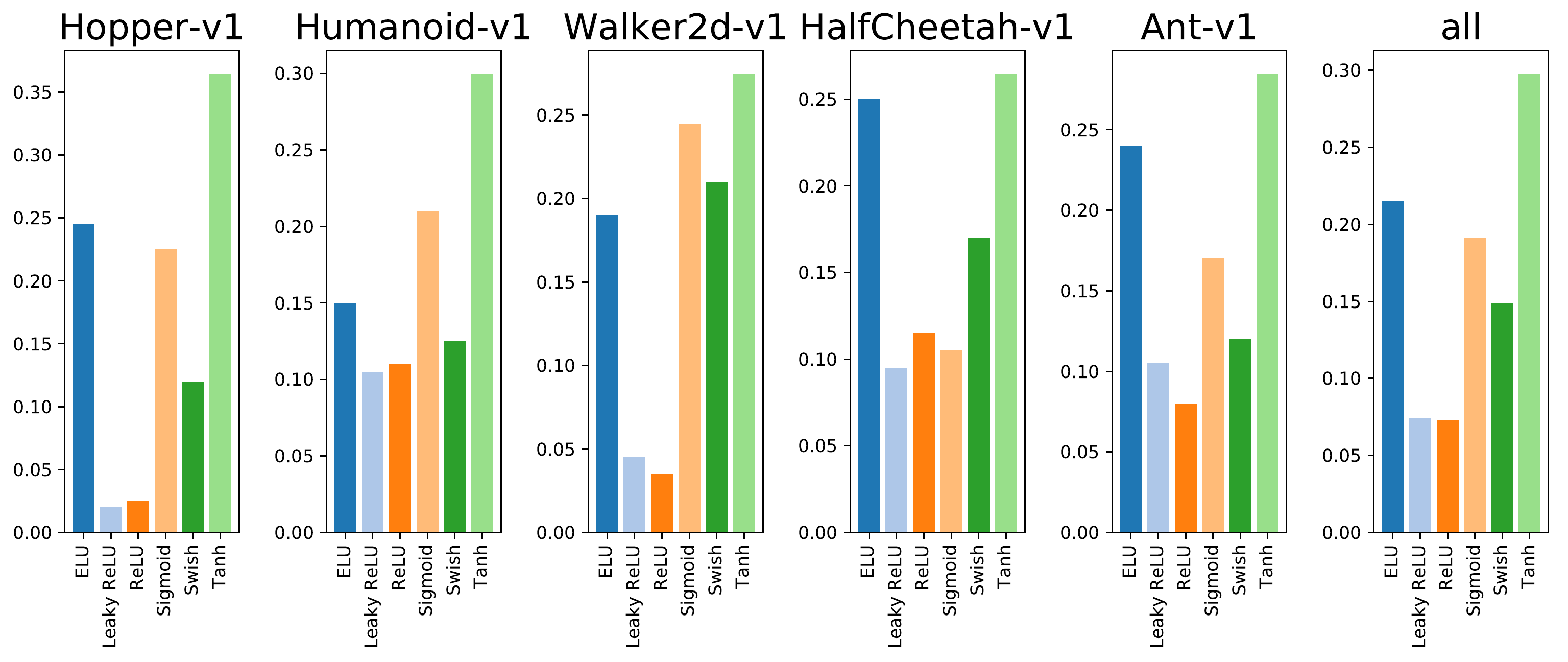}}
\caption{Analysis of choice \choicet{activation}: 95th percentile of performance scores conditioned on choice (left) and distribution of choices in top 5\% of configurations (right).}
\label{fig:final_arch2__gin_study_design_choice_value_activation}
\end{center}
\end{figure}

We report aggregate statistics of the experiment in Table~\ref{tab:final_arch2_overview} as well as training curves in Figure~\ref{fig:final_arch2_training_curves}.
For each of the investigated choices in this experiment, we further provide a per-choice analysis in Figures~\ref{fig:final_arch2__gin_study_design_choice_value_action_postprocessing}-\ref{fig:final_arch2__gin_study_design_choice_value_activation}.
\begin{table}[ht]
\begin{center}
\caption{Performance quantiles across choice configurations.}
\label{tab:final_arch2_overview}
\begin{tabular}{lrrrrr}
\toprule
{} & Ant-v1 & HalfCheetah-v1 & Hopper-v1 & Humanoid-v1 & Walker2d-v1 \\
\midrule
90th percentile &   2098 &           1513 &      1133 &        1817 &         528 \\
95th percentile &   2494 &           2120 &      1349 &        2382 &         637 \\
99th percentile &   3138 &           3031 &      1582 &        3202 &         934 \\
Max             &   4112 &           4358 &      1875 &        3987 &        1265 \\
\bottomrule
\end{tabular}

\end{center}
\end{table}

\clearpage

%% file: final_stability/main.tex
\clearpage
\section{Experiment \texttt{Normalization and clipping}}
\label{exp_final_stability}
\subsection{Design}
\label{exp_design_final_stability}
For each of the 5 environments, we sampled 2000 choice configurations where we sampled the following choices independently and uniformly from the following ranges:
\begin{itemize}
    \item \choicet{ppoepsilon}: \{0.1, 0.2, 0.3, 0.5\}
    \item \choicet{norminput}: \{Average, None\}
    \begin{itemize}
        \item For the case ``\choicet{norminput} = Average'', we further sampled the sub-choices:
        \begin{itemize}
            \item \choicet{clipinput}: \{1.0, 2.0, 5.0, 10.0, None\}
        \end{itemize}
    \end{itemize}
    \item \choicet{clipgrad}: \{0.5, 1.0, 2.0, 5.0, None\}
    \item \choicet{normadv}: \{False, True\}
    \item \choicet{adamlr}: \{3e-05, 0.0001, 0.0003, 0.001\}
    \item \choicet{normreward}: \{Average, None\}
\end{itemize}
All the other choices were set to the default values as described in Appendix~\ref{sec:default_settings}.

For each of the sampled choice configurations, we train 3 agents with different random seeds and compute the performance metric as described in Section~\ref{sec:performance}.
\subsection{Results}
\label{exp_results_final_stability}
\begin{figure}[ht]
\begin{center}
\centerline{\includegraphics[width=1\textwidth]{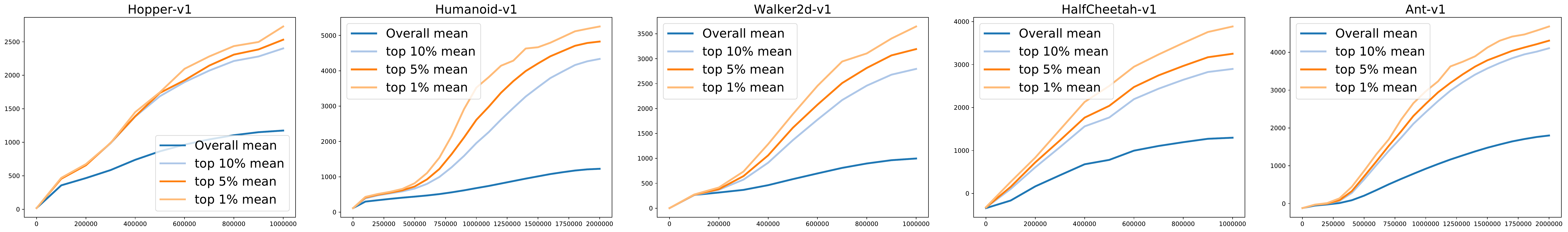}}
\caption{Training curves.}
\label{fig:final_stability_training_curves}
\end{center}
\end{figure}

\begin{figure}[ht]
\begin{center}
\centerline{\includegraphics[width=0.45\textwidth]{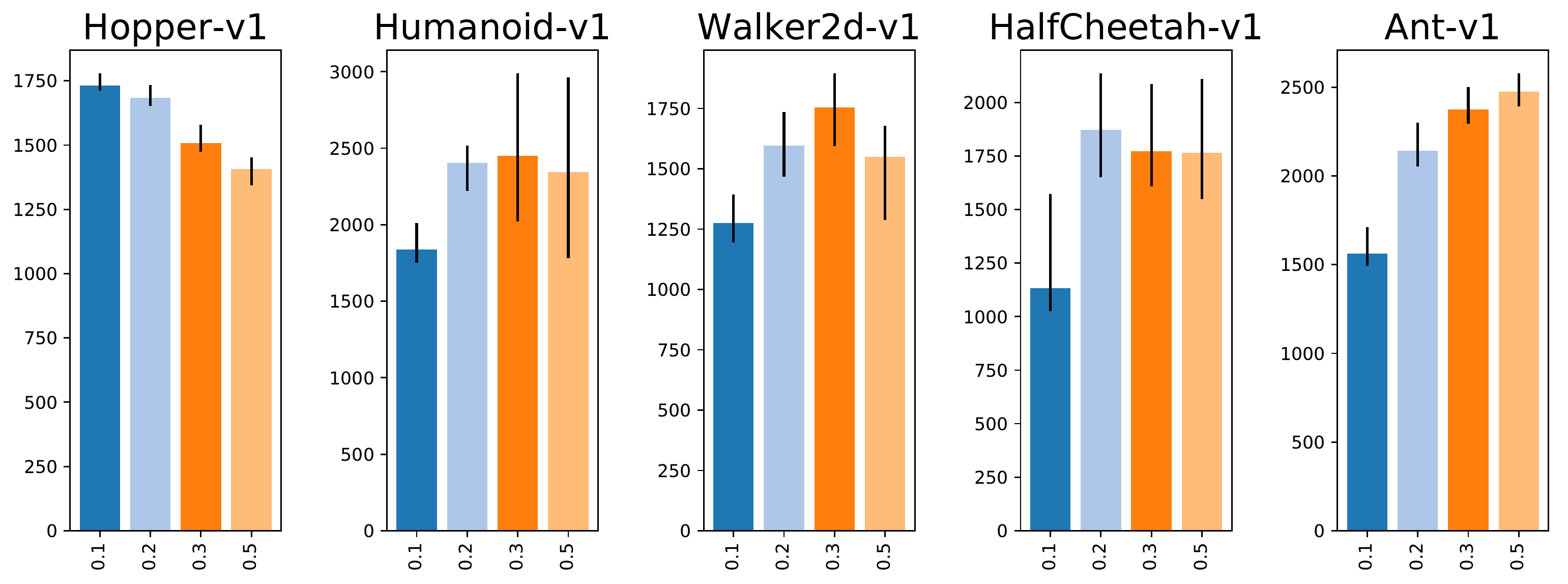}\hspace{1cm}\includegraphics[width=0.45\textwidth]{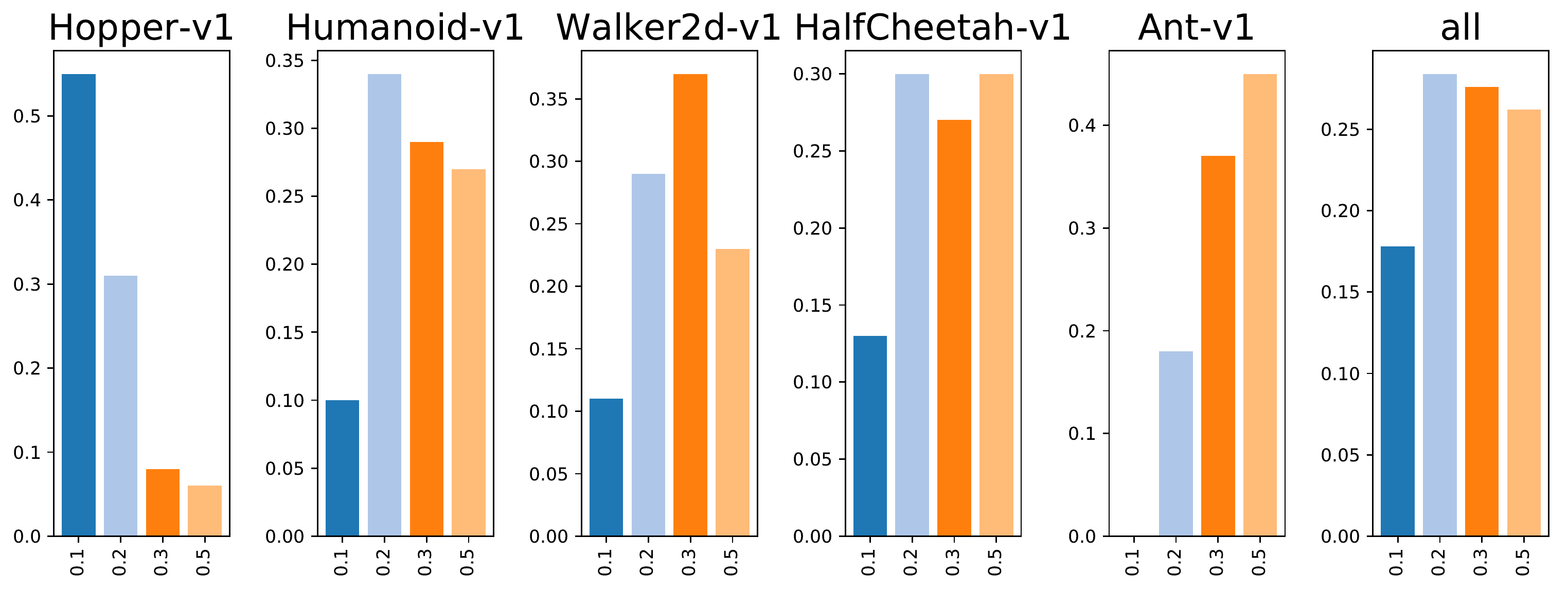}}
\caption{Analysis of choice \choicet{ppoepsilon}: 95th percentile of performance scores conditioned on choice (left) and distribution of choices in top 5\% of configurations (right).}
\label{fig:final_stability__gin_study_design_choice_value_ppo_epsilon}
\end{center}
\end{figure}

\begin{figure}[ht]
\begin{center}
\centerline{\includegraphics[width=0.45\textwidth]{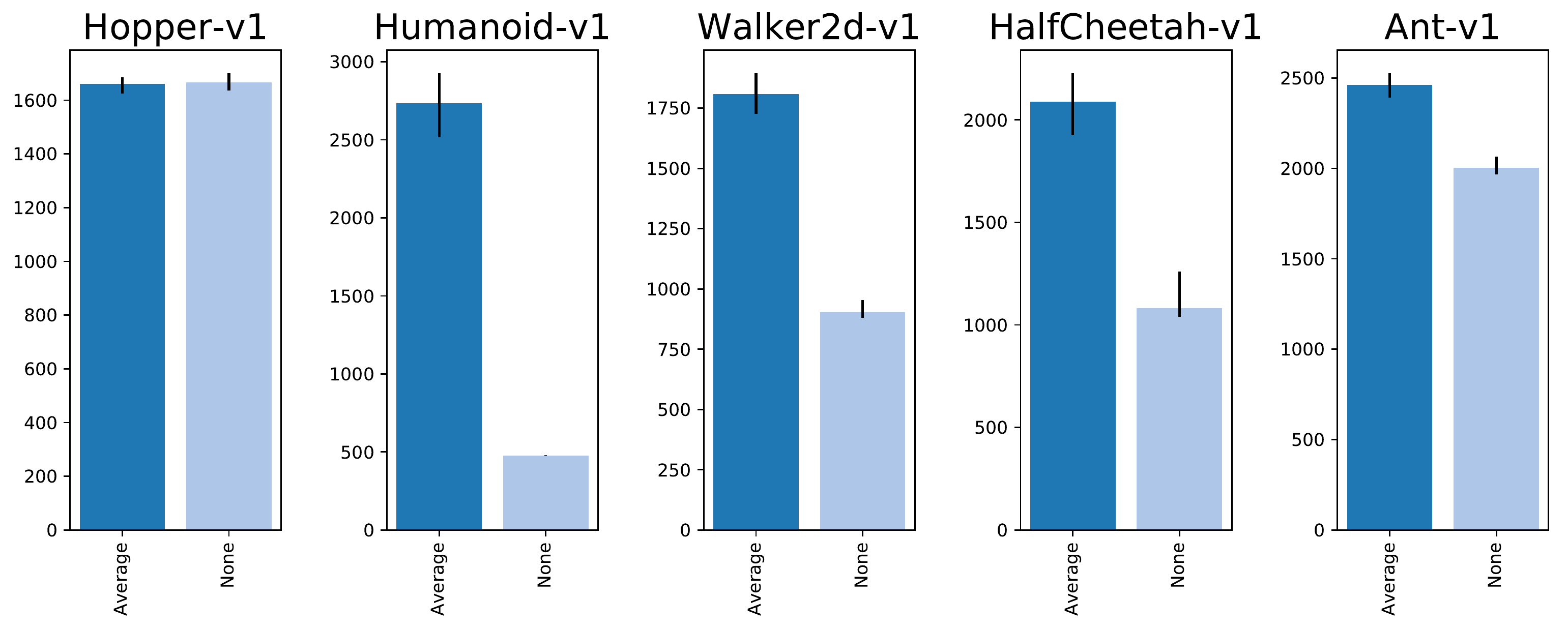}\hspace{1cm}\includegraphics[width=0.45\textwidth]{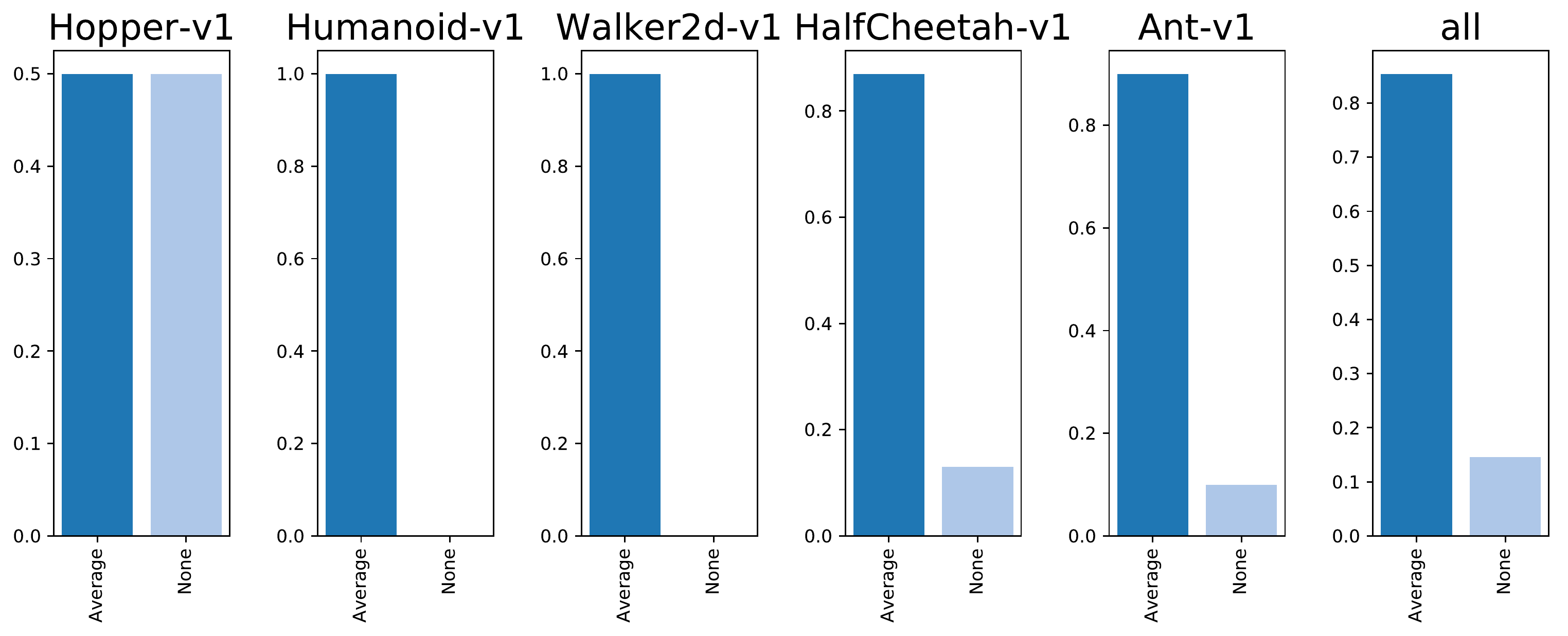}}
\caption{Analysis of choice \choicet{norminput}: 95th percentile of performance scores conditioned on choice (left) and distribution of choices in top 5\% of configurations (right).}
\label{fig:final_stability__gin_study_design_choice_value_input_normalization}
\end{center}
\end{figure}

\begin{figure}[ht]
\begin{center}
\centerline{\includegraphics[width=0.45\textwidth]{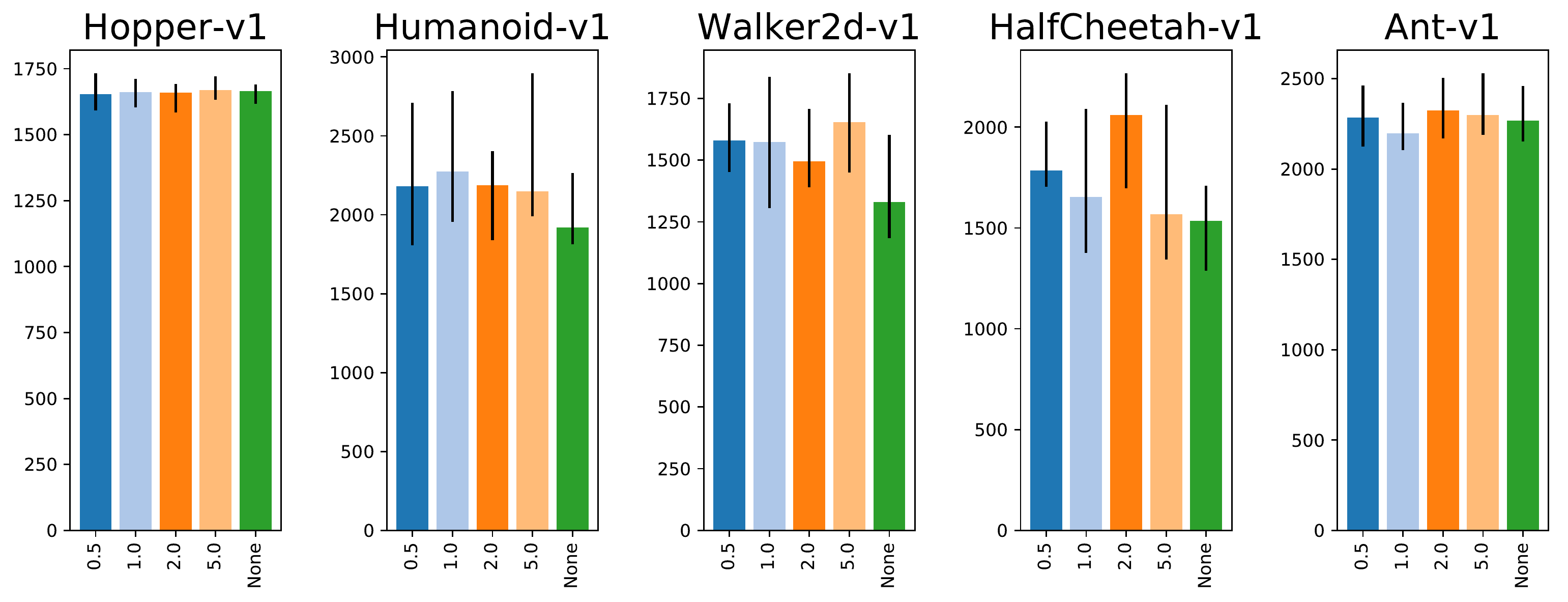}\hspace{1cm}\includegraphics[width=0.45\textwidth]{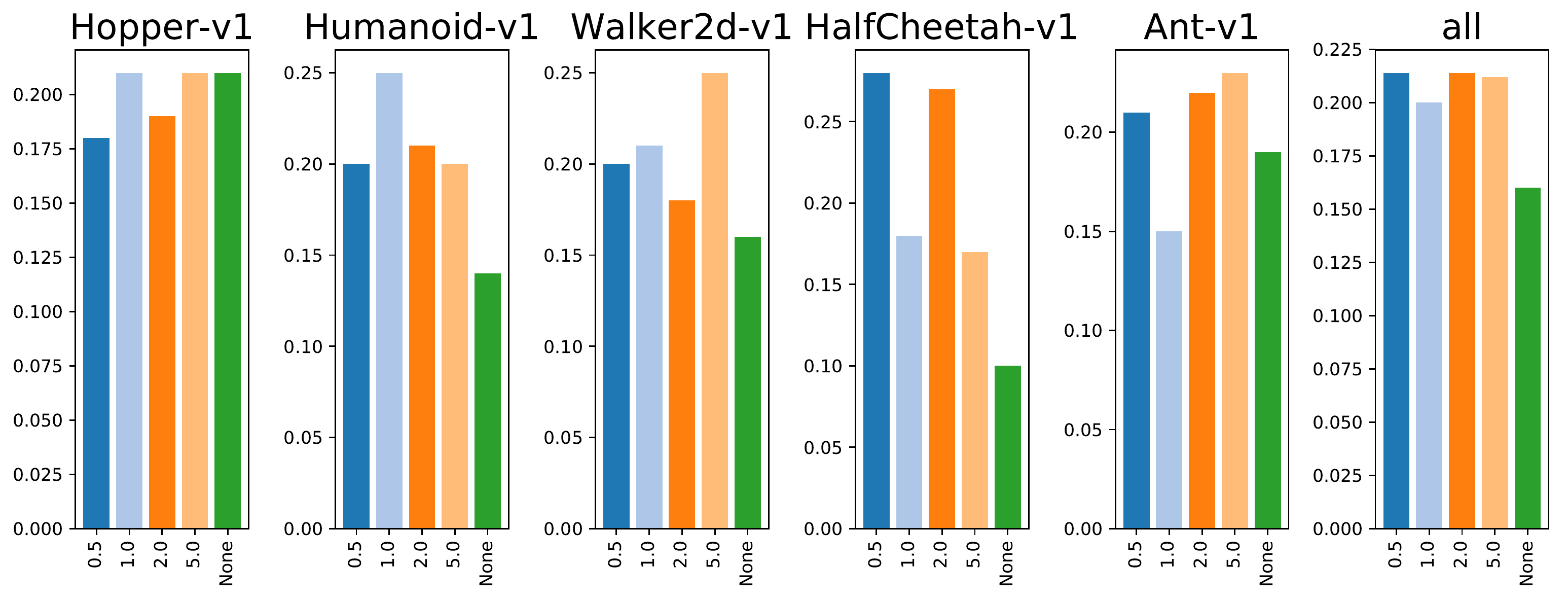}}
\caption{Analysis of choice \choicet{clipgrad}: 95th percentile of performance scores conditioned on choice (left) and distribution of choices in top 5\% of configurations (right).}
\label{fig:final_stability__gin_study_design_choice_value_gradient_clipping}
\end{center}
\end{figure}

\begin{figure}[ht]
\begin{center}
\centerline{\includegraphics[width=0.45\textwidth]{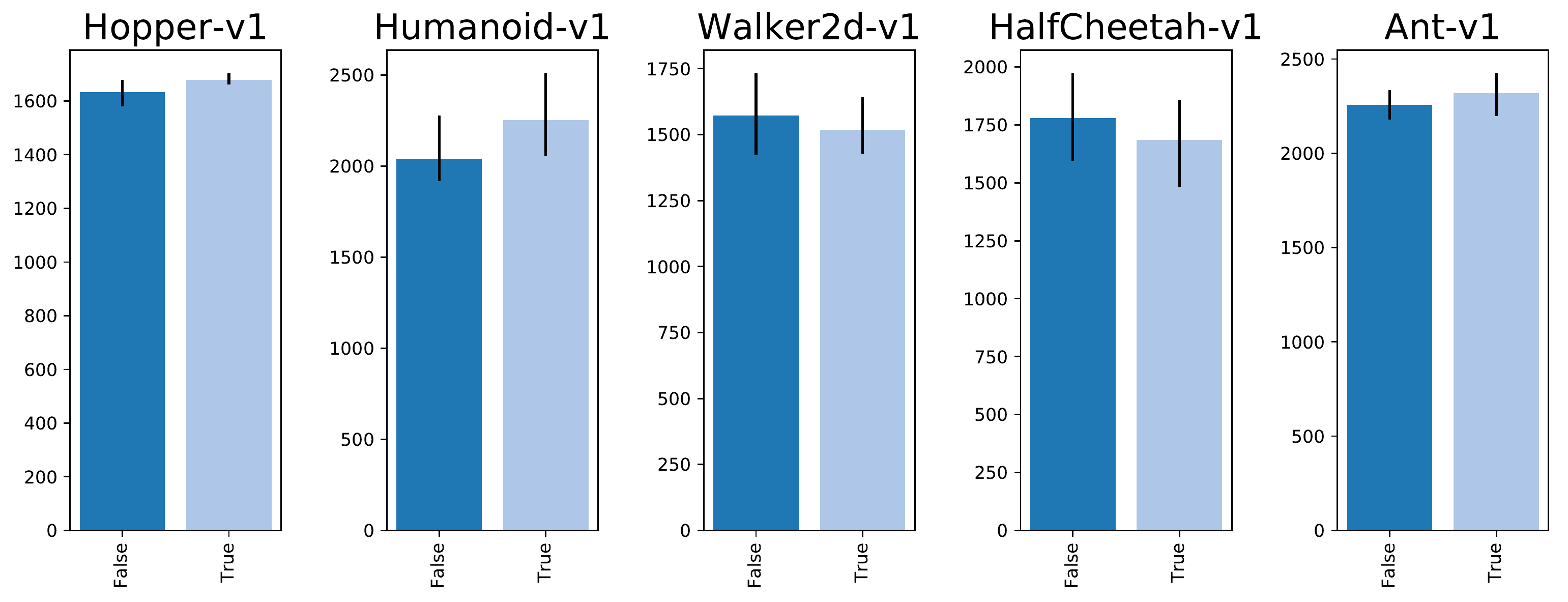}\hspace{1cm}\includegraphics[width=0.45\textwidth]{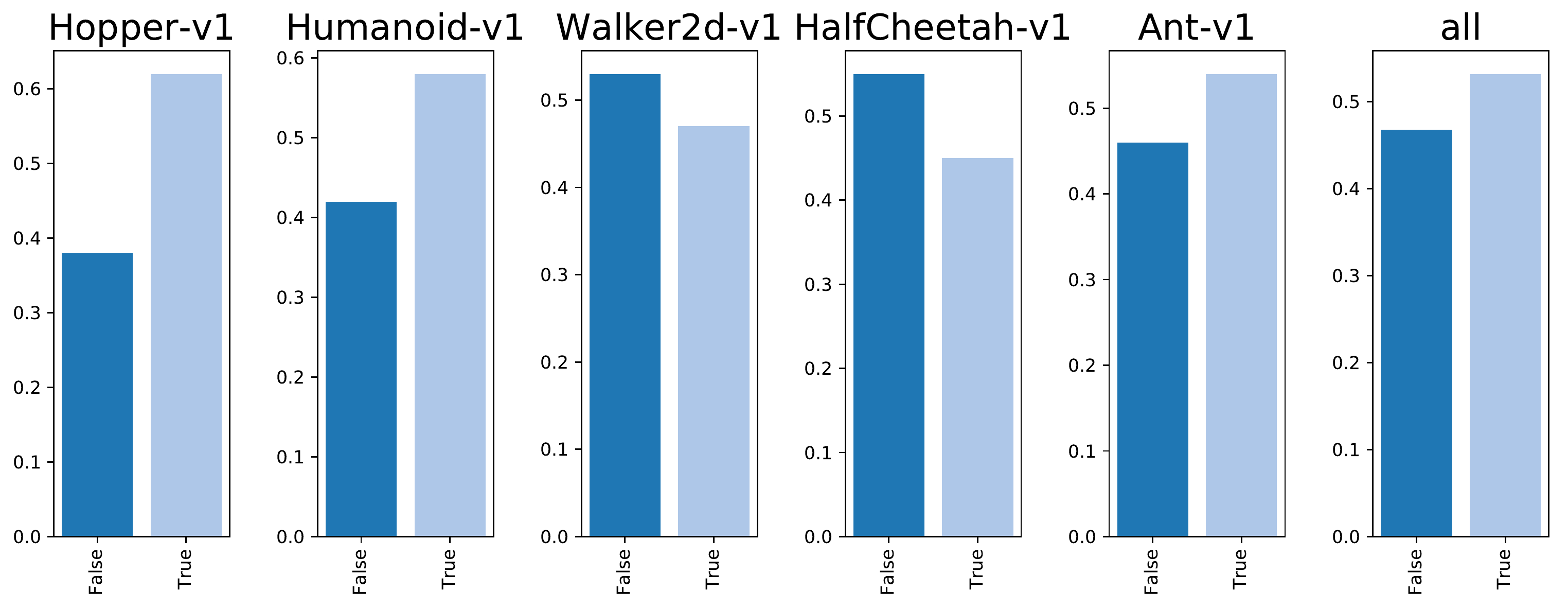}}
\caption{Analysis of choice \choicet{normadv}: 95th percentile of performance scores conditioned on choice (left) and distribution of choices in top 5\% of configurations (right).}
\label{fig:final_stability__gin_study_design_choice_value_normalize_advantages}
\end{center}
\end{figure}

\begin{figure}[ht]
\begin{center}
\centerline{\includegraphics[width=0.45\textwidth]{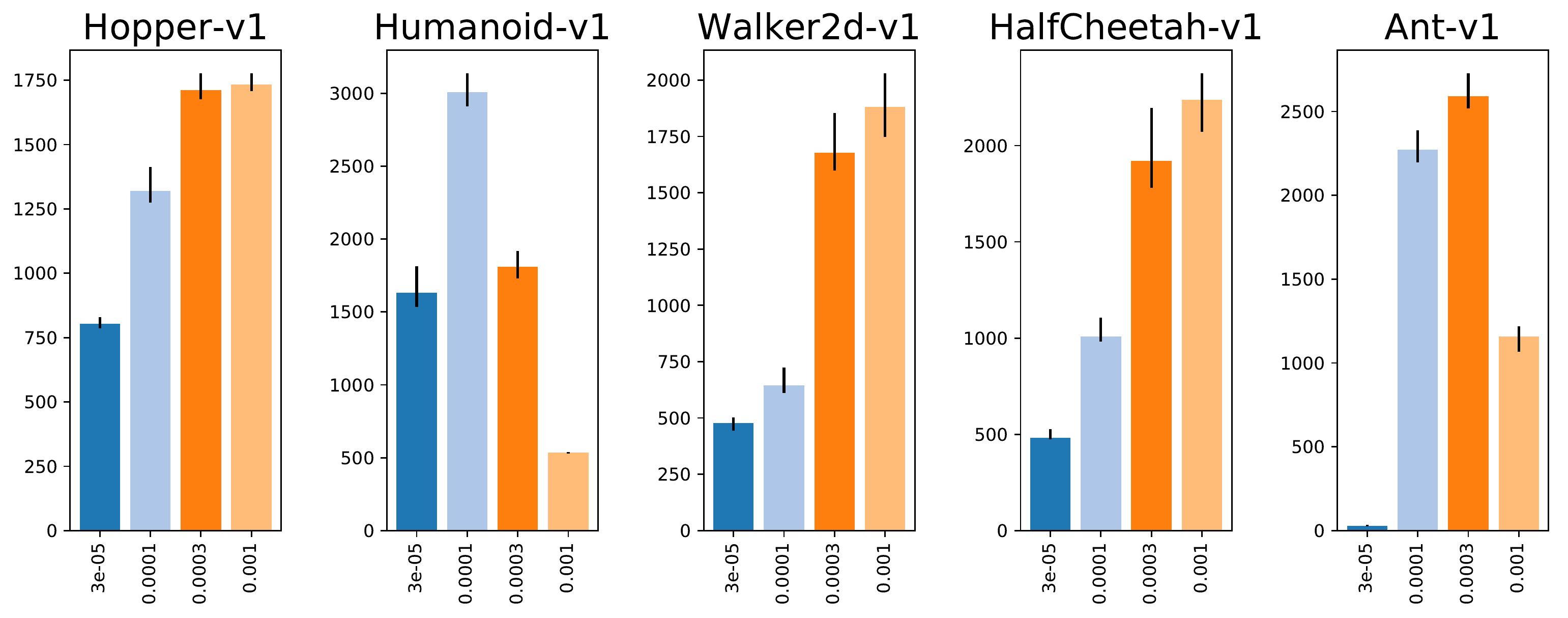}\hspace{1cm}\includegraphics[width=0.45\textwidth]{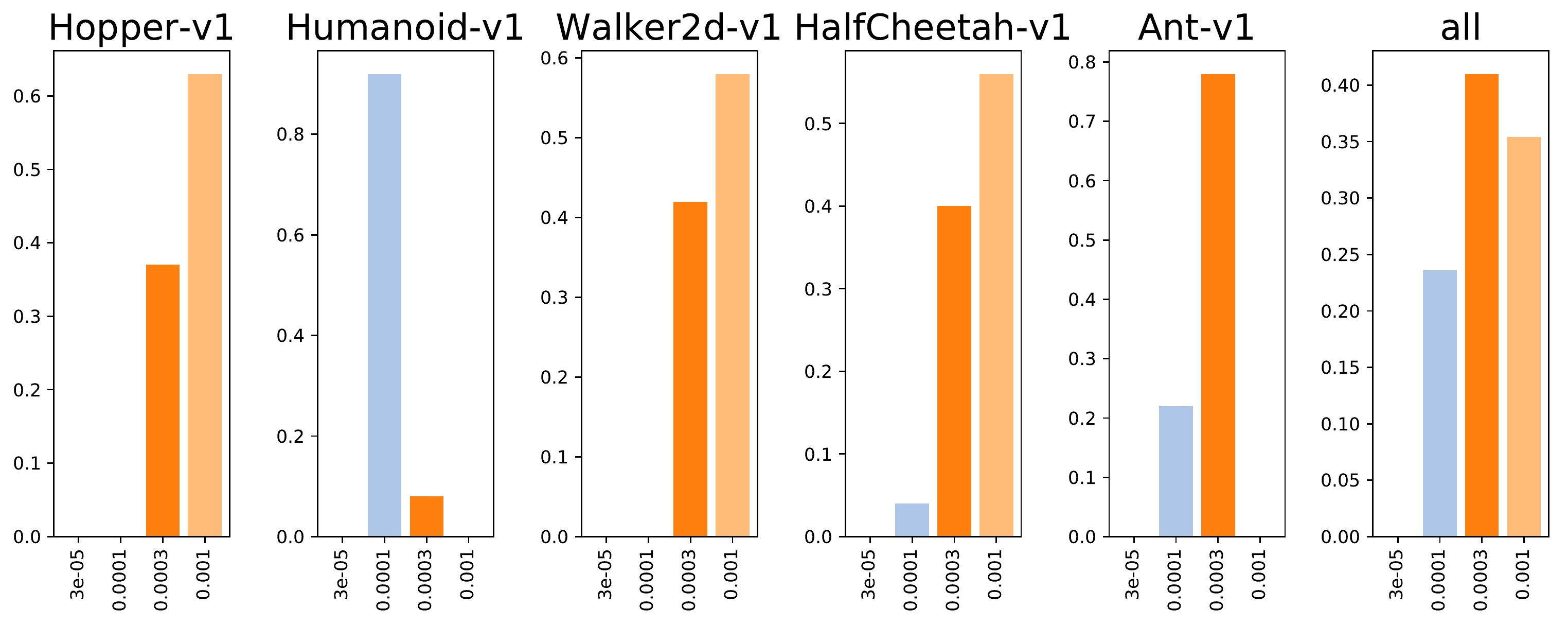}}
\caption{Analysis of choice \choicet{adamlr}: 95th percentile of performance scores conditioned on choice (left) and distribution of choices in top 5\% of configurations (right).}
\label{fig:final_stability__gin_study_design_choice_value_learning_rate}
\end{center}
\end{figure}

\begin{figure}[ht]
\begin{center}
\centerline{\includegraphics[width=0.45\textwidth]{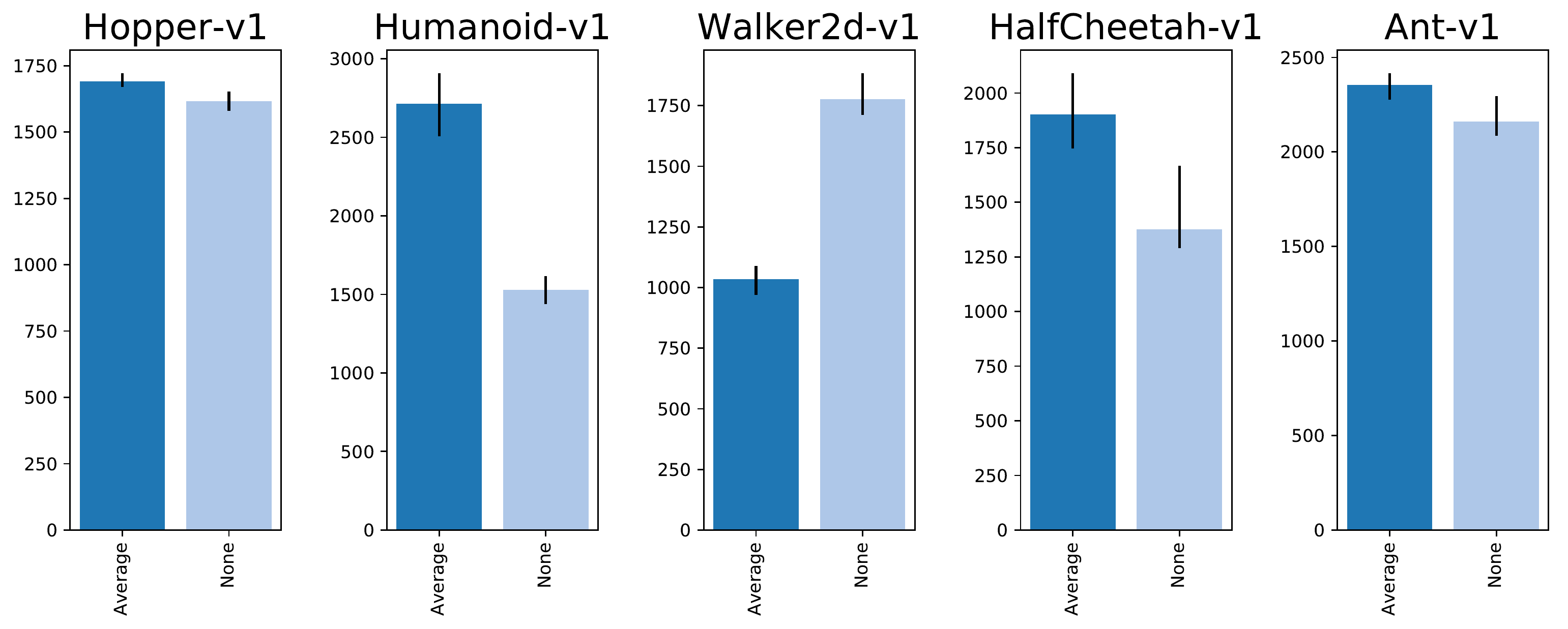}\hspace{1cm}\includegraphics[width=0.45\textwidth]{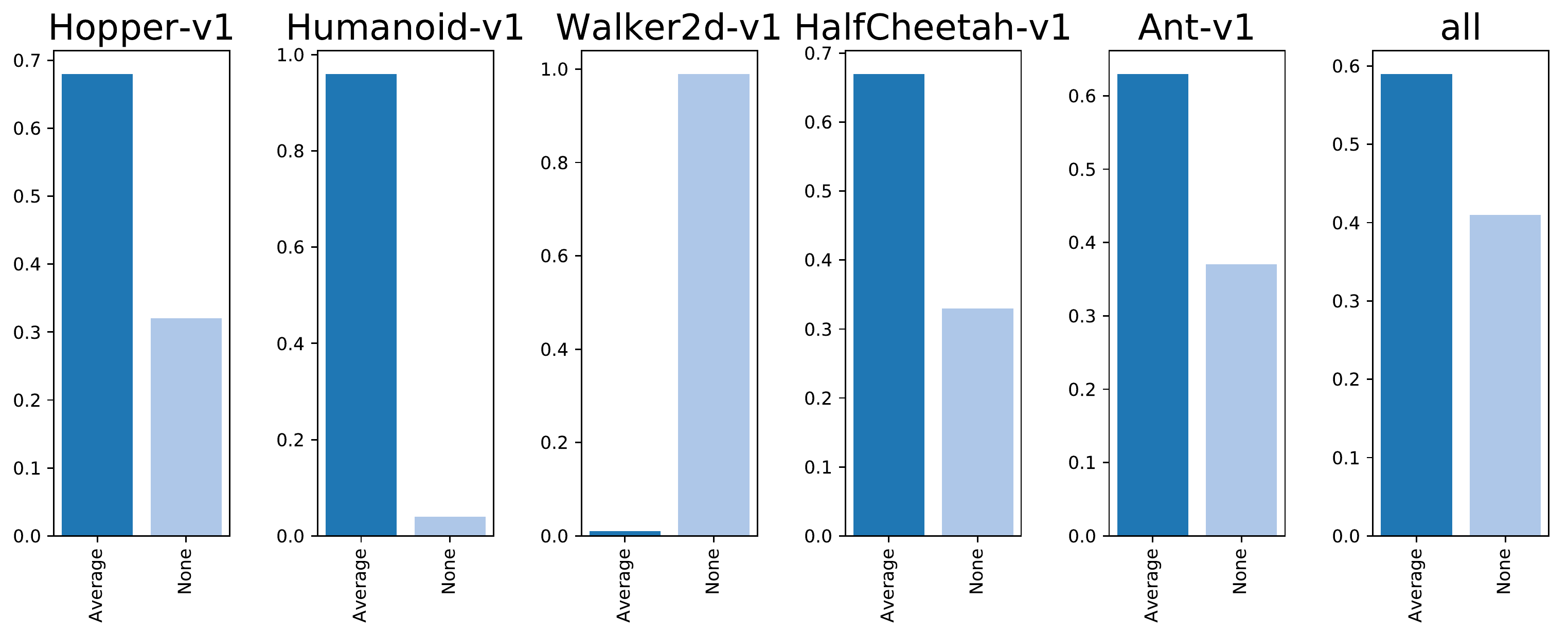}}
\caption{Analysis of choice \choicet{normreward}: 95th percentile of performance scores conditioned on choice (left) and distribution of choices in top 5\% of configurations (right).}
\label{fig:final_stability__gin_study_design_choice_value_reward_normalization}
\end{center}
\end{figure}

\begin{figure}[ht]
\begin{center}
\centerline{\includegraphics[width=0.45\textwidth]{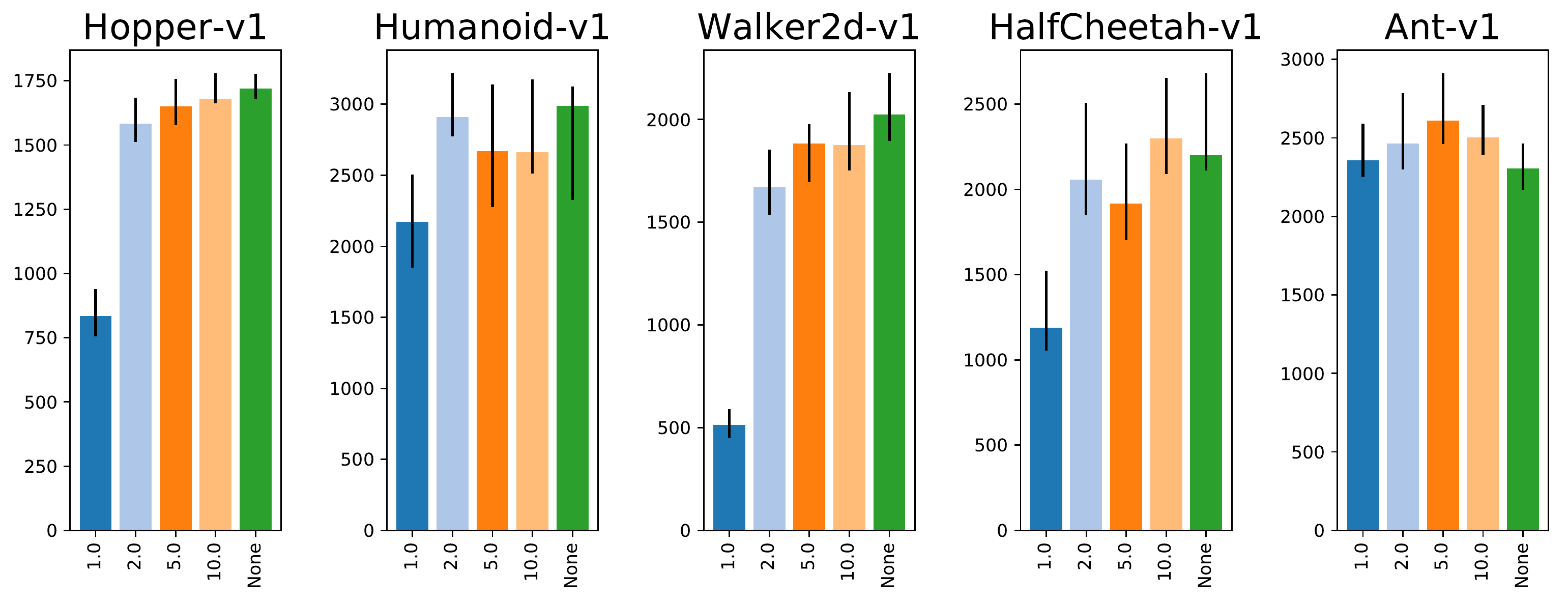}\hspace{1cm}\includegraphics[width=0.45\textwidth]{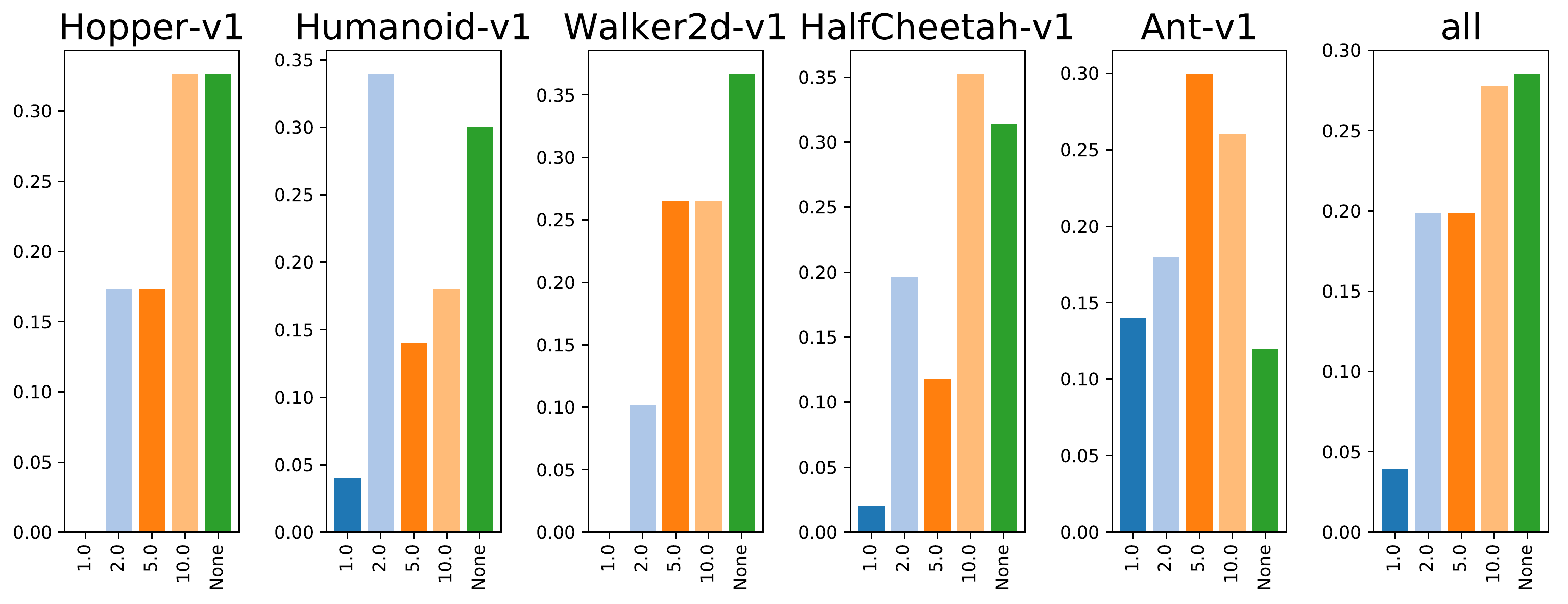}}
\caption{Analysis of choice \choicet{clipinput}: 95th percentile of performance scores conditioned on sub-choice (left) and distribution of sub-choices in top 5\% of configurations (right).}
\label{fig:final_stability__gin_study_design_choice_value_sub_input_normalization_avg_compfalse_input_clipping}
\end{center}
\end{figure}

We report aggregate statistics of the experiment in Table~\ref{tab:final_stability_overview} as well as training curves in Figure~\ref{fig:final_stability_training_curves}.
For each of the investigated choices in this experiment, we further provide a per-choice analysis in Figures~\ref{fig:final_stability__gin_study_design_choice_value_ppo_epsilon}-\ref{fig:final_stability__gin_study_design_choice_value_sub_input_normalization_avg_compfalse_input_clipping}.
\begin{table}[ht]
\begin{center}
\caption{Performance quantiles across choice configurations.}
\label{tab:final_stability_overview}
\begin{tabular}{lrrrrr}
\toprule
{} & Ant-v1 & HalfCheetah-v1 & Hopper-v1 & Humanoid-v1 & Walker2d-v1 \\
\midrule
90th percentile &   2058 &           1265 &      1533 &        1649 &        1143 \\
95th percentile &   2287 &           1716 &      1662 &        2165 &        1564 \\
99th percentile &   2662 &           2465 &      1809 &        3100 &        2031 \\
Max             &   3333 &           3515 &      2074 &        3482 &        2371 \\
\bottomrule
\end{tabular}

\end{center}
\end{table}

\clearpage

%% file: final_advantages/main.tex
\clearpage
\section{Experiment \texttt{Advantage Estimation}}
\label{exp_final_advantages}
\subsection{Design}
\label{exp_design_final_advantages}
For each of the 5 environments, we sampled 4000 choice configurations where we sampled the following choices independently and uniformly from the following ranges:
\begin{itemize}
    \item \choicet{numenvs}: \{64, 128, 256\}
    \item \choicet{valueloss}: \{Huber, MSE\}
    \begin{itemize}
        \item For the case ``\choicet{valueloss} = Huber'', we further sampled the sub-choices:
        \begin{itemize}
            \item \choicet{huberdelta}: \{0.001, 0.01, 0.1, 1.0\}
        \end{itemize}
    \end{itemize}
    \item \choicet{ppovalueclip}: \{0.001, 0.01, 0.1, 1.0, None\}
    \item \choicet{advantageestimator}: \{GAE, N-step, V-Trace\}
    \begin{itemize}
        \item For the case ``\choicet{advantageestimator} = GAE'', we further sampled the sub-choices:
        \begin{itemize}
            \item \choicet{gaelambda}: \{0.8, 0.9, 0.95, 0.99\}
        \end{itemize}
        \item For the case ``\choicet{advantageestimator} = N-step'', we further sampled the sub-choices:
        \begin{itemize}
            \item \choicet{nstep}: \{1, 3, 10, 1000000\}
        \end{itemize}
        \item For the case ``\choicet{advantageestimator} = V-Trace'', we further sampled the sub-choices:
        \begin{itemize}
            \item \choicet{vtraceaelambda}: \{0.8, 0.9, 0.95, 0.99, 1.0\}
            \item \choicet{vtraceaecrho}: \{1.0, 1.2, 1.5, 2.0\}
        \end{itemize}
    \end{itemize}
    \item \choicet{adamlr}: \{3e-05, 0.0001, 0.0003, 0.001, 0.003\}
\end{itemize}
All the other choices were set to the default values as described in Appendix~\ref{sec:default_settings}.

For each of the sampled choice configurations, we train 3 agents with different random seeds and compute the performance metric as described in Section~\ref{sec:performance}.
\subsection{Results}
\label{exp_results_final_advantages}
\begin{figure}[ht]
\begin{center}
\centerline{\includegraphics[width=1\textwidth]{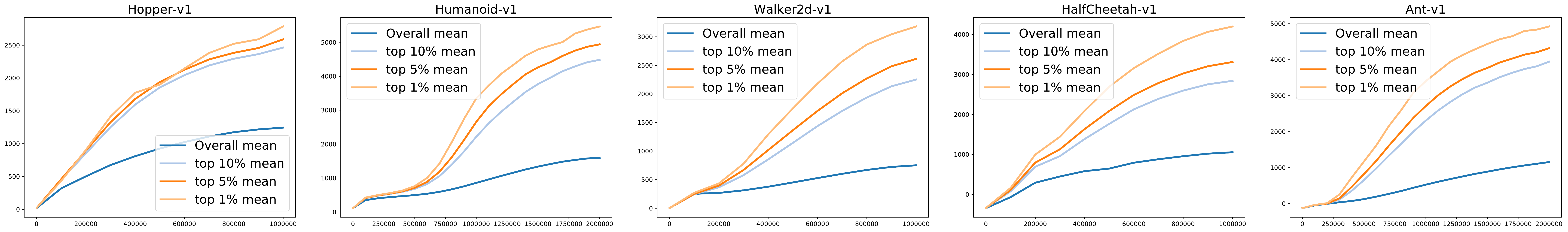}}
\caption{Training curves.}
\label{fig:final_advantages_training_curves}
\end{center}
\end{figure}

\begin{figure}[ht]
\begin{center}
\centerline{\includegraphics[width=1\textwidth]{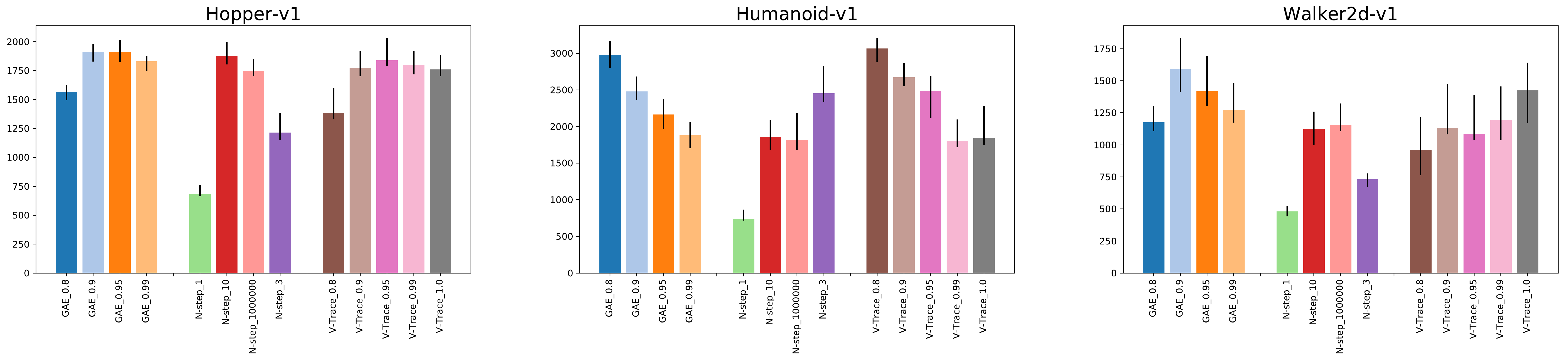}}
\centerline{\includegraphics[width=0.65\textwidth]{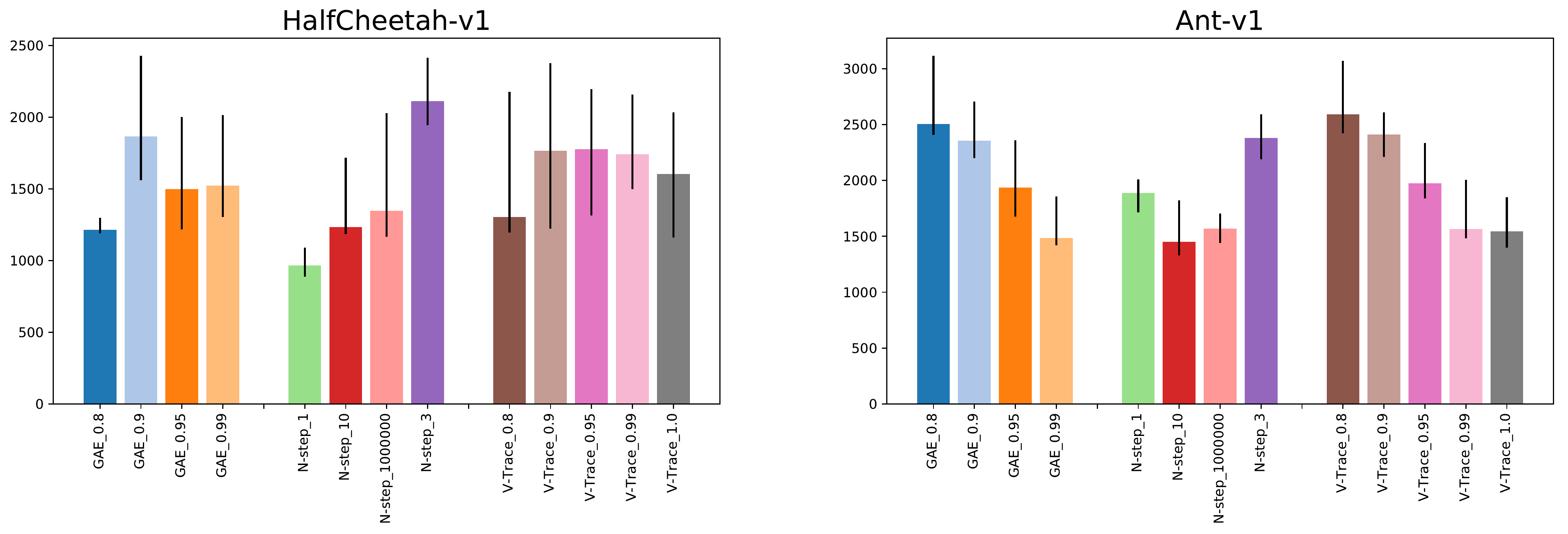}}
\caption{Comparison of 95th percentile of the performance of different advantage estimators conditioned on their hyperparameters.}
\label{fig:final_advantages_custom_advantage_estimator}
\end{center}
\end{figure}

\begin{figure}[ht]
\begin{center}
\centerline{\includegraphics[width=0.45\textwidth]{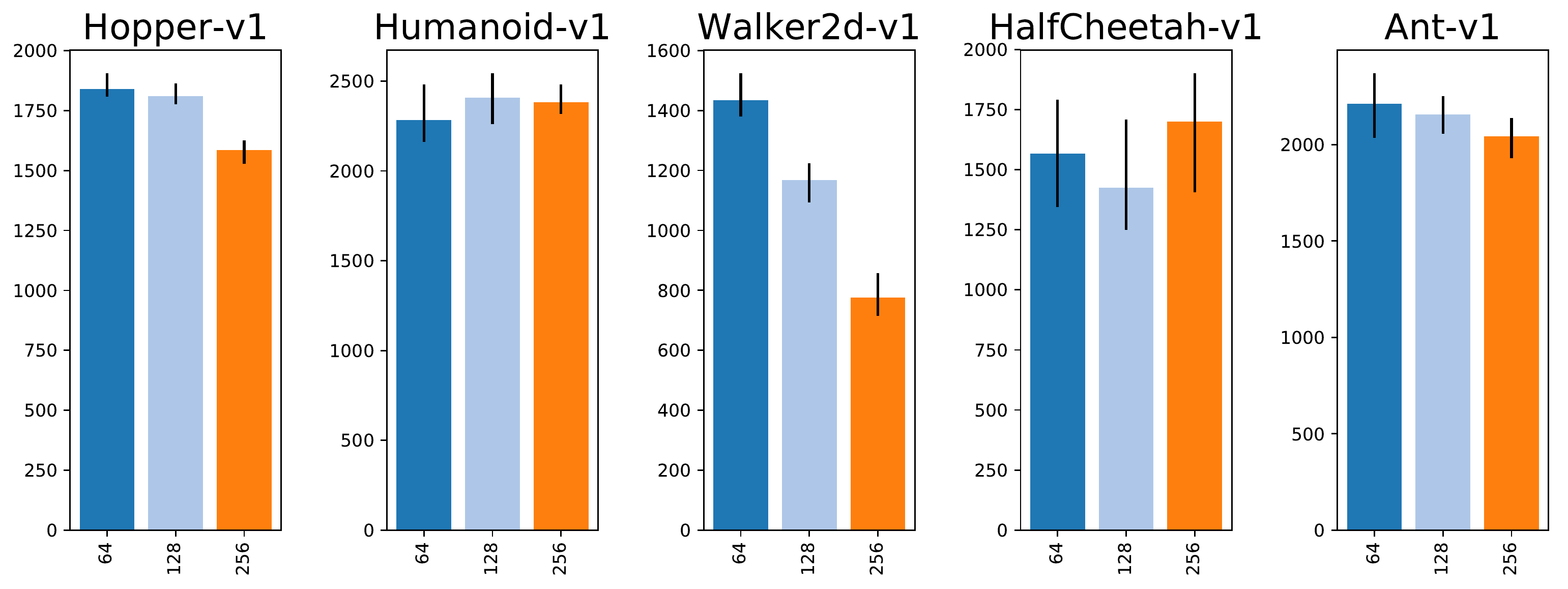}\hspace{1cm}\includegraphics[width=0.45\textwidth]{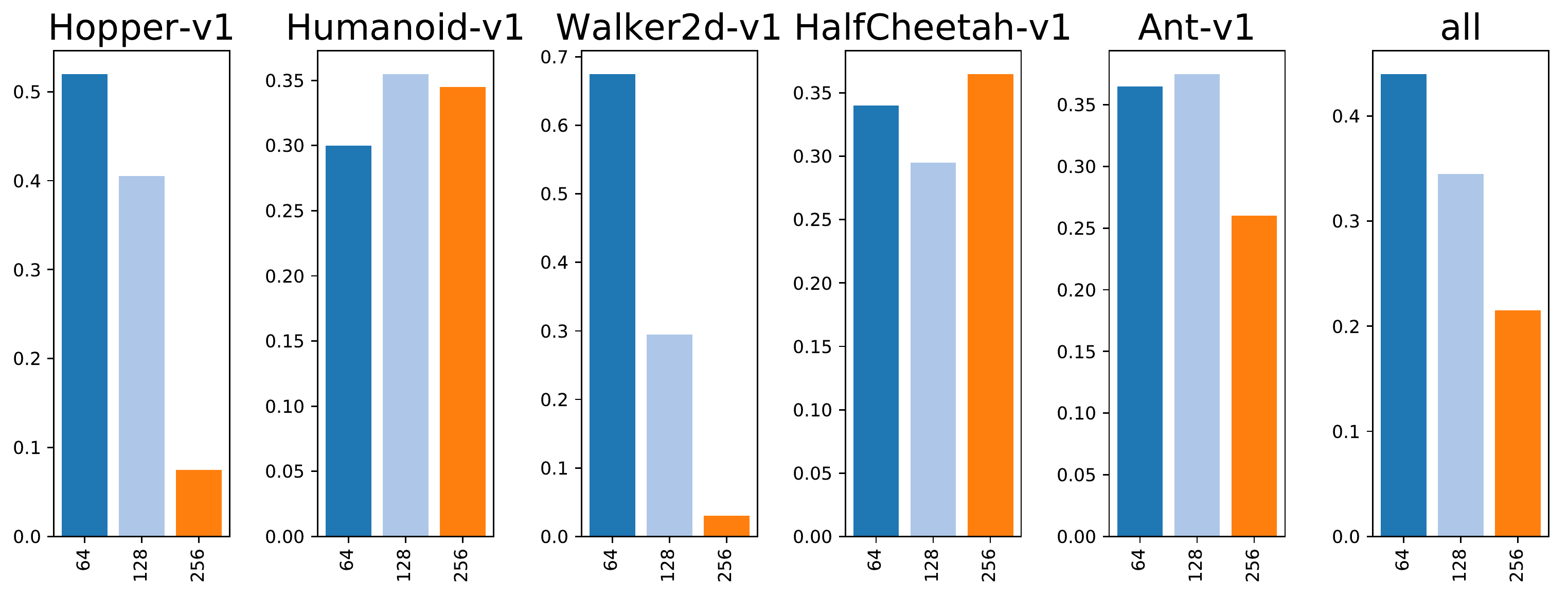}}
\caption{Analysis of choice \choicet{numenvs}: 95th percentile of performance scores conditioned on choice (left) and distribution of choices in top 5\% of configurations (right).}
\label{fig:final_advantages__gin_study_design_choice_value_num_actors_in_learner}
\end{center}
\end{figure}

\begin{figure}[ht]
\begin{center}
\centerline{\includegraphics[width=0.45\textwidth]{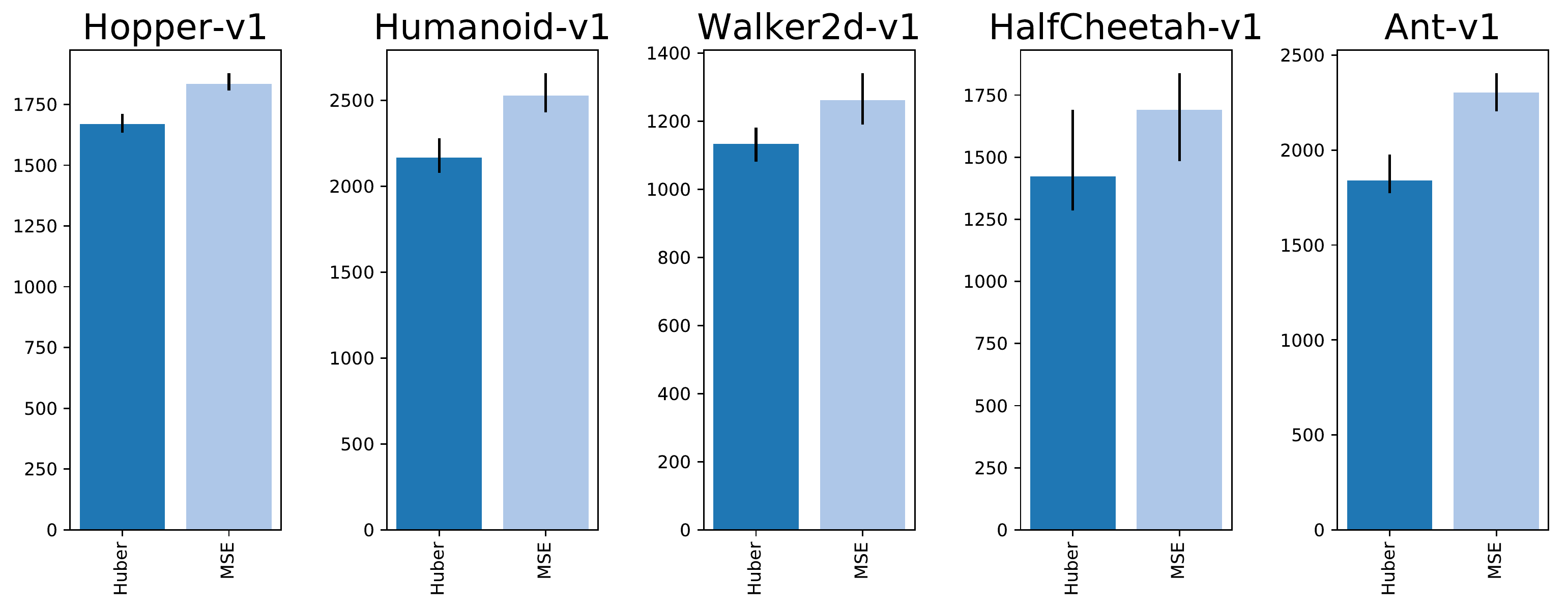}\hspace{1cm}\includegraphics[width=0.45\textwidth]{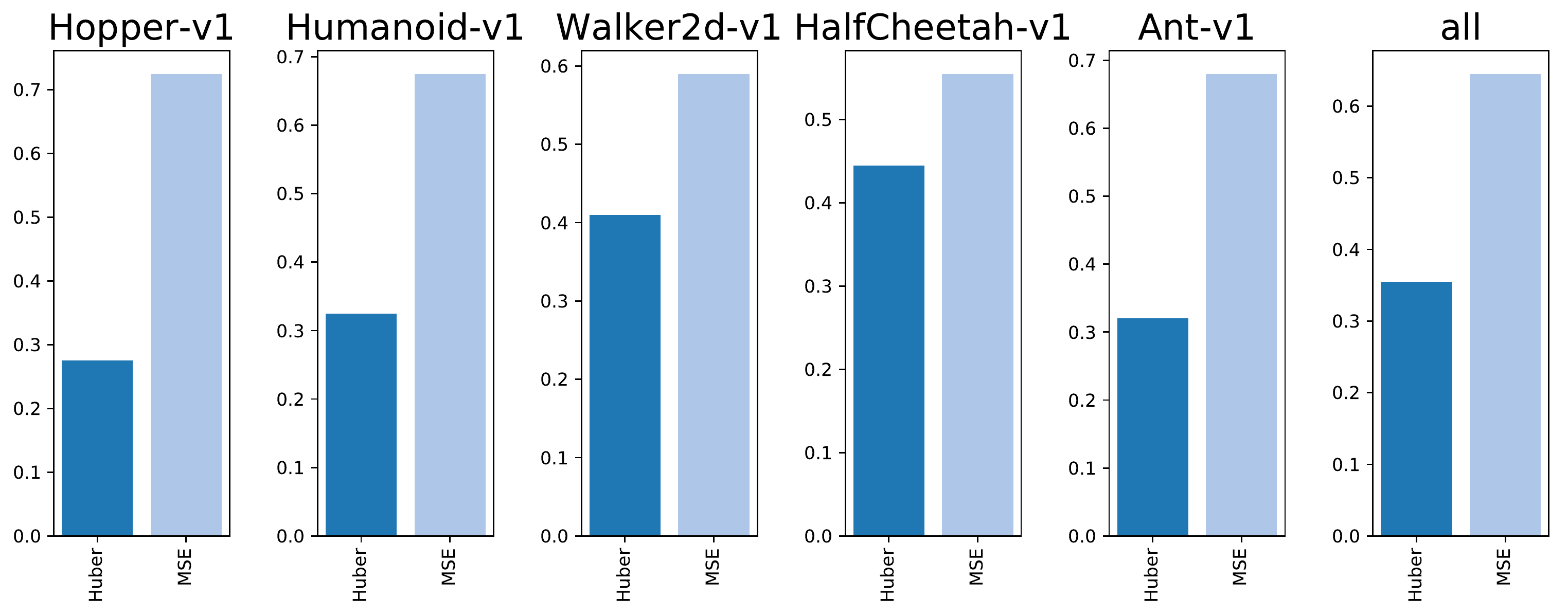}}
\caption{Analysis of choice \choicet{valueloss}: 95th percentile of performance scores conditioned on choice (left) and distribution of choices in top 5\% of configurations (right).}
\label{fig:final_advantages__gin_study_design_choice_value_value_loss}
\end{center}
\end{figure}

\begin{figure}[ht]
\begin{center}
\centerline{\includegraphics[width=0.45\textwidth]{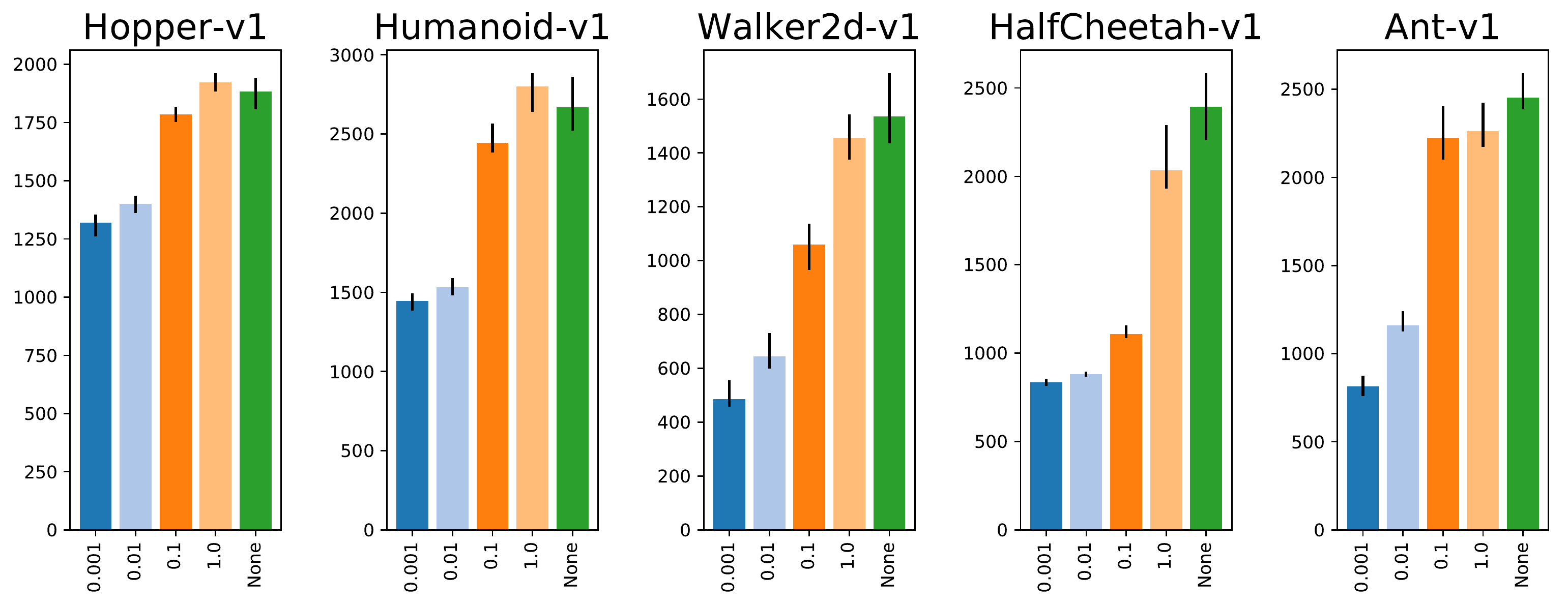}\hspace{1cm}\includegraphics[width=0.45\textwidth]{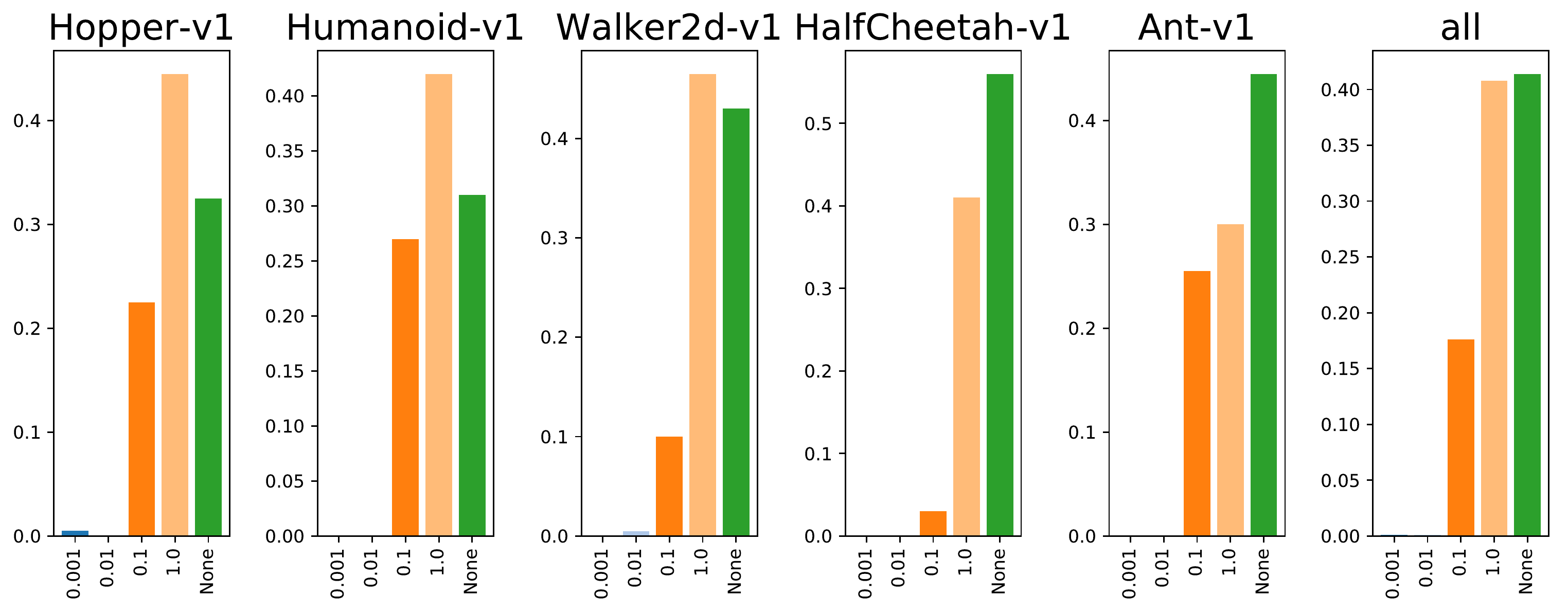}}
\caption{Analysis of choice \choicet{ppovalueclip}: 95th percentile of performance scores conditioned on choice (left) and distribution of choices in top 5\% of configurations (right).}
\label{fig:final_advantages__gin_study_design_choice_value_ppo_style_value_clipping_epsilon}
\end{center}
\end{figure}

\begin{figure}[ht]
\begin{center}
\centerline{\includegraphics[width=0.45\textwidth]{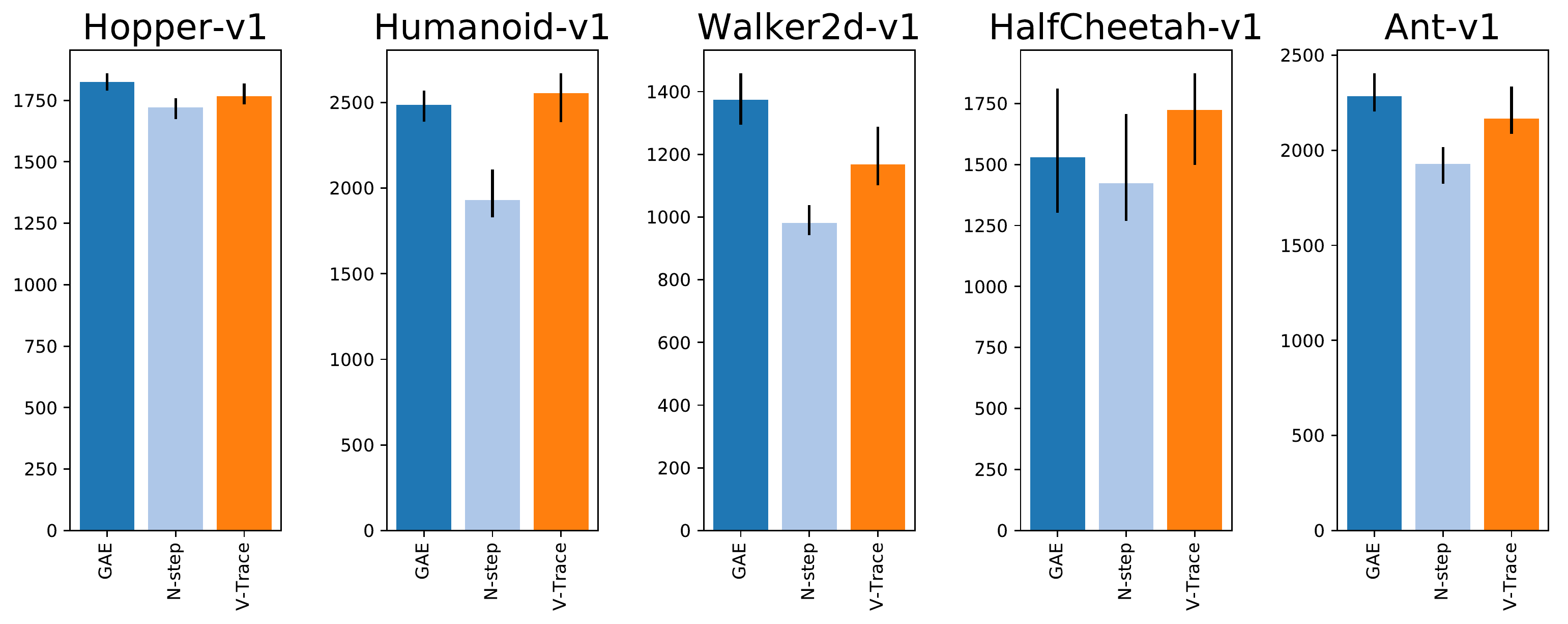}\hspace{1cm}\includegraphics[width=0.45\textwidth]{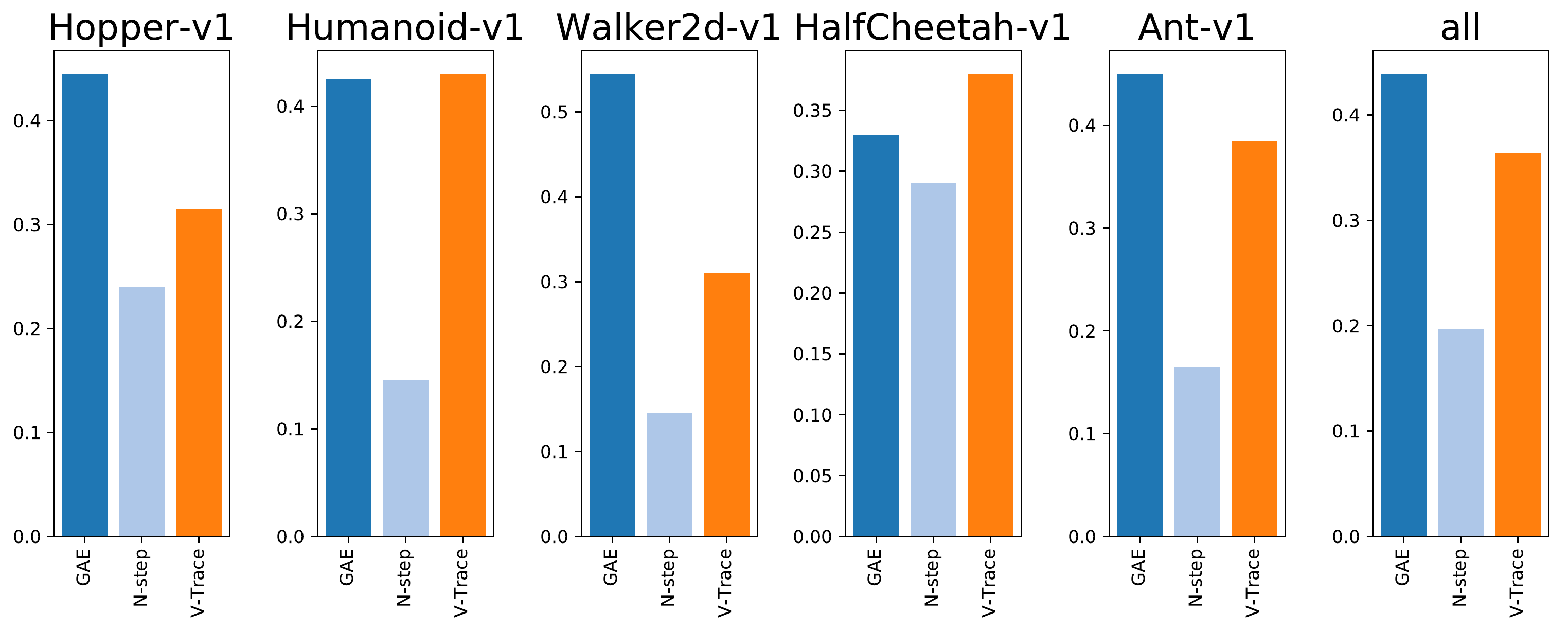}}
\caption{Analysis of choice \choicet{advantageestimator}: 95th percentile of performance scores conditioned on choice (left) and distribution of choices in top 5\% of configurations (right).}
\label{fig:final_advantages__gin_study_design_choice_value_advantage_estimator}
\end{center}
\end{figure}

\begin{figure}[ht]
\begin{center}
\centerline{\includegraphics[width=0.45\textwidth]{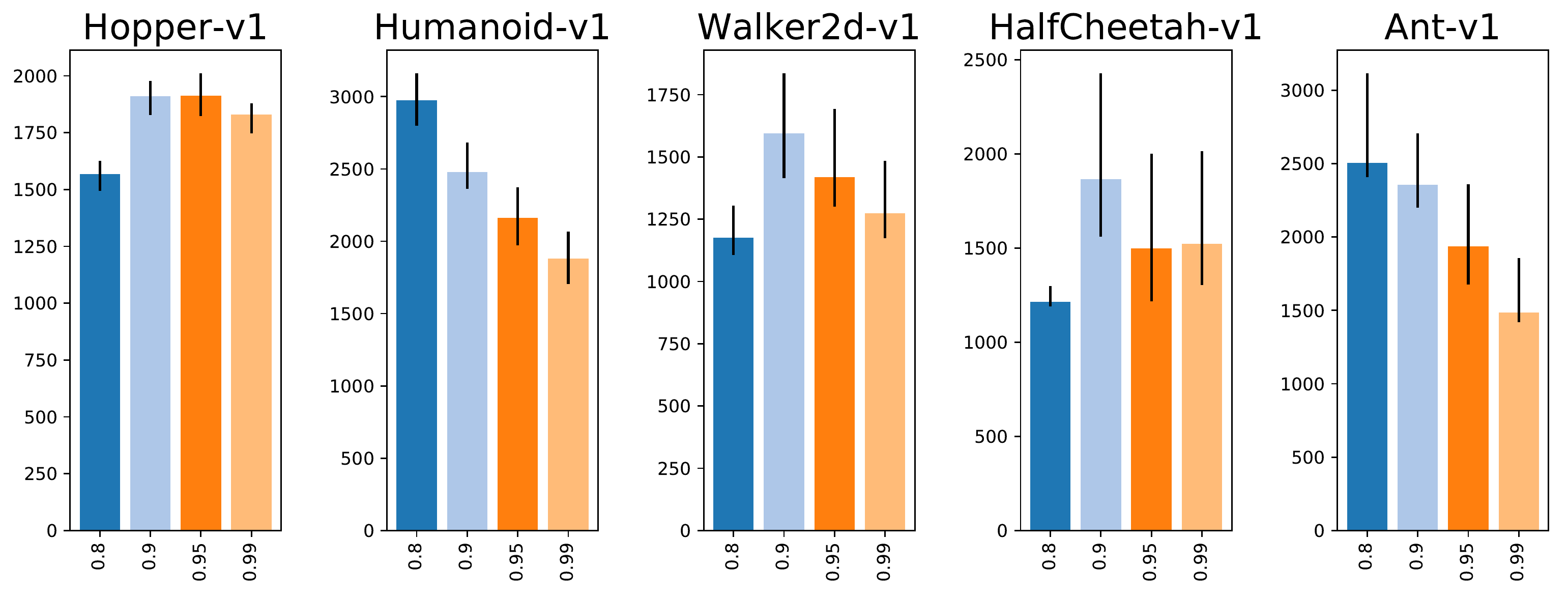}\hspace{1cm}\includegraphics[width=0.45\textwidth]{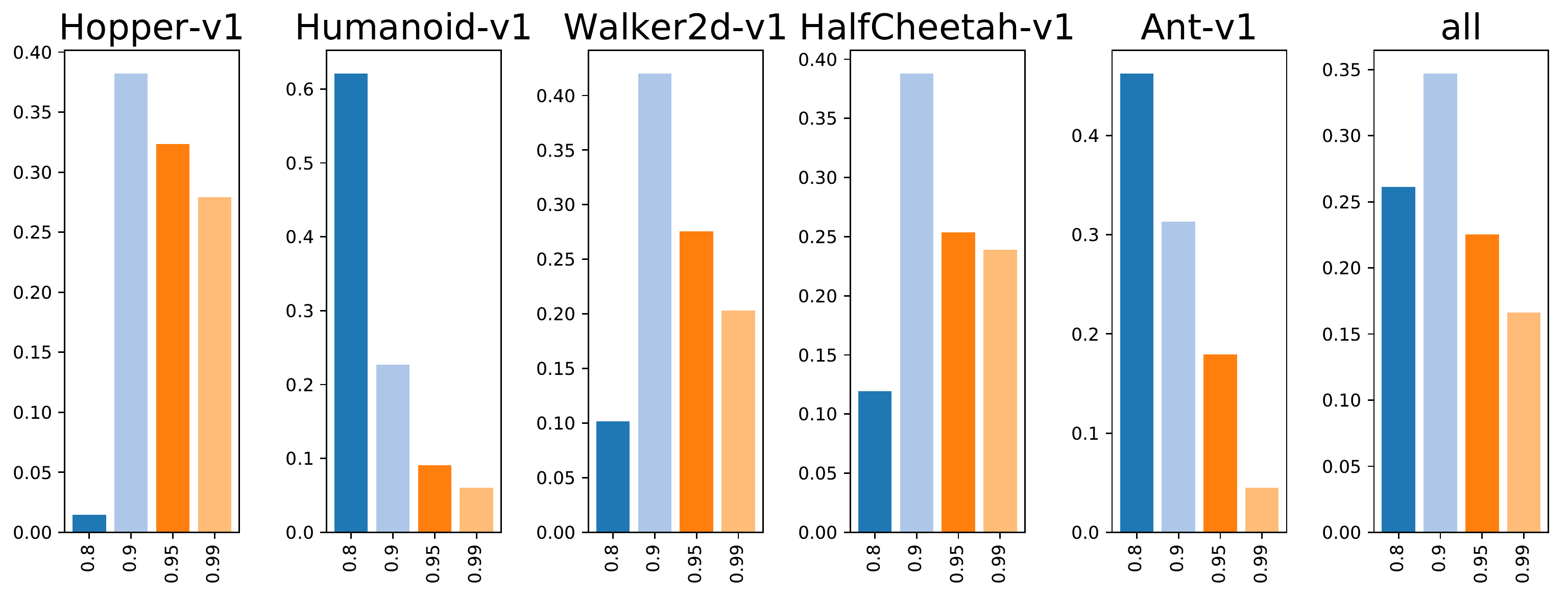}}
\caption{Analysis of choice \choicet{gaelambda}: 95th percentile of performance scores conditioned on sub-choice (left) and distribution of sub-choices in top 5\% of configurations (right).}
\label{fig:final_advantages__gin_study_design_choice_value_sub_advantage_estimator_gae_gae_lambda}
\end{center}
\end{figure}

\begin{figure}[ht]
\begin{center}
\centerline{\includegraphics[width=0.45\textwidth]{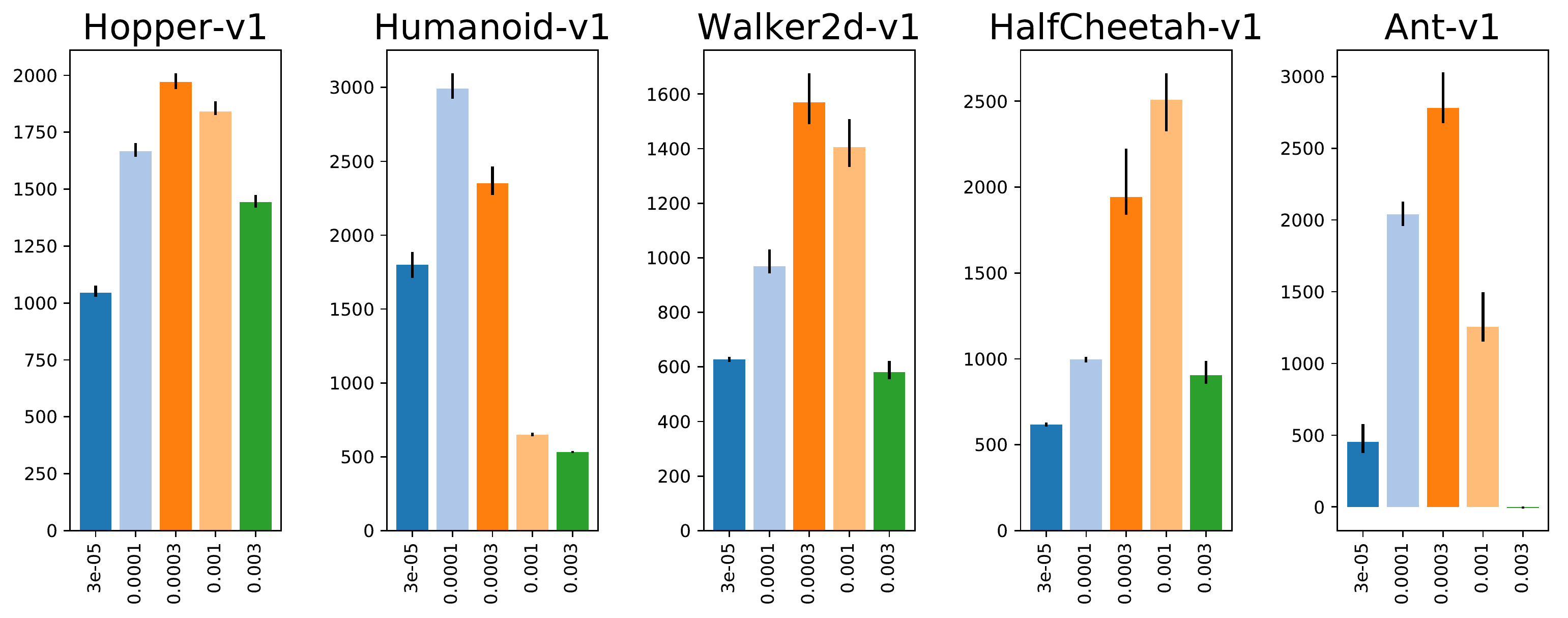}\hspace{1cm}\includegraphics[width=0.45\textwidth]{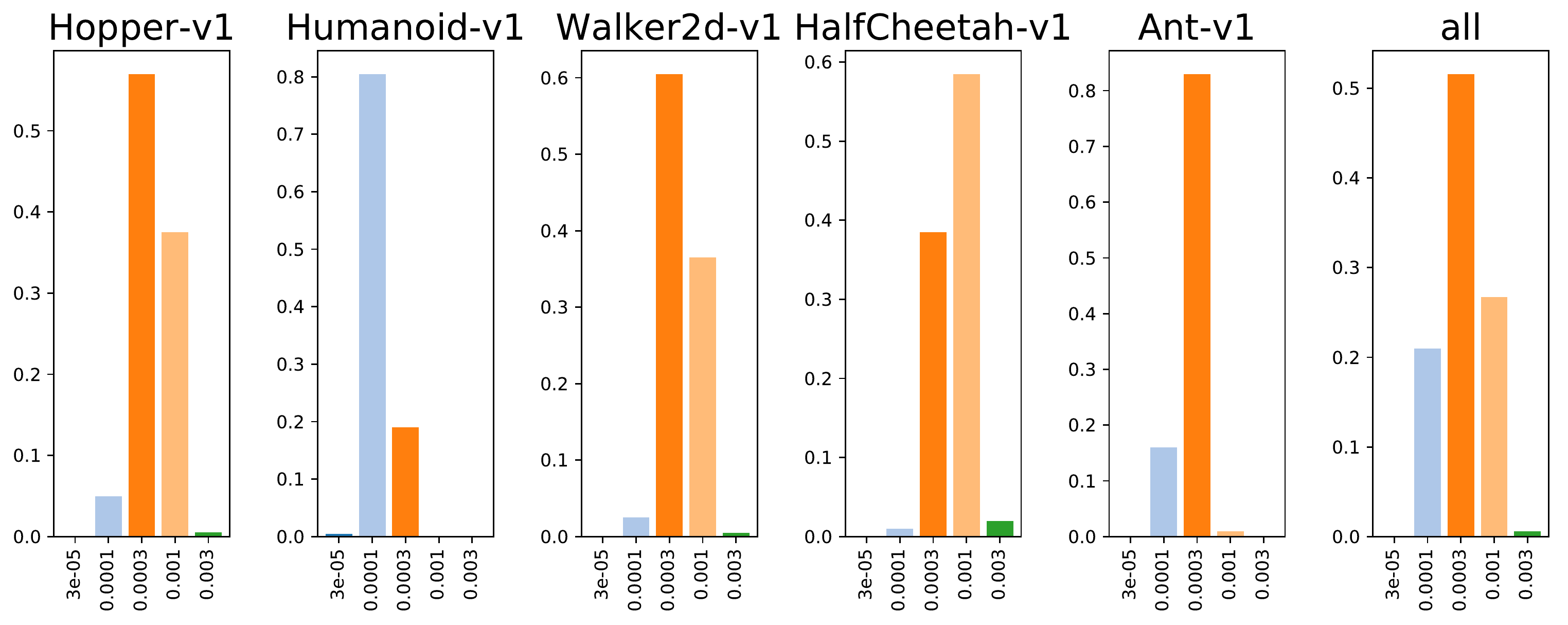}}
\caption{Analysis of choice \choicet{adamlr}: 95th percentile of performance scores conditioned on choice (left) and distribution of choices in top 5\% of configurations (right).}
\label{fig:final_advantages__gin_study_design_choice_value_learning_rate}
\end{center}
\end{figure}

\begin{figure}[ht]
\begin{center}
\centerline{\includegraphics[width=0.45\textwidth]{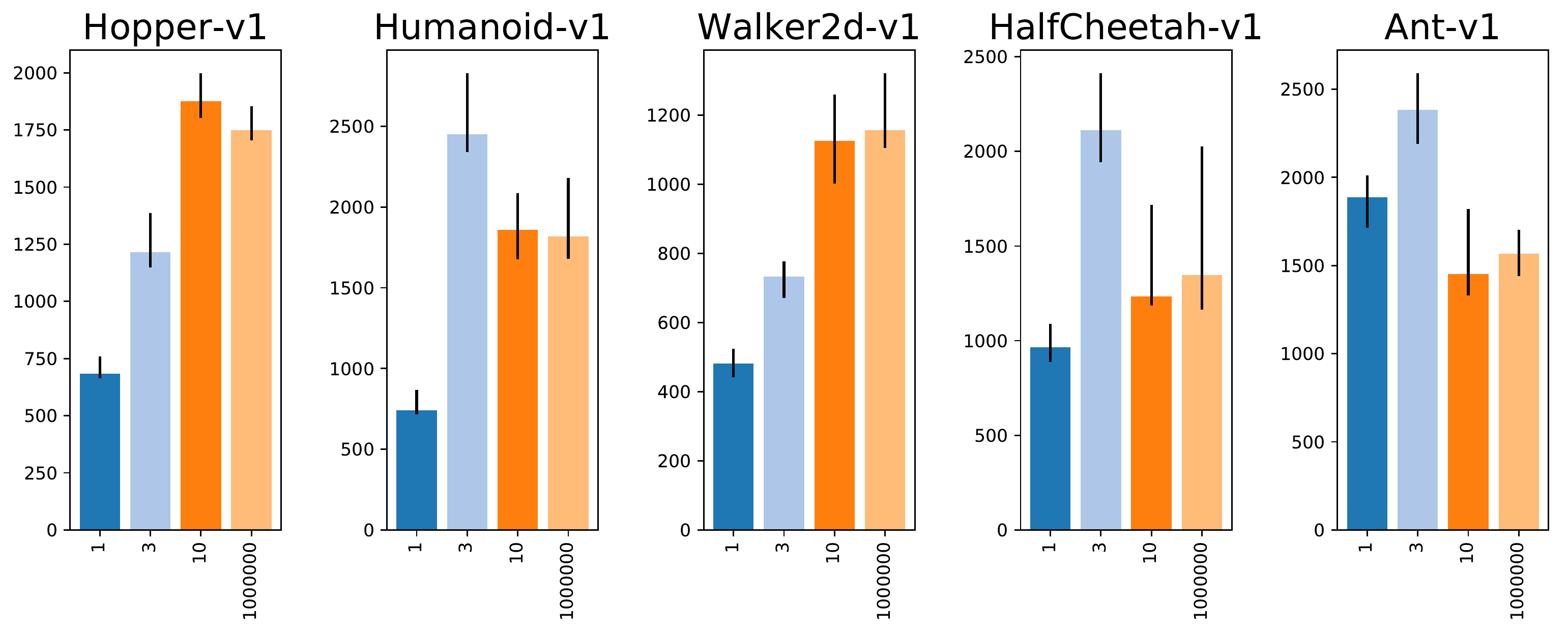}\hspace{1cm}\includegraphics[width=0.45\textwidth]{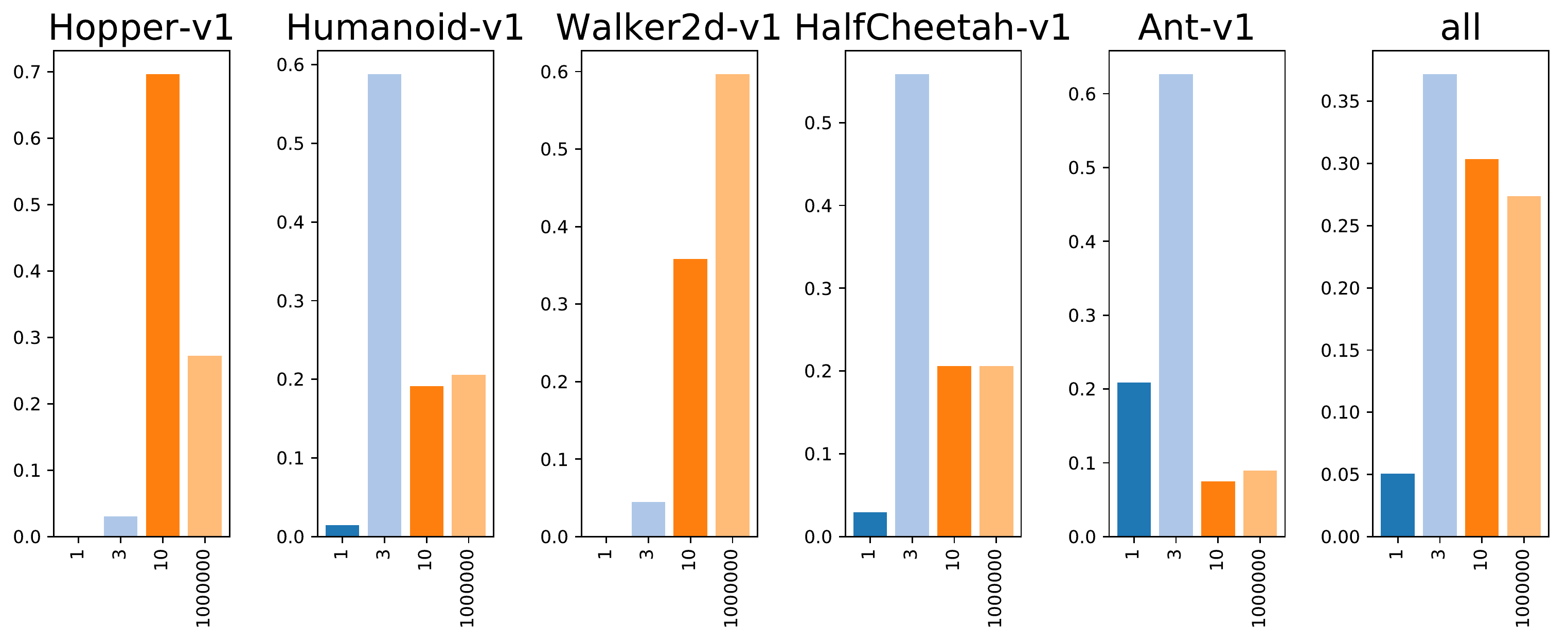}}
\caption{Analysis of choice \choicet{nstep}: 95th percentile of performance scores conditioned on sub-choice (left) and distribution of sub-choices in top 5\% of configurations (right).}
\label{fig:final_advantages__gin_study_design_choice_value_sub_advantage_estimator_n_step_n}
\end{center}
\end{figure}

\begin{figure}[ht]
\begin{center}
\centerline{\includegraphics[width=0.45\textwidth]{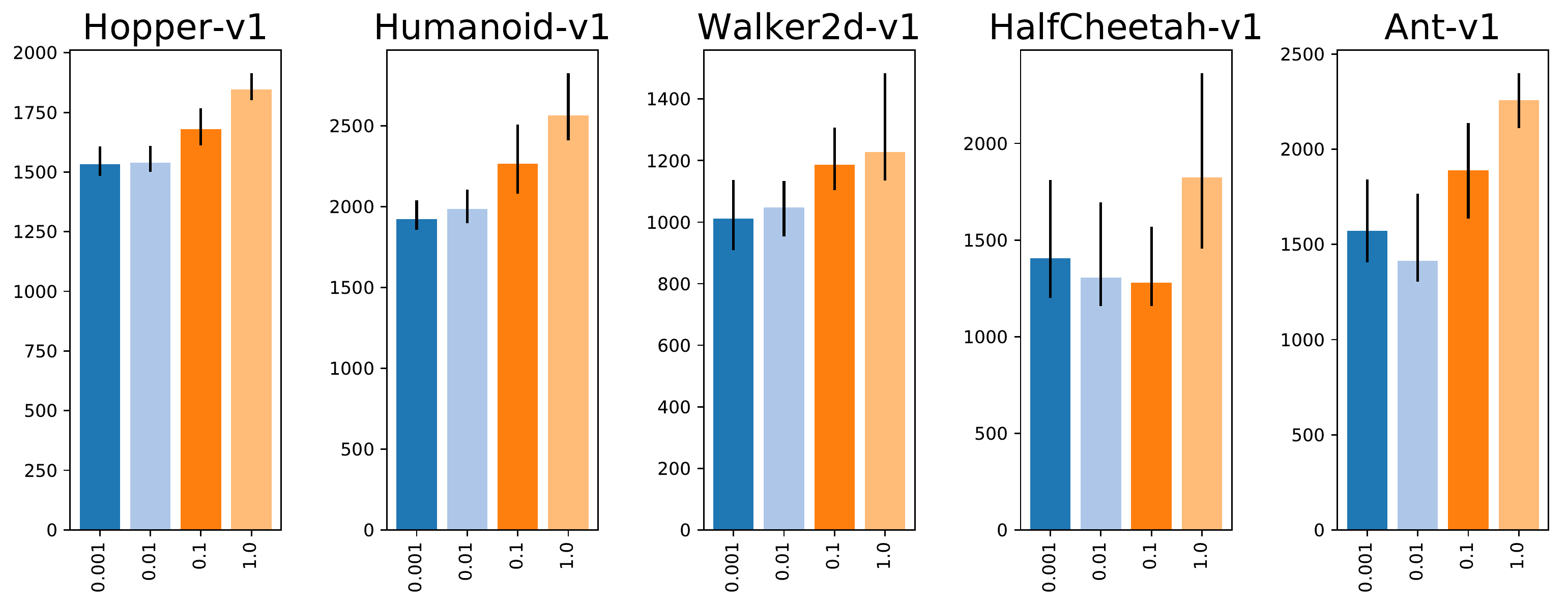}\hspace{1cm}\includegraphics[width=0.45\textwidth]{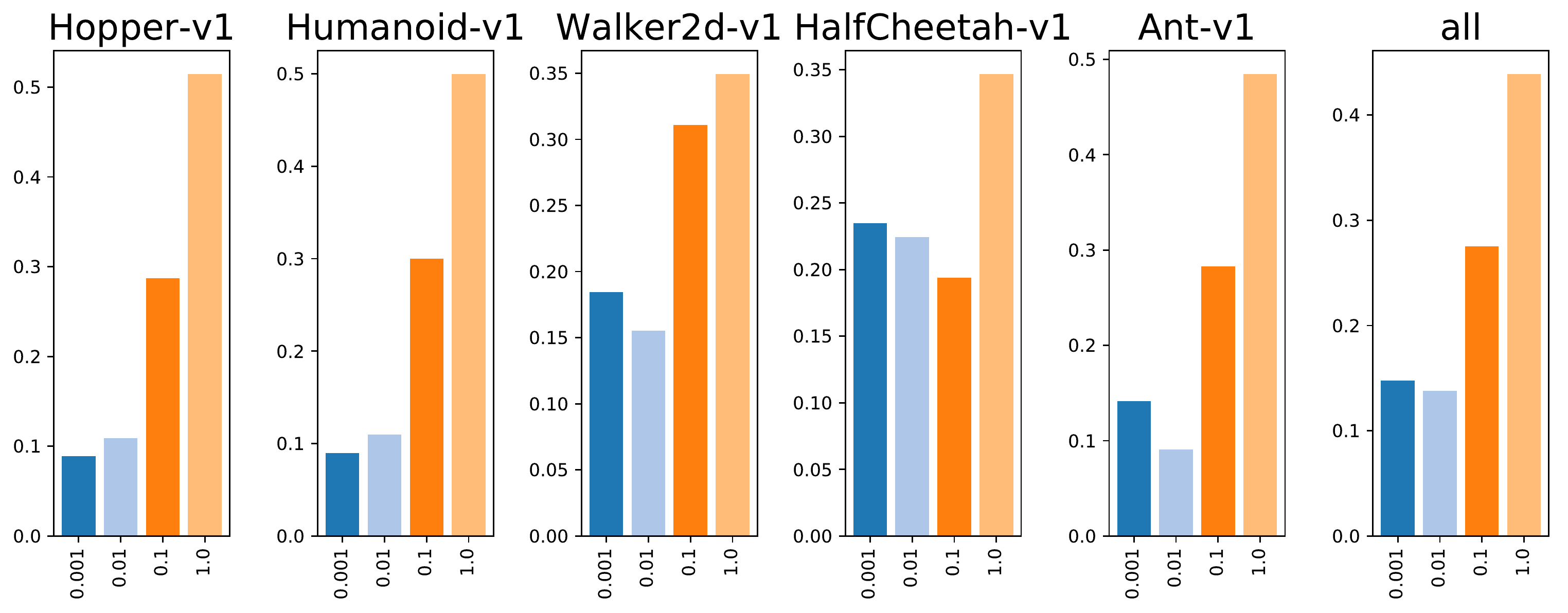}}
\caption{Analysis of choice \choicet{huberdelta}: 95th percentile of performance scores conditioned on sub-choice (left) and distribution of sub-choices in top 5\% of configurations (right).}
\label{fig:final_advantages__gin_study_design_choice_value_sub_value_loss_huber_delta}
\end{center}
\end{figure}

\begin{figure}[ht]
\begin{center}
\centerline{\includegraphics[width=0.45\textwidth]{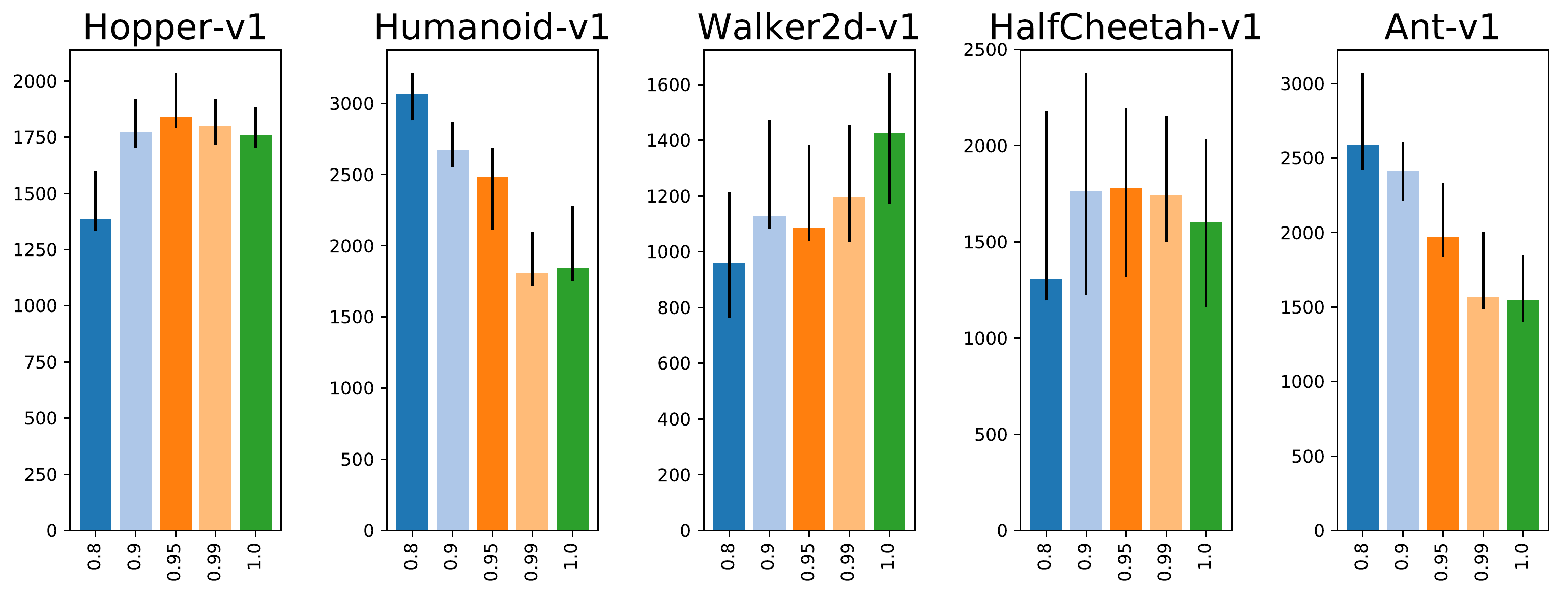}\hspace{1cm}\includegraphics[width=0.45\textwidth]{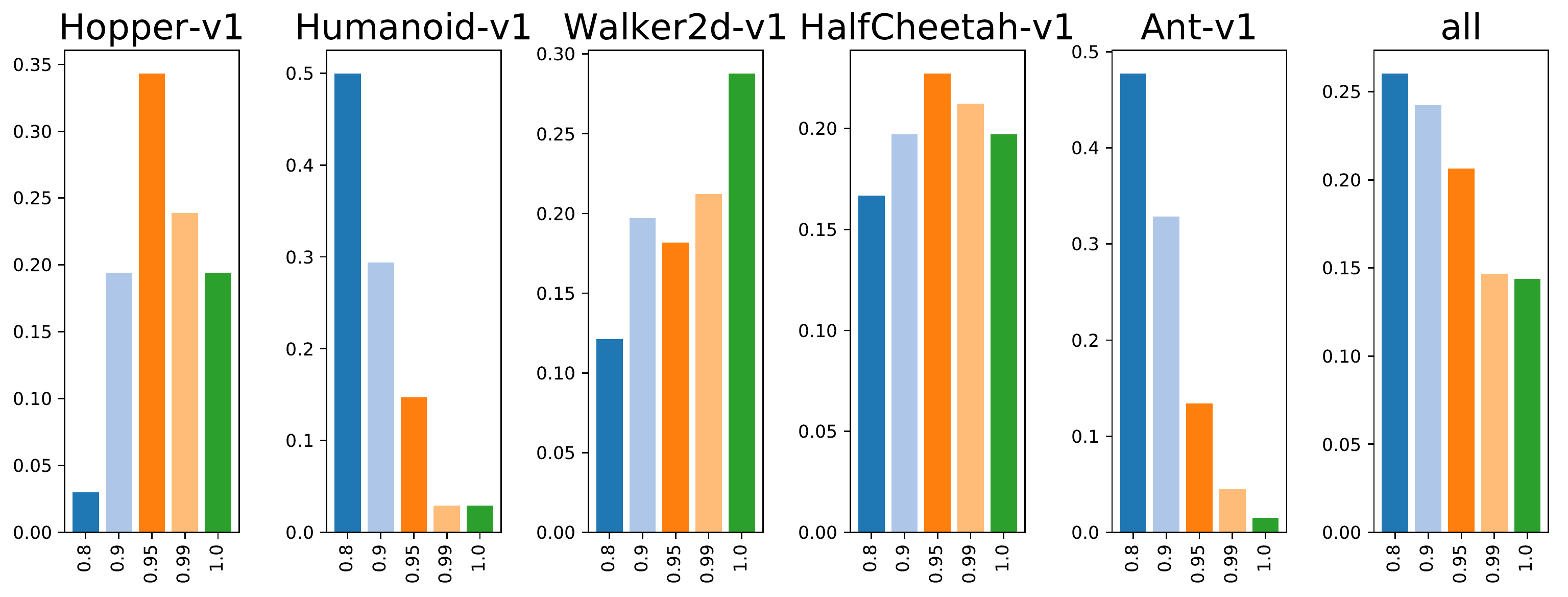}}
\caption{Analysis of choice \choicet{vtraceaelambda}: 95th percentile of performance scores conditioned on sub-choice (left) and distribution of sub-choices in top 5\% of configurations (right).}
\label{fig:final_advantages__gin_study_design_choice_value_sub_advantage_estimator_v_trace_lambda}
\end{center}
\end{figure}

\begin{figure}[ht]
\begin{center}
\centerline{\includegraphics[width=0.45\textwidth]{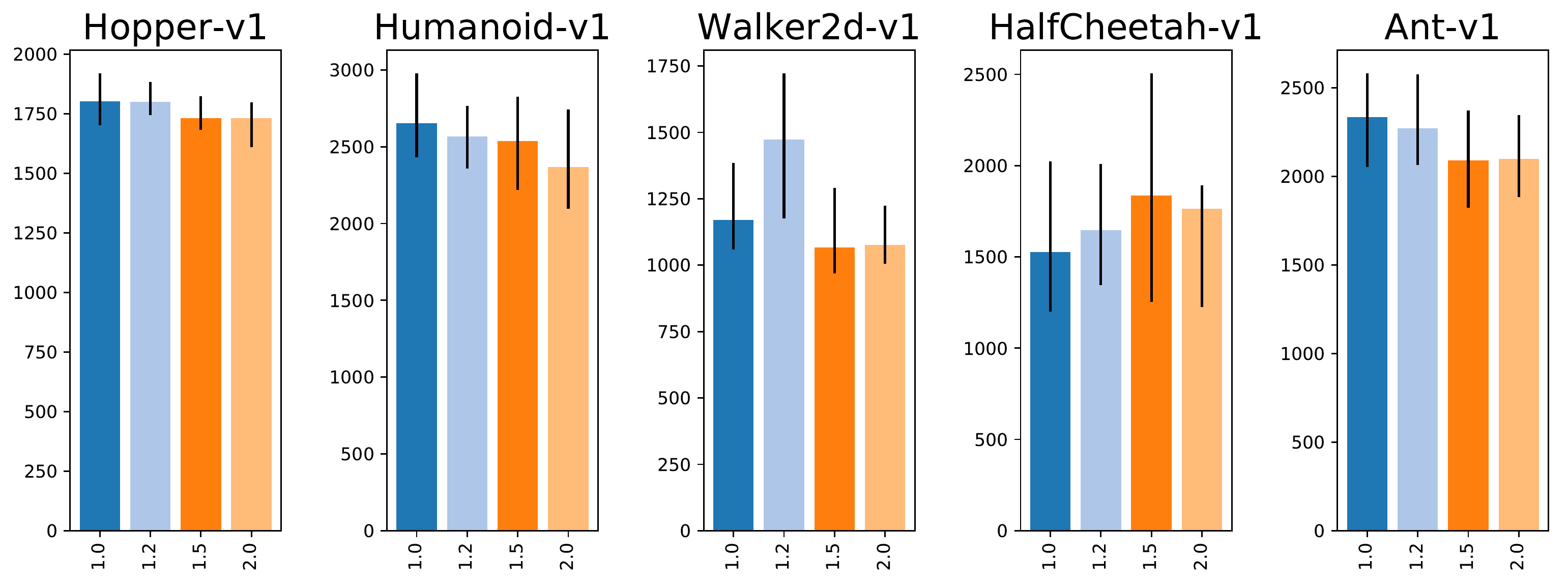}\hspace{1cm}\includegraphics[width=0.45\textwidth]{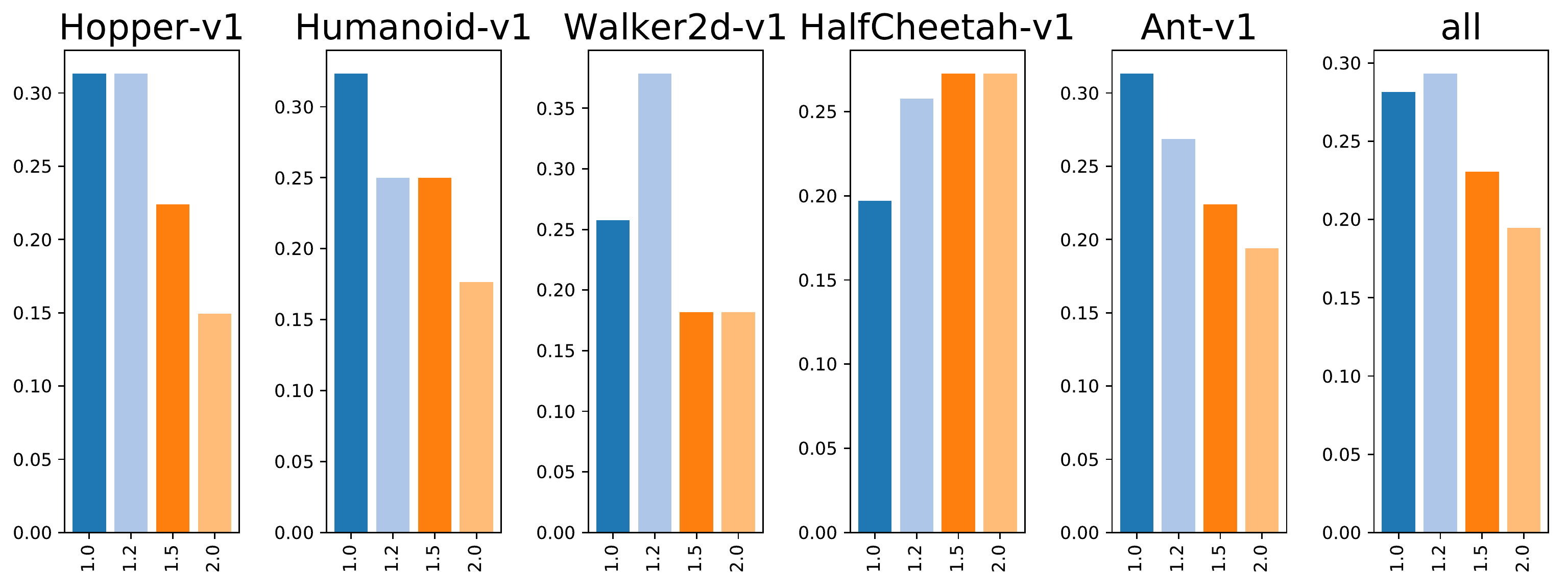}}
\caption{Analysis of choice \choicet{vtraceaecrho}: 95th percentile of performance scores conditioned on sub-choice (left) and distribution of sub-choices in top 5\% of configurations (right).}
\label{fig:final_advantages__gin_study_design_choice_value_sub_advantage_estimator_v_trace_max_importance_weight_in_advantage_estimation}
\end{center}
\end{figure}

We report aggregate statistics of the experiment in Table~\ref{tab:final_advantages_overview} as well as training curves in Figure~\ref{fig:final_advantages_training_curves}.
For each of the investigated choices in this experiment, we further provide a per-choice analysis in Figures~\ref{fig:final_advantages_custom_advantage_estimator}-\ref{fig:final_advantages__gin_study_design_choice_value_sub_advantage_estimator_v_trace_max_importance_weight_in_advantage_estimation}.
\begin{table}[ht]
\begin{center}
\caption{Performance quantiles across choice configurations.}
\label{tab:final_advantages_overview}
\begin{tabular}{lrrrrr}
\toprule
{} & Ant-v1 & HalfCheetah-v1 & Hopper-v1 & Humanoid-v1 & Walker2d-v1 \\
\midrule
90th percentile &   1705 &           1128 &      1626 &        1922 &         947 \\
95th percentile &   2114 &           1535 &      1777 &        2374 &        1185 \\
99th percentile &   2781 &           2631 &      2001 &        3013 &        1697 \\
Max             &   3775 &           3613 &      2215 &        3564 &        2309 \\
\bottomrule
\end{tabular}

\end{center}
\end{table}

\clearpage

%% file: final_setup/main.tex
\clearpage
\section{Experiment \texttt{Training setup}}
\label{exp_final_setup}
\subsection{Design}
\label{exp_design_final_setup}
For each of the 5 environments, we sampled 2000 choice configurations where we sampled the following choices independently and uniformly from the following ranges:
\begin{itemize}
    \item \choicet{stepsize}: \{512, 1024, 2048, 4096\}
    \item \choicet{batchhandling}: \{\texttt{Fixed trajectories}, \texttt{Shuffle trajectories}, \texttt{Shuffle transitions}, \texttt{Shuffle transitions (recompute advantages)}\}
    \item \choicet{numepochsperstep}: \{1, 3, 10\}
    \item \choicet{numenvs}: \{64, 128, 256\}
    \item \choicet{adamlr}: \{3e-05, 0.0001, 0.0003, 0.001, 0.003\}
    \item \choicet{batchsize}: \{64, 128, 256\}
\end{itemize}
All the other choices were set to the default values as described in Appendix~\ref{sec:default_settings}.

For each of the sampled choice configurations, we train 3 agents with different random seeds and compute the performance metric as described in Section~\ref{sec:performance}.
\subsection{Results}
\label{exp_results_final_setup}
\begin{figure}[ht]
\begin{center}
\centerline{\includegraphics[width=1\textwidth]{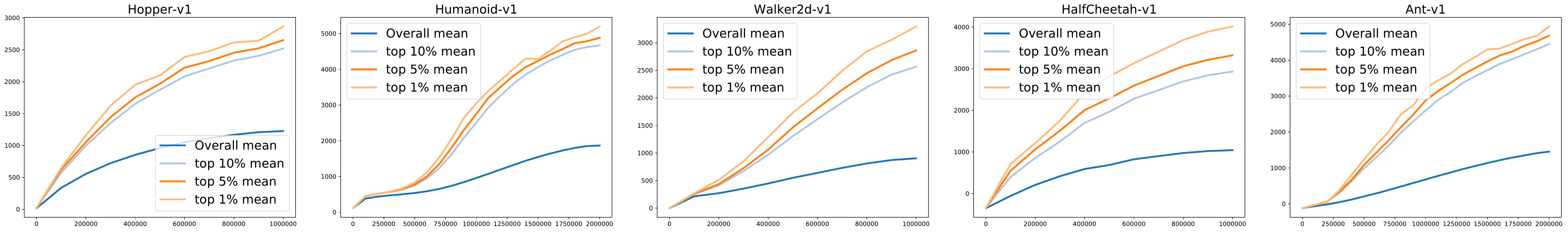}}
\caption{Training curves.}
\label{fig:final_setup_training_curves}
\end{center}
\end{figure}

\begin{figure}[ht]
\begin{center}
\centerline{\includegraphics[width=0.45\textwidth]{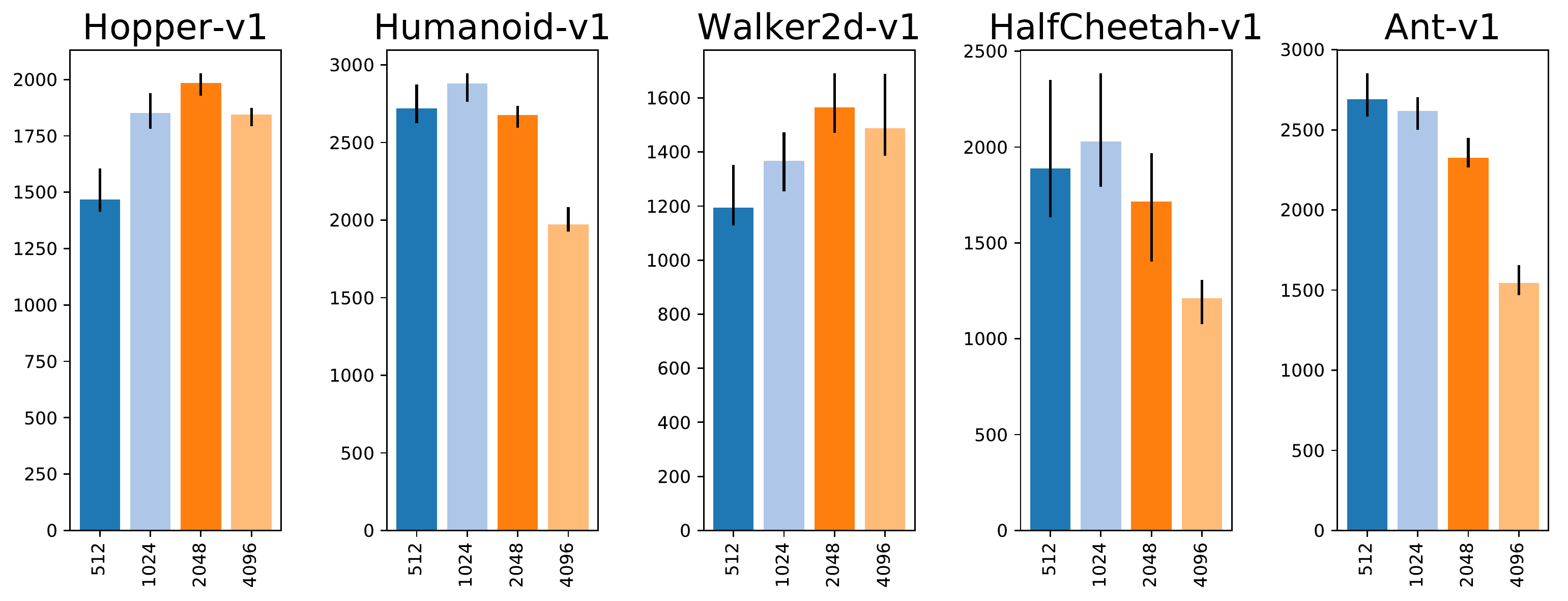}\hspace{1cm}\includegraphics[width=0.45\textwidth]{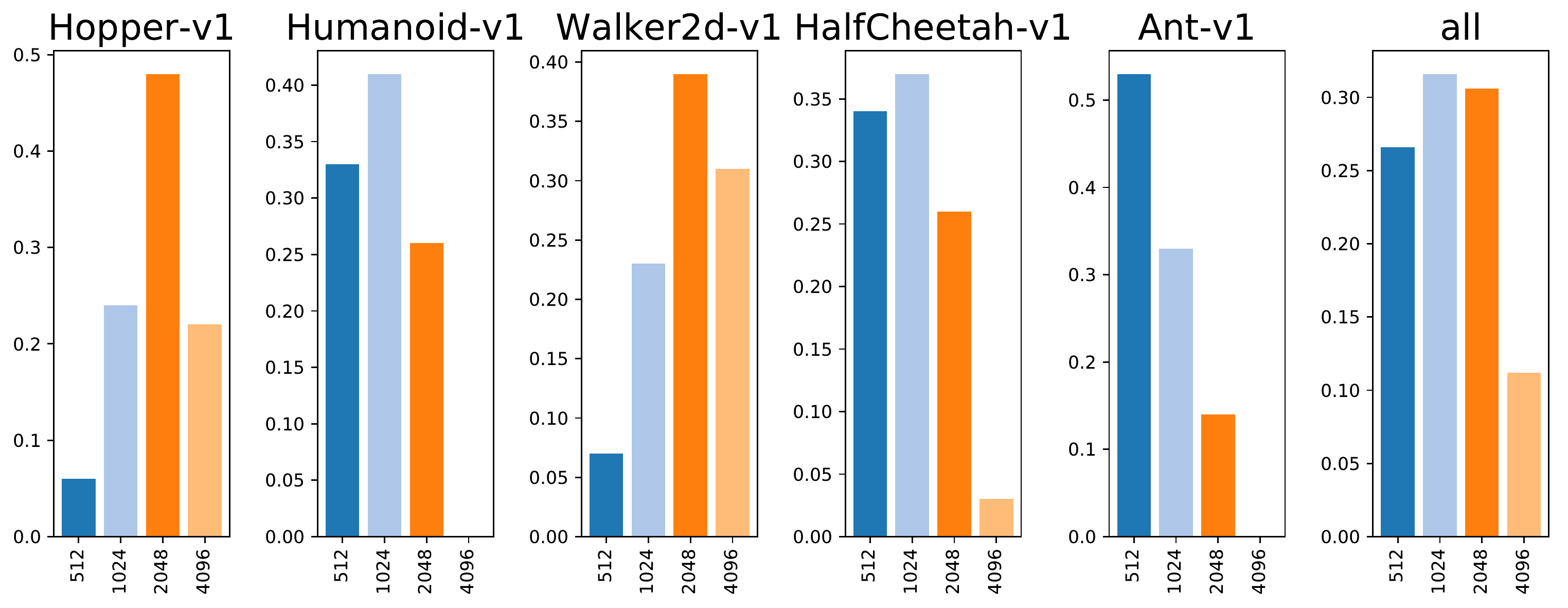}}
\caption{Analysis of choice \choicet{stepsize}: 95th percentile of performance scores conditioned on choice (left) and distribution of choices in top 5\% of configurations (right).}
\label{fig:final_setup__gin_study_design_choice_value_step_size_transitions}
\end{center}
\end{figure}

\begin{figure}[ht]
\begin{center}
\centerline{\includegraphics[width=0.45\textwidth]{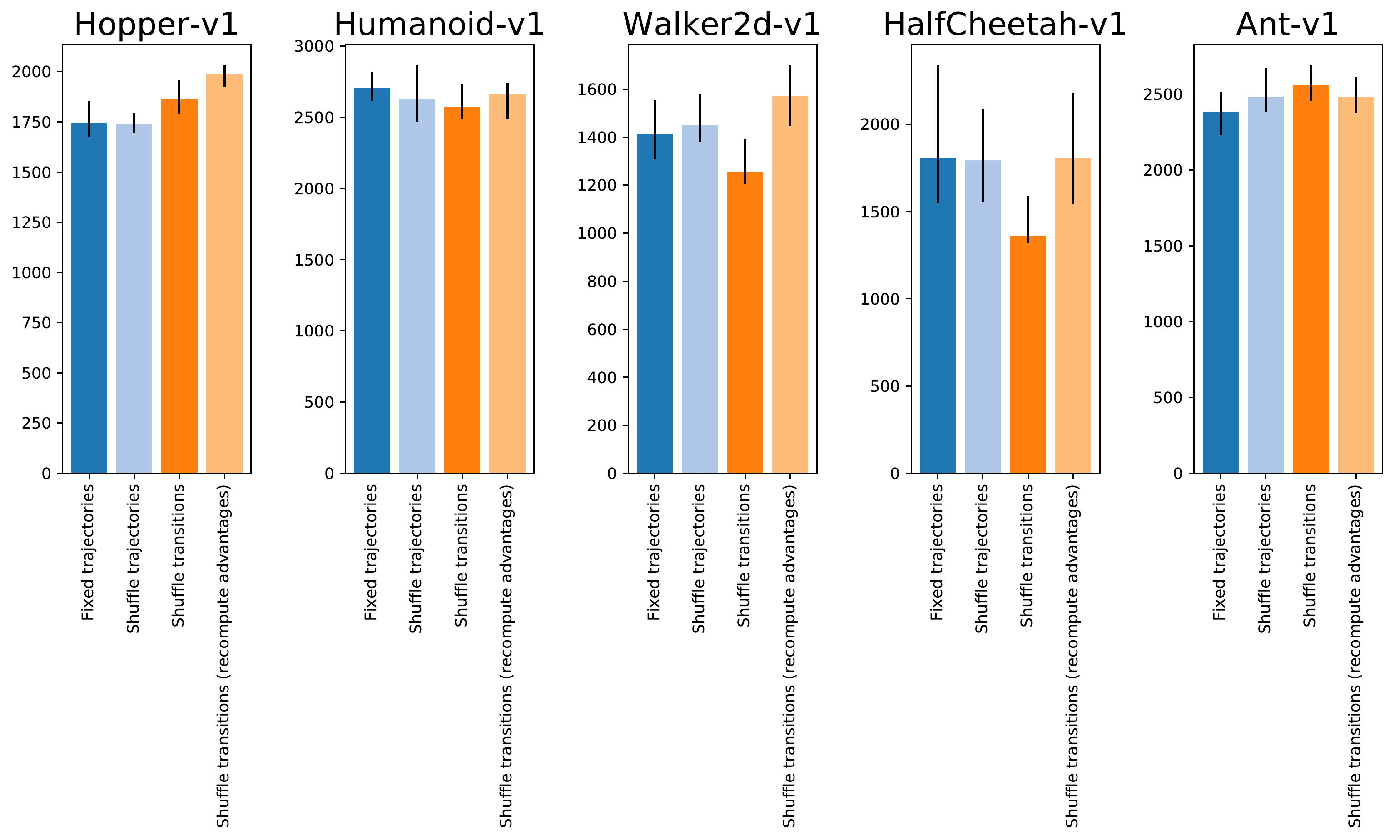}\hspace{1cm}\includegraphics[width=0.45\textwidth]{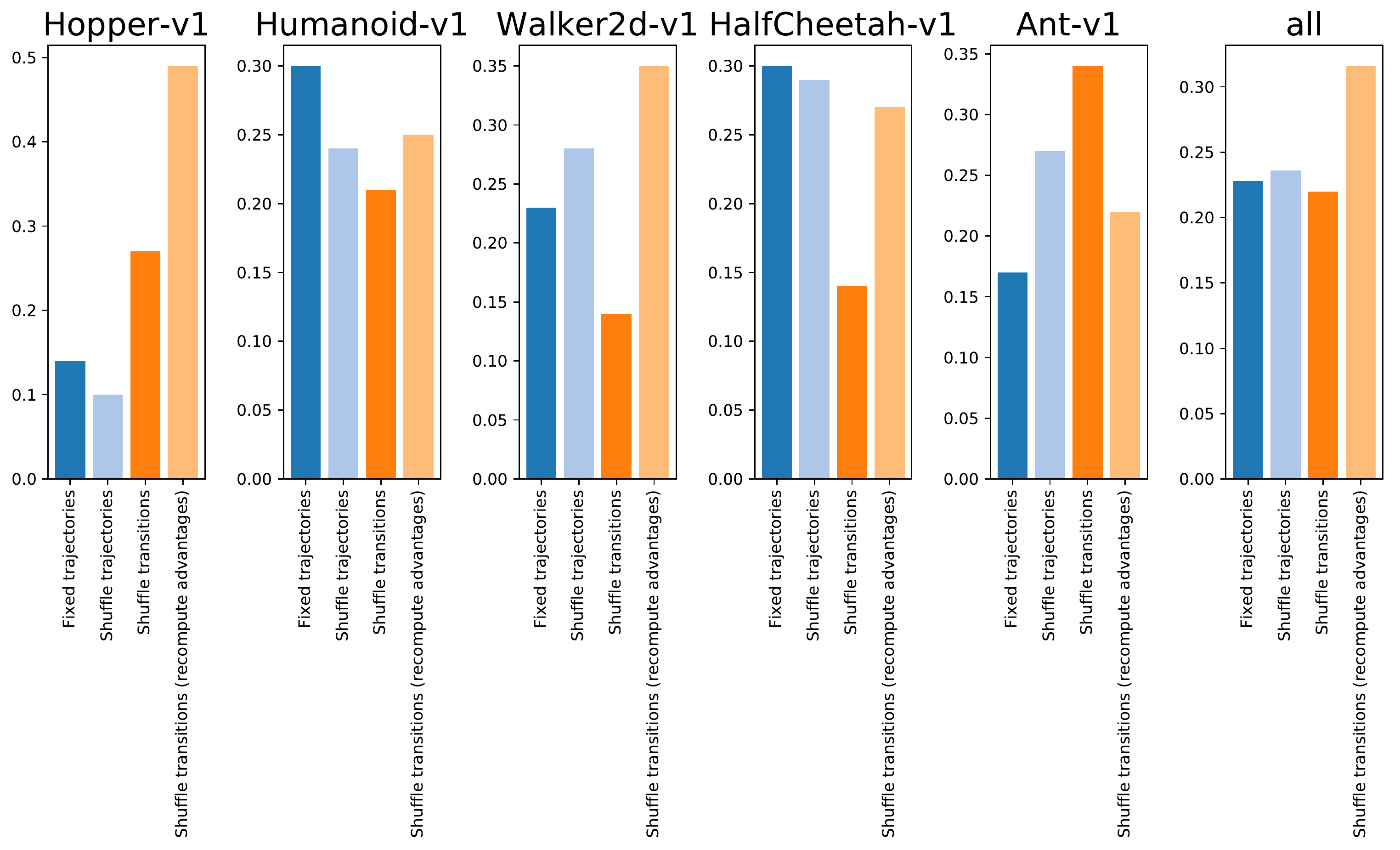}}
\caption{Analysis of choice \choicet{batchhandling}: 95th percentile of performance scores conditioned on choice (left) and distribution of choices in top 5\% of configurations (right).}
\label{fig:final_setup__gin_study_design_choice_value_batch_mode}
\end{center}
\end{figure}

\begin{figure}[ht]
\begin{center}
\centerline{\includegraphics[width=0.45\textwidth]{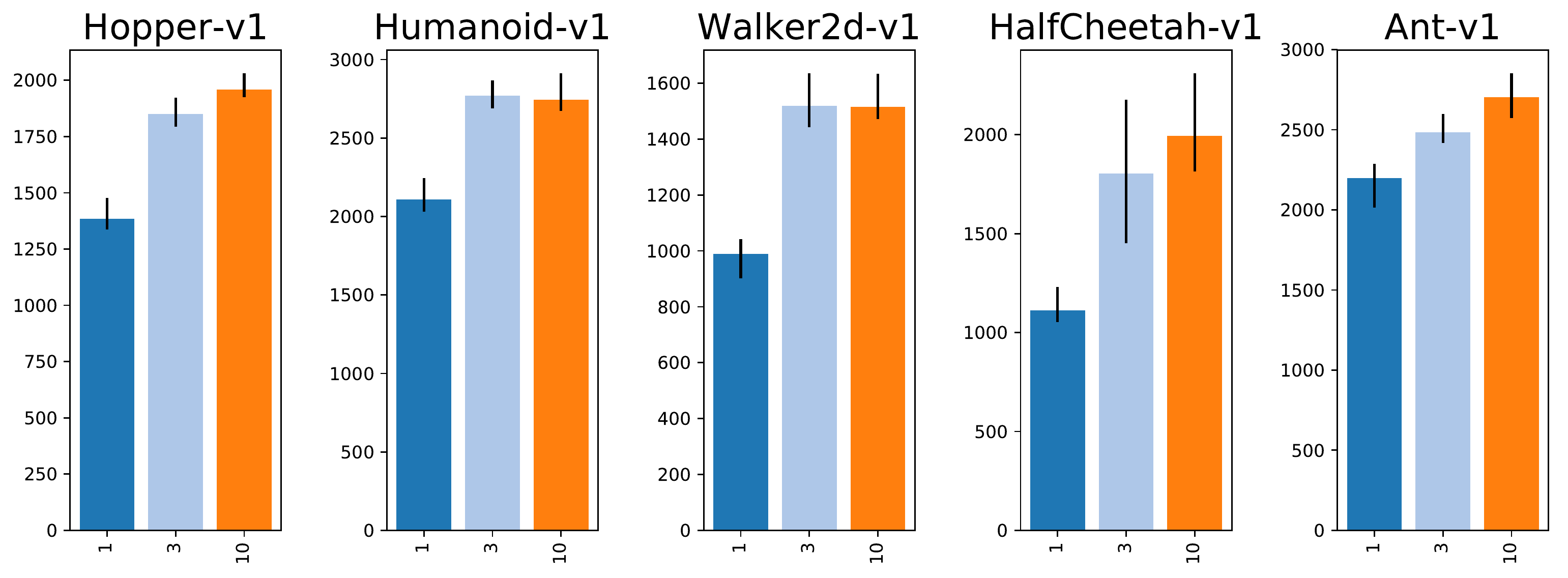}\hspace{1cm}\includegraphics[width=0.45\textwidth]{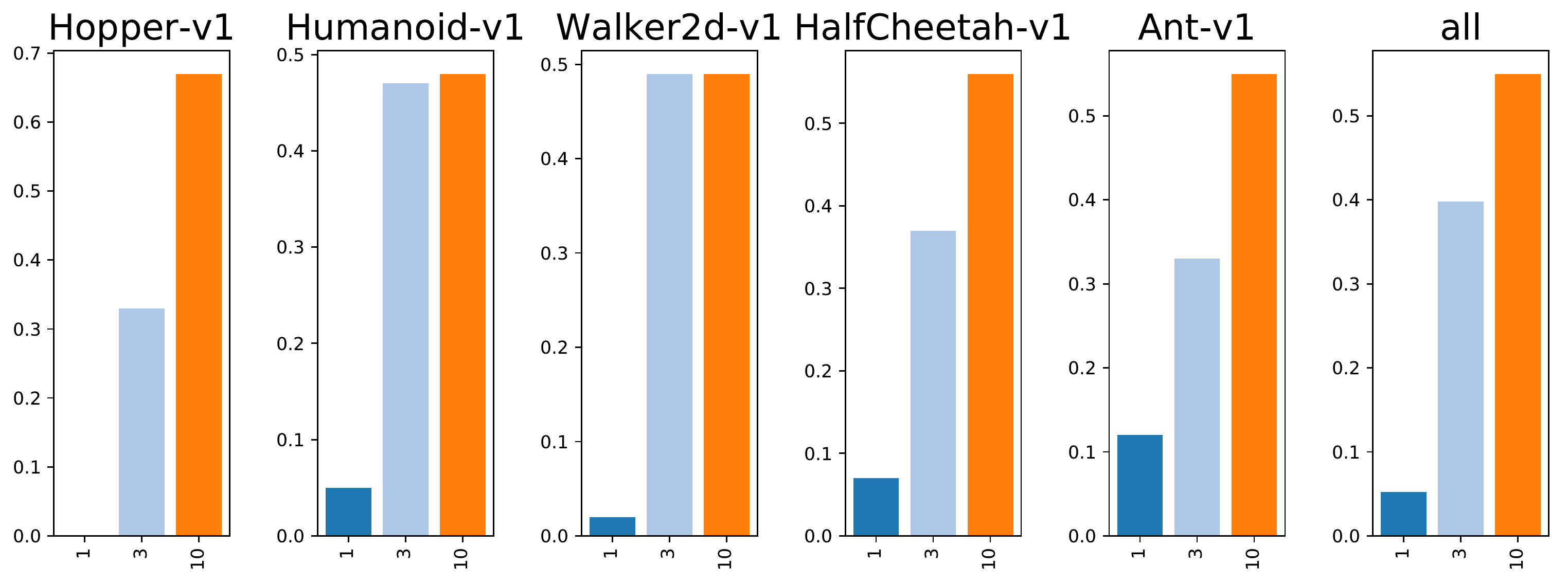}}
\caption{Analysis of choice \choicet{numepochsperstep}: 95th percentile of performance scores conditioned on choice (left) and distribution of choices in top 5\% of configurations (right).}
\label{fig:final_setup__gin_study_design_choice_value_epochs_per_step}
\end{center}
\end{figure}

\begin{figure}[ht]
\begin{center}
\centerline{\includegraphics[width=0.45\textwidth]{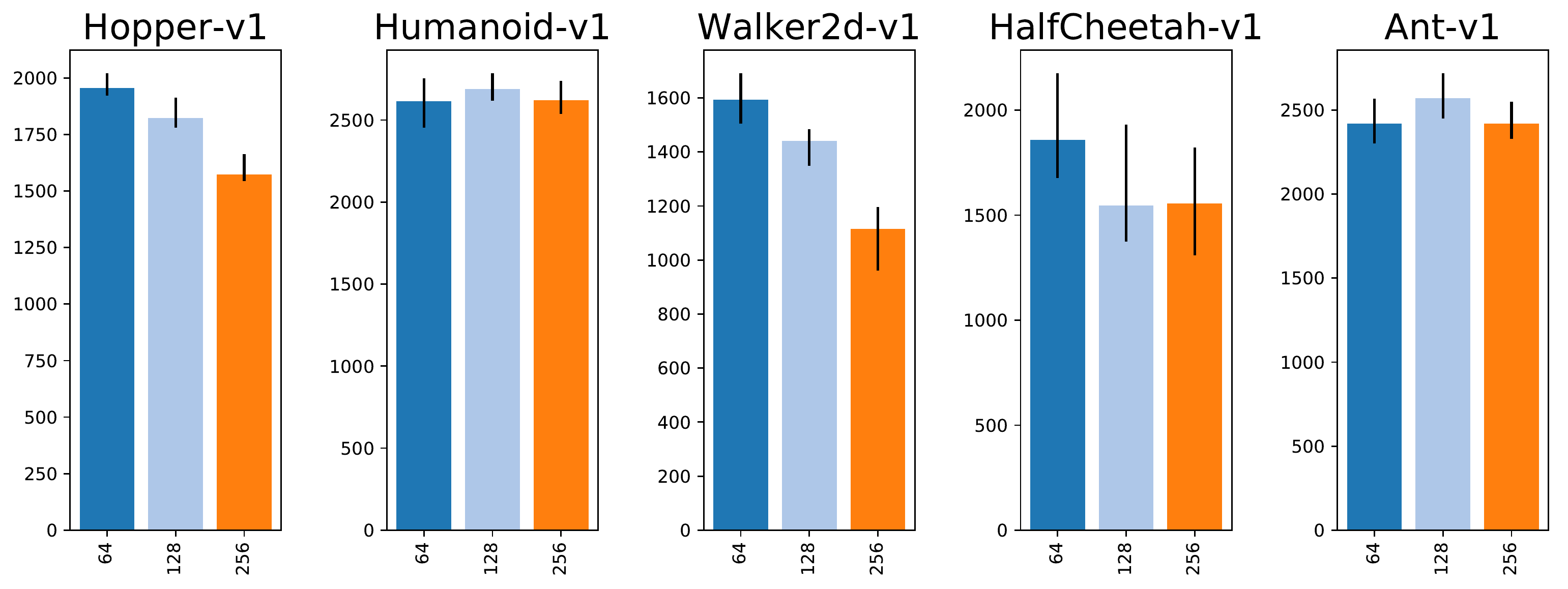}\hspace{1cm}\includegraphics[width=0.45\textwidth]{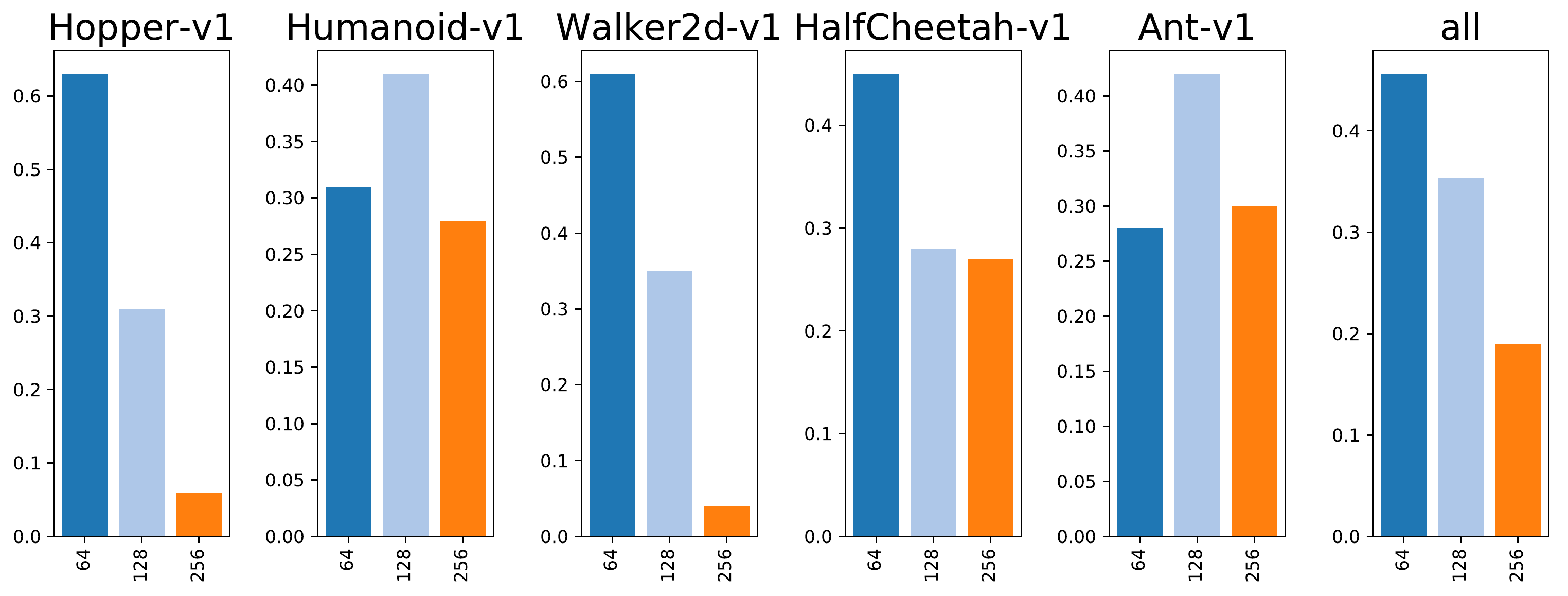}}
\caption{Analysis of choice \choicet{numenvs}: 95th percentile of performance scores conditioned on choice (left) and distribution of choices in top 5\% of configurations (right).}
\label{fig:final_setup__gin_study_design_choice_value_num_actors_in_learner}
\end{center}
\end{figure}

\begin{figure}[ht]
\begin{center}
\centerline{\includegraphics[width=0.45\textwidth]{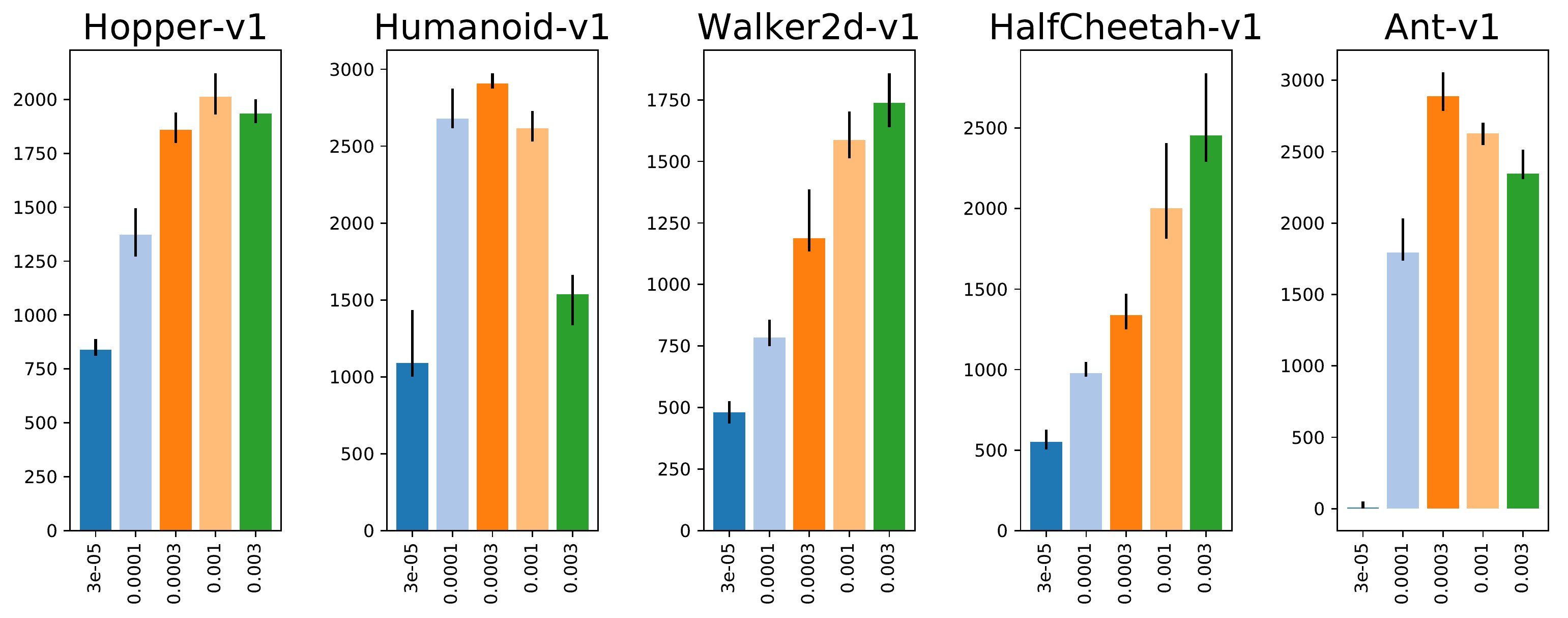}\hspace{1cm}\includegraphics[width=0.45\textwidth]{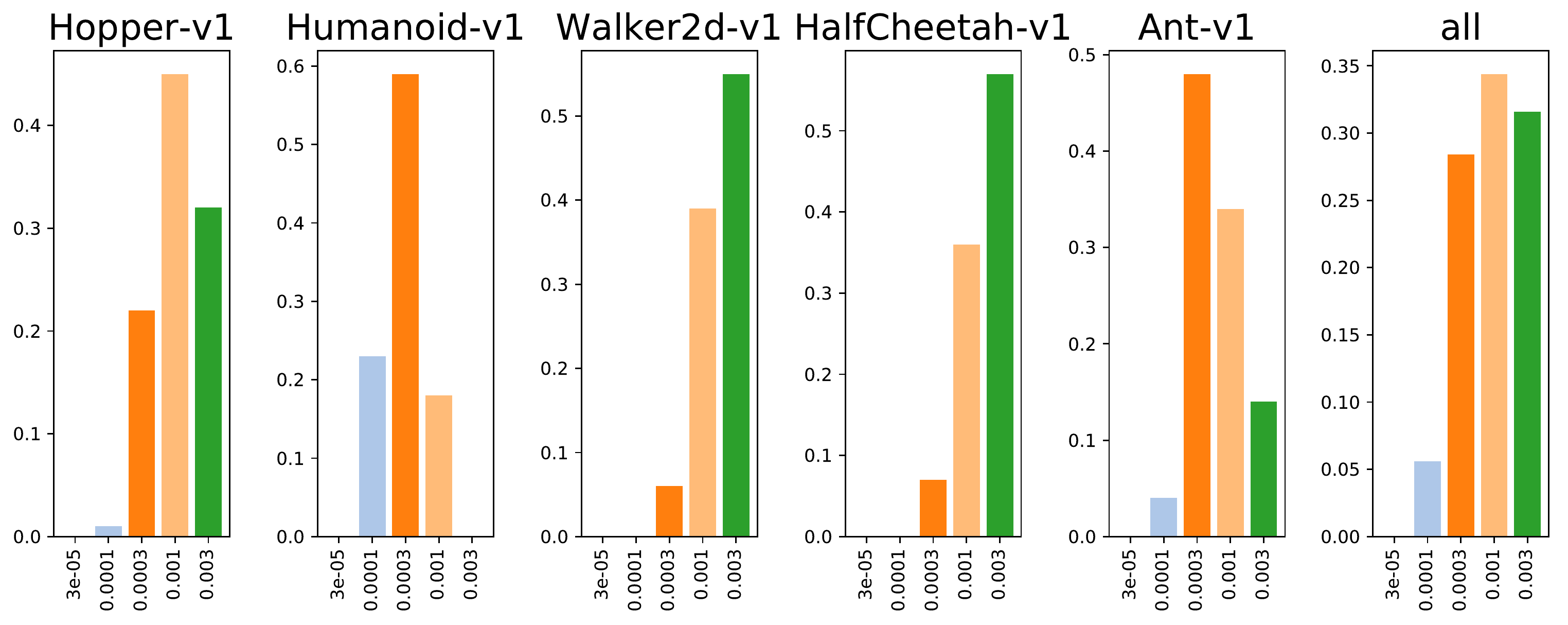}}
\caption{Analysis of choice \choicet{adamlr}: 95th percentile of performance scores conditioned on choice (left) and distribution of choices in top 5\% of configurations (right).}
\label{fig:final_setup__gin_study_design_choice_value_learning_rate}
\end{center}
\end{figure}

\begin{figure}[ht]
\begin{center}
\centerline{\includegraphics[width=0.45\textwidth]{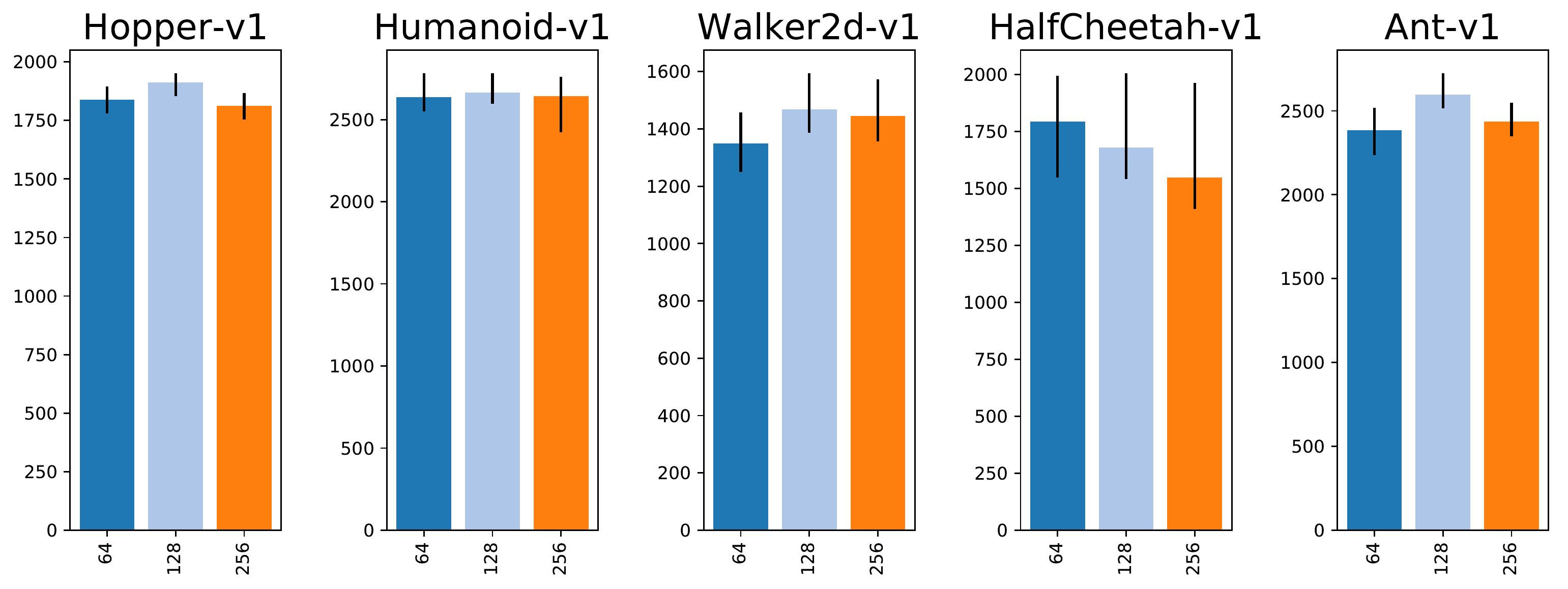}\hspace{1cm}\includegraphics[width=0.45\textwidth]{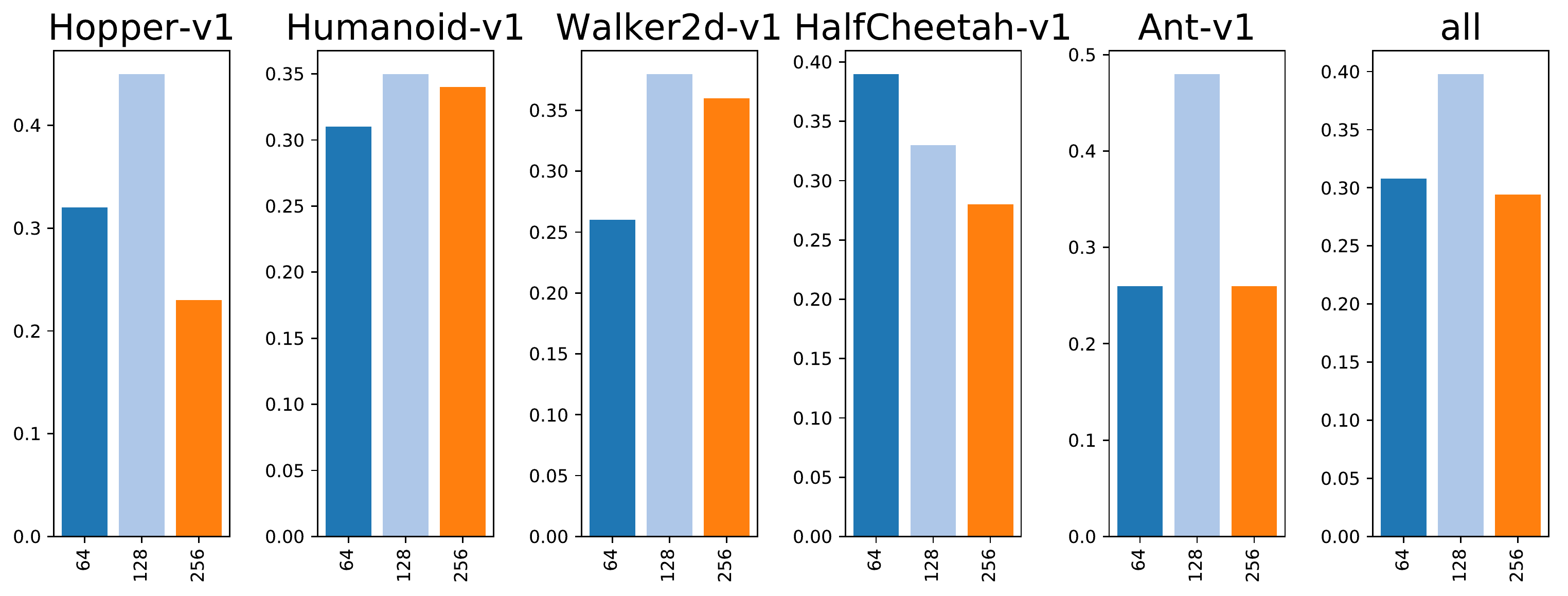}}
\caption{Analysis of choice \choicet{batchsize}: 95th percentile of performance scores conditioned on choice (left) and distribution of choices in top 5\% of configurations (right).}
\label{fig:final_setup__gin_study_design_choice_value_batch_size_transitions}
\end{center}
\end{figure}

\begin{figure}[ht]
\begin{center}
\centerline{\includegraphics[width=0.45\textwidth]{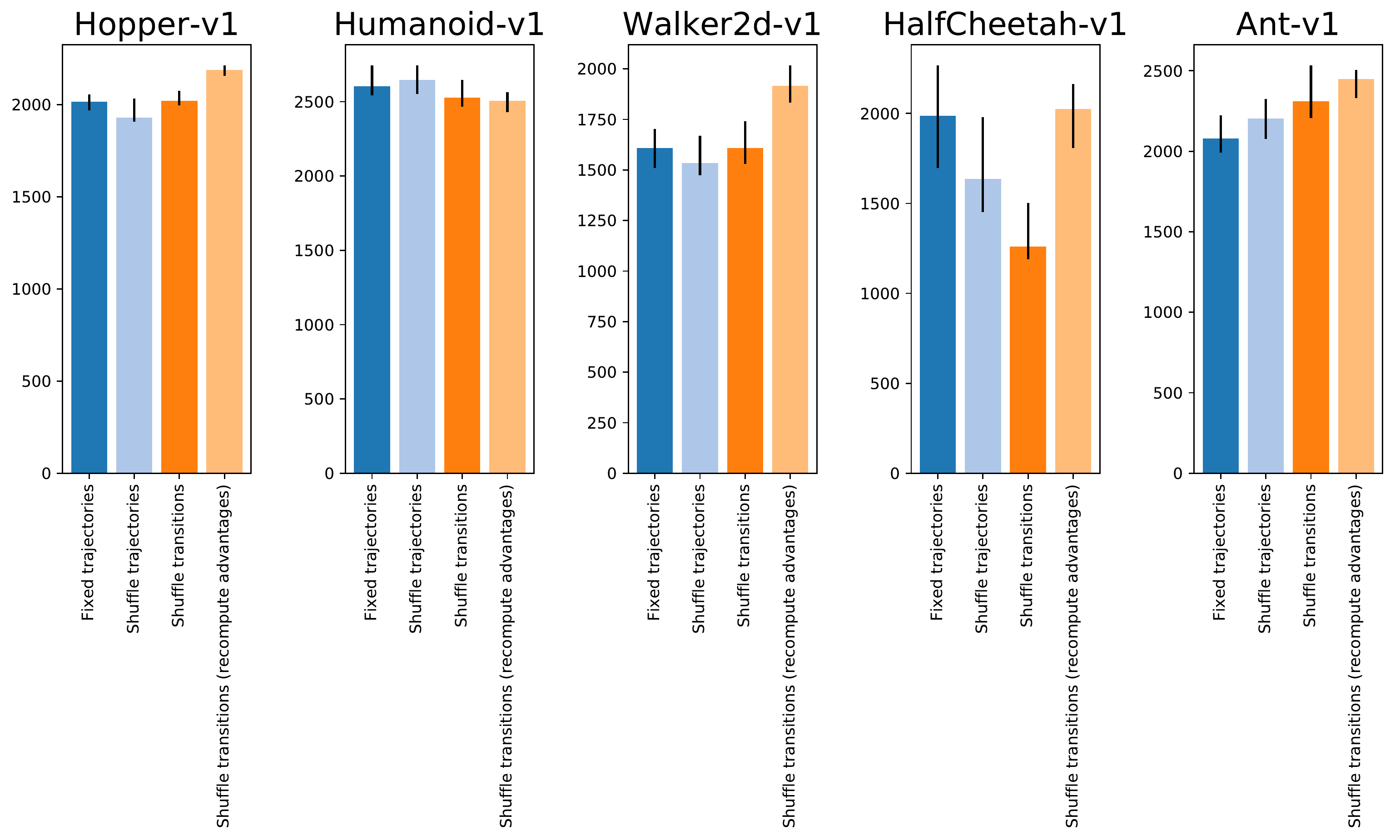}\hspace{1cm}\includegraphics[width=0.45\textwidth]{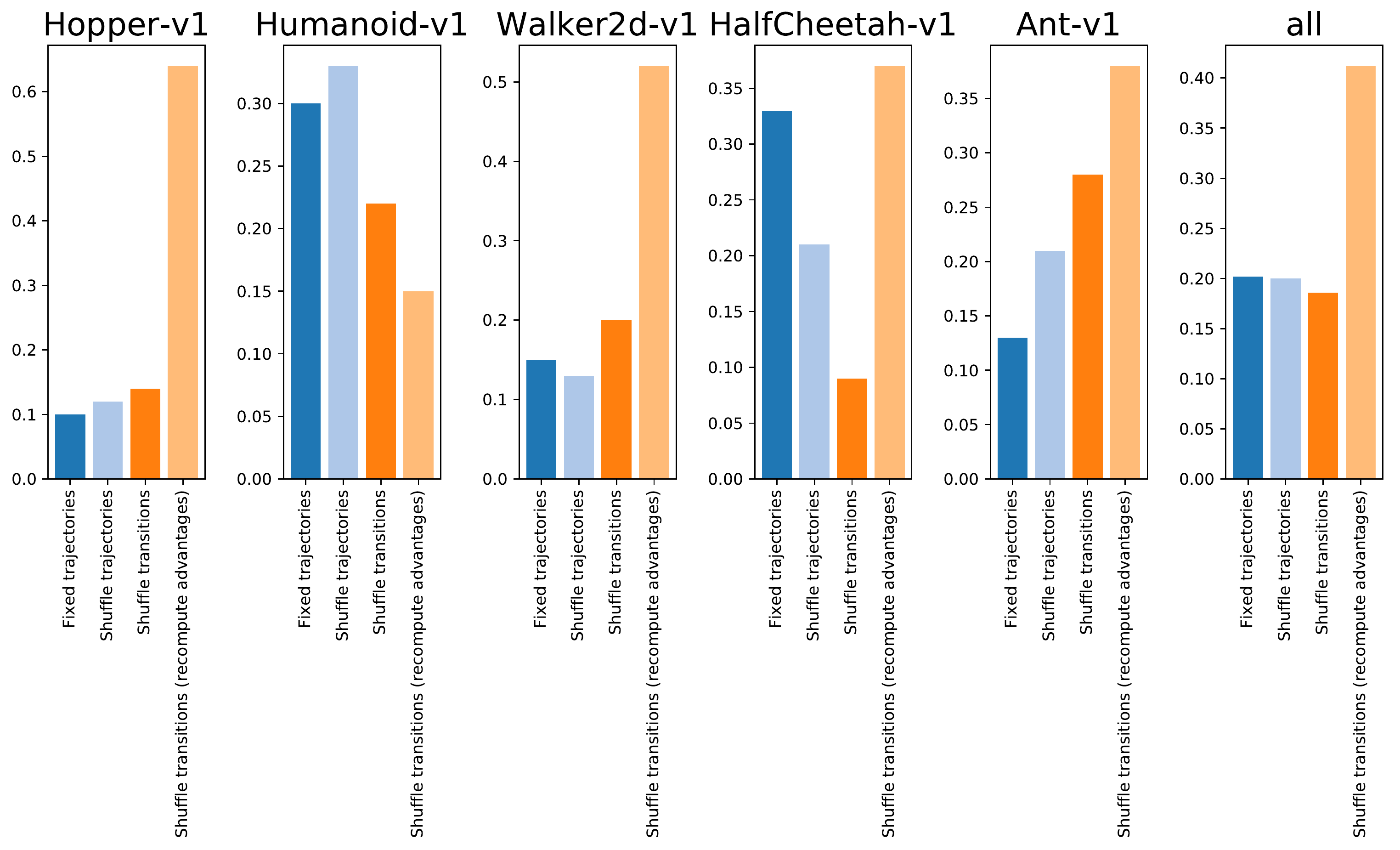}}
\caption{Analysis of choice \choicet{batchhandling}: 95th percentile of performance scores conditioned on choice (left) and distribution of choices in top 5\% of configurations(right). In order to obtain narrower confidence intervals in this experiment we only sweep \choicet{batchhandling}, \choicet{numenvs}, \choicet{adamlr}.}
\label{fig:final_setup2__gin_study_design_choice_value_batch_mode}
\end{center}
\end{figure}

We report aggregate statistics of the experiment in Table~\ref{tab:final_setup_overview} as well as training curves in Figure~\ref{fig:final_setup_training_curves}.
For each of the investigated choices in this experiment, we further provide a per-choice analysis in Figures~\ref{fig:final_setup__gin_study_design_choice_value_step_size_transitions}-\ref{fig:final_setup2__gin_study_design_choice_value_batch_mode}.
\begin{table}[ht]
\begin{center}
\caption{Performance quantiles across choice configurations.}
\label{tab:final_setup_overview}
\begin{tabular}{lrrrrr}
\toprule
{} & Ant-v1 & HalfCheetah-v1 & Hopper-v1 & Humanoid-v1 & Walker2d-v1 \\
\midrule
90th percentile &   2203 &           1316 &      1695 &        2310 &        1190 \\
95th percentile &   2484 &           1673 &      1853 &        2655 &        1431 \\
99th percentile &   2907 &           2665 &      2060 &        3014 &        1844 \\
Max             &   3563 &           3693 &      2434 &        3502 &        2426 \\
\bottomrule
\end{tabular}

\end{center}
\end{table}

\clearpage

%% file: final_time/main.tex
\clearpage
\section{Experiment \texttt{Time}}
\label{exp_final_time}
\subsection{Design}
\label{exp_design_final_time}
For each of the 5 environments, we sampled 2000 choice configurations where we sampled the following choices independently and uniformly from the following ranges:
\begin{itemize}
    \item \choicet{discount}: \{0.95, 0.97, 0.99, 0.999\}
    \item \choicet{frameskip}: \{1, 2, 5\}
    \item \choicet{handleabandon}: \{False, True\}
    \item \choicet{adamlr}: \{3e-05, 0.0001, 0.0003, 0.001\}
\end{itemize}
All the other choices were set to the default values as described in Appendix~\ref{sec:default_settings}.

For each of the sampled choice configurations, we train 3 agents with different random seeds and compute the performance metric as described in Section~\ref{sec:performance}.
\subsection{Results}
\label{exp_results_final_time}
\begin{figure}[ht]
\begin{center}
\centerline{\includegraphics[width=1\textwidth]{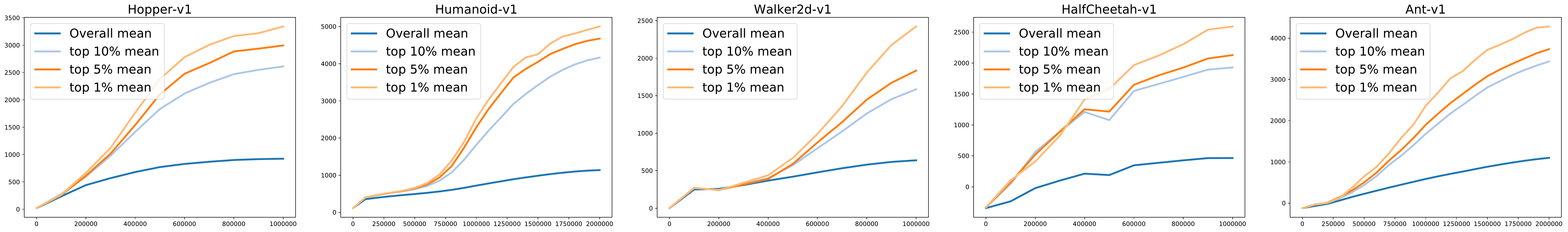}}
\caption{Training curves.}
\label{fig:final_time_training_curves}
\end{center}
\end{figure}

\begin{figure}[ht]
\begin{center}
\centerline{\includegraphics[width=0.45\textwidth]{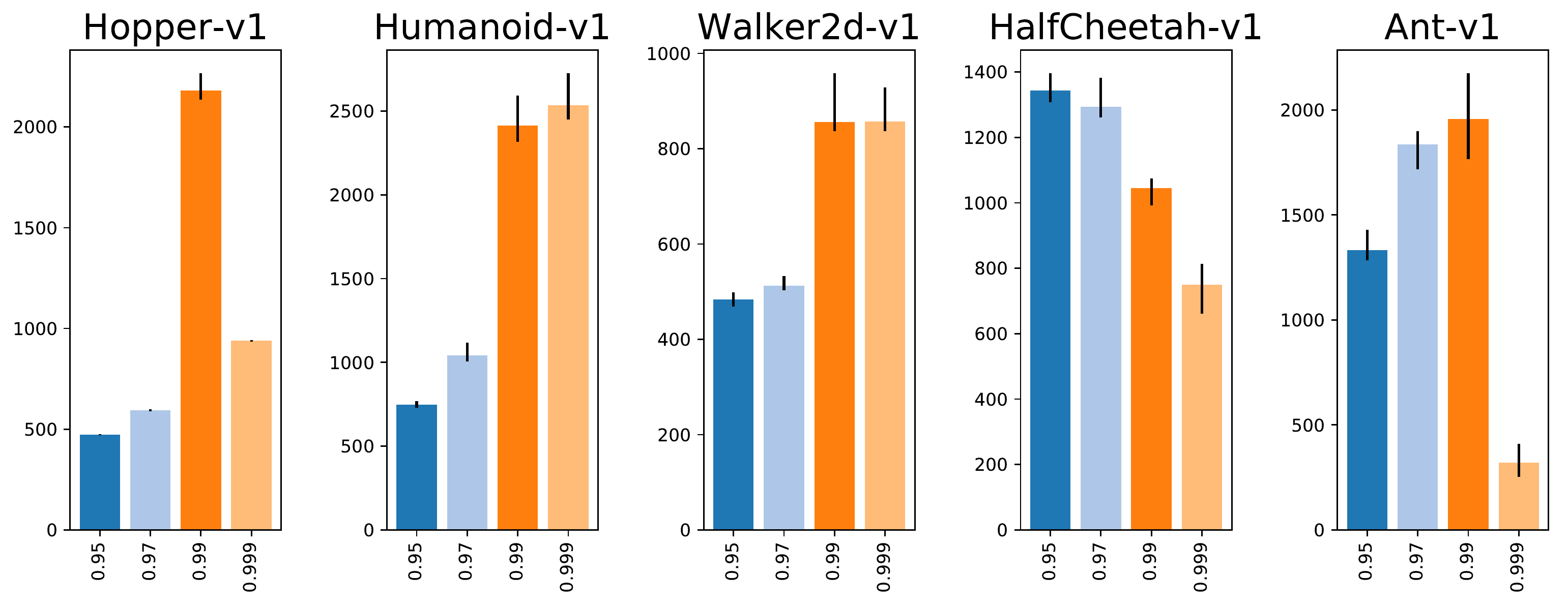}\hspace{1cm}\includegraphics[width=0.45\textwidth]{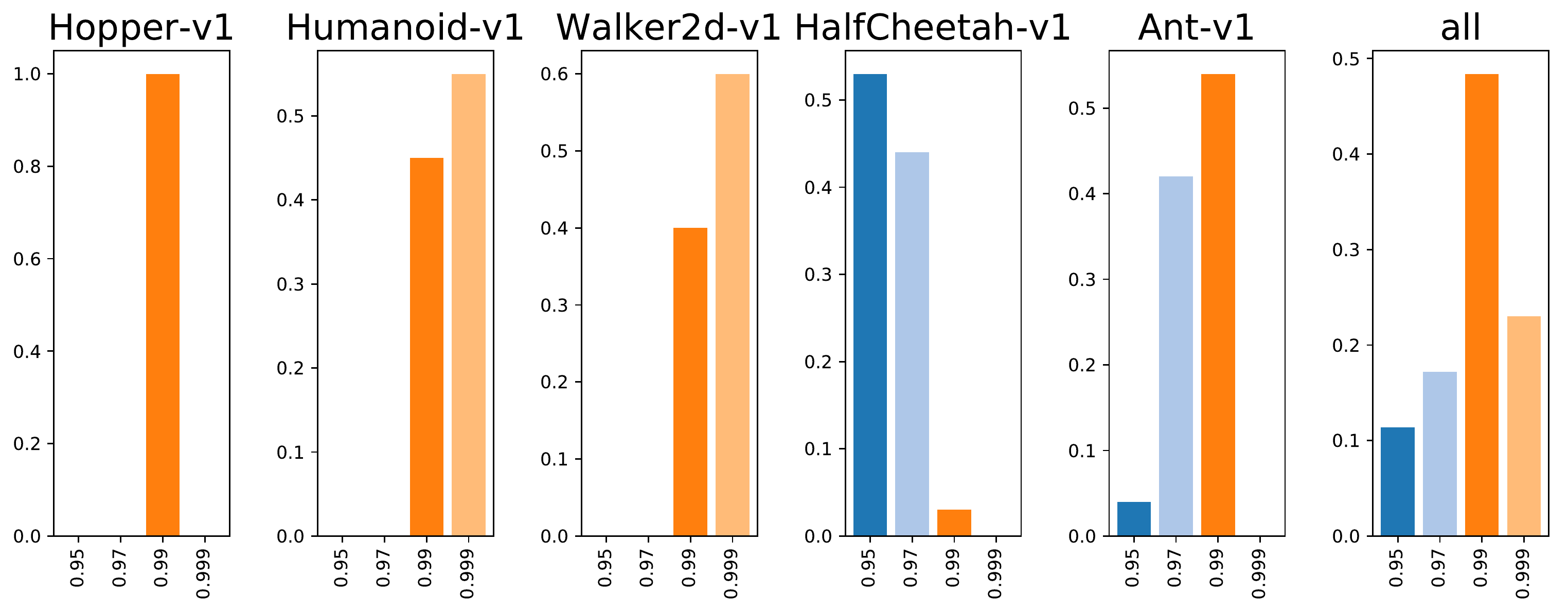}}
\caption{Analysis of choice \choicet{discount}: 95th percentile of performance scores conditioned on choice (left) and distribution of choices in top 5\% of configurations (right).}
\label{fig:final_time__gin_study_design_choice_value_discount_factor}
\end{center}
\end{figure}

\begin{figure}[ht]
\begin{center}
\centerline{\includegraphics[width=0.45\textwidth]{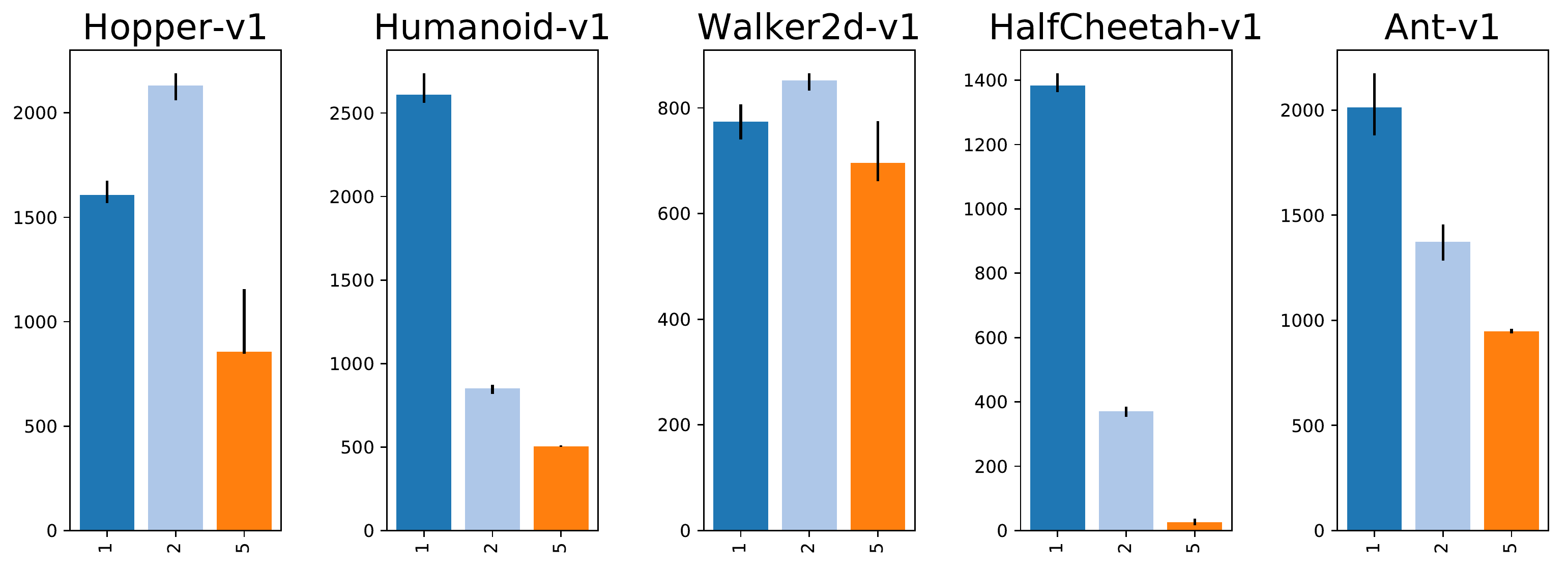}\hspace{1cm}\includegraphics[width=0.45\textwidth]{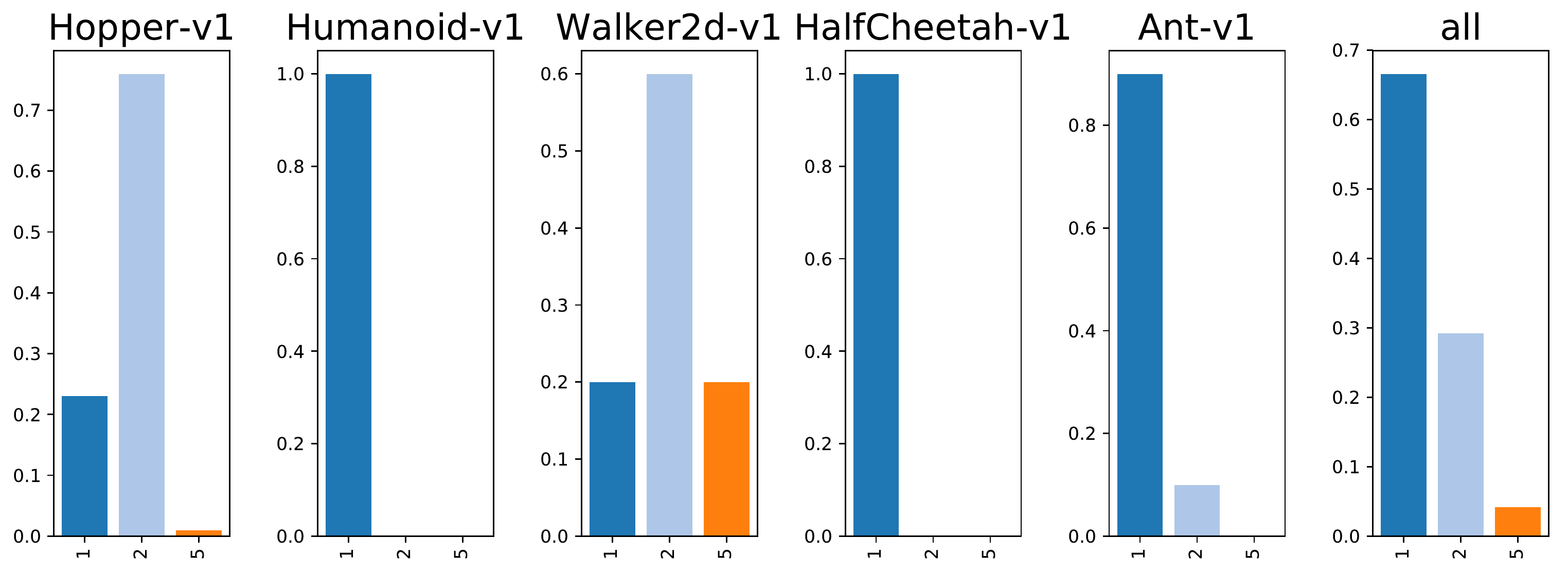}}
\caption{Analysis of choice \choicet{frameskip}: 95th percentile of performance scores conditioned on choice (left) and distribution of choices in top 5\% of configurations (right).}
\label{fig:final_time__gin_study_design_choice_value_frame_skip}
\end{center}
\end{figure}

\begin{figure}[ht]
\begin{center}
\centerline{\includegraphics[width=0.45\textwidth]{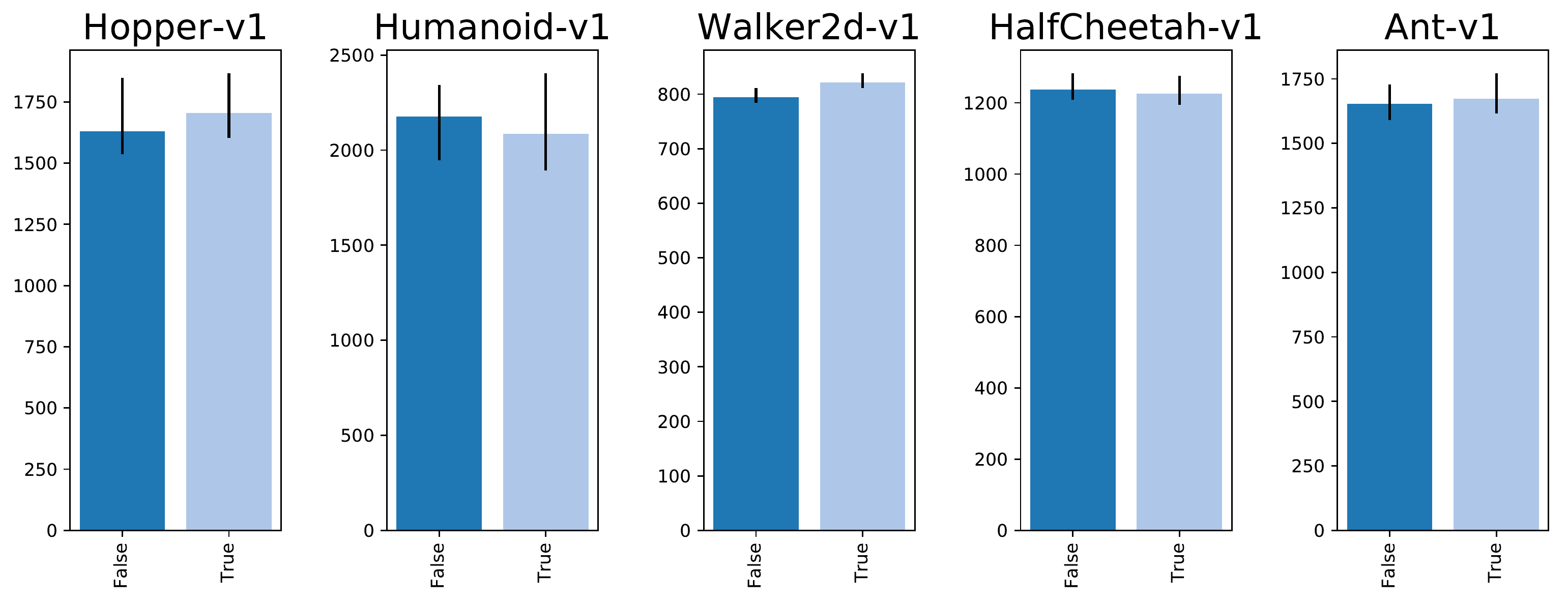}\hspace{1cm}\includegraphics[width=0.45\textwidth]{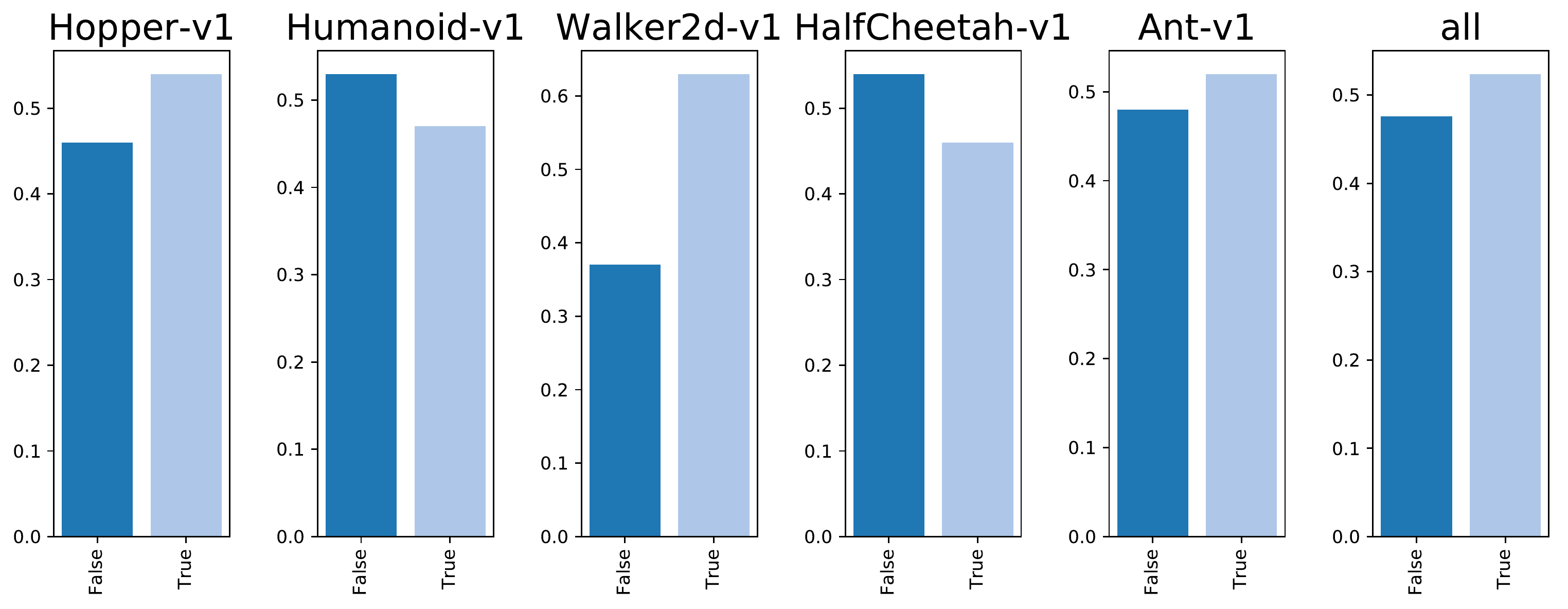}}
\caption{Analysis of choice \choicet{handleabandon}: 95th percentile of performance scores conditioned on choice (left) and distribution of choices in top 5\% of configurations (right).}
\label{fig:final_time__gin_study_design_choice_value_handle_abandoned_episodes_properly}
\end{center}
\end{figure}

\begin{figure}[ht]
\begin{center}
\centerline{\includegraphics[width=0.45\textwidth]{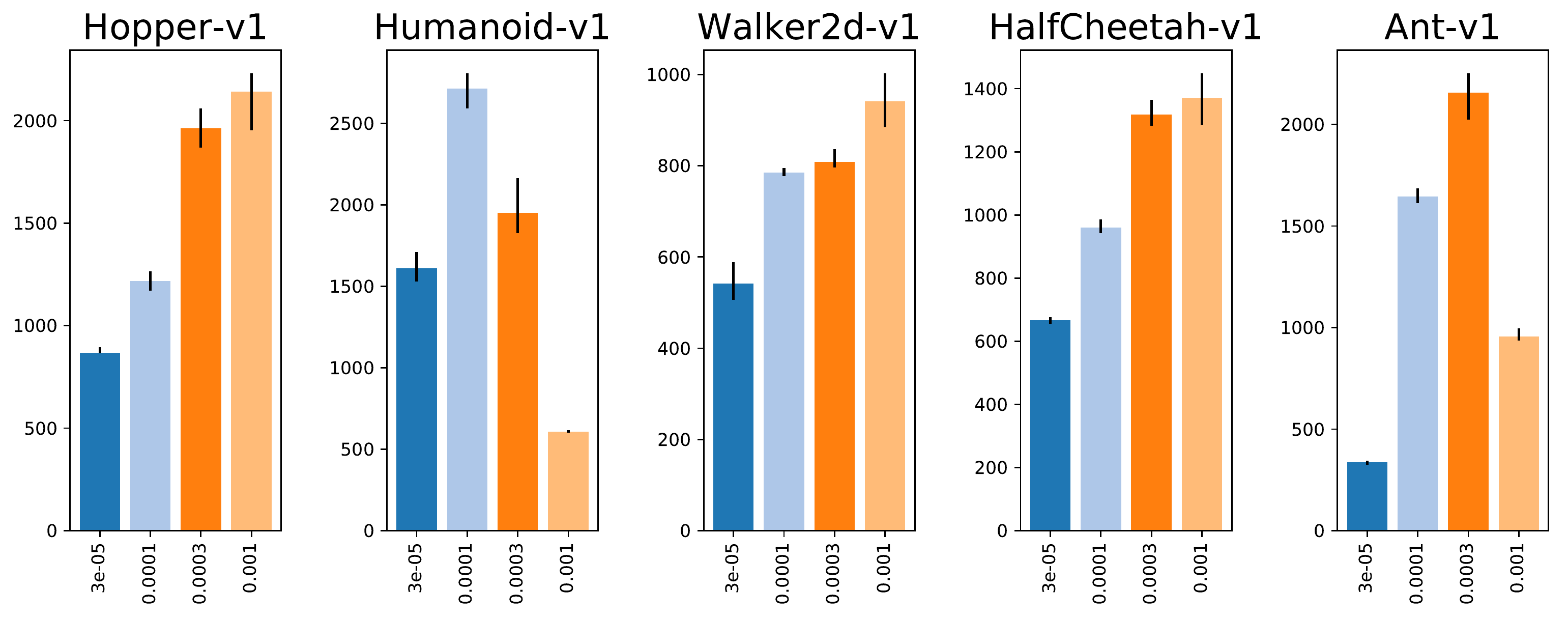}\hspace{1cm}\includegraphics[width=0.45\textwidth]{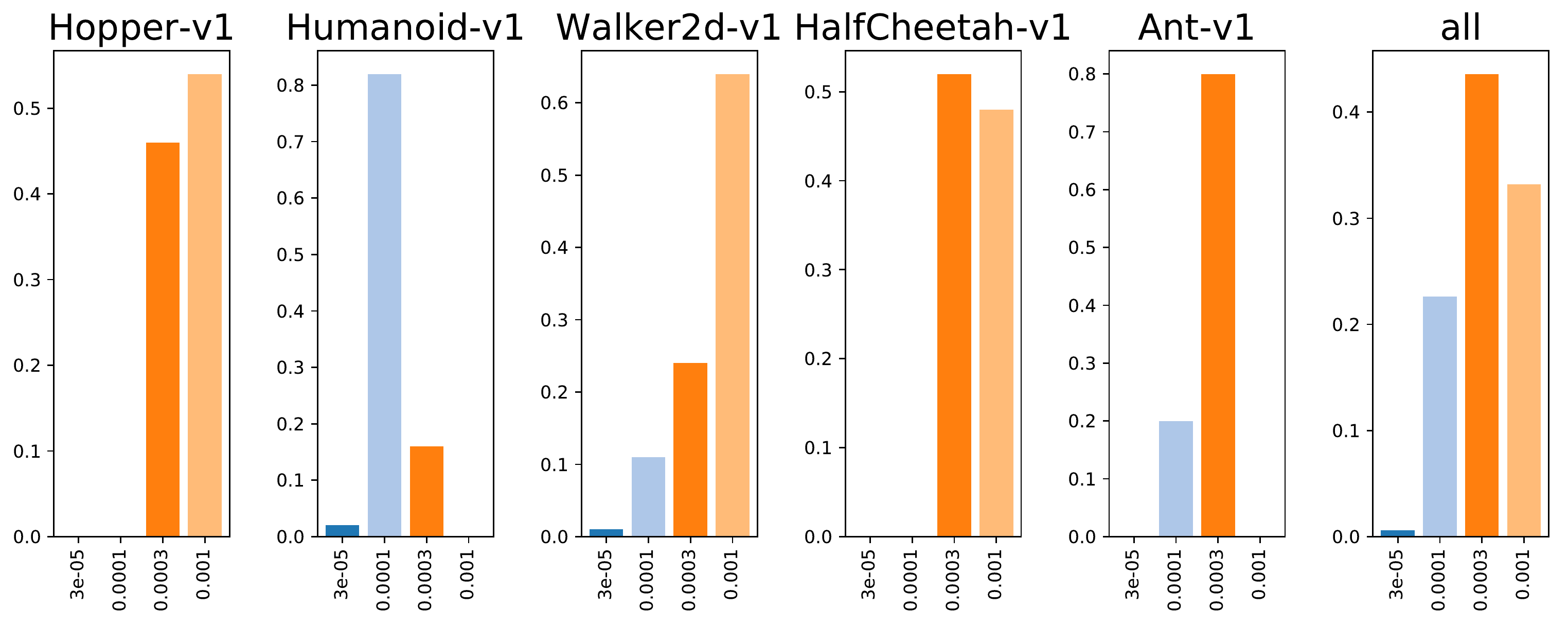}}
\caption{Analysis of choice \choicet{adamlr}: 95th percentile of performance scores conditioned on choice (left) and distribution of choices in top 5\% of configurations (right).}
\label{fig:final_time__gin_study_design_choice_value_learning_rate}
\end{center}
\end{figure}

We report aggregate statistics of the experiment in Table~\ref{tab:final_time_overview} as well as training curves in Figure~\ref{fig:final_time_training_curves}.
For each of the investigated choices in this experiment, we further provide a per-choice analysis in Figures~\ref{fig:final_time__gin_study_design_choice_value_discount_factor}-\ref{fig:final_time__gin_study_design_choice_value_learning_rate}.
\begin{table}[ht]
\begin{center}
\caption{Performance quantiles across choice configurations.}
\label{tab:final_time_overview}
\begin{tabular}{lrrrrr}
\toprule
{} & Ant-v1 & HalfCheetah-v1 & Hopper-v1 & Humanoid-v1 & Walker2d-v1 \\
\midrule
90th percentile &   1462 &           1063 &      1243 &        1431 &         761 \\
95th percentile &   1654 &           1235 &      1675 &        2158 &         810 \\
99th percentile &   2220 &           1423 &      2204 &        2769 &         974 \\
Max             &   2833 &           1918 &      2434 &        3106 &        1431 \\
\bottomrule
\end{tabular}

\end{center}
\end{table}

\clearpage

%% file: final_optimize/main.tex
\clearpage
\section{Experiment \texttt{Optimizers}}
\label{exp_final_optimize}
\subsection{Design}
\label{exp_design_final_optimize}
For each of the 5 environments, we sampled 2000 choice configurations where we sampled the following choices independently and uniformly from the following ranges:
\begin{itemize}
    \item \choicet{lrdecay}: \{0.0, 1.0\}
    \item \choicet{optimizer}: \{Adam, RMSProp\}
    \begin{itemize}
        \item For the case ``\choicet{optimizer} = Adam'', we further sampled the sub-choices:
        \begin{itemize}
            \item \choicet{adammom}: \{0.0, 0.9\}
            \item \choicet{adameps}: \{1e-09, 1e-08, 1e-07, 1e-06, 1e-05, 0.0001\}
            \item \choicet{adamlr}: \{3e-05, 0.0001, 0.0003, 0.001\}
        \end{itemize}
        \item For the case ``\choicet{optimizer} = RMSProp'', we further sampled the sub-choices:
        \begin{itemize}
            \item \choicet{rmscent}: \{False, True\}
            \item \choicet{rmsmom}: \{0.0, 0.9\}
            \item \choicet{rmseps}: \{1e-09, 1e-08, 1e-07, 1e-06, 1e-05, 0.0001\}
            \item \choicet{rmslr}: \{3e-05, 0.0001, 0.0003, 0.001\}
        \end{itemize}
    \end{itemize}
\end{itemize}
All the other choices were set to the default values as described in Appendix~\ref{sec:default_settings}.

For each of the sampled choice configurations, we train 3 agents with different random seeds and compute the performance metric as described in Section~\ref{sec:performance}.
\subsection{Results}
\label{exp_results_final_optimize}
\begin{figure}[ht]
\begin{center}
\centerline{\includegraphics[width=1\textwidth]{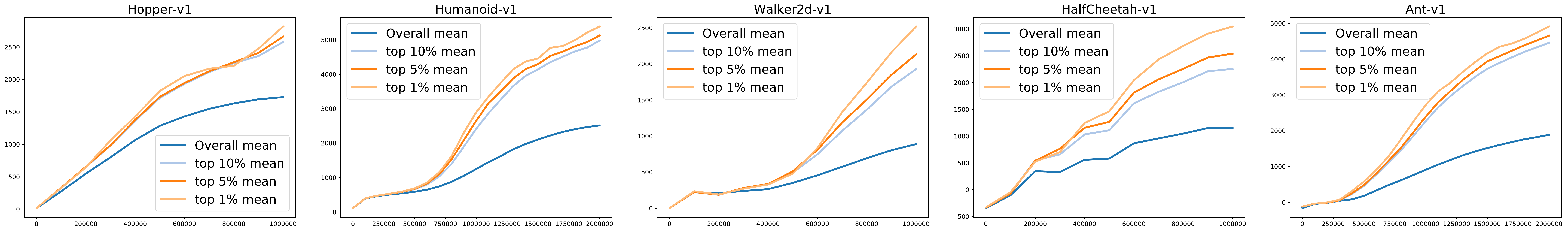}}
\caption{Training curves.}
\label{fig:final_optimize_training_curves}
\end{center}
\end{figure}

\begin{figure}[ht]
\begin{center}
\centerline{\includegraphics[width=0.45\textwidth]{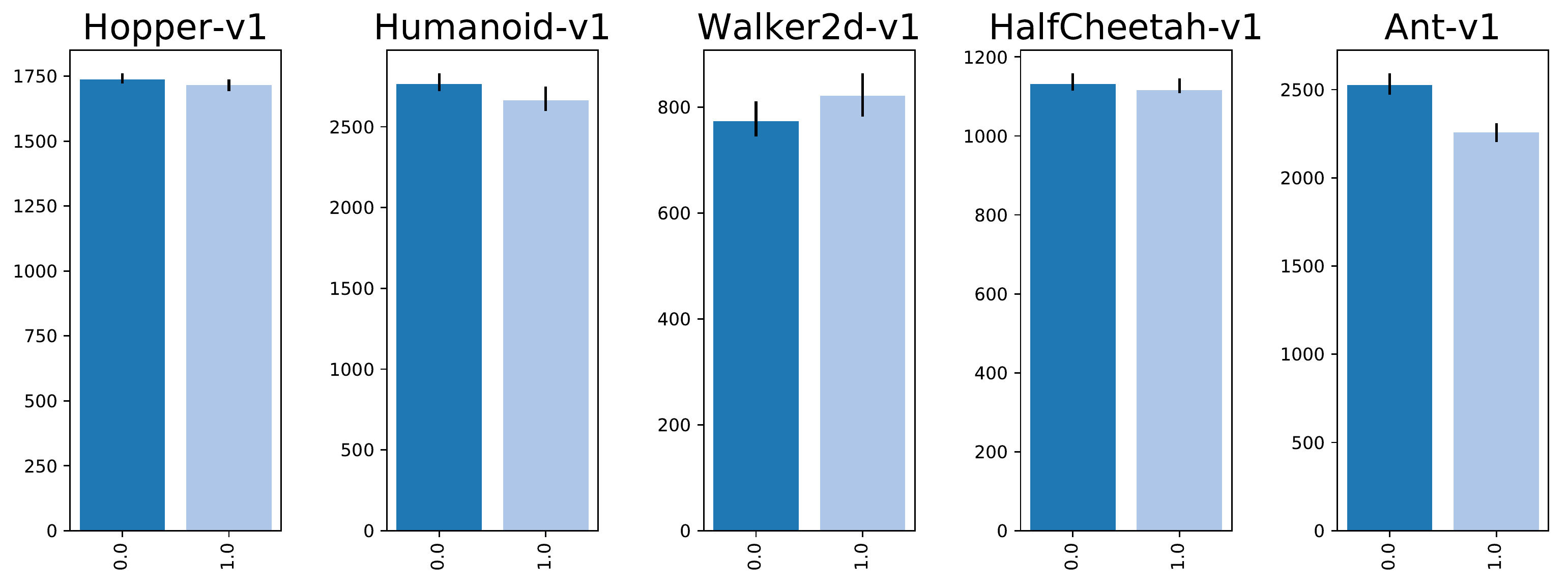}\hspace{1cm}\includegraphics[width=0.45\textwidth]{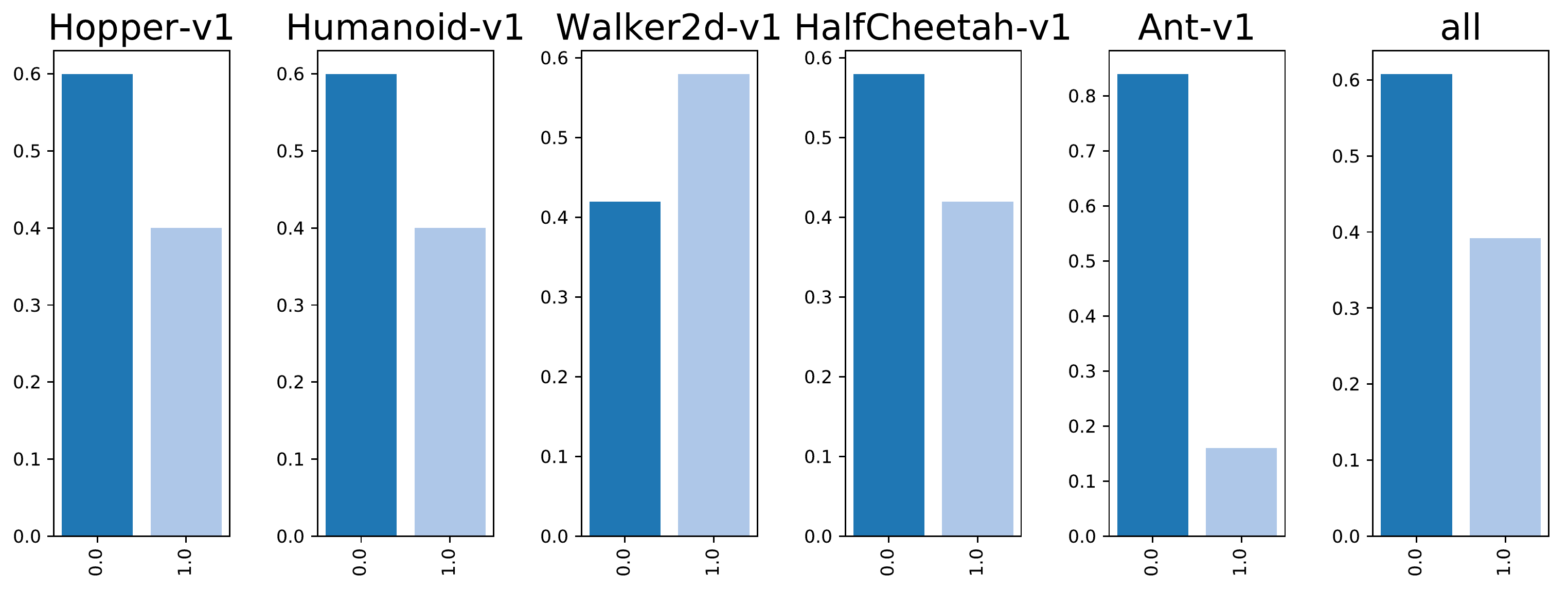}}
\caption{Analysis of choice \choicet{lrdecay}: 95th percentile of performance scores conditioned on choice (left) and distribution of choices in top 5\% of configurations (right).}
\label{fig:final_optimize__gin_study_design_choice_value_learning_rate_decay}
\end{center}
\end{figure}

\begin{figure}[ht]
\begin{center}
\centerline{\includegraphics[width=0.45\textwidth]{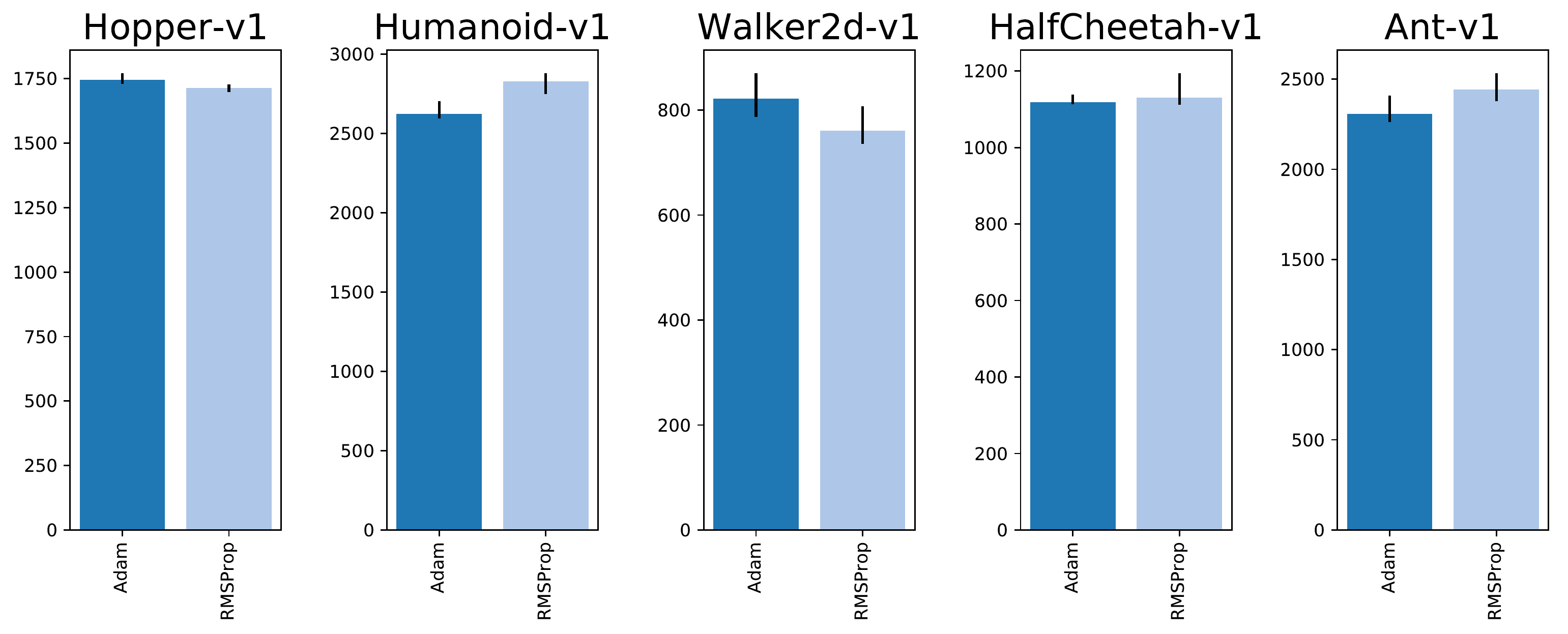}\hspace{1cm}\includegraphics[width=0.45\textwidth]{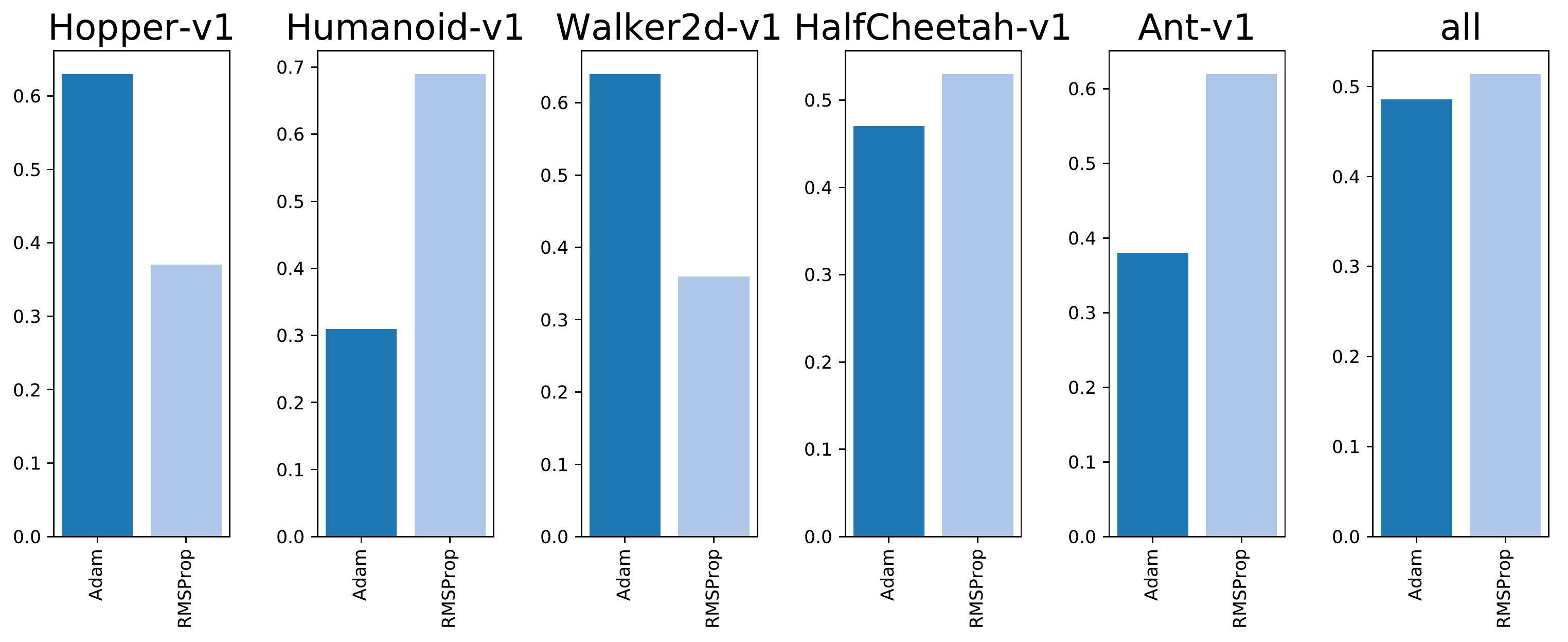}}
\caption{Analysis of choice \choicet{optimizer}: 95th percentile of performance scores conditioned on choice (left) and distribution of choices in top 5\% of configurations (right).}
\label{fig:final_optimize__gin_study_design_choice_value_optimizer}
\end{center}
\end{figure}

\begin{figure}[ht]
\begin{center}
\centerline{\includegraphics[width=0.45\textwidth]{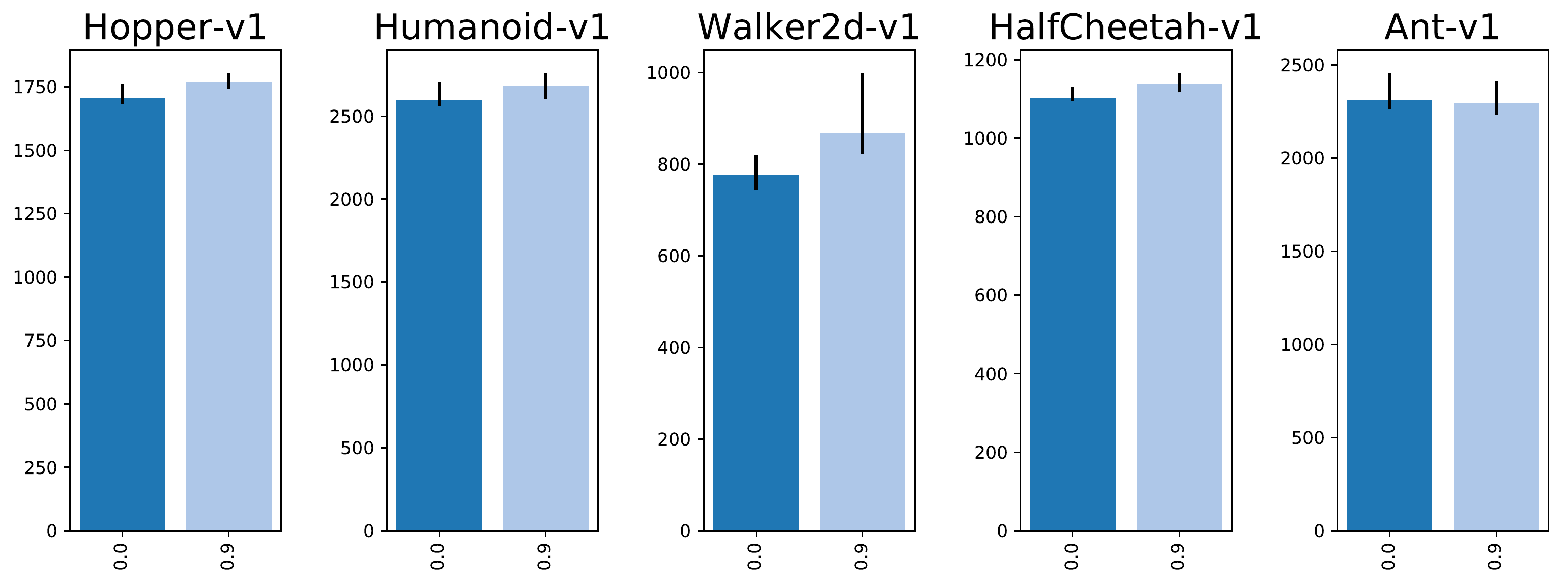}\hspace{1cm}\includegraphics[width=0.45\textwidth]{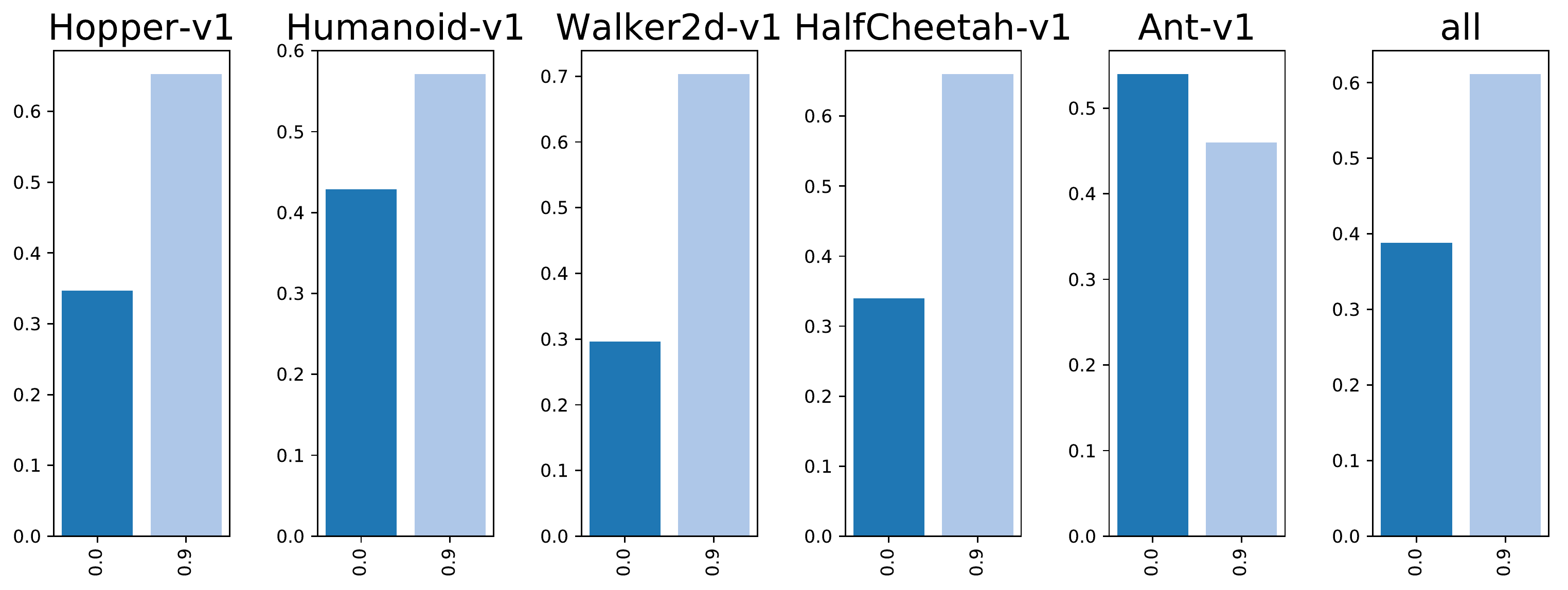}}
\caption{Analysis of choice \choicet{adammom}: 95th percentile of performance scores conditioned on sub-choice (left) and distribution of sub-choices in top 5\% of configurations (right).}
\label{fig:final_optimize__gin_study_design_choice_value_sub_optimizer_adam_momentum}
\end{center}
\end{figure}

\begin{figure}[ht]
\begin{center}
\centerline{\includegraphics[width=0.45\textwidth]{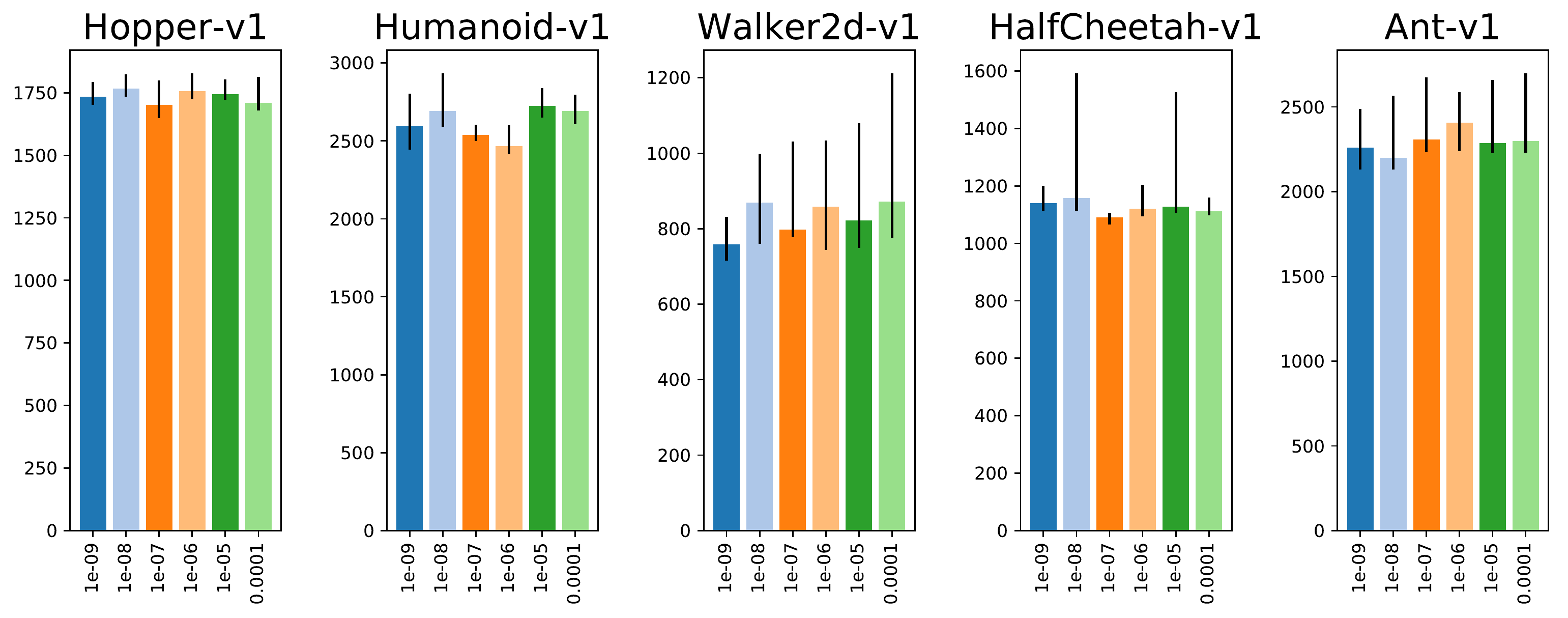}\hspace{1cm}\includegraphics[width=0.45\textwidth]{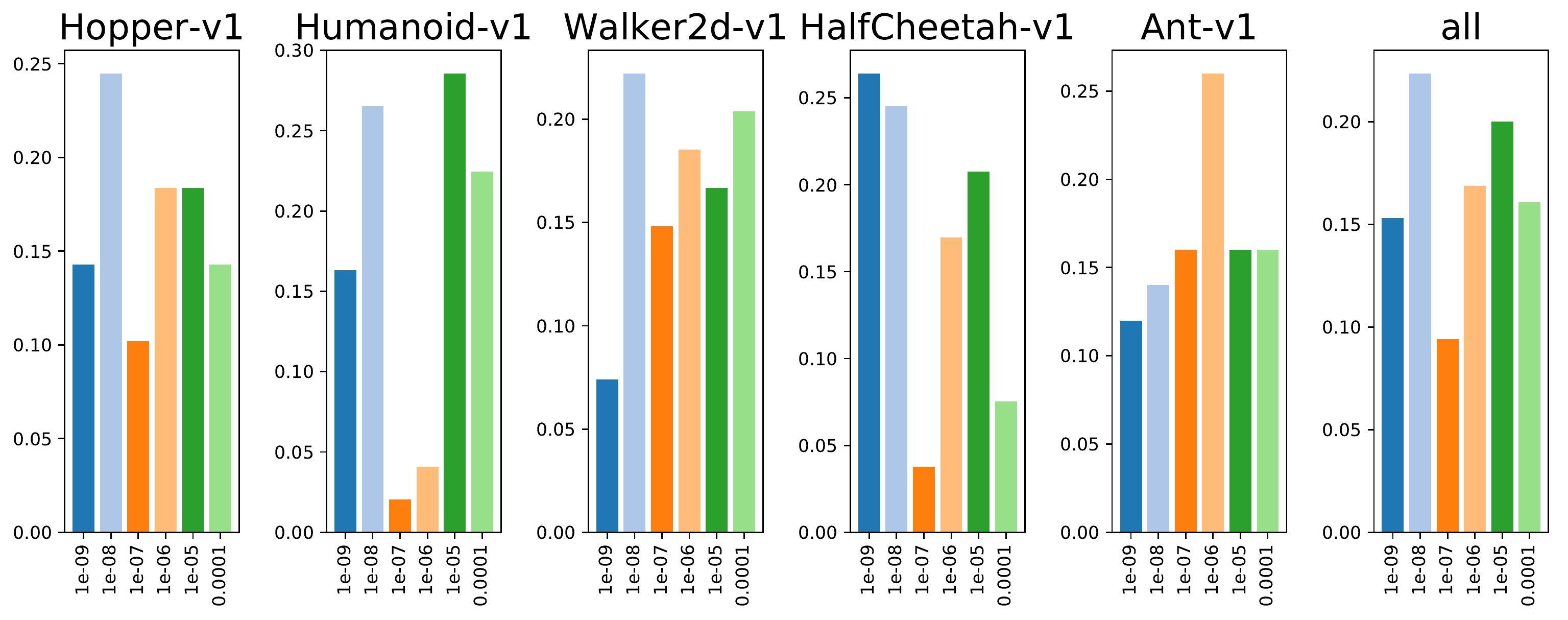}}
\caption{Analysis of choice \choicet{adameps}: 95th percentile of performance scores conditioned on sub-choice (left) and distribution of sub-choices in top 5\% of configurations (right).}
\label{fig:final_optimize__gin_study_design_choice_value_sub_optimizer_adam_epsilon}
\end{center}
\end{figure}

\begin{figure}[ht]
\begin{center}
\centerline{\includegraphics[width=0.45\textwidth]{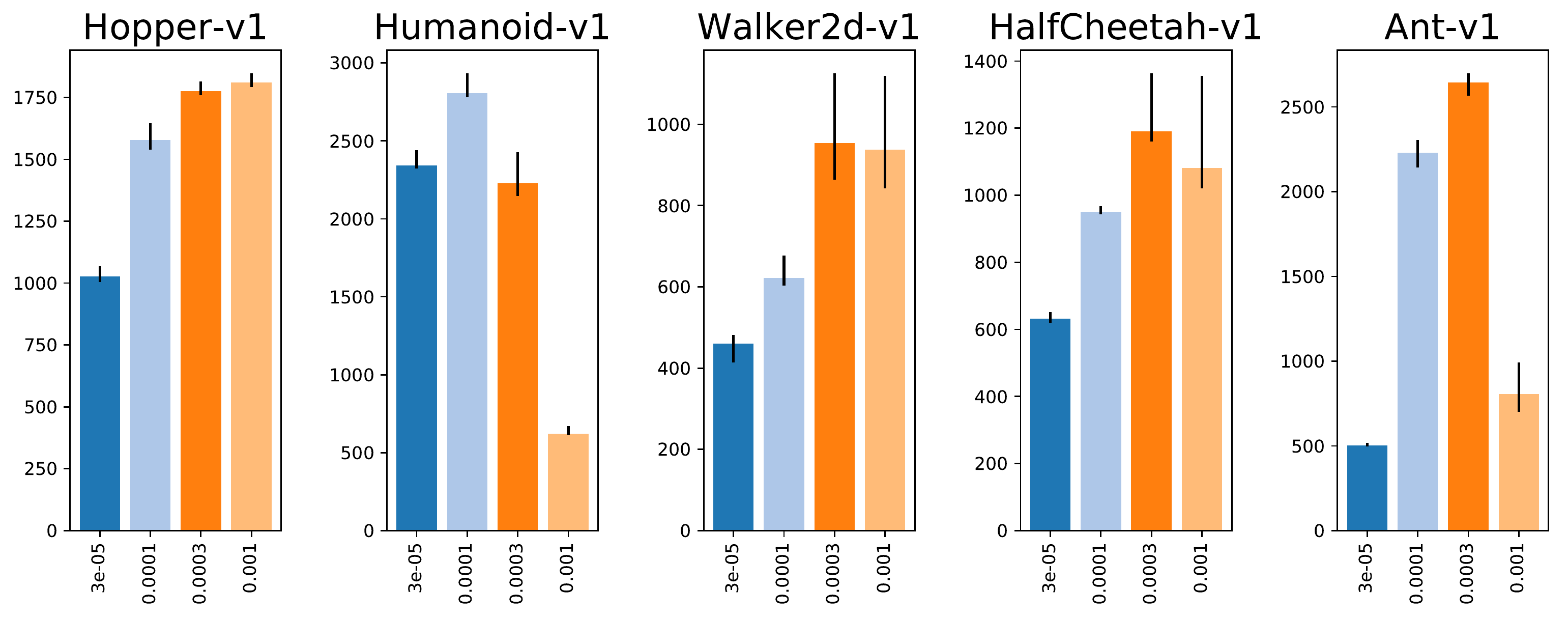}\hspace{1cm}\includegraphics[width=0.45\textwidth]{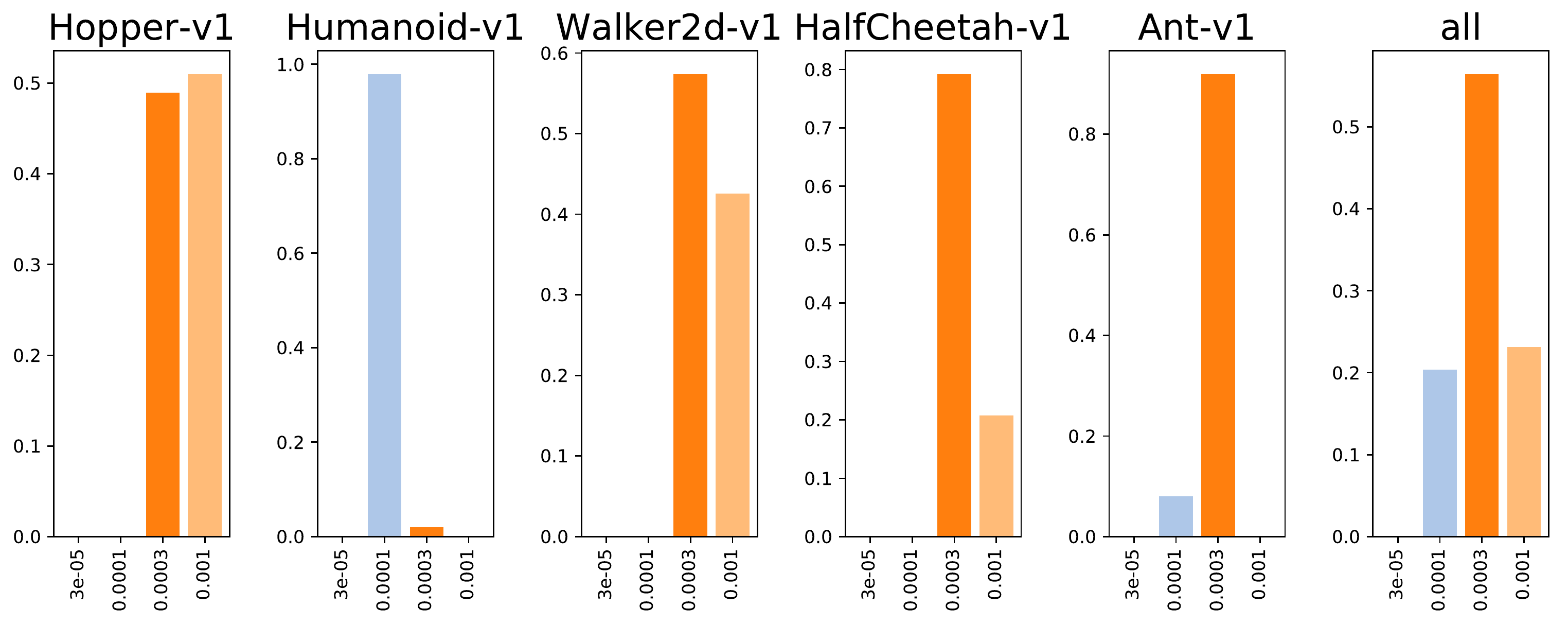}}
\caption{Analysis of choice \choicet{adamlr}: 95th percentile of performance scores conditioned on sub-choice (left) and distribution of sub-choices in top 5\% of configurations (right).}
\label{fig:final_optimize__gin_study_design_choice_value_sub_optimizer_adam_learning_rate}
\end{center}
\end{figure}

\begin{figure}[ht]
\begin{center}
\centerline{\includegraphics[width=0.45\textwidth]{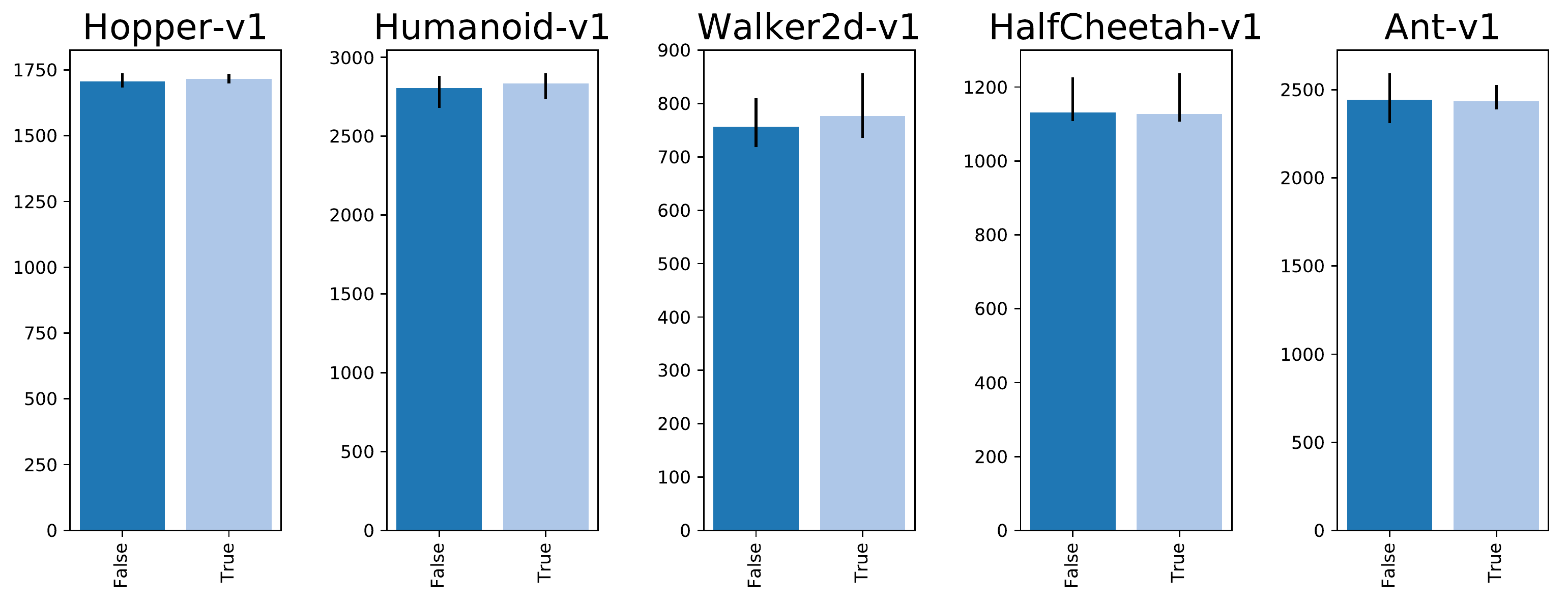}\hspace{1cm}\includegraphics[width=0.45\textwidth]{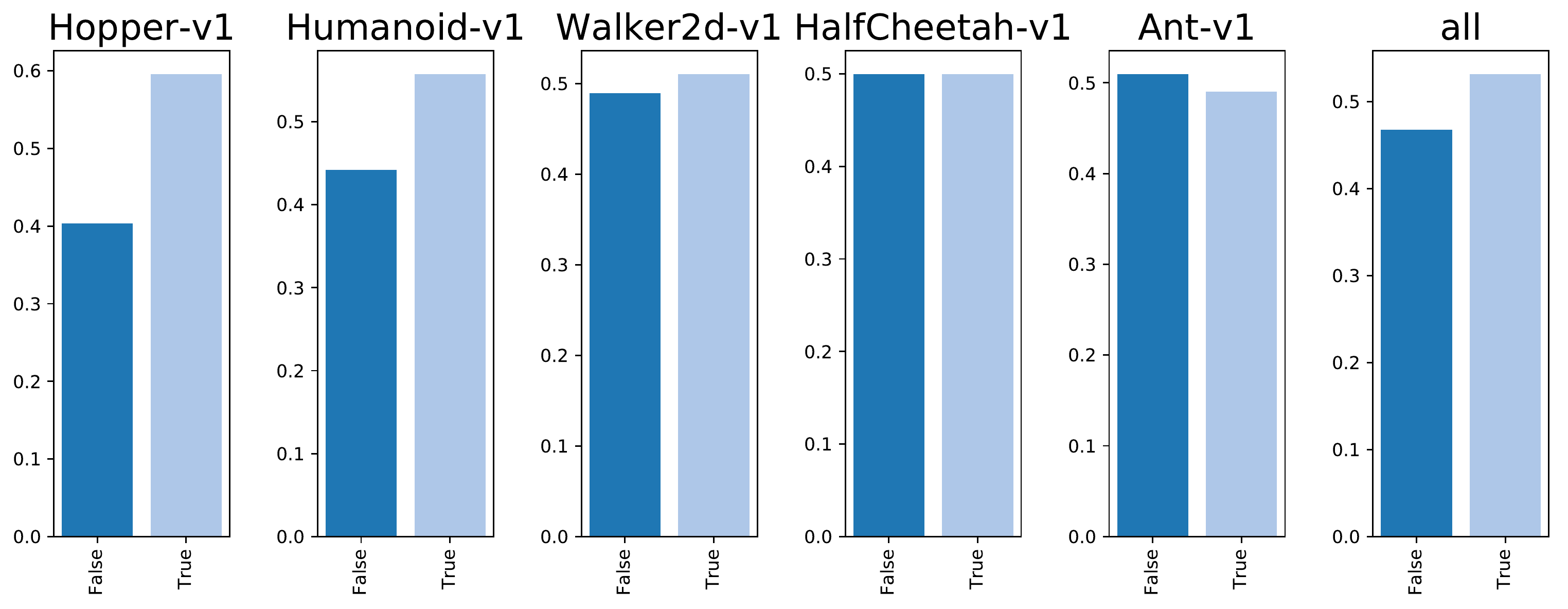}}
\caption{Analysis of choice \choicet{rmscent}: 95th percentile of performance scores conditioned on sub-choice (left) and distribution of sub-choices in top 5\% of configurations (right).}
\label{fig:final_optimize__gin_study_design_choice_value_sub_optimizer_rmsprop_centered}
\end{center}
\end{figure}

\begin{figure}[ht]
\begin{center}
\centerline{\includegraphics[width=0.45\textwidth]{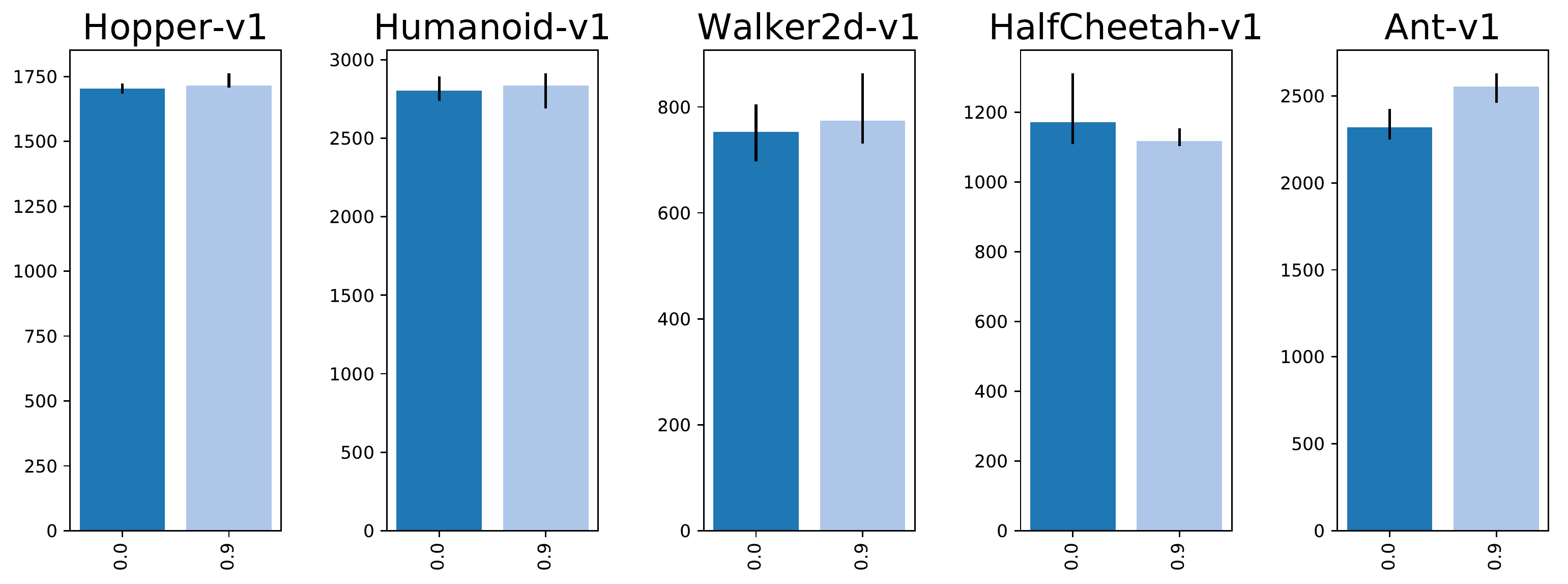}\hspace{1cm}\includegraphics[width=0.45\textwidth]{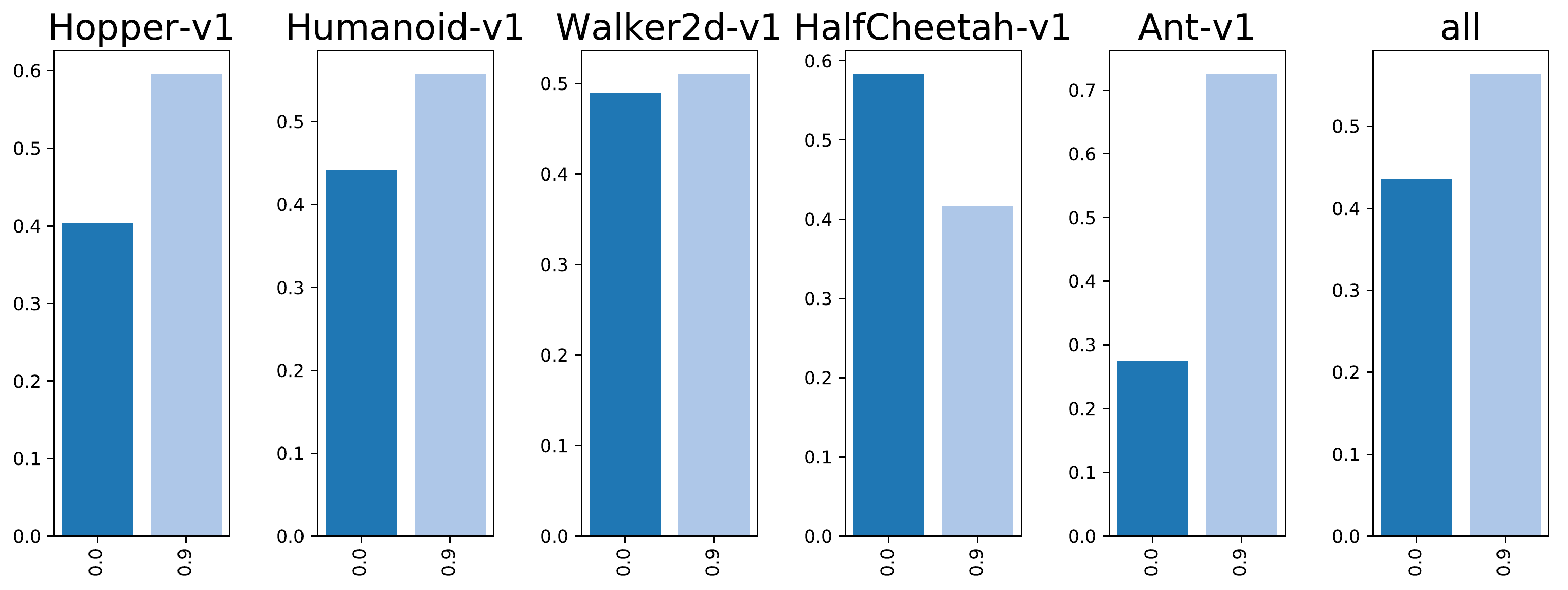}}
\caption{Analysis of choice \choicet{rmsmom}: 95th percentile of performance scores conditioned on sub-choice (left) and distribution of sub-choices in top 5\% of configurations (right).}
\label{fig:final_optimize__gin_study_design_choice_value_sub_optimizer_rmsprop_momentum}
\end{center}
\end{figure}

\begin{figure}[ht]
\begin{center}
\centerline{\includegraphics[width=0.45\textwidth]{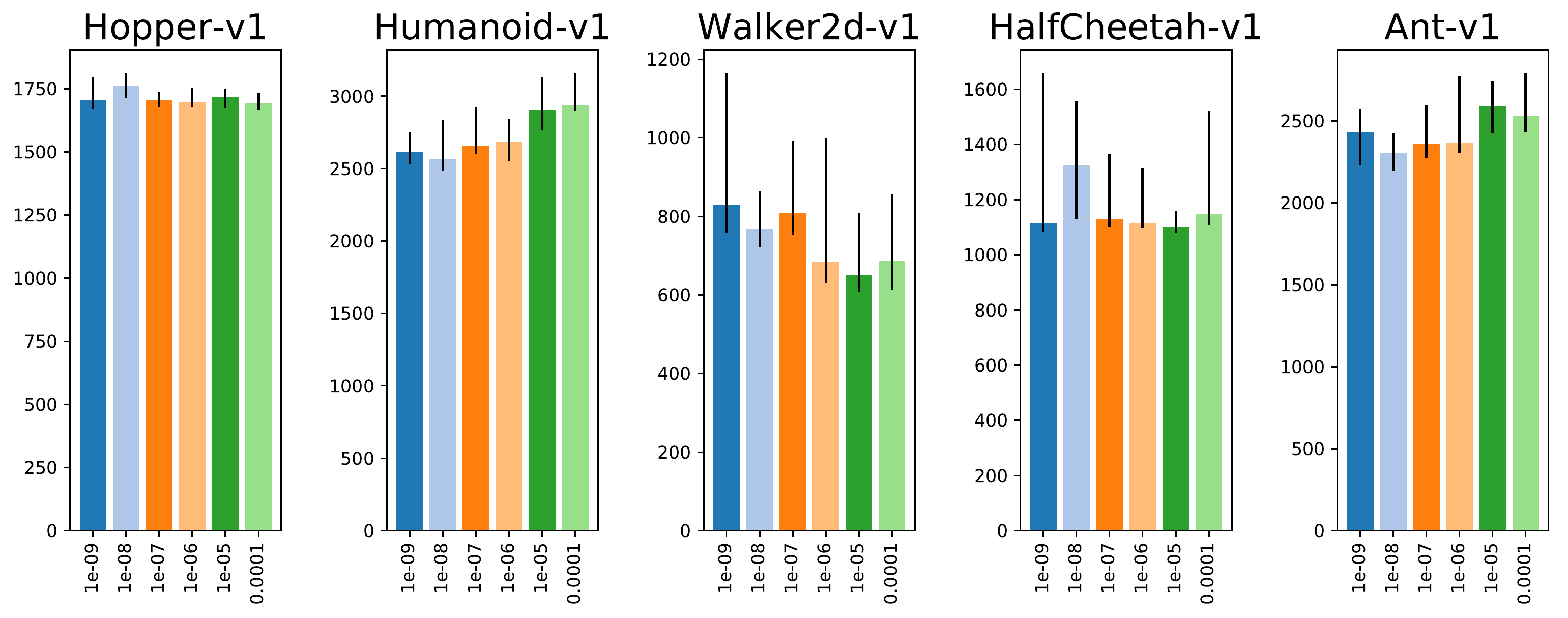}\hspace{1cm}\includegraphics[width=0.45\textwidth]{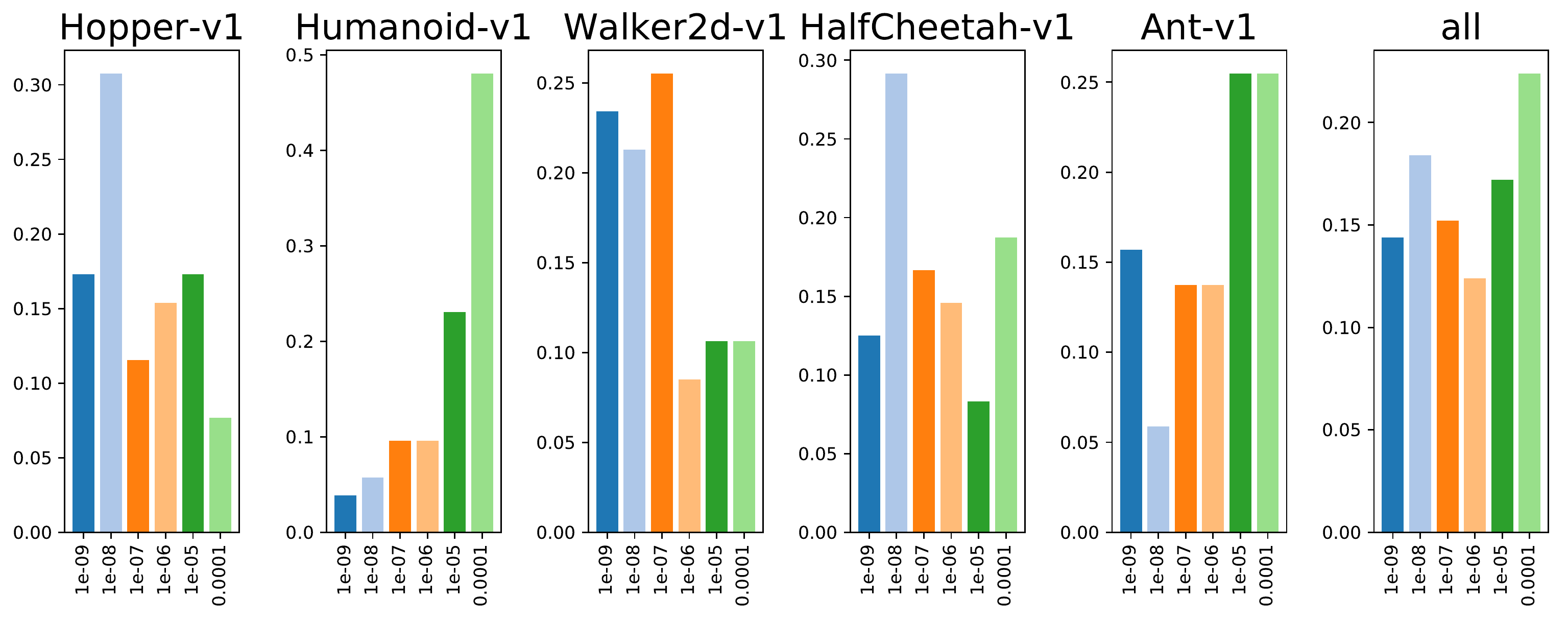}}
\caption{Analysis of choice \choicet{rmseps}: 95th percentile of performance scores conditioned on sub-choice (left) and distribution of sub-choices in top 5\% of configurations (right).}
\label{fig:final_optimize__gin_study_design_choice_value_sub_optimizer_rmsprop_epsilon}
\end{center}
\end{figure}

\begin{figure}[ht]
\begin{center}
\centerline{\includegraphics[width=0.45\textwidth]{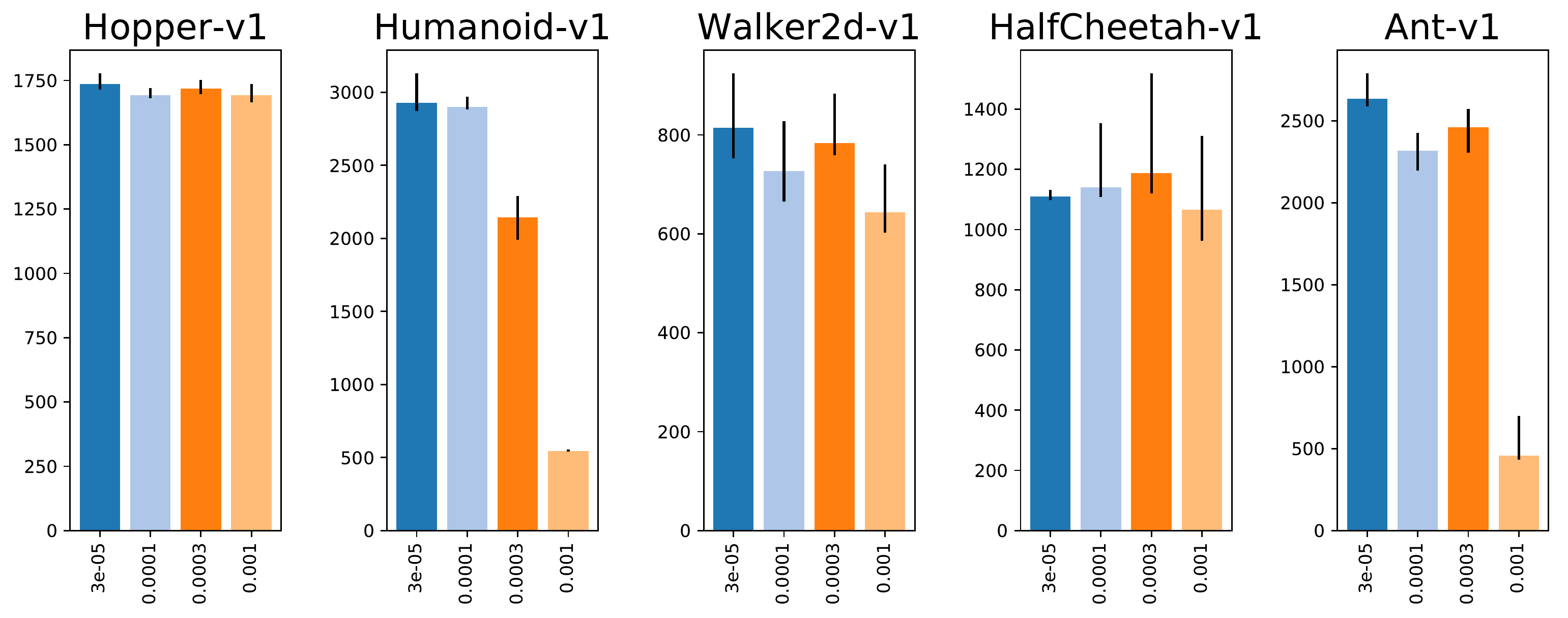}\hspace{1cm}\includegraphics[width=0.45\textwidth]{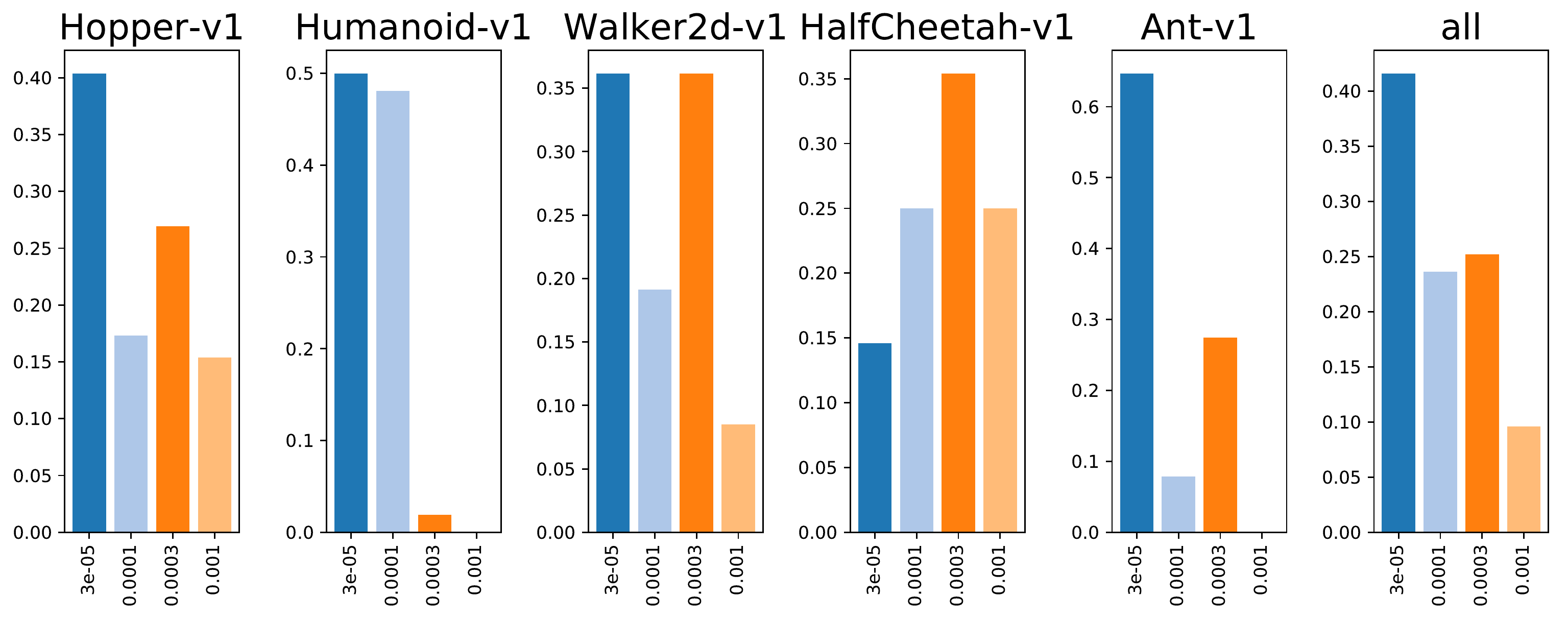}}
\caption{Analysis of choice \choicet{rmslr}: 95th percentile of performance scores conditioned on sub-choice (left) and distribution of sub-choices in top 5\% of configurations (right).}
\label{fig:final_optimize__gin_study_design_choice_value_sub_optimizer_rmsprop_learning_rate}
\end{center}
\end{figure}

We report aggregate statistics of the experiment in Table~\ref{tab:final_optimize_overview} as well as training curves in Figure~\ref{fig:final_optimize_training_curves}.
For each of the investigated choices in this experiment, we further provide a per-choice analysis in Figures~\ref{fig:final_optimize__gin_study_design_choice_value_learning_rate_decay}-\ref{fig:final_optimize__gin_study_design_choice_value_sub_optimizer_rmsprop_learning_rate}.
\begin{table}[ht]
\begin{center}
\caption{Performance quantiles across choice configurations.}
\label{tab:final_optimize_overview}
\begin{tabular}{lrrrrr}
\toprule
{} & Ant-v1 & HalfCheetah-v1 & Hopper-v1 & Humanoid-v1 & Walker2d-v1 \\
\midrule
90th percentile &   2180 &           1085 &      1675 &        2549 &         712 \\
95th percentile &   2388 &           1124 &      1728 &        2726 &         797 \\
99th percentile &   2699 &           1520 &      1826 &        2976 &        1079 \\
Max             &   2953 &           2532 &      1959 &        3332 &        1453 \\
\bottomrule
\end{tabular}

\end{center}
\end{table}

\begin{figure}[ht]
\begin{center}
\centerline{\includegraphics[width=1\textwidth]{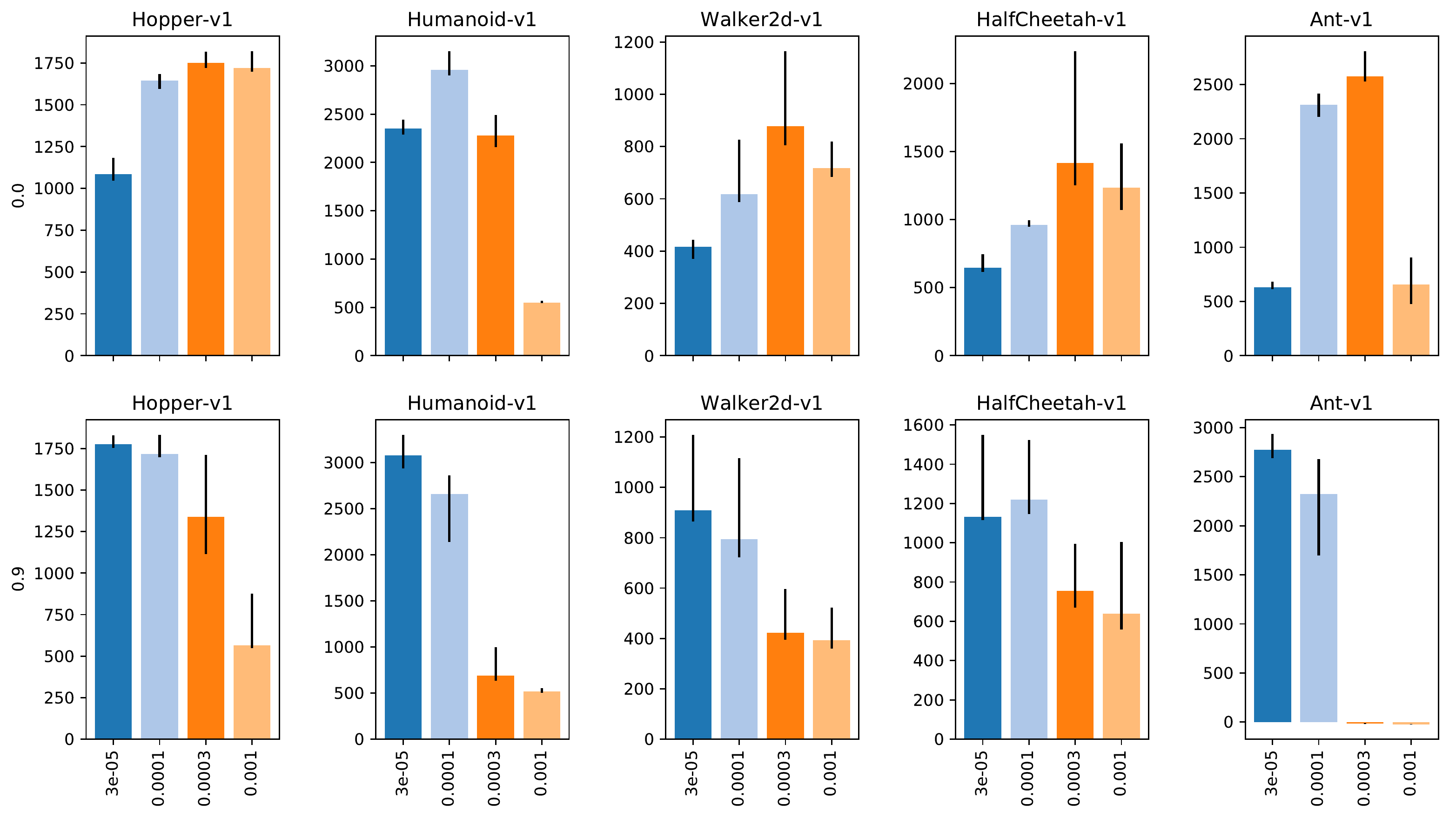}}
\caption{95th percentile of performance scores conditioned on \choicet{rmsmom}(rows) and \choicet{rmslr}(bars).}
\label{fig:final_optimize__correlation_rmsprop_momentum_vs_lr}
\end{center}
\end{figure}
\clearpage

%% file: final_regularizer/main.tex
\clearpage
\section{Experiment \texttt{Regularizers}}
\label{exp_final_regularizer}
\subsection{Design}
\label{exp_design_final_regularizer}
For each of the 5 environments, we sampled 4000 choice configurations where we sampled the following choices independently and uniformly from the following ranges:
\begin{itemize}
    \item \choicet{regularizationtype}: \{Constraint, No regularization, Penalty\}
    \begin{itemize}
        \item For the case ``\choicet{regularizationtype} = Constraint'', we further sampled the sub-choices:
        \begin{itemize}
            \item \choicet{regularizerconstraint}: \{$\kl(\mu||\pi)$, $\kl(\pi||\mu)$, $\kl(\texttt{ref}||\pi)$, decoupled $\kl(\mu||\pi)$, entropy\}
            \begin{itemize}
                \item For the case ``\choicet{regularizerconstraint} = $\kl(\mu||\pi)$'', we further sampled the sub-choices:
                \begin{itemize}[label={--}]
                    \item \choicet{regularizerconstraintklmupi}: \{0.005, 0.01, 0.02, 0.04, 0.08\}
                \end{itemize}
                \item For the case ``\choicet{regularizerconstraint} = $\kl(\pi||\mu)$'', we further sampled the sub-choices:
                \begin{itemize}[label={--}]
                    \item \choicet{regularizerconstraintklpimu}: \{0.005, 0.01, 0.02, 0.04, 0.08\}
                \end{itemize}
                \item For the case ``\choicet{regularizerconstraint} = $\kl(\texttt{ref}||\pi)$'', we further sampled the sub-choices:
                \begin{itemize}[label={--}]
                    \item \choicet{regularizerconstraintklrefpi}: \{10.0, 20.0, 40.0, 80.0, 160.0\}
                \end{itemize}
                \item For the case ``\choicet{regularizerconstraint} = decoupled $\kl(\mu||\pi)$'', we further sampled the sub-choices:
                \begin{itemize}[label={--}]
                    \item \choicet{regularizerconstraintklmupimean}: \{0.005, 0.01, 0.02, 0.04, 0.08\}
                    \item \choicet{regularizerconstraintklmupistd}: \{5e-05, 0.000125, 0.00025, 0.0005, 0.001, 0.002, 0.004\}
                \end{itemize}
                \item For the case ``\choicet{regularizerconstraint} = entropy'', we further sampled the sub-choices:
                \begin{itemize}[label={--}]
                    \item \choicet{regularizerconstraintentropy}: \{0.0, -5.0, -10.0, -15.0\}
                \end{itemize}
            \end{itemize}
        \end{itemize}
        \item For the case ``\choicet{regularizationtype} = Penalty'', we further sampled the sub-choices:
        \begin{itemize}
            \item \choicet{regularizerpenalty}: \{$\kl(\mu||\pi)$, $\kl(\pi||\mu)$, $\kl(\texttt{ref}||\pi)$, decoupled $\kl(\mu||\pi)$, entropy\}
            \begin{itemize}
                \item For the case ``\choicet{regularizerpenalty} = $\kl(\mu||\pi)$'', we further sampled the sub-choices:
                \begin{itemize}[label={--}]
                    \item \choicet{regularizerpenaltyklmupi}: \{0.003, 0.01, 0.03, 0.1, 0.3, 1.0\}
                \end{itemize}
                \item For the case ``\choicet{regularizerpenalty} = $\kl(\pi||\mu)$'', we further sampled the sub-choices:
                \begin{itemize}[label={--}]
                    \item \choicet{regularizerpenaltyklpimu}: \{0.003, 0.01, 0.03, 0.1, 0.3, 1.0\}
                \end{itemize}
                \item For the case ``\choicet{regularizerpenalty} = $\kl(\texttt{ref}||\pi)$'', we further sampled the sub-choices:
                \begin{itemize}[label={--}]
                    \item \choicet{regularizerpenaltyklrefpi}: \{3e-06, 1e-05, 3e-05, 0.0001, 0.0003, 0.001\}
                \end{itemize}
                \item For the case ``\choicet{regularizerpenalty} = decoupled $\kl(\mu||\pi)$'', we further sampled the sub-choices:
                \begin{itemize}[label={--}]
                    \item \choicet{regularizerpenaltyklmupimean}: \{0.003, 0.01, 0.03, 0.1, 0.3, 1.0\}
                    \item \choicet{regularizerpenaltyklmupistd}: \{0.1, 0.3, 1.0, 3.0, 10.0, 30.0, 100.0, 300.0\}
                \end{itemize}
                \item For the case ``\choicet{regularizerpenalty} = entropy'', we further sampled the sub-choices:
                \begin{itemize}[label={--}]
                    \item \choicet{regularizerpenaltyentropy}: \{1e-05, 3e-05, 0.0001, 0.0003, 0.001, 0.003\}
                \end{itemize}
            \end{itemize}
        \end{itemize}
    \end{itemize}
    \item \choicet{adamlr}: \{3e-05, 0.0001, 0.0003, 0.001, 0.003\}
\end{itemize}
All the other choices were set to the default values as described in Appendix~\ref{sec:default_settings}.

For each of the sampled choice configurations, we train 3 agents with different random seeds and compute the performance metric as described in Section~\ref{sec:performance}.
\subsection{Results}
\label{exp_results_final_regularizer}
\begin{figure}[ht]
\begin{center}
\centerline{\includegraphics[width=1\textwidth]{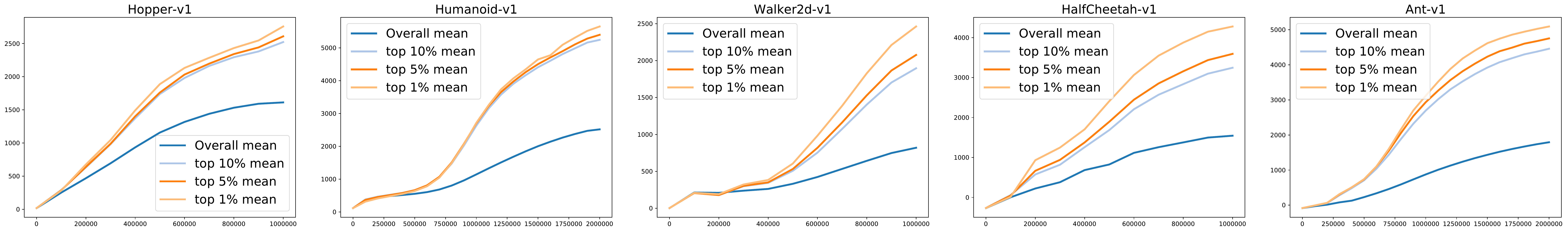}}
\caption{Training curves.}
\label{fig:final_regularizer_training_curves}
\end{center}
\end{figure}

\begin{figure}[ht]
\begin{center}
\centerline{\includegraphics[width=1\textwidth]{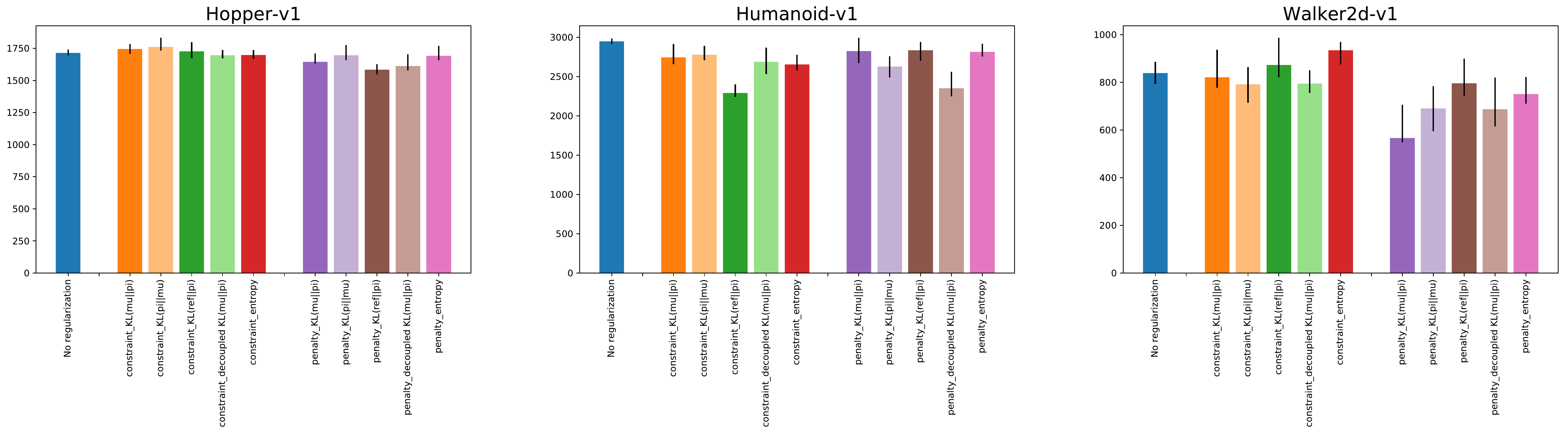}}
\centerline{\includegraphics[width=0.65\textwidth]{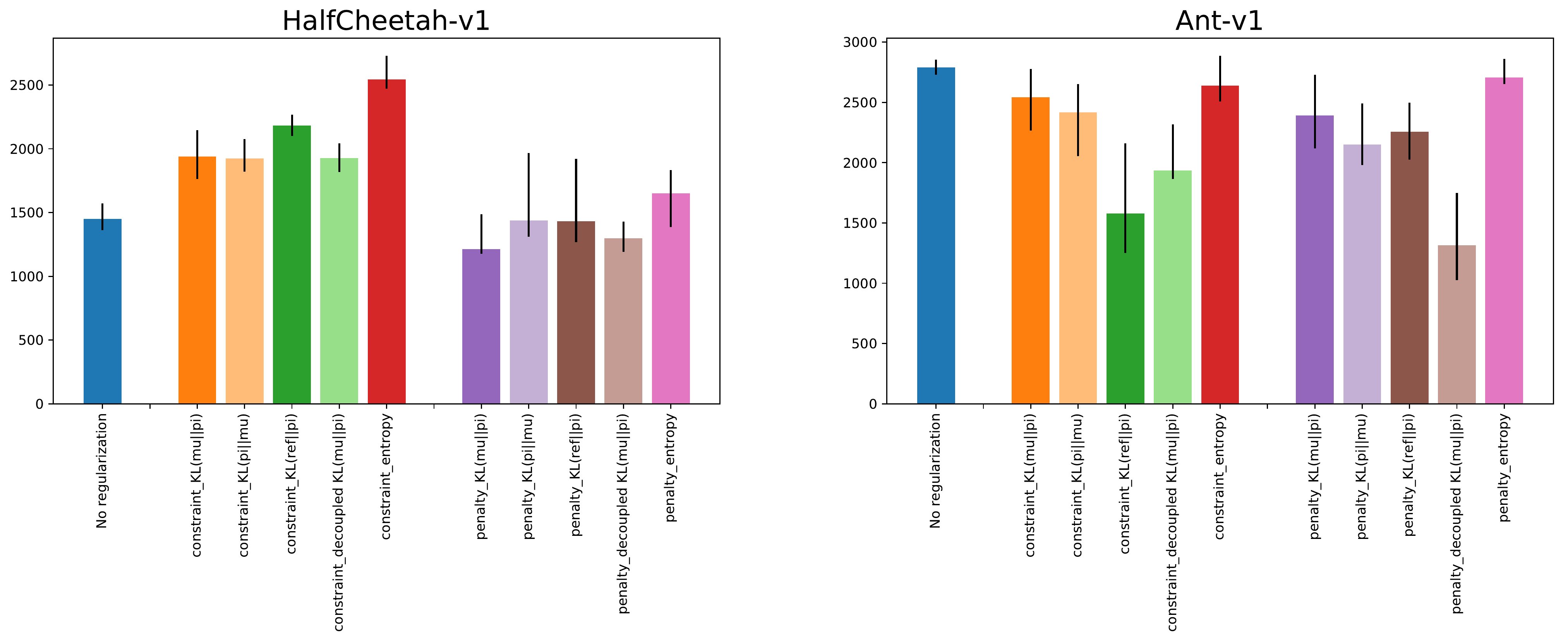}}
\caption{Comparison of 95th percentile of the performance of different regularization approaches conditioned on their type.}
\label{fig:final_regularizer_custom_regularization}
\end{center}
\end{figure}

\begin{figure}[ht]
\begin{center}
\centerline{\includegraphics[width=0.45\textwidth]{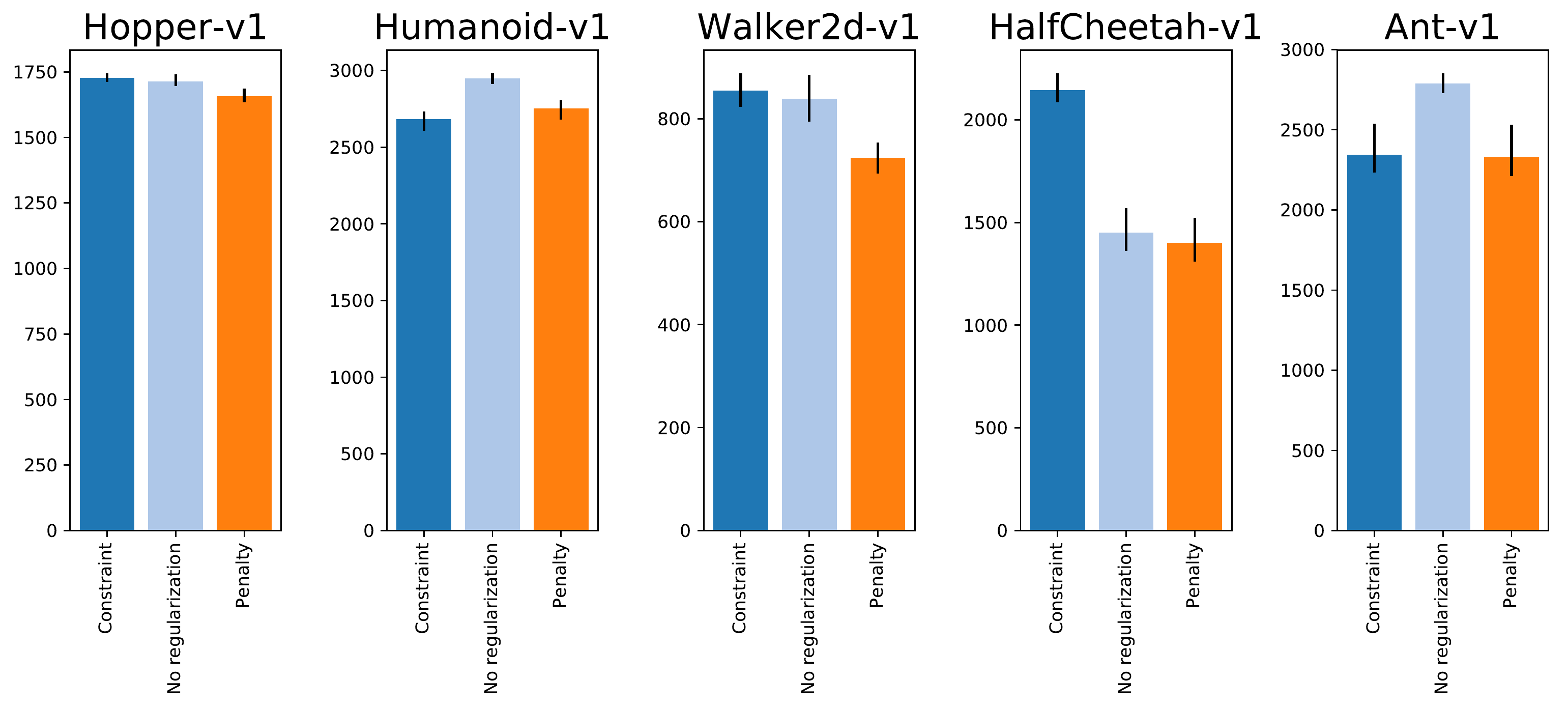}\hspace{1cm}\includegraphics[width=0.45\textwidth]{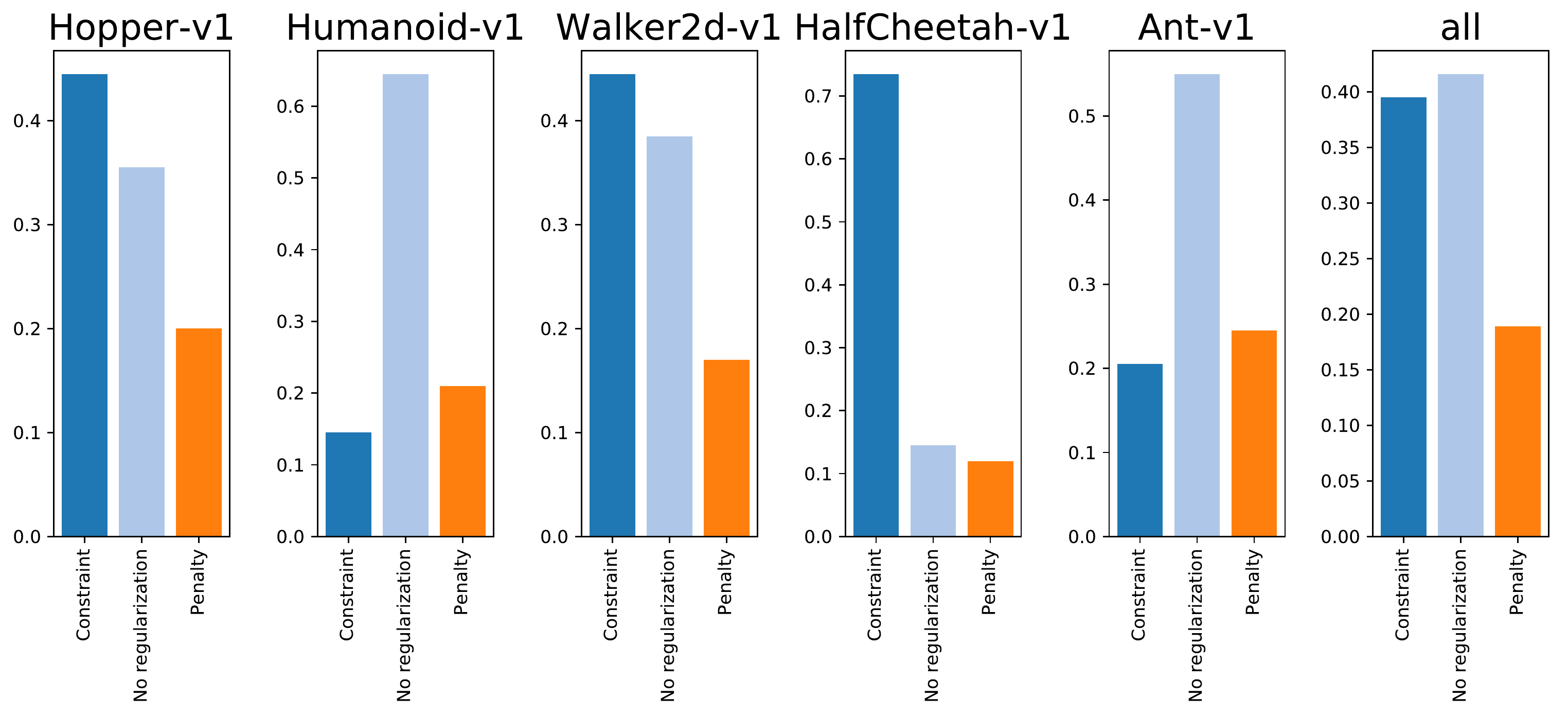}}
\caption{Analysis of choice \choicet{regularizationtype}: 95th percentile of performance scores conditioned on choice (left) and distribution of choices in top 5\% of configurations (right).}
\label{fig:final_regularizer__gin_study_design_choice_value_policy_regularization}
\end{center}
\end{figure}

\begin{figure}[ht]
\begin{center}
\centerline{\includegraphics[width=0.45\textwidth]{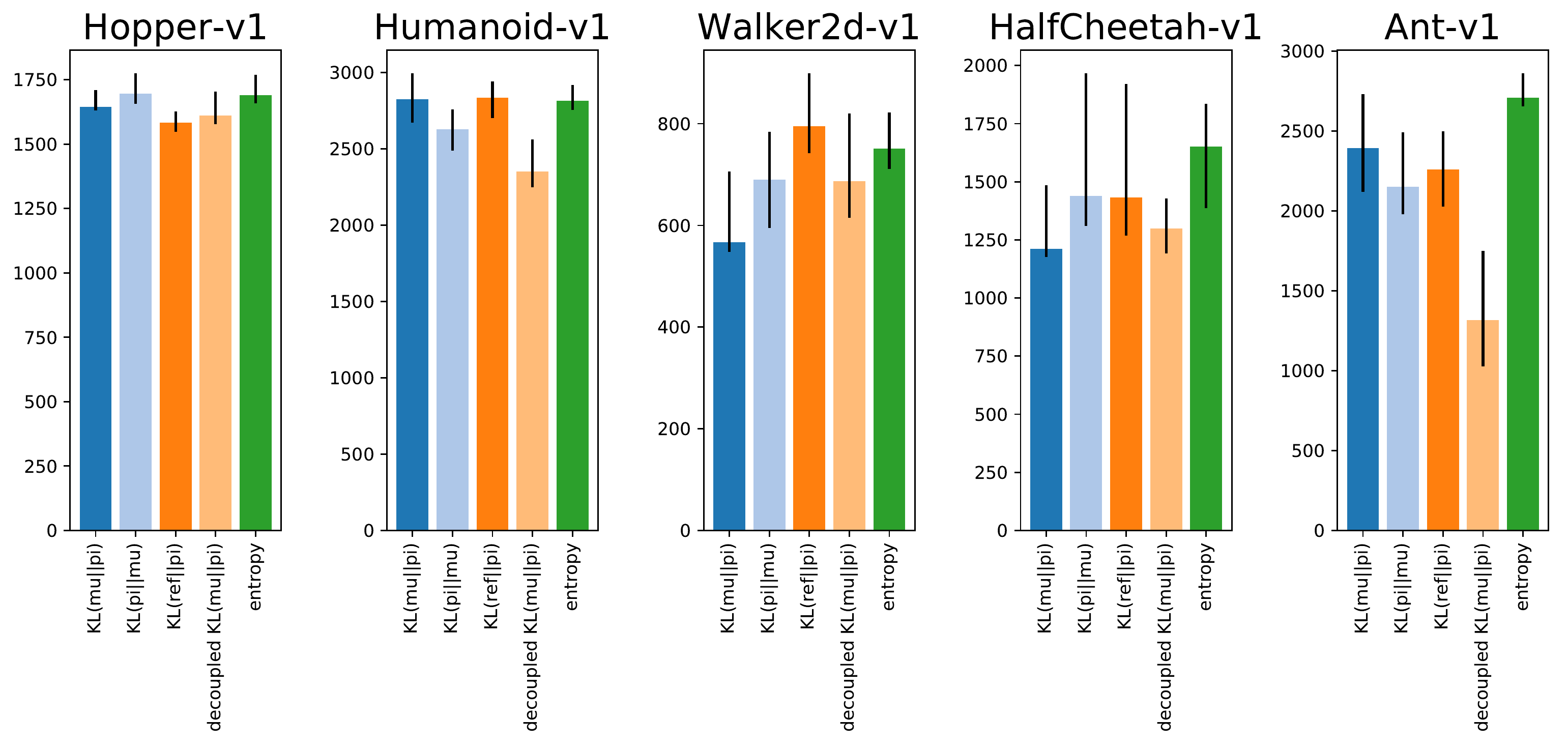}\hspace{1cm}\includegraphics[width=0.45\textwidth]{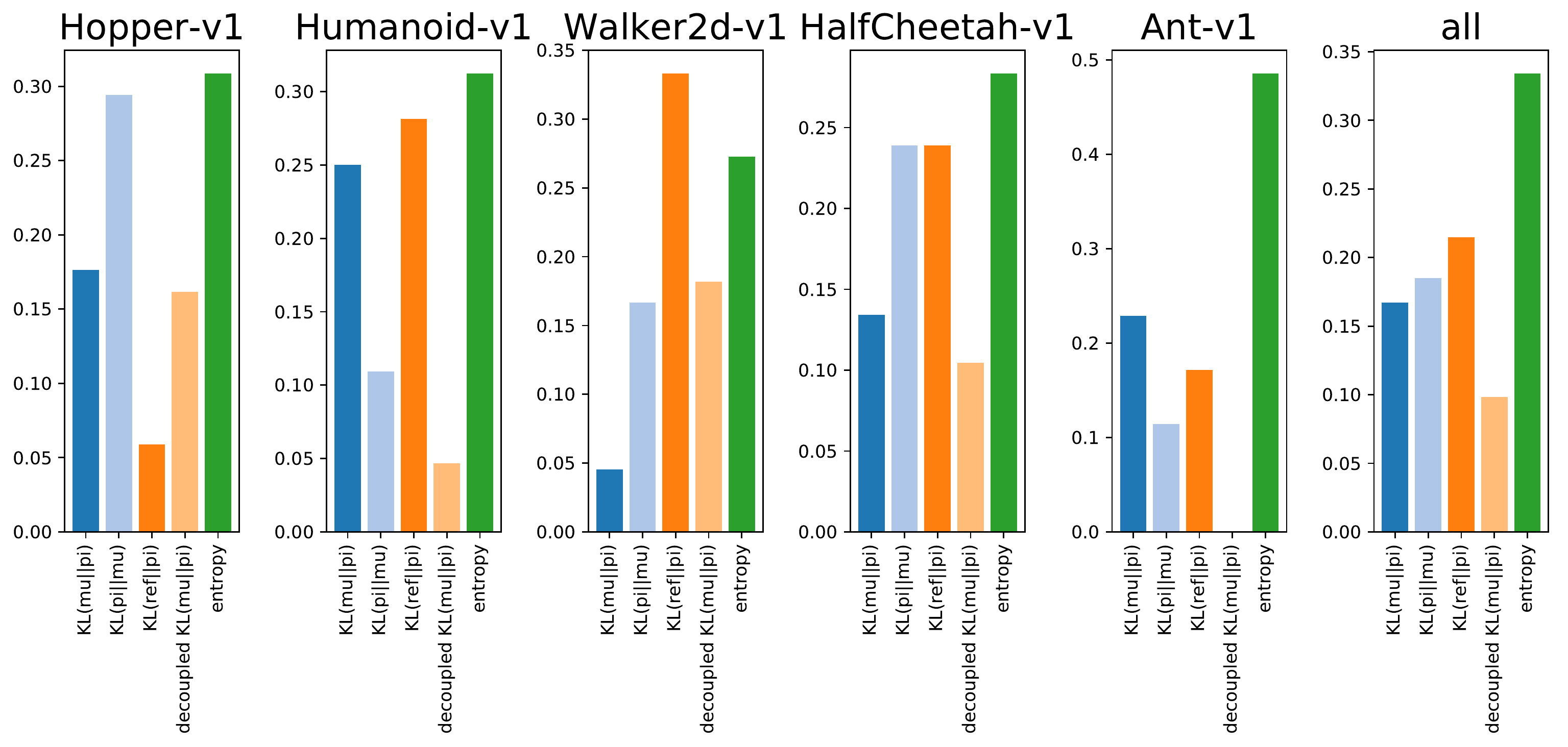}}
\caption{Analysis of choice \choicet{regularizerpenalty}: 95th percentile of performance scores conditioned on sub-choice (left) and distribution of sub-choices in top 5\% of configurations (right).}
\label{fig:final_regularizer__gin_study_design_choice_value_sub_policy_regularization_penalty_regularization_penalty}
\end{center}
\end{figure}

\begin{figure}[ht]
\begin{center}
\centerline{\includegraphics[width=0.45\textwidth]{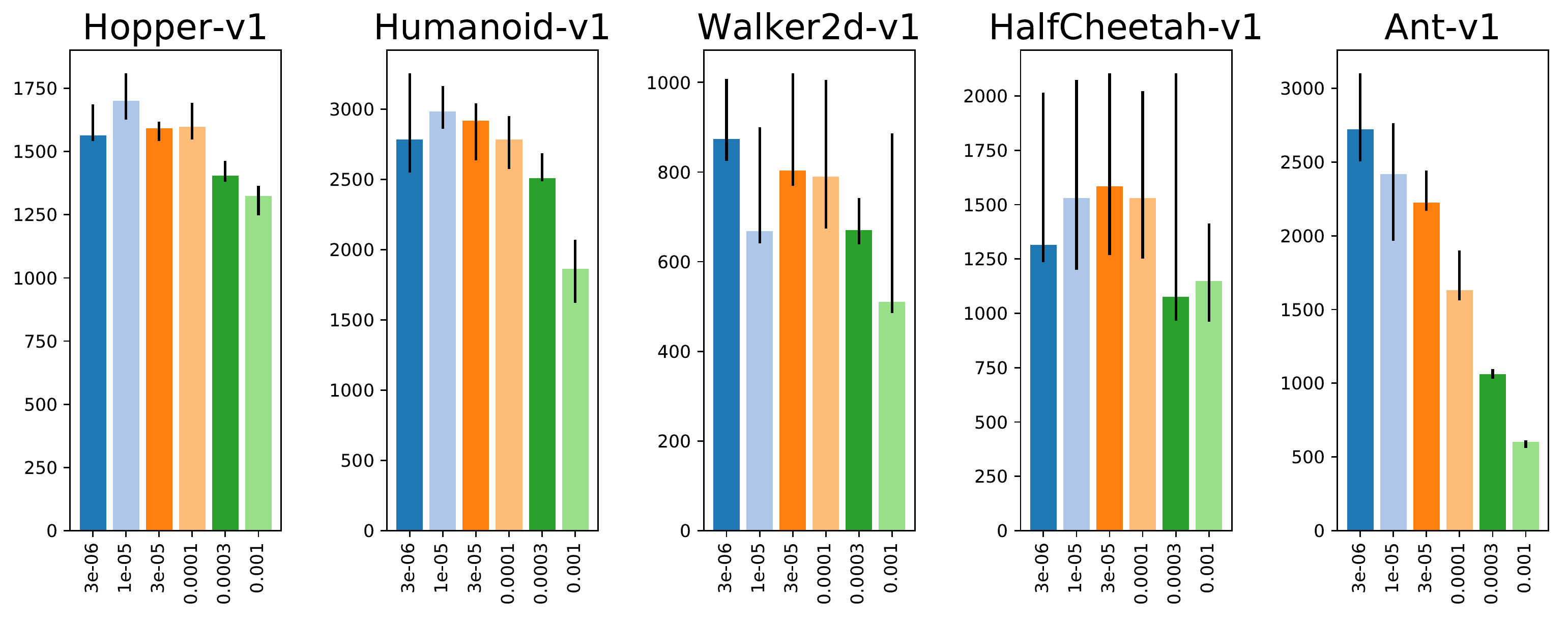}\hspace{1cm}\includegraphics[width=0.45\textwidth]{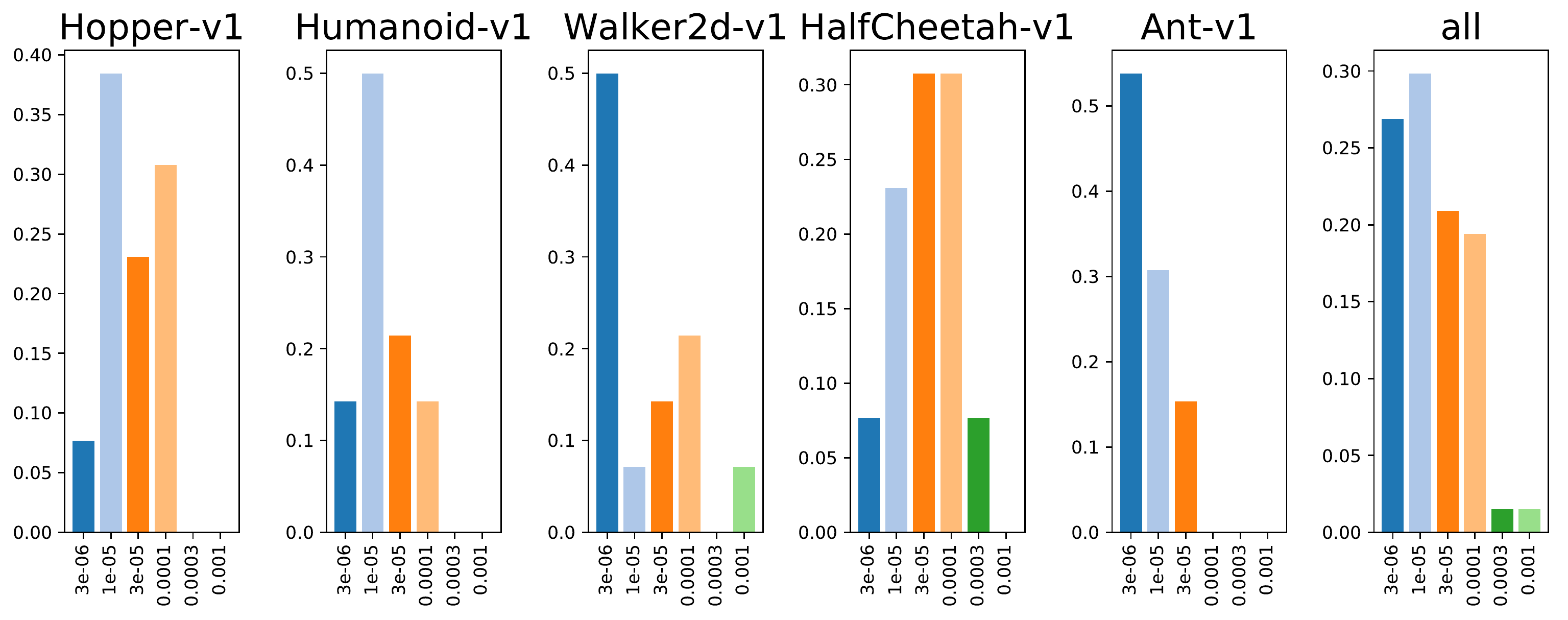}}
\caption{Analysis of choice \choicet{regularizerpenaltyklrefpi}: 95th percentile of performance scores conditioned on sub-choice (left) and distribution of sub-choices in top 5\% of configurations (right).}
\label{fig:final_regularizer__gin_study_design_choice_value_sub_regularization_penalty_klrefpi_coefficient}
\end{center}
\end{figure}

\begin{figure}[ht]
\begin{center}
\centerline{\includegraphics[width=0.45\textwidth]{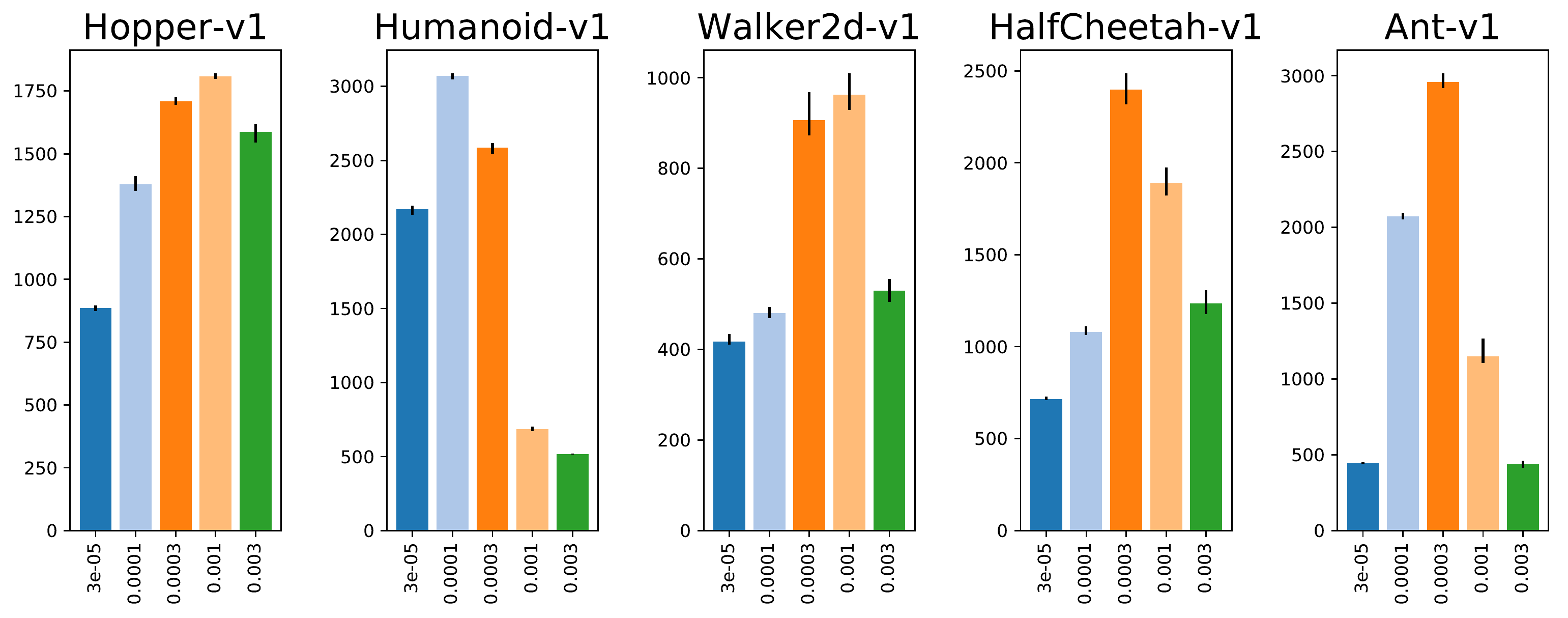}\hspace{1cm}\includegraphics[width=0.45\textwidth]{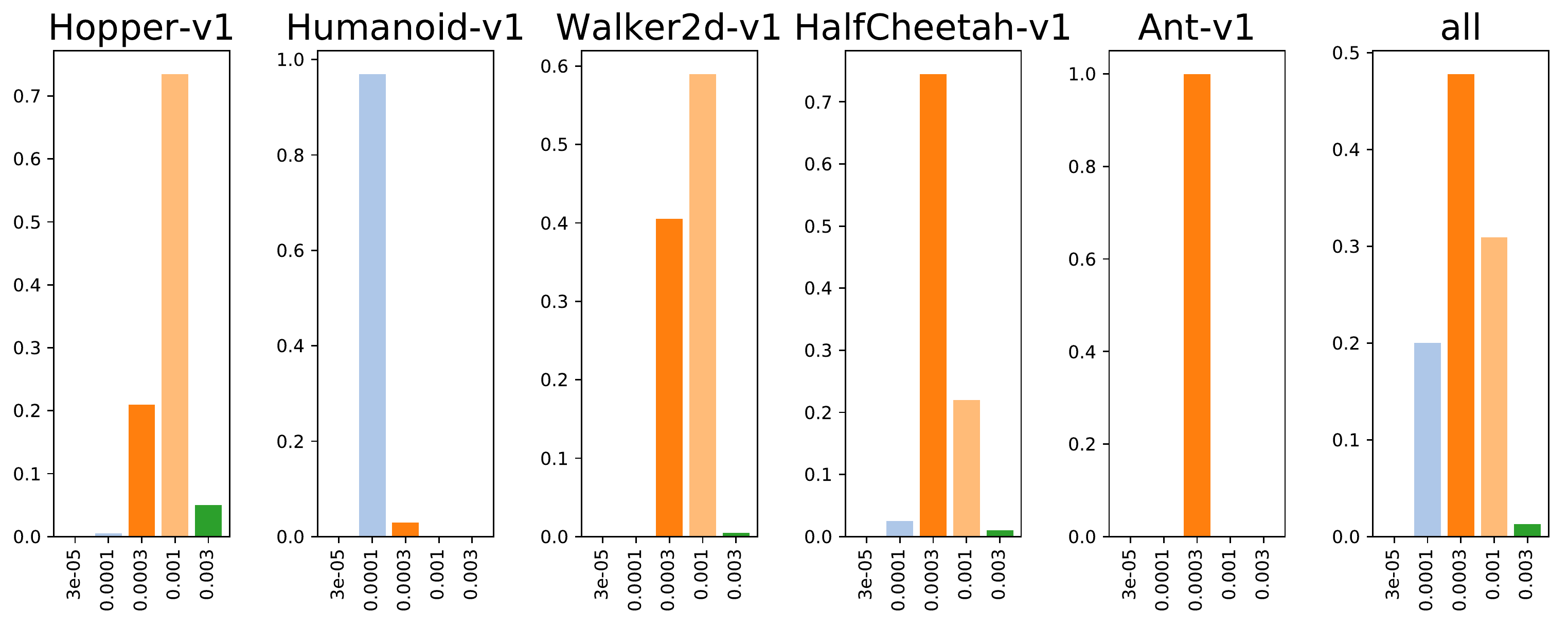}}
\caption{Analysis of choice \choicet{adamlr}: 95th percentile of performance scores conditioned on choice (left) and distribution of choices in top 5\% of configurations (right).}
\label{fig:final_regularizer__gin_study_design_choice_value_learning_rate}
\end{center}
\end{figure}

\begin{figure}[ht]
\begin{center}
\centerline{\includegraphics[width=0.45\textwidth]{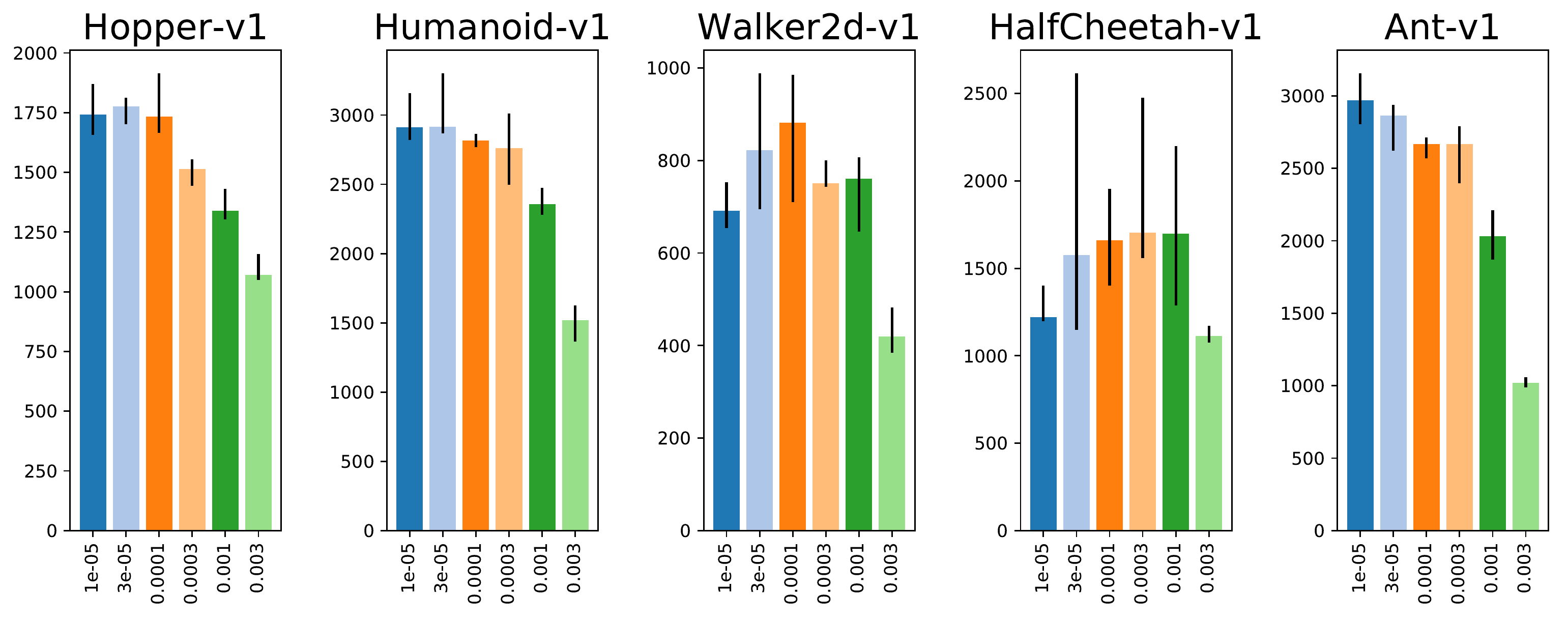}\hspace{1cm}\includegraphics[width=0.45\textwidth]{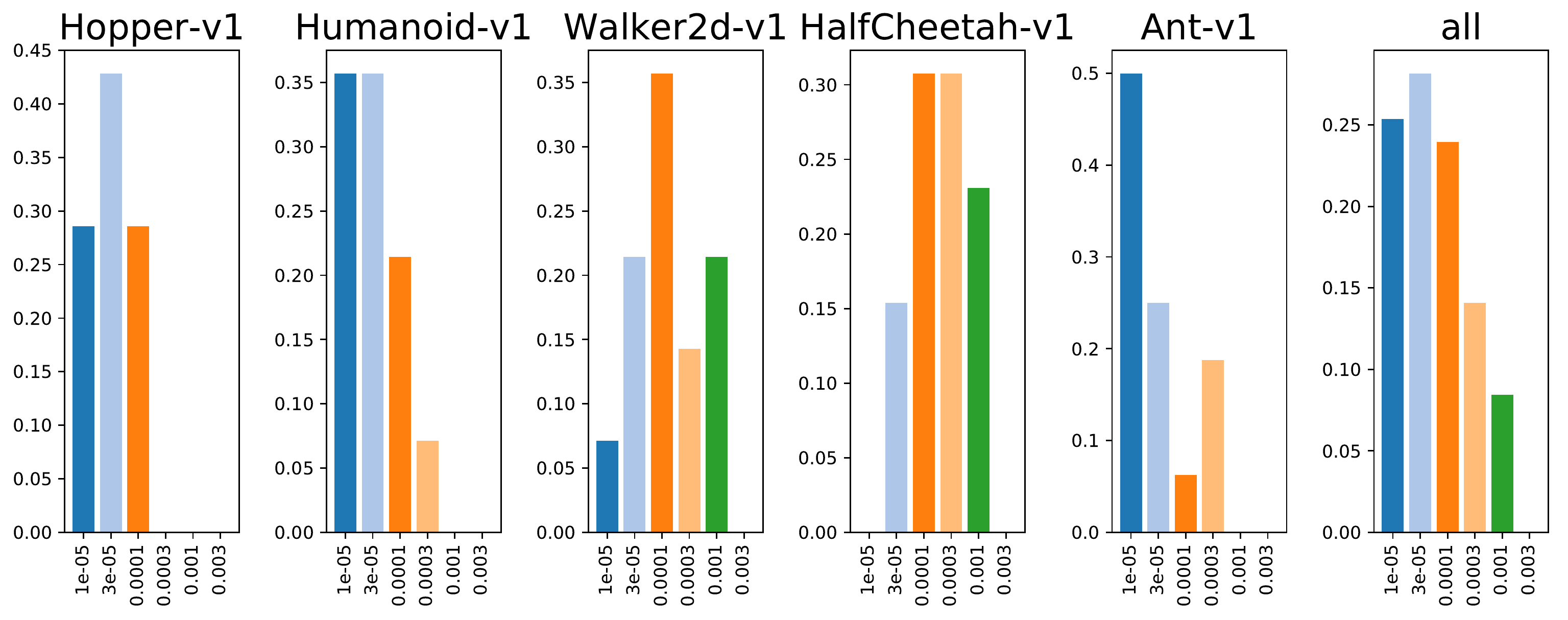}}
\caption{Analysis of choice \choicet{regularizerpenaltyentropy}: 95th percentile of performance scores conditioned on sub-choice (left) and distribution of sub-choices in top 5\% of configurations (right).}
\label{fig:final_regularizer__gin_study_design_choice_value_sub_regularization_penalty_entropy_coefficient}
\end{center}
\end{figure}

\begin{figure}[ht]
\begin{center}
\centerline{\includegraphics[width=0.45\textwidth]{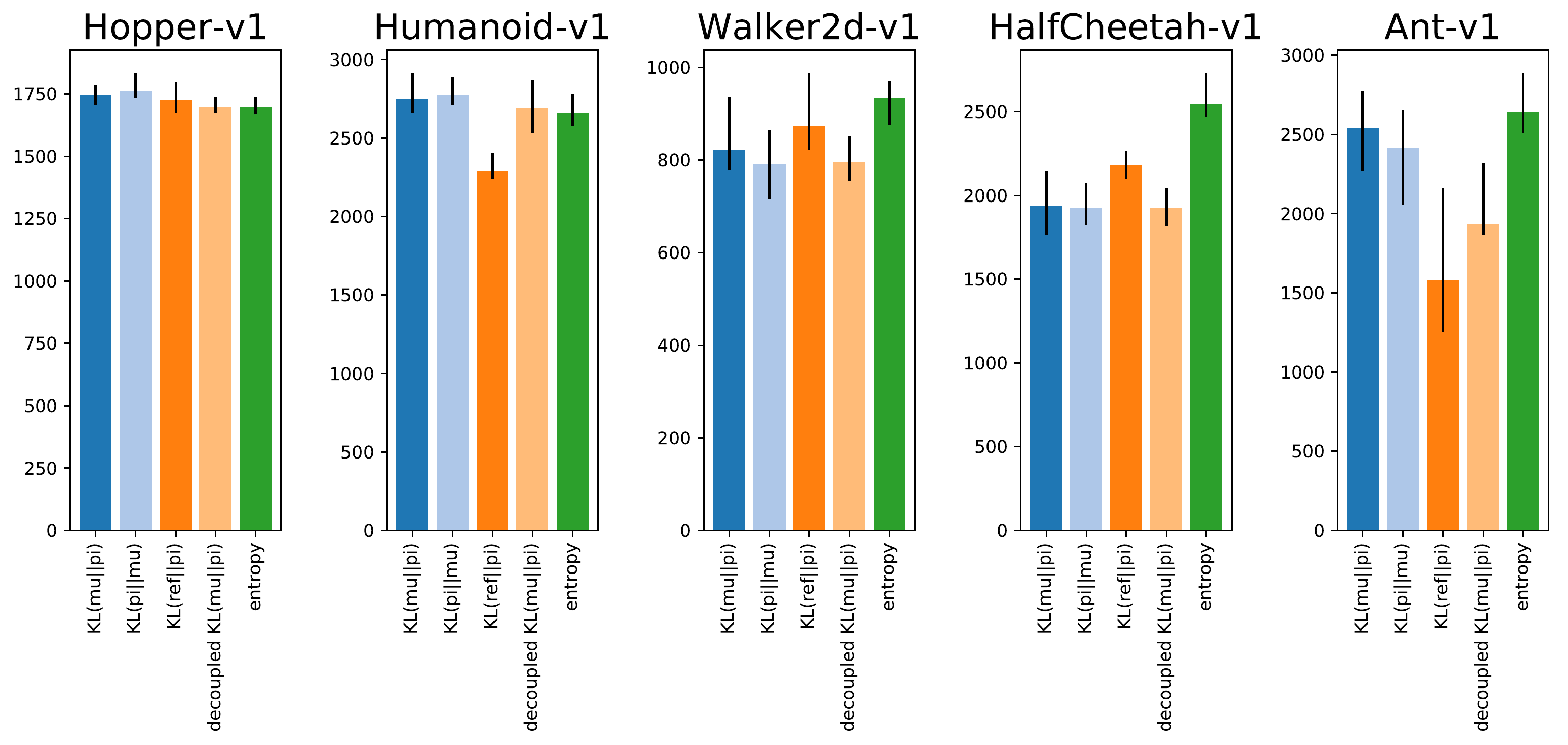}\hspace{1cm}\includegraphics[width=0.45\textwidth]{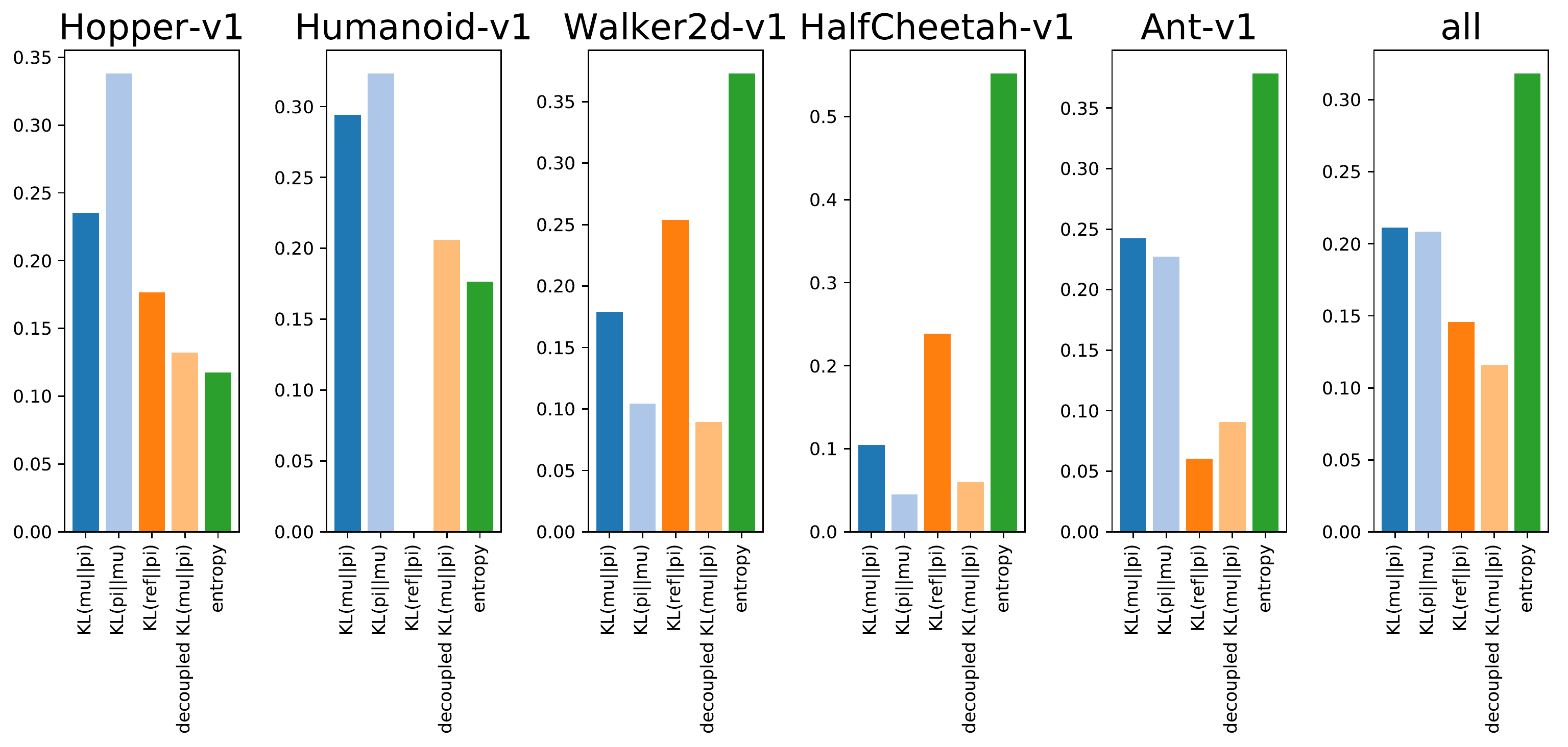}}
\caption{Analysis of choice \choicet{regularizerconstraint}: 95th percentile of performance scores conditioned on sub-choice (left) and distribution of sub-choices in top 5\% of configurations (right).}
\label{fig:final_regularizer__gin_study_design_choice_value_sub_policy_regularization_constraint_regularization_constraint}
\end{center}
\end{figure}

\begin{figure}[ht]
\begin{center}
\centerline{\includegraphics[width=0.45\textwidth]{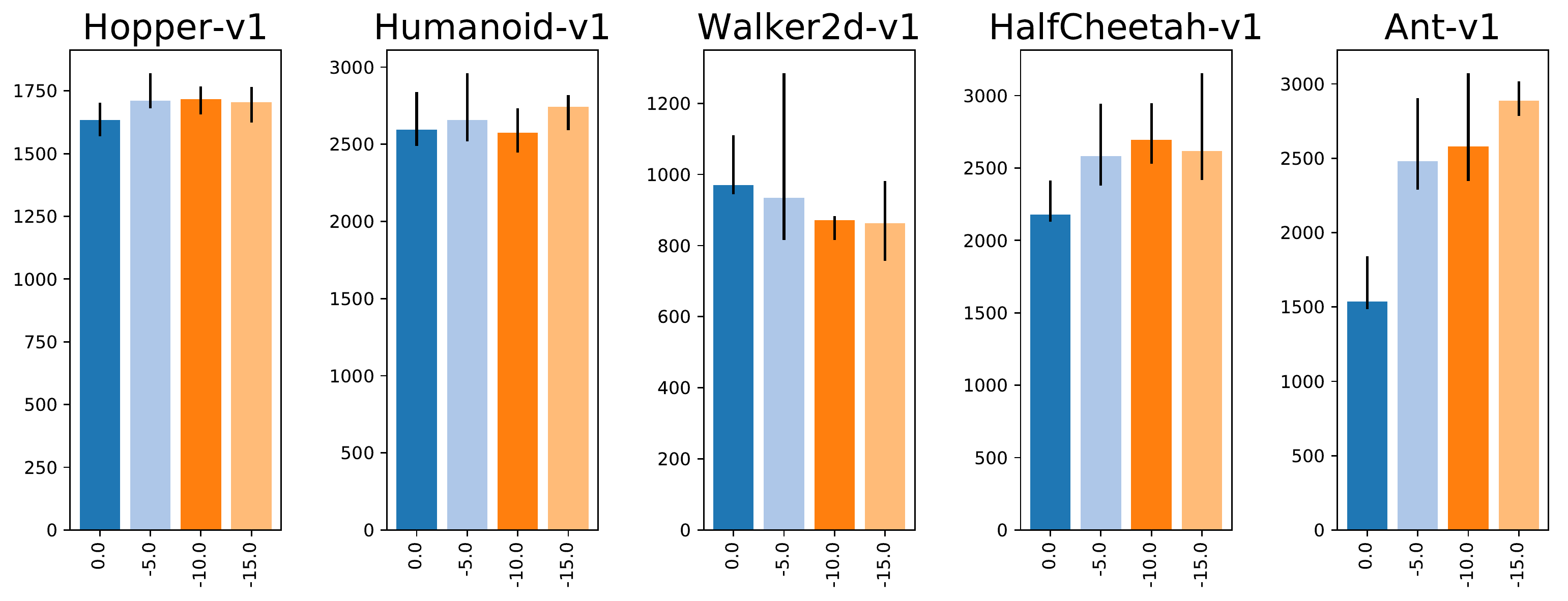}\hspace{1cm}\includegraphics[width=0.45\textwidth]{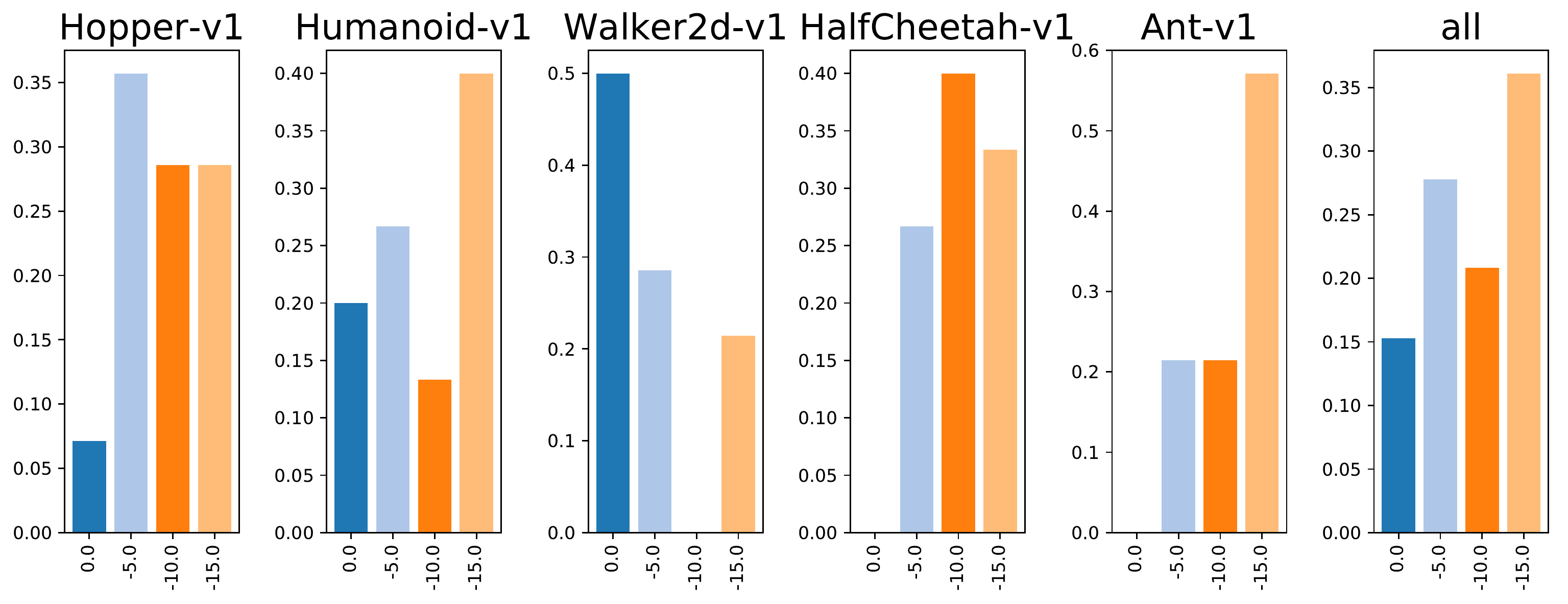}}
\caption{Analysis of choice \choicet{regularizerconstraintentropy}: 95th percentile of performance scores conditioned on sub-choice (left) and distribution of sub-choices in top 5\% of configurations (right).}
\label{fig:final_regularizer__gin_study_design_choice_value_sub_regularization_constraint_entropy_entropy_threshold}
\end{center}
\end{figure}

\begin{figure}[ht]
\begin{center}
\centerline{\includegraphics[width=0.45\textwidth]{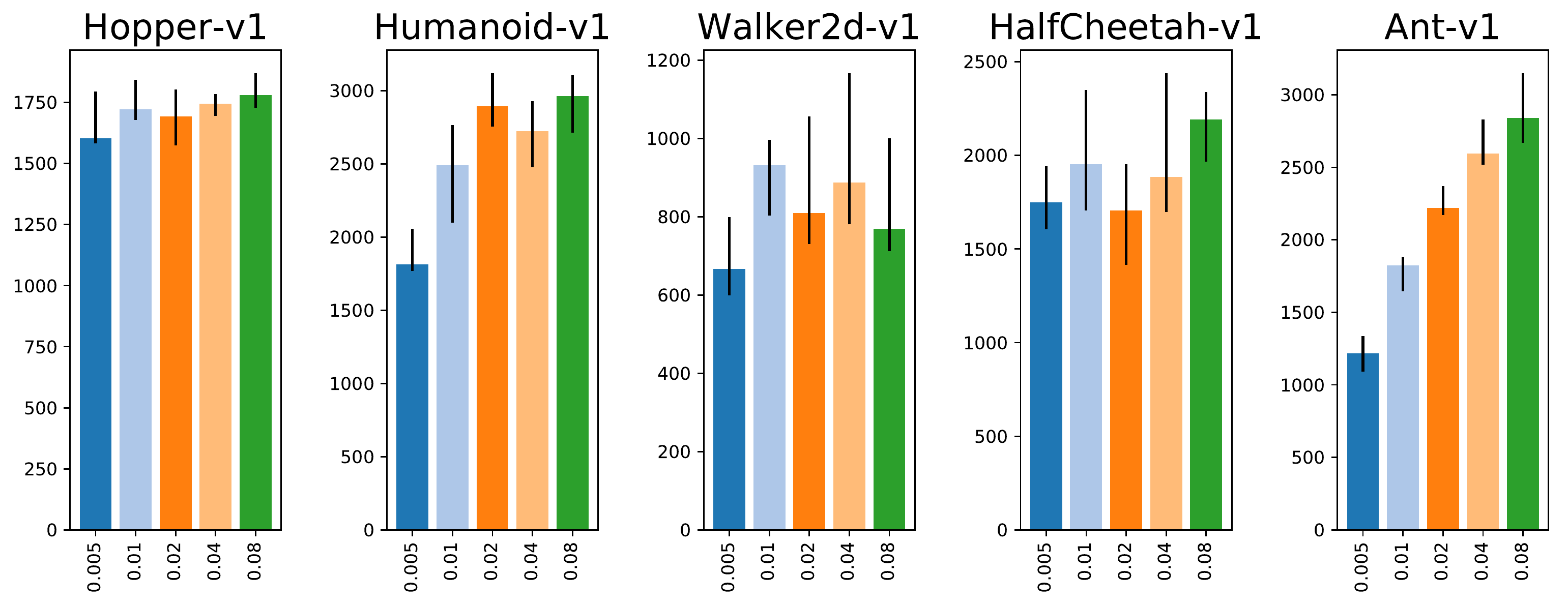}\hspace{1cm}\includegraphics[width=0.45\textwidth]{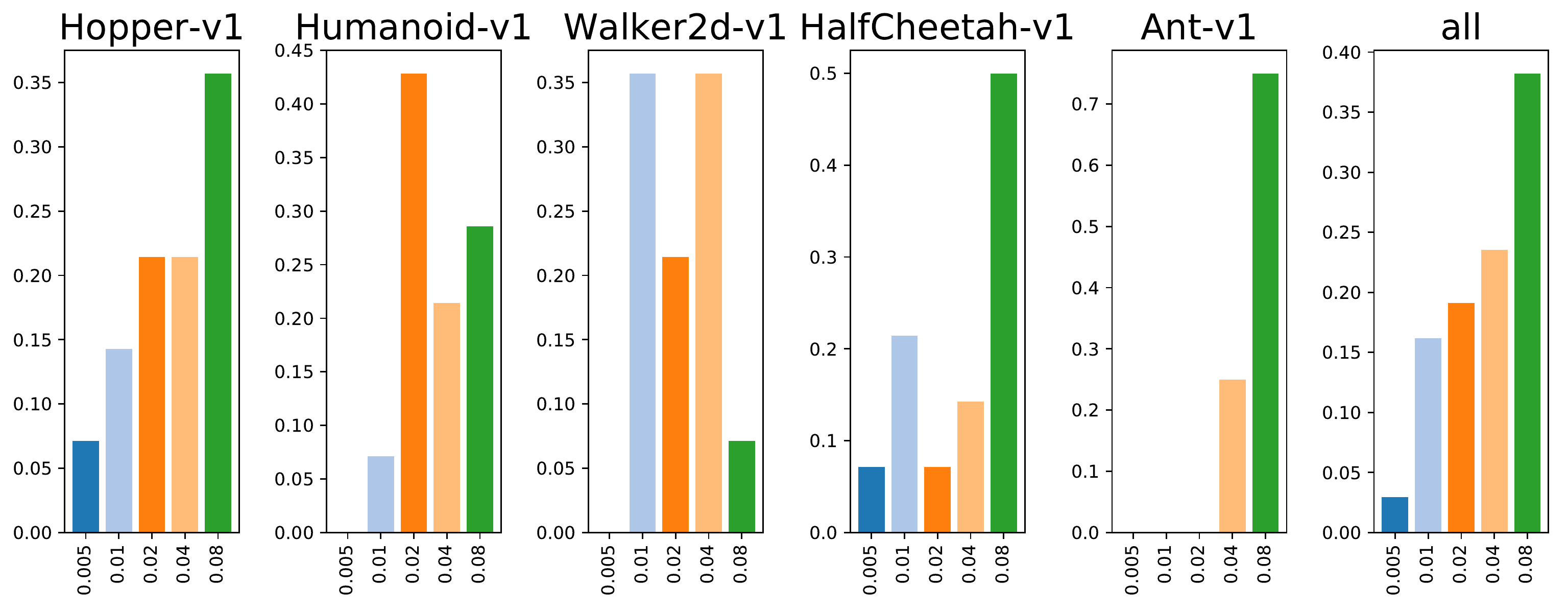}}
\caption{Analysis of choice \choicet{regularizerconstraintklmupi}: 95th percentile of performance scores conditioned on sub-choice (left) and distribution of sub-choices in top 5\% of configurations (right).}
\label{fig:final_regularizer__gin_study_design_choice_value_sub_regularization_constraint_klmupi_kl_mu_pi_threshold}
\end{center}
\end{figure}

\begin{figure}[ht]
\begin{center}
\centerline{\includegraphics[width=0.45\textwidth]{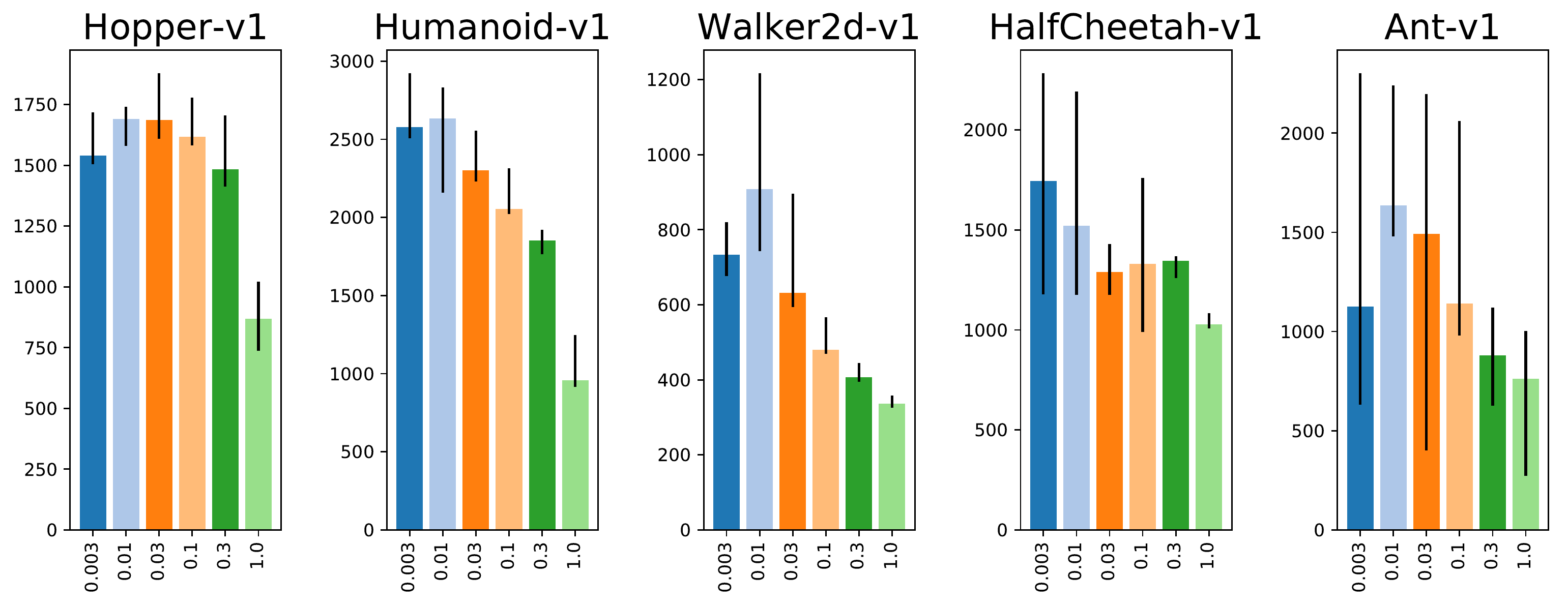}\hspace{1cm}\includegraphics[width=0.45\textwidth]{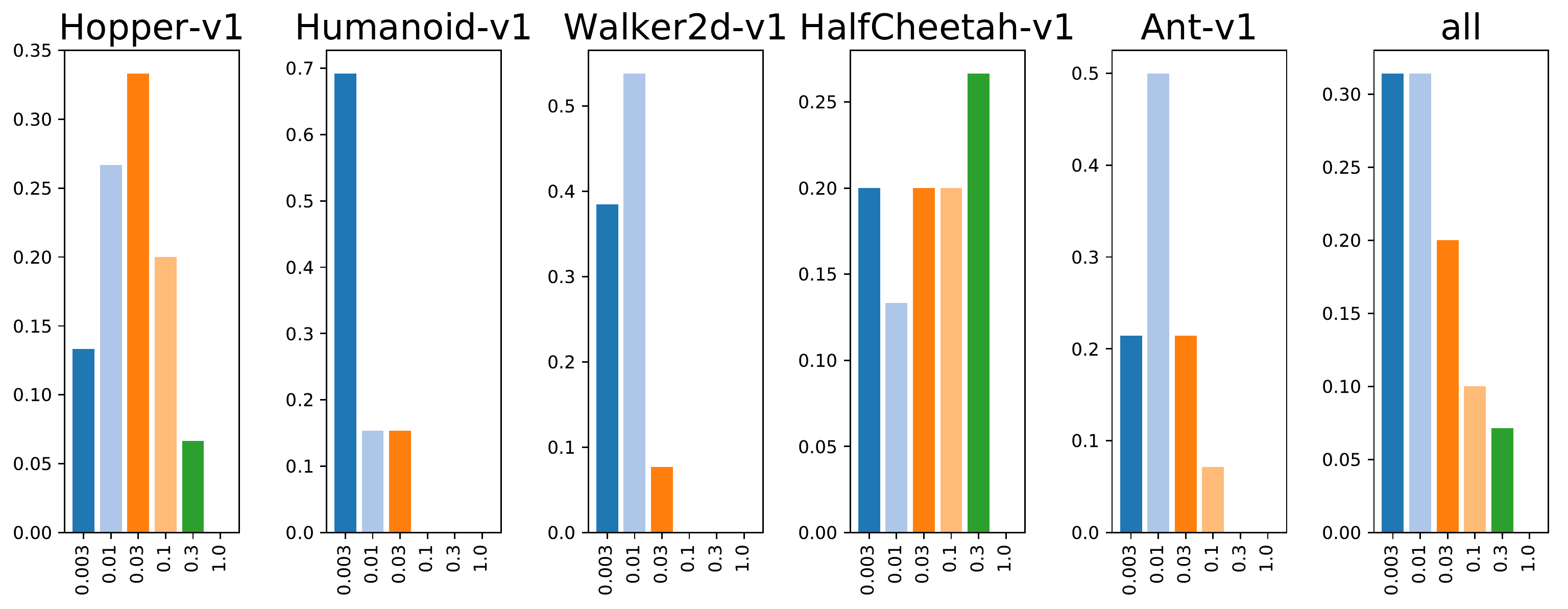}}
\caption{Analysis of choice \choicet{regularizerpenaltyklmupimean}: 95th percentile of performance scores conditioned on sub-choice (left) and distribution of sub-choices in top 5\% of configurations (right).}
\label{fig:final_regularizer__gin_study_design_choice_value_sub_regularization_penalty_decoupled_klmupi_mean_coefficient}
\end{center}
\end{figure}

\begin{figure}[ht]
\begin{center}
\centerline{\includegraphics[width=0.45\textwidth]{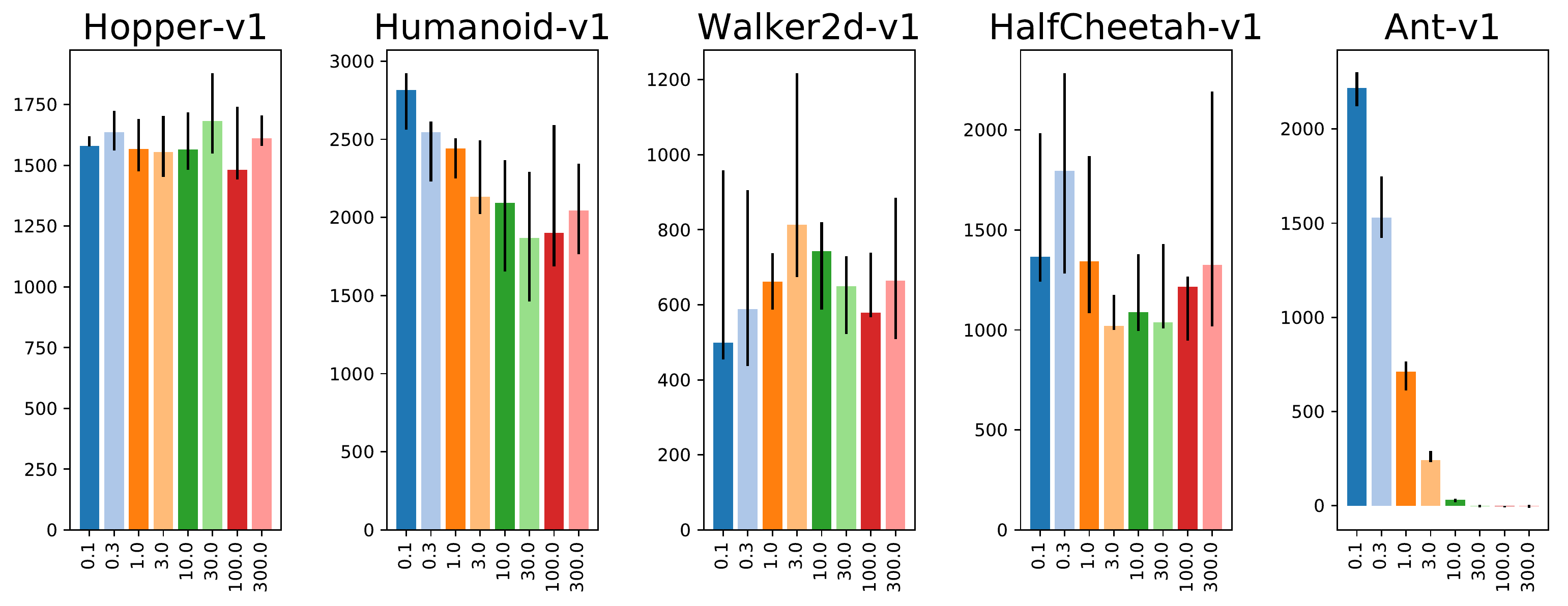}\hspace{1cm}\includegraphics[width=0.45\textwidth]{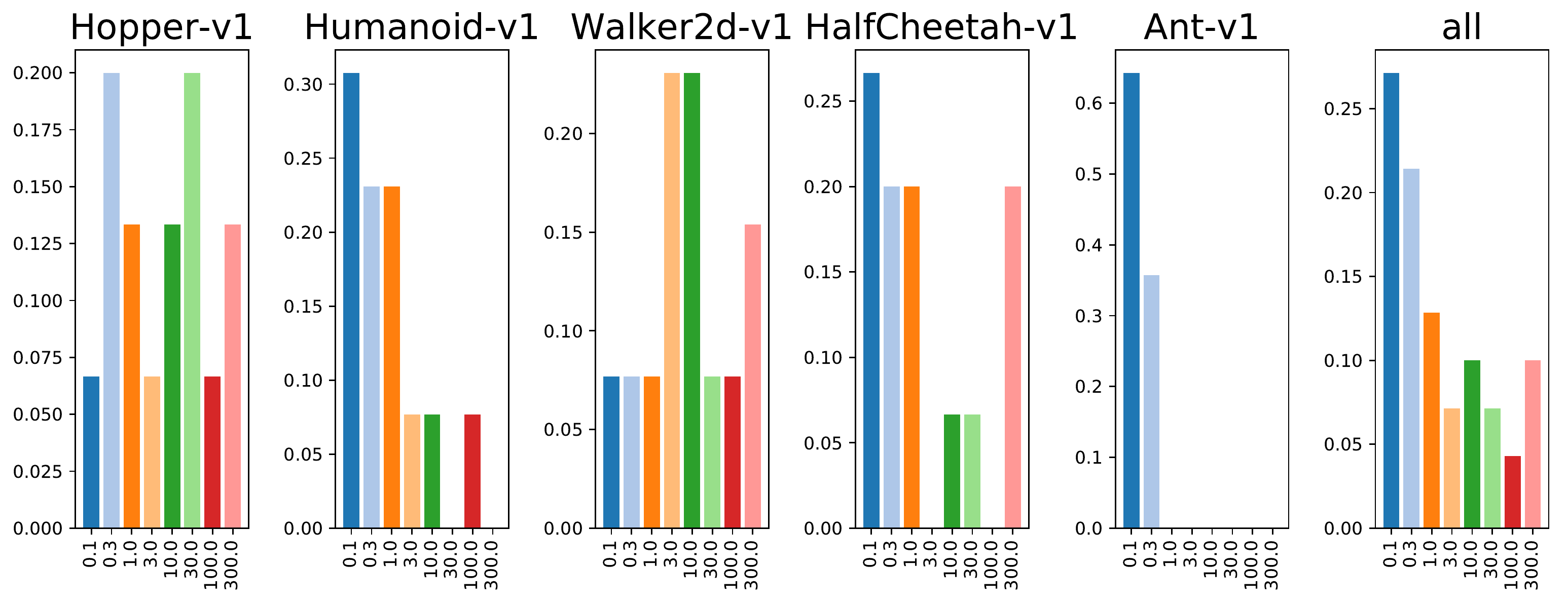}}
\caption{Analysis of choice \choicet{regularizerpenaltyklmupistd}: 95th percentile of performance scores conditioned on sub-choice (left) and distribution of sub-choices in top 5\% of configurations (right).}
\label{fig:final_regularizer__gin_study_design_choice_value_sub_regularization_penalty_decoupled_klmupi_std_coefficient}
\end{center}
\end{figure}

\begin{figure}[ht]
\begin{center}
\centerline{\includegraphics[width=0.45\textwidth]{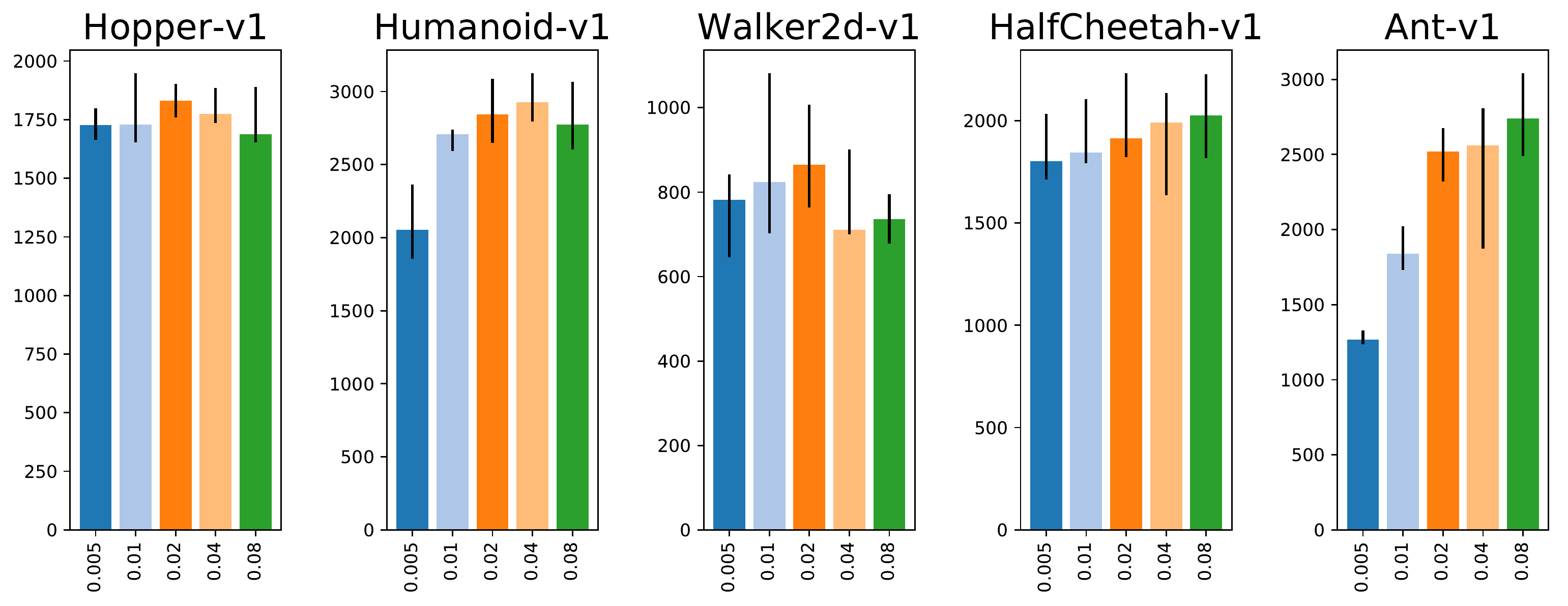}\hspace{1cm}\includegraphics[width=0.45\textwidth]{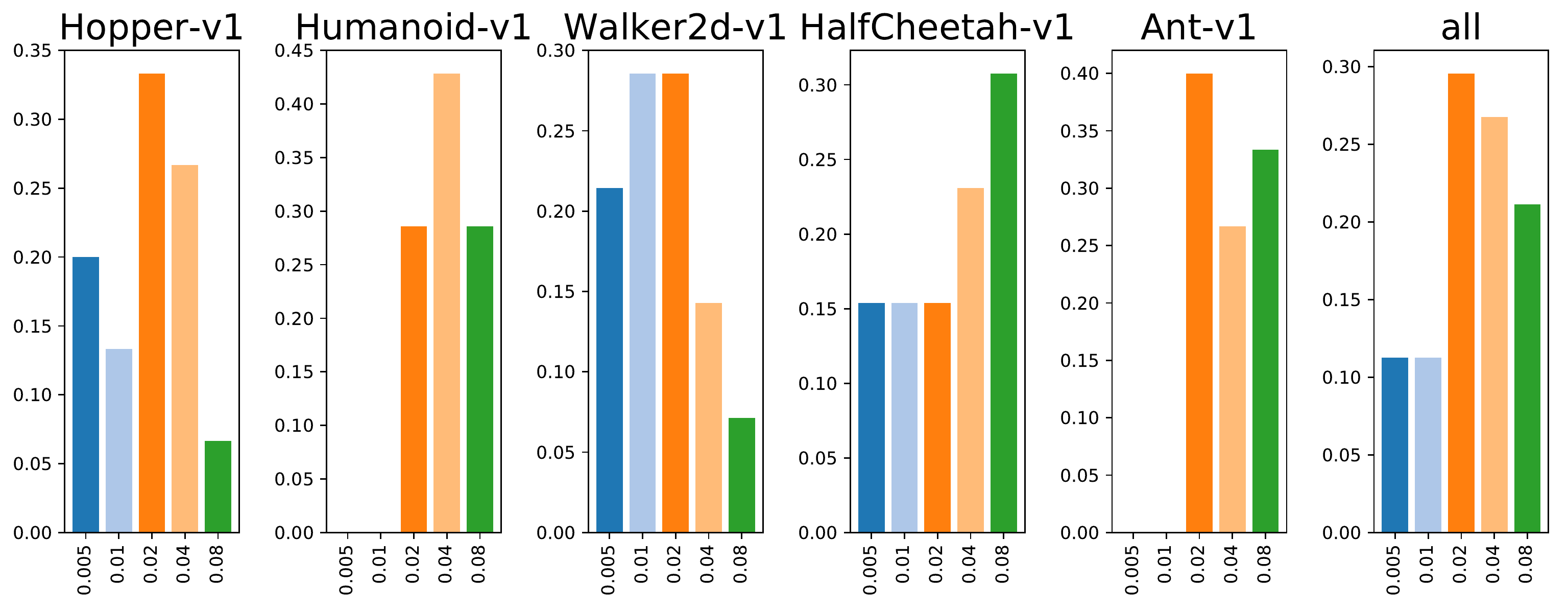}}
\caption{Analysis of choice \choicet{regularizerconstraintklpimu}: 95th percentile of performance scores conditioned on sub-choice (left) and distribution of sub-choices in top 5\% of configurations (right).}
\label{fig:final_regularizer__gin_study_design_choice_value_sub_regularization_constraint_klpimu_kl_pi_mu_threshold}
\end{center}
\end{figure}

\begin{figure}[ht]
\begin{center}
\centerline{\includegraphics[width=0.45\textwidth]{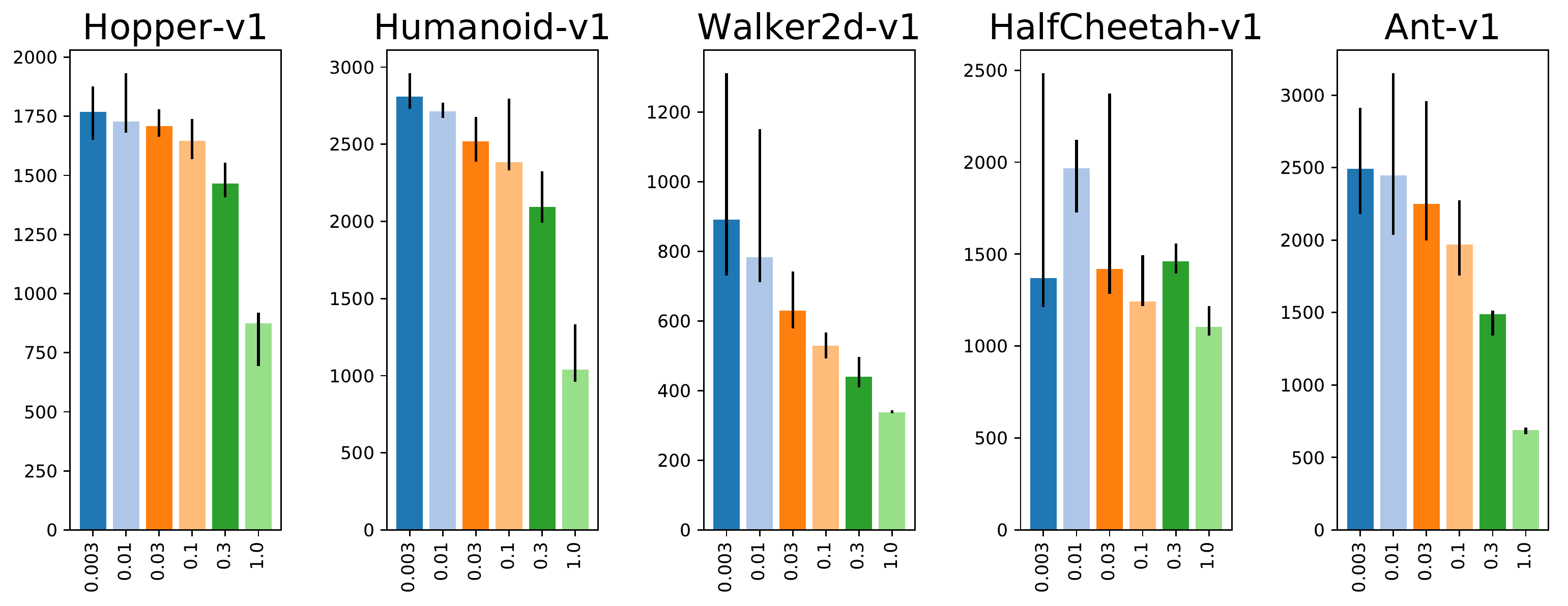}\hspace{1cm}\includegraphics[width=0.45\textwidth]{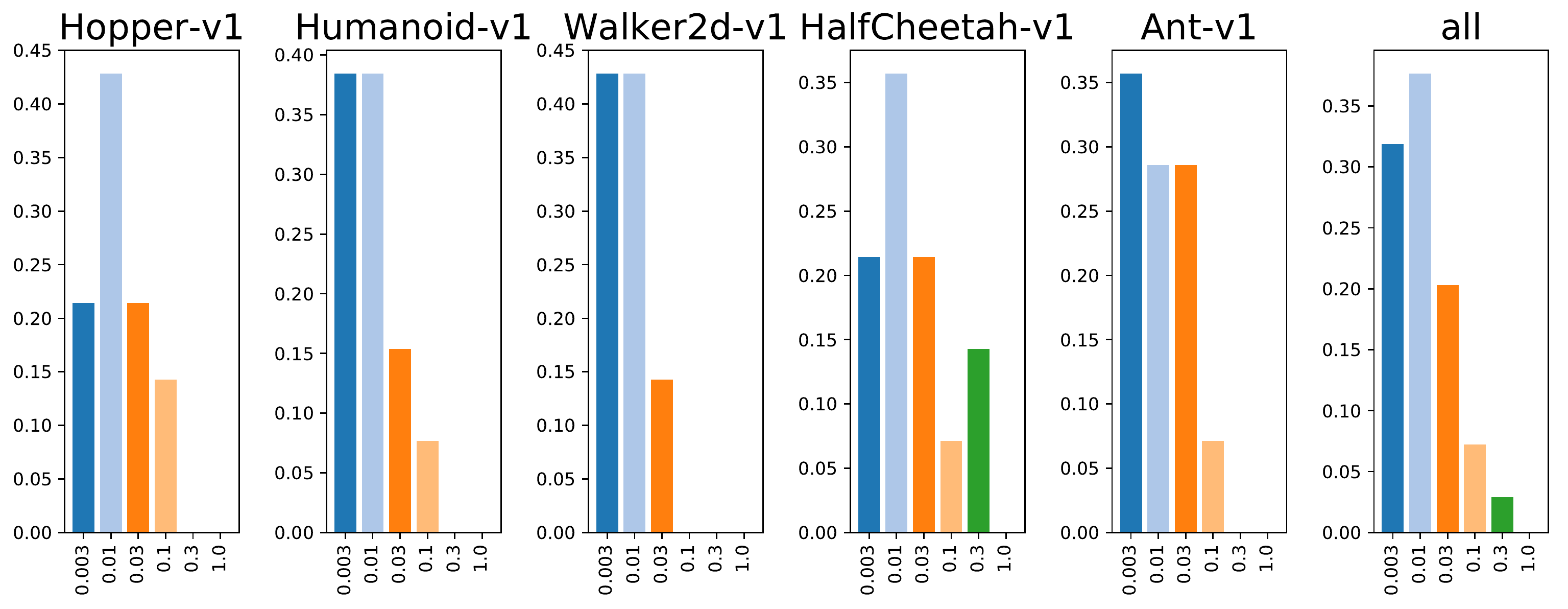}}
\caption{Analysis of choice \choicet{regularizerpenaltyklpimu}: 95th percentile of performance scores conditioned on sub-choice (left) and distribution of sub-choices in top 5\% of configurations (right).}
\label{fig:final_regularizer__gin_study_design_choice_value_sub_regularization_penalty_klpimu_coefficient}
\end{center}
\end{figure}

\begin{figure}[ht]
\begin{center}
\centerline{\includegraphics[width=0.45\textwidth]{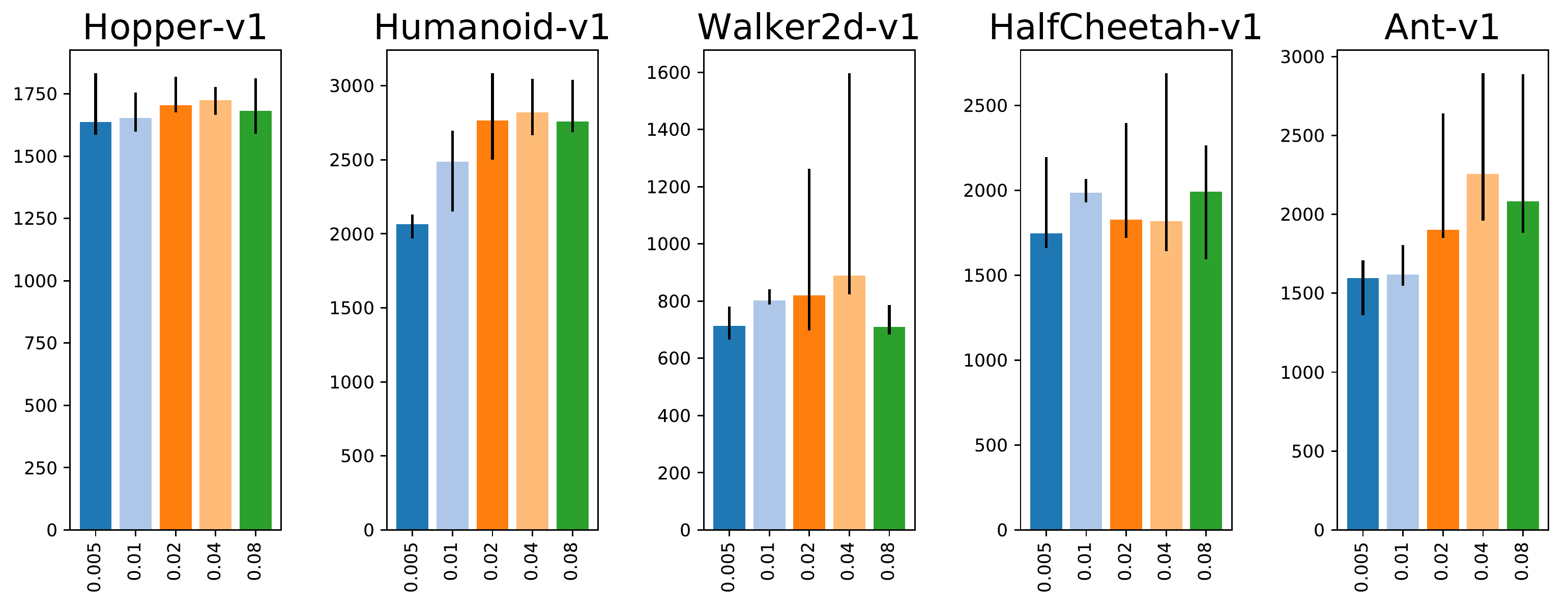}\hspace{1cm}\includegraphics[width=0.45\textwidth]{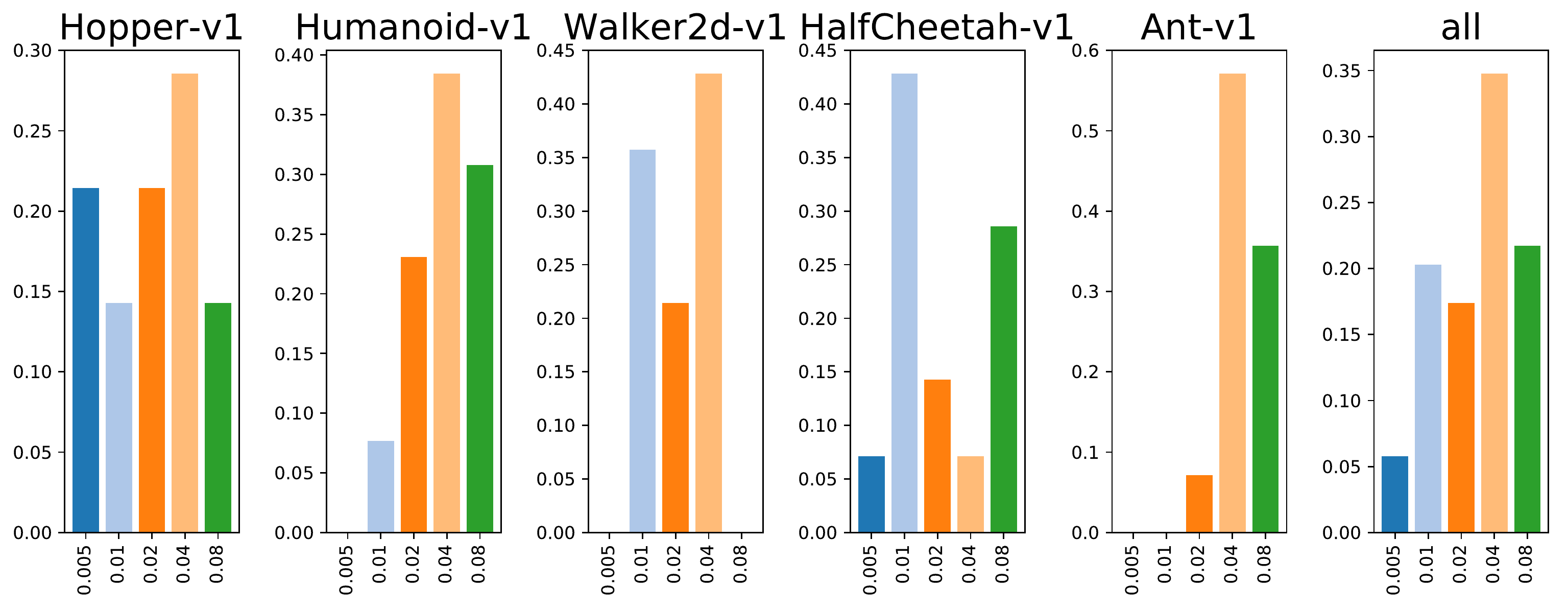}}
\caption{Analysis of choice \choicet{regularizerconstraintklmupimean}: 95th percentile of performance scores conditioned on sub-choice (left) and distribution of sub-choices in top 5\% of configurations (right).}
\label{fig:final_regularizer__gin_study_design_choice_value_sub_regularization_constraint_decoupled_klmupi_kl_mu_pi_mean_threshold}
\end{center}
\end{figure}

\begin{figure}[ht]
\begin{center}
\centerline{\includegraphics[width=0.45\textwidth]{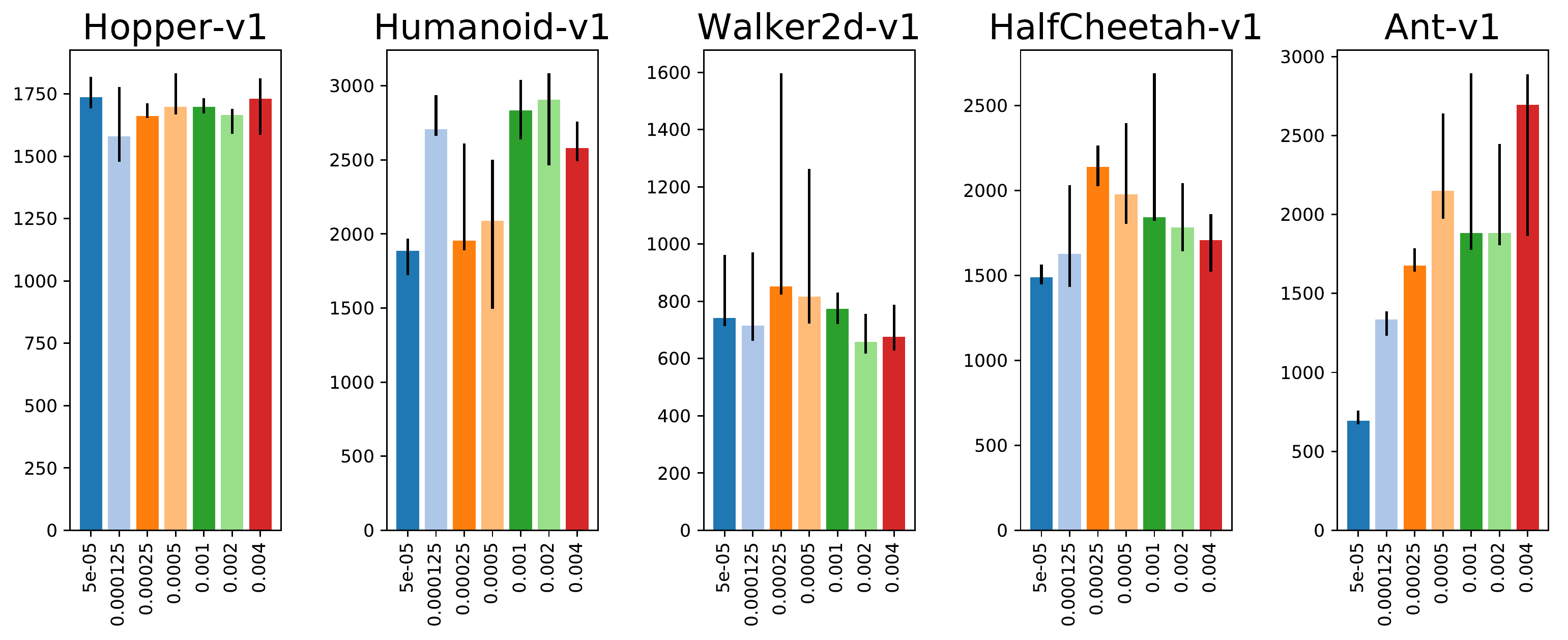}\hspace{1cm}\includegraphics[width=0.45\textwidth]{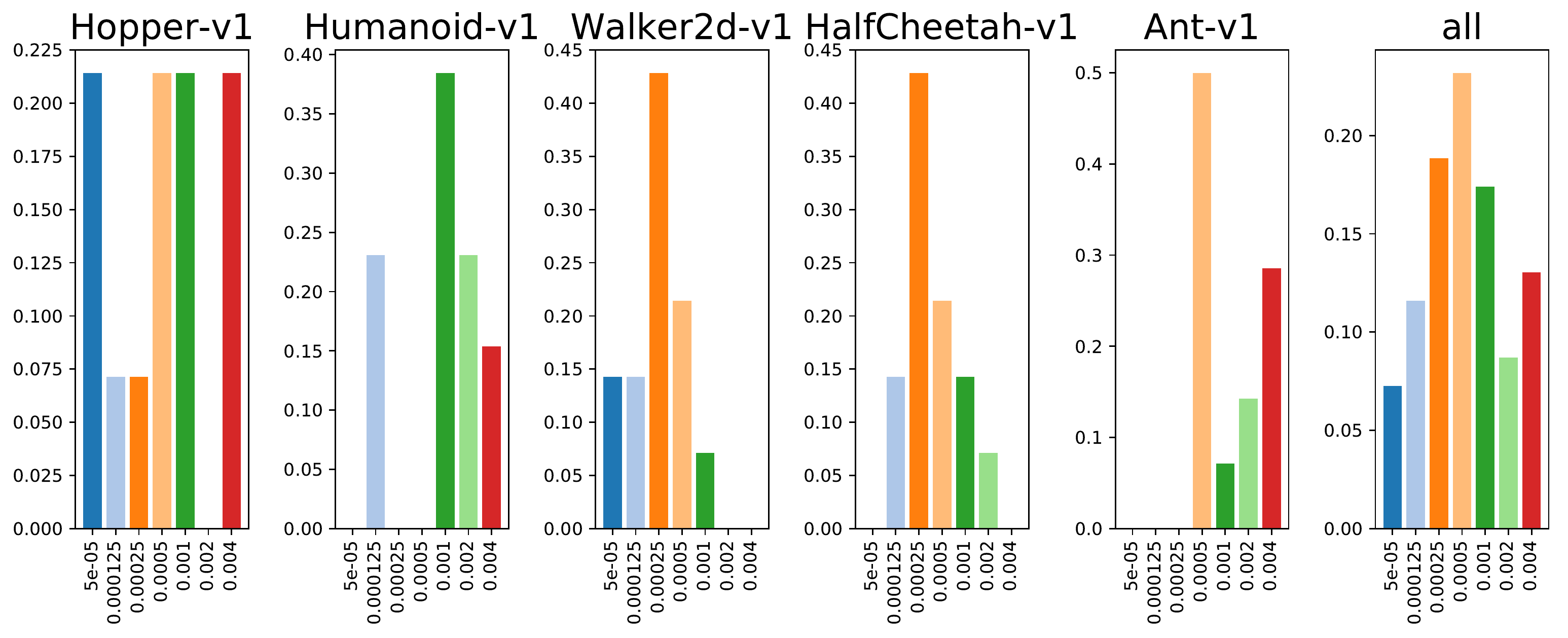}}
\caption{Analysis of choice \choicet{regularizerconstraintklmupistd}: 95th percentile of performance scores conditioned on sub-choice (left) and distribution of sub-choices in top 5\% of configurations (right).}
\label{fig:final_regularizer__gin_study_design_choice_value_sub_regularization_constraint_decoupled_klmupi_kl_mu_pi_std_threshold}
\end{center}
\end{figure}

\begin{figure}[ht]
\begin{center}
\centerline{\includegraphics[width=0.45\textwidth]{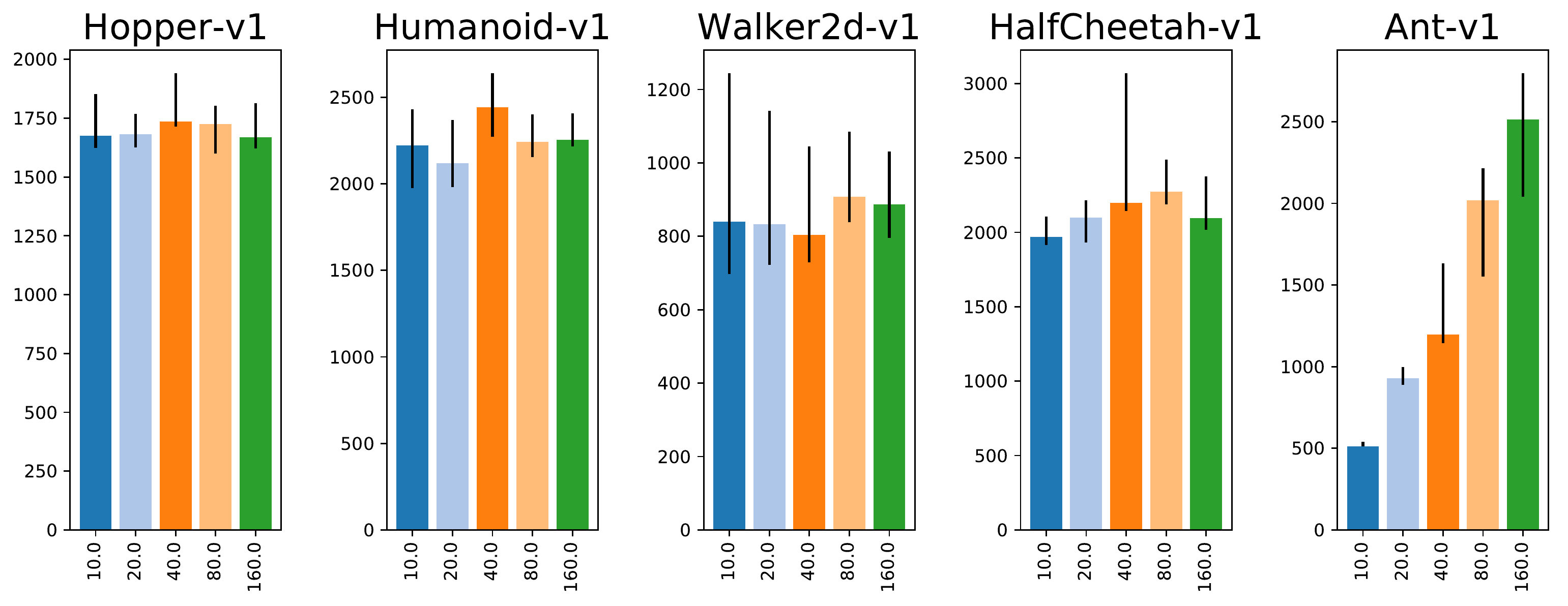}\hspace{1cm}\includegraphics[width=0.45\textwidth]{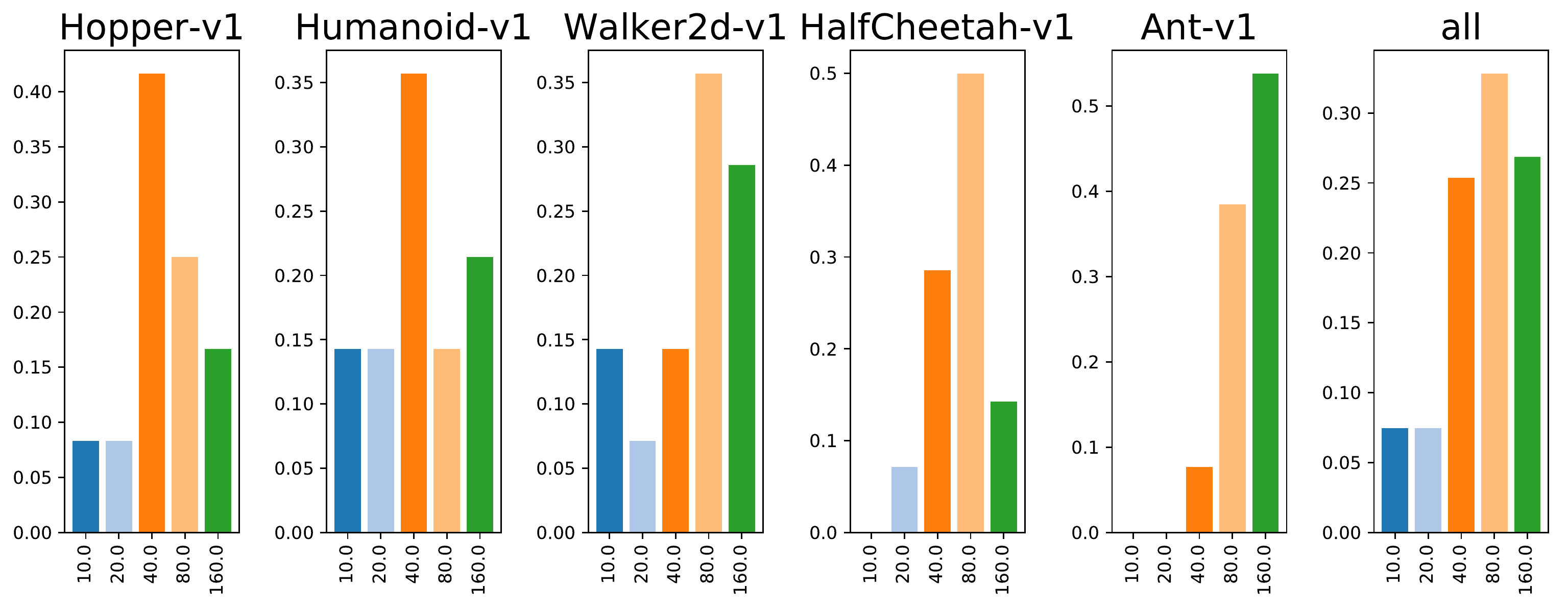}}
\caption{Analysis of choice \choicet{regularizerconstraintklrefpi}: 95th percentile of performance scores conditioned on sub-choice (left) and distribution of sub-choices in top 5\% of configurations (right).}
\label{fig:final_regularizer__gin_study_design_choice_value_sub_regularization_constraint_klrefpi_kl_ref_pi_threshold}
\end{center}
\end{figure}

\begin{figure}[ht]
\begin{center}
\centerline{\includegraphics[width=0.45\textwidth]{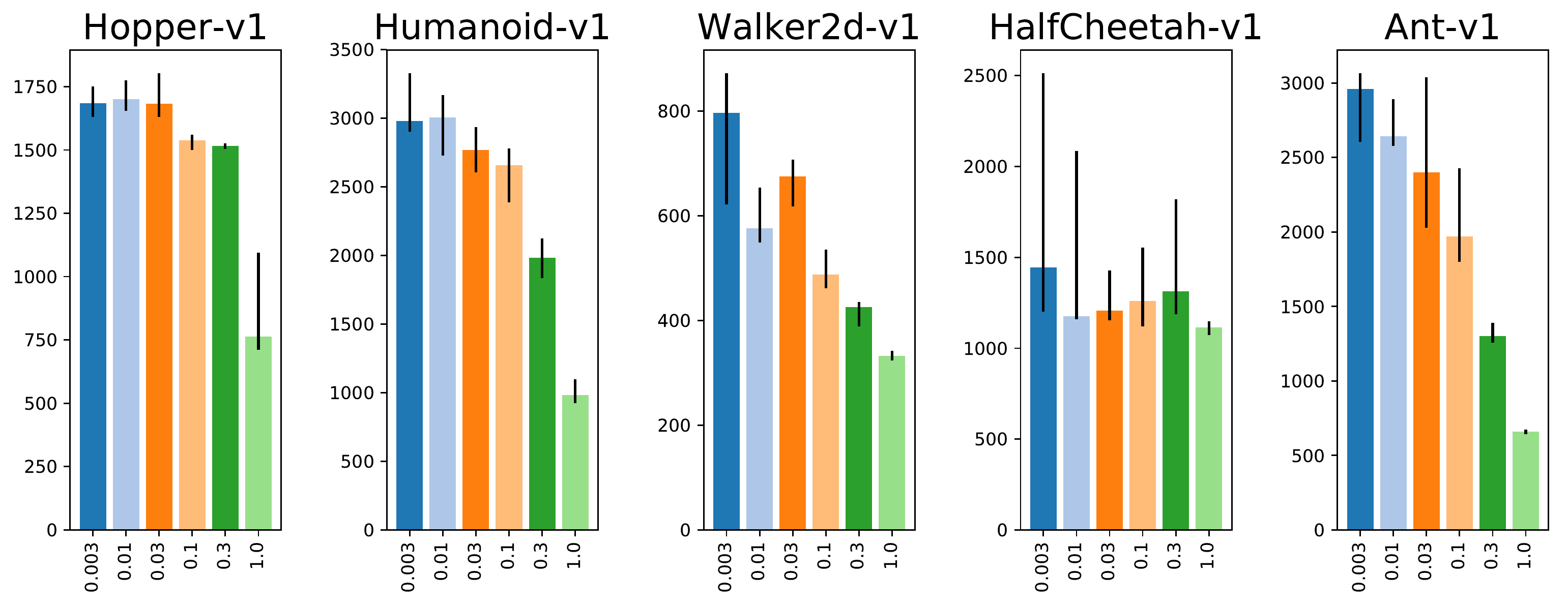}\hspace{1cm}\includegraphics[width=0.45\textwidth]{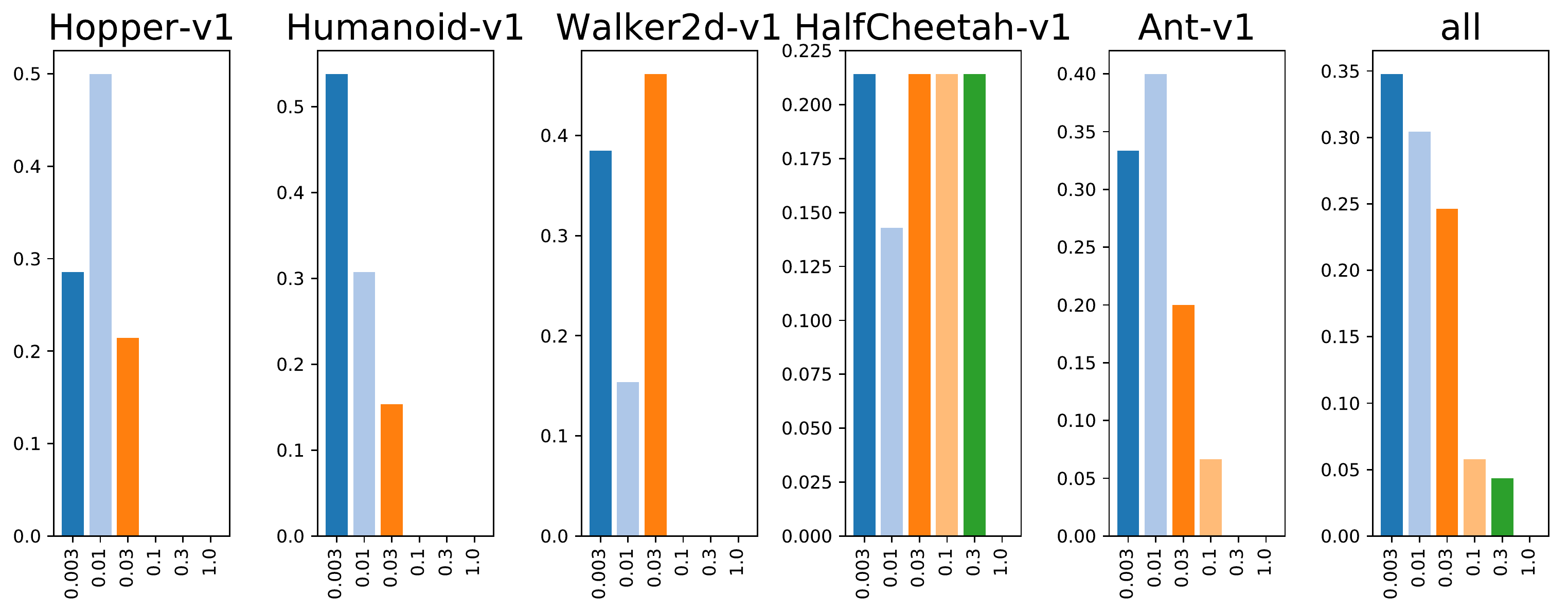}}
\caption{Analysis of choice \choicet{regularizerpenaltyklmupi}: 95th percentile of performance scores conditioned on sub-choice (left) and distribution of sub-choices in top 5\% of configurations (right).}
\label{fig:final_regularizer__gin_study_design_choice_value_sub_regularization_penalty_klmupi_coefficient}
\end{center}
\end{figure}

We report aggregate statistics of the experiment in Table~\ref{tab:final_regularizer_overview} as well as training curves in Figure~\ref{fig:final_regularizer_training_curves}.
For each of the investigated choices in this experiment, we further provide a per-choice analysis in Figures~\ref{fig:final_regularizer_custom_regularization}-\ref{fig:final_regularizer__gin_study_design_choice_value_sub_regularization_penalty_klmupi_coefficient}.
\begin{table}[ht]
\begin{center}
\caption{Performance quantiles across choice configurations.}
\label{tab:final_regularizer_overview}
\begin{tabular}{lrrrrr}
\toprule
{} & Ant-v1 & HalfCheetah-v1 & Hopper-v1 & Humanoid-v1 & Walker2d-v1 \\
\midrule
90th percentile &   2158 &           1477 &      1639 &        2624 &         705 \\
95th percentile &   2600 &           1870 &      1707 &        2832 &         814 \\
99th percentile &   2956 &           2413 &      1812 &        3071 &        1016 \\
Max             &   3202 &           3156 &      1979 &        3348 &        1597 \\
\bottomrule
\end{tabular}

\end{center}
\end{table}

\clearpage